\documentclass{tlp}
\usepackage{amsmath}
\usepackage{aopmath}
\usepackage{amssymb}
\usepackage{url}
\usepackage{xcolor}
\usepackage{textcomp}
\usepackage{IEEEtrantools}
\usepackage{enumitem}
\usepackage{rotating}
\usepackage{multirow}
\usepackage{cancel}
\usepackage{xspace}
\usepackage{footnote}
\makesavenoteenv{tabular}
\makesavenoteenv{table}

\setlist{ leftmargin={15pt} }

\usepackage{tikz}
\usetikzlibrary{external}
\usetikzlibrary{shapes.multipart}
\usetikzlibrary{fit,backgrounds,shapes,decorations.markings,decorations.pathmorphing,arrows,shadows,patterns,shapes.callouts,calc,intersections}
\tikzstyle{arg}        = [circle, thick, minimum size=0.2cm, draw=black, font=\scriptsize, fill=white, preaction={
  fill=gray,
  transform canvas={xshift=0.8mm,yshift=-0.8mm}
}]
\tikzstyle{vide}        = [thick, minimum size=0.2cm, draw=black, font=\scriptsize, fill=white, preaction={
  fill=gray,
  transform canvas={xshift=0.8mm,yshift=-0.8mm}
}]
\tikzstyle{ttvide}        = [thick, minimum size=0.2cm, draw=white, font=\normalsize, fill=white
]
\tikzstyle{tikzpict}   = [>=latex,text depth=0.25ex]

\usepackage{graphicx}
\usepackage{subfig}

\def\at{\mathit{At}}
\def\lb{\mathit{Lb}}
\def\ExtLb{\mathit{Lb_{ext}}}
\def\ExpLit{\mathit{At_{ext}}}
\def\ExtAt{\mathit{At_{ext}}}
\def\litExt{\mathit{Lit_{ext}}}
\def\R{\mathit{r}}
\def\lit{\mathit{Lit}}

\def\atp{\at^{p}}
\def\atn{\at^{n}}
\def\assume{\mathit{assume}}

\def\qed{~\hfill$\Box$}%

\def\negClas{\neg}
\def\negSet{\negClas}
\DeclareMathOperator{\Not}{\mathit{not}}
\DeclareMathOperator{\lparrow}{\ \leftarrow \ }
\let\oldempty\emptyset
\def\emptyset{\set{}}

\newcommand{\eqdef}{%
  \mathrel{\vbox{\offinterlineskip\ialign{%
    \hfil##\hfil\cr%
    $\scriptscriptstyle\mathrm{def}$\cr%
    \noalign{\kern1pt}%
    $=$\cr%
    \noalign{\kern-0.1pt}%
}}}}

\newcommand{\tuple}[1]{\ensuremath{\langle #1 \rangle}}
\newcommand{\setb}[1]{\ensuremath{\big\{\ #1\ \big\}}}
\newcommand{\setm}[2]{\ensuremath{\{\ #1\ \big|\ #2\ \}}}
\newcommand{\body}[1]{\ensuremath{\mathit{body}(#1)}}
\newcommand{\bodyp}[1]{\ensuremath{\mathit{body}^{+}(#1)}}
\newcommand{\bodyn}[1]{\ensuremath{\mathit{body}^{-}(#1)}}
\newcommand{\head}[1]{\ensuremath{\mathit{head}(#1)}}
\newcommand{\atom}[1]{\ensuremath{\mathit{atom}(#1)}}
\newcommand{\atoms}[1]{\ensuremath{\mathit{atoms}(#1)}}
\newcommand{\TA}[2]{\ensuremath{{\cal T\hspace{-3.5pt}A}_{\hspace{-1pt}#1}\hspace{-1pt}(#2)}}
\newcommand{\NANT}[1]{\ensuremath{N\hspace{-2.5pt}A\hspace{-1.5pt}N\hspace{-2.5pt}T\hspace{-1pt}(#1)}}
\newcommand{\WF}[1]{\ensuremath{W\hspace{-2.5pt}F_{#1}}}
\newcommand{\NR}[2]{\ensuremath{N\hspace{-2.5pt}R(#1,#2)}}
\newcommand{\Assumptions}[2]{\ensuremath{\mathit{Assumptions}(#1,#2)}}

\definecolor{darkred}{rgb}{0.7,0.0,0.0}
\newcommand{\changed}[1]{\colorlet{saved}{.}\color{black}#1\color{saved}\xspace}
\newcommand{\newChanged}[1]{#1}
\newcommand{\CHANGED}{\colorlet{saved}{.}\color{black}}
\newcommand{\END}{\color{saved}}

\newtheorem{example}{Example}
\newtheorem{definition}{Definition}
\newtheorem{observation}{Observation}
\newenvironment{examplecont}[1]{\begin{example}[Ex.~\ref{#1} continued]}{\end{example}}
\newenvironment{examplecontpage}[1]{\begin{example}[Ex.~\ref{#1} continued, page~\pageref{#1}]}{\end{example}}

\def\lfp{\mathtt{lfp}}
\def\gfp{\mathtt{gfp}}

\def\on{\text{\it on}}
\def\off{\text{\it off}}
\def\swa{\text{\it swa}}
\def\swb{\text{\it swb}}
\def\swc{\text{\it swc}}
\def\swd{\text{\it swd}}

\newcommand{\argument}[3]{\ensuremath{(#1,#2) \vdash #3}}
\newcommand{\posArg}{\textquotesingle$+$\textquotesingle}
\newcommand{\negArg}{\textquotesingle$-$\textquotesingle}

\def\spock{\text{\tt spock}}
\def\ouroboros{\text{\tt Ouroboros}}
\def\stepping{\text{\tt stepping}}
\def\sealion{\text{\tt SeaLion}}
\def\dwasp{\text{\sc dwasp}}
\def\wasp{\text{\sc wasp}}
\newcommand{\appRule}[1]{\ensuremath{\mathsf{applicable}(#1)}}
\newcommand{\blockRule}[1]{\ensuremath{\mathsf{blocked}(#1)}}
\newcommand{\ok}[1]{\ensuremath{\mathsf{ok}(#1)}}
\newcommand{\ko}[1]{\ensuremath{\mathsf{ko}(#1)}}
\newcommand{\abP}[1]{\ensuremath{\mathsf{unsatisfied}(#1)}}
\newcommand{\abC}[1]{\ensuremath{\mathsf{unsupported}(#1)}}
\newcommand{\abL}[1]{\ensuremath{\mathsf{unfounded}(#1)}}

\newcommand{\transKern}[1]{\ensuremath{\mathcal{T}_k[#1]}}
\newcommand{\transEx}[1]{\ensuremath{\mathcal{T}_{ex}[#1]}}
\newcommand{\inAS}[1]{\ensuremath{\mathsf{in}(#1)}}
\newcommand{\notinAS}[1]{\ensuremath{\mathsf{out}(#1)}}
\newcommand{\atomTrans}[1]{\ensuremath{\mathsf{atom}(#1)}}
\newcommand{\ruleTrans}[1]{\ensuremath{\mathsf{rule}(#1)}}
\newcommand{\headTrans}[2]{\ensuremath{\mathsf{head}(#1,#2)}}
\newcommand{\bodypTrans}[2]{\ensuremath{\mathsf{bodyP}(#1,#2)}}
\newcommand{\bodynTrans}[2]{\ensuremath{\mathsf{bodyN}(#1,#2)}}
\newcommand{\trueHead}[1]{\ensuremath{\mathsf{headSatisfied}(#1)}}
\newcommand{\supported}[1]{\ensuremath{\mathsf{supported}(#1)}}
\newcommand{\otherHeadTrue}[2]{\ensuremath{\mathsf{oHeadTrue}(#1,#2)}}
\newcommand{\noAS}{\ensuremath{\mathsf{noAS}}}
\newcommand{\metaP}{\ensuremath{\mathcal{P}_{meta}}}
\newcommand{\inputP}[1]{\ensuremath{\mathcal{P}_{inp}[#1]}}
\newcommand{\notAbL}[1]{\ensuremath{\mathsf{founded}(#1)}}
\newcommand{\headOutLoop}[1]{\ensuremath{\mathsf{headNotinLoop}(#1)}}
\newcommand{\bodyInLoop}[1]{\ensuremath{\mathsf{BodyPInLoop}(#1)}}
\newcommand{\inputPOur}[1]{\ensuremath{\varrho_{inp}[#1]}}
\newcommand{\interPOur}[1]{\ensuremath{\varrho_{int}[#1]}}
\newcommand{\metaPOur}{\ensuremath{\varrho_{meta}}}
\newcommand{\posCase}{\ensuremath{\mathit{Pos}}}
\newcommand{\negCase}{\ensuremath{\mathit{Neg}}}
\newcommand{\backTheory}{\ensuremath{\mathcal{B}}}
\newcommand{\abAtoms}[1]{\ensuremath{\mathit{Er}(#1)}}
\newcommand{\diagnosis}{\ensuremath{\mathcal{D}}}
\newcommand{\diagnosisSet}{\ensuremath{\mathbf{D}}}
\newcommand{\query}{\ensuremath{\mathcal{Q}}}
\newcommand{\diagnosisP}{\ensuremath{\mathbf{D^P}}}
\newcommand{\diagnosisN}{\ensuremath{\mathbf{D^N}}}
\newcommand{\diagnosisO}{\ensuremath{\mathbf{D^{\oldempty}}}}
\newcommand{\debP}[1]{\ensuremath{\mathcal{P}_{deb}[#1]}}
\newcommand{\debPGround}[1]{\ensuremath{\mathcal{P}_{deb}^G[#1]}}
\newcommand{\debugAtom}[2]{\ensuremath{\_debug(#1,#2)}}
\newcommand{\vars}[1]{\textit{\bf vars}_{#1}}

\newcommand{\core}{\ensuremath{C}}
\newcommand{\debAt}[1]{\ensuremath{At_{deb}(#1)}}
\newcommand{\diagnosisDwasp}{\ensuremath{\mathcal{D}_{\dwasp}}}

\newcommand{\ifonlyif}{iff}
\newcommand{\wrt}{w.r.t.}

\newcounter{programcount}
\newcounter{answersetcount}
\newcommand{\newprogram}{\refstepcounter{programcount}\ensuremath{P_{\arabic{programcount}}}}
\newcommand{\newanswerset}{\refstepcounter{answersetcount}\ensuremath{M_{\arabic{answersetcount}}}}
\newcommand{\programr}[2]{\ensuremath{P_{#1{#2}}}}
\newcommand{\answersetr}[2]{\ensuremath{M_{#1{#2}}}}
\DeclareMathOperator{\sneg}{\sim\!}
\newcommand{\evalues}{\ensuremath{\mathbf{E}_{\lb}}}
\newcommand{\wvalues}{\ensuremath{\mathbf{W}_{\lb}}}

\newcommand{\causes}{\ensuremath{\mathbf{C}_{\lb}}}
\newcommand{\values}{\ensuremath{\mathbf{V}_{\!\lb}}}

\DeclareMathOperator{\cdotl}{\ensuremath{\!\cdot\!}}
\newcommand{\Label}[1]{\mathit{label}(#1)}

\newcommand{\cgraph}[1]{\ensuremath{\mathit{graph}{(#1)}}}
\newcommand{\cgraphs}[1]{\ensuremath{\mathit{graphs}{(#1)}}}
\def\rA{a}
\def\rB{b}
\def\rC{c}
\def\rH{h}
\def\rL{l}
\def\R{r}
\def\P{P}
\def\nphead{k}

\def\npbody{n}
\def\nnbody{m}

\newcommand{\cI}{{\ensuremath{I}}}
\newcommand{\cJ}{{\ensuremath{J}}}
\newcommand{\wleq}{\sqsubseteq} 
\newcommand{\wless}{\sqsubset} 
\newcommand{\Atoms}{\mathit{Atoms}} 
\def\botI{\ensuremath{\mathbf{0}}}
\def\topI{\ensuremath{\mathbf{1}}}

\newcommand{\eI}{{\ensuremath{\cal I}}}
\newcommand{\eJ}{{\ensuremath{\cal J}}}

\newcommand{\cP}{{\ensuremath{P}}}

\newcommand{\eWpP}[1]{\ensuremath{\hat{\Gamma}_{#1}}}
\newcommand{\eWp}{\eWpP{\cP}}

\newcommand{\wlfpP}[1]{{\ensuremath{\mathfrak{T}_{#1}}}}
\newcommand{\wgfpP}[1]{{\ensuremath{\mathfrak{TU}_{#1}}}}

\newcommand{\wlfp}{{\ensuremath{\mathfrak{T}_\wP}}}
\newcommand{\wgfp}{{\ensuremath{\mathfrak{TU}_\wP}}}

\newcommand{\elfpP}[1]{{\ensuremath{\mathbb{L}_{#1}}}}
\newcommand{\egfpP}[1]{{\ensuremath{\mathbb{U}_{#1}}}}

\newcommand{\elfp}{{\elfpP{\cP}}}
\newcommand{\egfp}{{\egfpP{\cP}}}

\newcommand{\ereduct}[2]{{\ensuremath{#1^{#2}}}}

\newcommand{\dead}{\mathit{dead}}
\newcommand{\wet}{\mathit{wet}}
\newcommand{\shoot}{\mathit{shoot}}
\newcommand{\bulletproof}{\mathit{bulletproof}}
\newcommand{\damaged}{\mathit{damaged}}
\newcommand{\ab}{\mathit{ab}}
\newcommand{\harvey}{\mathit{harvey}}

\newcommand{\heads}{\mathit{head}}
\newcommand{\tails}{\mathit{tails}}

\newcommand{\trigger}{\mathit{trigger}}
\newcommand{\bullett}{\mathit{bullet}}
\newcommand{\gunpowder}{\mathit{gunpowder}}
\newcommand{\impact}{\mathit{impact}}
\newcommand{\haemorrhage}{\mathit{haemorrhage}}
\newcommand{\suzy}{\mathit{suzy}}

\newtheorem{notation}{Notation}

\newcounter{causalanswersetcount}
\newcommand{\newcausalanswerset}{\refstepcounter{causalanswersetcount}\ensuremath{\cI_{\arabic{causalanswersetcount}}}}
\newcommand{\causalanswersetr}[2]{\ensuremath{\cI_{#1{#2}}}}

\def\eM{\mathbb{M}}

\newcommand{\wI}{{\ensuremath{\tilde{\eI}}}}
\newcommand{\wJ}{{\ensuremath{\tilde{\eJ}}}}
\newcommand{\WpP}[1]{\ensuremath{\tilde{\Gamma}_{#1}}}
\newcommand{\Wp}{\WpP{\cP}}

\newcommand{\wP}{{\ensuremath{\mathfrak{P}}}}

\newcommand{\lambdap}{\ensuremath{\lambda}}

\def\Undef{\hbox{\normalfont undef} \ }

\def\KeepFacts{\mathit{KeepFacts}}
\def\MissingFacts{\mathit{MissingFacts}}
\def\KeepRules{\mathit{KeepRules}}
\def\RemoveRules{\mathit{RemoveRules}}
\def\RemoveFacts{\mathit{RemoveFacts}}
\def\NoFacts{\mathit{NoFacts}}

\begin{document}
\bibliographystyle{acmtrans}

\title[Answering the ``why'' in Answer Set Programming]{Answering the ``why'' in Answer Set Programming -- 
A Survey of Explanation Approaches}

\author[J. Fandinno and C. Schulz]
{JORGE FANDINNO \\
Institut de Recherche en Informatique de Toulouse (IRIT)\\
Universit{\'e} de Toulouse, CNRS\\
E-mail: jorge.fandinno@irit.fr
\and CLAUDIA SCHULZ \\
Ubiquitous Knowledge Processing (UKP) Lab\\
Technische Universit\"{a}t Darmstadt\\
E-mail: schulz@ukp.informatik.tu-darmstadt.de
}

\label{firstpage}
\pagerange{\pageref{firstpage}--\pageref{lastpage}}
\volume{\textbf{10} (3):}
\jdate{March 2002}
\setcounter{page}{1}
\pubyear{2002}


\maketitle

\begin{abstract}
Artificial Intelligence (AI) approaches to problem-solving and decision-making are becoming more and more
complex, leading to a decrease in the understandability of solutions. 
The European Union's new General Data Protection Regulation tries to tackle this problem by stipulating a 
``right to explanation'' for decisions made by AI systems.
One of the AI paradigms that may be affected by this new regulation is
Answer Set Programming~(ASP).
Thanks to the emergence of efficient solvers, 
ASP has recently been used for problem-solving in a variety of domains, including
medicine, cryptography, and biology.
To ensure the successful application of ASP as a problem-solving paradigm in the future,
explanations of ASP solutions are crucial.
In this survey, we give an overview of approaches that provide an answer to the question of
\emph{why} an answer set is a solution to a given problem, notably off-line justifications, causal graphs,  
argumentative explanations and \mbox{why-not} provenance, and highlight their similarities and differences.
Moreover, we review methods explaining why a set of literals is \emph{not} an answer set or why no solution
exists at all.
\emph{Under consideration in Theory and Practice of Logic Programming (TPLP)}
\end{abstract}
\begin{keywords}
answer set, explanation, justification, debugging
\end{keywords}

\section{Introduction}\label{sec:introduction}

With the increasing use of Artificial Intelligence methods in applications affecting all parts of our lives, the need for explainability of such methods is becoming ever more important.
The European Union recently put forward
a new General Data Protection Regulation (GDPR)~\cite{GDPregulation}, outlining how personal data may be collected, stored, and -- most importantly -- processed.
The GDPR reflects the current suspicion of the public towards automatic methods influencing our lives. 
It states\footnote{Article 22} that anyone has the right to reject a ``decision based solely on automated processing''  that ``significantly affects'' this person.
\changed{This new regulation may not come as a surprise since
most Artificial Intelligence methods are `black-boxes', that is, they produce accurate decisions, but without the means for humans to understand \emph{why} a decision was computed.}
According to \citeNS{Goodman2016european}, an implication of the GDPR is that, in the future, automatically computed decisions will only \changed{be} acceptable if they are explainable in a human-understandable manner.
The GDPR states that such an explanation \changed{needs to be} made of ``meaningful information about the logic involved'' in the automatic decision-making and should be communicated to the person concerned in a ``concise, intelligible and easily accessible form'' \cite{Goodman2016european}.

\changed{A popular Artificial Intelligengce \newChanged{paradigm} for decision-making and problem-solving is Answer Set Programming (ASP) \cite{BrewkaET2011,Lifschitz08}. It has proven useful in a variety of application areas, such as biology \cite{GebserSTV2011}, psychology \cite{Inclezan2015a,BalducciniG2010}, medicine \cite{ErdemO2015}, and music composition \cite{BoennBVF2011}.
 ASP is a declarative programming language used to specify a problem in terms of general inference rules and constraints, along with concrete information about the application scenario. 
For example, \citeNS{RiccaGAMLIL2012} present the problem of allocating employees of the large Gioia Tauro seaport into functional teams. To solve this problem, rules and constraints are formulated concerning, amongst others, team requirements and employees' shift constraints, along with factual knowledge about available employees.
The reasoning engine of ASP then infers possible team configurations, or more generally, solutions to the problem.
Such solutions are called \emph{stable models} or \emph{answer sets} \cite{GL88,GelfondL1991}.
Since the computation of answer sets relies on a `guess and check' procedure,
the question as to \emph{why} an answer set is a solution to the given problem can -- intuitively -- only be answered with ``because it fulfils the requirements of an answer set''. Clearly, this explanation does not provide ``meaningful information about the logic involved'', as required by the GDPR.
}

In ASP, the need for human-understandable explanations \changed{as to \emph{why} an answer set was computed,}
was recognised long before the new GDPR was put forward \cite{BrainV2008}.
Explanation approaches \changed{for ASP} have thus been developed for the past twenty years, each focusing on different aspects.
Some explain \emph{why} a literal is or is not contained in an answer set, using either the dependencies between literals or the \mbox{(non-)} application of rules as an explanation.
Other approaches provide explanations of the whole logic program, in other words, the explanation is not specific to one particular answer set. 
\changed{We will here refer to such explanations of logic programs that have some (potentially unexpected) answer set as \emph{justifications}. A different type of explanation is given by \emph{debugging}
approaches for ASP, which focus on explaining errors in logic programs.}
Such errors become apparent either if an unexpected answer set is computed or if the answer set computation fails, i.e. if the logic program is inconsistent. \changed{Debugging approaches thus aim to answer the question \emph{why} an unexpected answer set is computed or \emph{why} no answer set exists at all.}

In this survey paper, we outline and compare the most prominent justification approaches for ASP, notably,
off-line justifications \cite{PontelliSE2009}, LABAS justifications \cite{SchulzT2016}, causal justifications \cite{CabalarFF2014,CabalarF16}, and why-not provenance \cite{DamasioAA2013}.
Further related approaches outlined here are the 
formal theory of justifications \cite{DeneckerS1993,DeneckerBS2015} and
rule-based justifications \cite{BeatrixLGS2016}.
We will see that justifications obtained using these
\CHANGED
approaches significantly differ due to their ideological underpinnings.
\END
For example, causal justifications are inspired by causal reasoning, LABAS justifications by argumentative reasoning, \mbox{why-not} provenance by ideas from databases, and \mbox{off-line} justifications by Prolog tabled computations~\cite{roychoudhuryRR00}.
These ideological differences manifest themselves in the construction and layout of justifications, leading to variations in, for instance, the elements used in a justification (e.g. rules versus literals) and the treatment of negation (e.g. assuming versus further explaining negation-as-failure literals).

\changed{Besides explanation approaches for consistent logic programs under the answer set semantics, i.e. justification approaches, we 
review and discuss approaches for explaining inconsistent logic programs under the answer set semantics, i.e. debugging approaches, notably,}
\spock\ \cite{BrainGPSTW2007a,BrainGPSTW2007b,GebserPST08}, \ouroboros\ \cite{OetschPT10},
the interactive debugging approach by \citeNS{Shchekotykhin2015} that is built on top of \spock,
\dwasp\ \cite{AlvianoDFLR2013,AlvianoDLR15}, and \stepping\ \cite{OetschPT2017}.
We will see that these approaches \changed{form} three groups, which use different strategies for detecting errors in a logic program causing the inconsistency.
These strategies also lead to different \changed{types of} errors being pointed out to the user. \ 
\spock, \ouroboros\, and the interactive \spock\ approach use a program transformation to report unsatisfied rules, unsupported atoms, and unfounded atoms.
In contrast, \dwasp\ makes use of the solve-under-assumption and unsatisfiable core features of the \wasp\ solver \cite{AlvianoDFLR2013,AlvianoDLR15}, indicating faulty rules causing the inconsistency.
The \stepping\ approach uses the third strategy, namely a step-wise assignment of truth values to literals until a contradiction arises, which is then pointed out to the user.

The paper is structured as follows. We recall some background on logic programs and their semantics in Section~\ref{sec:background}.
\CHANGED
We then review ASP justification approaches in Section~\ref{sec:justifications} and
ASP debugging approaches in Section~\ref{sec:debugging}.
In Section~\ref{sec:related}, we give a brief historical overview of justifications for logic programs and discuss related work.
Finally, Section~\ref{sec:conclusion} concludes the paper, pointing out some issues with current approaches that provide interesting future work for the ASP community.
\END
%
%

\section{Syntax and Semantics of Logic Programs}\label{sec:background}

 \newChanged{In this section, we review the syntax and notation for disjunctive logic programs.
We also review the stable and the well-founded semantics for this class of programs,
which will be the basis for the works presented through the rest of the paper.}

We assume the existence of some \changed{(possibly empty or infinite)} set of atoms $\at$ and an operator $\Not$, denoting negation-as-failure (NAF)\footnote{sometimes called `default negation' in the literature}.
\mbox{$\lit \eqdef \at \cup \setm{ \Not a }{ a \in \at }$} denotes 
the set of literals over $\at$.
Literals of the form $a$ and $\Not a$ are respectively called \emph{positive} and \emph{negative}.
Given a literal $\rL \in \lit$, by~$\overline{\rL}$, we denote its complement, that is,
$\overline{\rL} \eqdef \Not a$ iff $\rL = a$
and
$\overline{\rL} \eqdef a$ iff $\rL = \Not  a$.
A \emph{rule} is an expression of the form
\begin{gather}
h_1 \vee \dotsc \vee h_k \lparrow b_1 \wedge \dotsc \wedge b_n \wedge \Not c_1 \wedge \dotsc \Not c_m
\label{eq:rule}
\end{gather}
where each $h_i$, $b_i$ and $c_i$ is an atom.
Given some rule $\R$ of the form of~\eqref{eq:rule},
by $\head{\R} \eqdef \set{ h_1, \dotsc h_k }$,
we denote the set of head atoms of the rule~$r$.
Similarly, by
\mbox{$\bodyp{\R} \eqdef \set{ b_1, \dotsc b_n }$}
and
\mbox{$\bodyn{\R} \eqdef \set{ c_1, \dotsc c_k }$},
we respectively denote the positive and negative body of $r$.
For a set of atoms $M \subseteq \at$
we denote the negative literals corresponding to atoms in $M$ by
\mbox{$\Not M \eqdef \setm{ \Not a }{ a \in M }$}.
Furthermore, by
$\body{\R} \eqdef \bodyp{\R} \cup \Not \bodyn{\R}$, we denote the body literals of $r$.
A rule is called \emph{normal} if it satisfies
\mbox{$\head{\R} = \set{h_1}$}
and \emph{positive} if
\mbox{$\bodyn{\R} = \emptyset$} holds.
A positive normal rule is called \changed{\emph{definite}.}
If \mbox{$\body{\R} = \emptyset$}, the rule is called a \emph{fact}\footnote{\CHANGED This includes disjunctive facts of the form $h_1 \vee \dotsc \vee h_k$.}
and we usually represent it omitting the symbol $\lparrow$.
We therefore sometimes use the term `fact' to refer to the literal(s) in a fact's head.
When dealing with normal rules, we sometimes denote by~$\head{r}$ the atom $h_1$ instead of the singleton set~$\set{h_1}$.
A rule with $\head{\R} = \emptyset$ is called \emph{constraint}.

A (logic) program $\P$ is a set of rules of the form of~\eqref{eq:rule}.
A program is called \emph{normal} (resp. \emph{positive} or \emph{\CHANGED definite\END}) \ifonlyif\ all its rules are
\emph{normal} (resp. positive or \CHANGED definite\END).

Given a set of atoms $M \subseteq \at$,
we write $\overline{M} \eqdef \at\backslash M$ for the set containing all atoms not belonging to~$M$.
We say that an atom $a$ is \emph{true} or \emph{holds} w.r.t. $M \subseteq \at$ when $a \in M$,
we say that it is \emph{false} otherwise. Similarly, we say that a negative literal $\Not a$ is \emph{true} or \emph{holds} w.r.t. $M \subseteq \at$ when $a \notin M$ and that it is \emph{false} otherwise. 
A rule $\R \in \P$ is
\emph{applicable} \wrt\ $M \subseteq \at$ \ifonlyif\ 
$\bodyp{\R} \subseteq M$ and
$\bodyn{\R} \cap M = \emptyset$, that is, when all body literals are true w.r.t. $M$.
A rule~$\R$ is \emph{satisfied} by $M$ \ifonlyif\
$\head{\R} \cap M \neq \emptyset$ 
whenever $\R$ is applicable.
$M \subseteq \at$ 
is \emph{closed} under $\P$ \ifonlyif\ every rule $\R \in \P$ is satisfied by~$M$.

\paragraph{Answer set semantics.}
\CHANGED
Intuitively, for an atom $a$, the literal $\Not a$ expresses that $a$ is false by default, i.e. unless it is proven to be true. The following definition of reduct and answer set~\cite{GL88} capture this intuition.\footnote{\CHANGED\citeN{GL88} define `stable models' rather than answer sets.
Later, \citeN{GelfondL1991} extended this definition to logic programs with explicit negation and with disjunction in the head, introducing the terms `answer set'. Since then, both terms are frequently used interchangeably.
We will here use the term answer set. }
\END
The \emph{reduct} of a program $\P$ w.r.t. a set of atoms $M \subseteq \at$, in symbols $\P^M$, is the result of applying the following two steps:%
\begin{enumerate}
\item removing all rules \newChanged{$\R$} such that $a \in M$ for some $a \in \bodyn{\R}$,
\item removing all negative literals from the remaining rules.
\end{enumerate}
The result is a positive program $\P^M$.
Then, a set of atoms $M \subseteq \at$ is an \emph{answer set} of a program~$\P$ \ifonlyif\ it is a \mbox{$\subseteq$-minimal} closed set under~$\P^M$.
A logic program is called \emph{consistent} if it has at least one answer set, and \emph{inconsistent} otherwise.
\CHANGED
Intuitively, a set of atoms is an answer set if all atoms in it are justified by the rules of the program under the assumption that all negative literals are evaluated \wrt\ this answer set.
\END

\begin{example}\label{ex:background}
\CHANGED
Let  $\newprogram\label{prg:background}$  be the logic program consisting of the following rules:
\begin{gather*}
\begin{IEEEeqnarraybox}{l C l}
p &\lparrow& q \wedge \Not r
\\
r &\lparrow& \Not p
\end{IEEEeqnarraybox}
\hspace{2cm}
\begin{IEEEeqnarraybox}{l C l}
s &\lparrow& t
\\
t &\lparrow& s
\end{IEEEeqnarraybox}
\hspace{2cm}
\begin{IEEEeqnarraybox}{l C l}
q
\\
\end{IEEEeqnarraybox}
\end{gather*}
and let $\newanswerset\label{as:1:prg:background}$ be the set of atoms $\set{p,q}$.
Then, the reduct of $\programr\ref{prg:background}$ \wrt\ $\answersetr\ref{as:1:prg:background}$
is the program $\programr\ref{prg:background}^{\answersetr\ref{as:1:prg:background}}$:
\begin{gather*}
\begin{IEEEeqnarraybox}{l C l}
p &\lparrow& q
\\
\end{IEEEeqnarraybox}
\hspace{2cm}
\begin{IEEEeqnarraybox}{l C l}
s &\lparrow& t
\\
t &\lparrow& s
\end{IEEEeqnarraybox}
\hspace{2cm}
\begin{IEEEeqnarraybox}{l C l}
q
\\
\end{IEEEeqnarraybox}
\end{gather*}
whose $\subseteq$-minimal closed set is precisely $\set{p,q}$.
Hence, $\answersetr\ref{as:1:prg:background}$ is an answer set of 
$\programr\ref{prg:background}$.
Intuitively, $q$ is in the answer since it is a fact in the program, while $p$ is in the answer set due to the rule $p \lparrow q \wedge \Not r$ and the fact that $q$ is true and $r$ is assumed to be false \wrt\ $\answersetr\ref{as:1:prg:background}$.
Note that $s$ and $t$ mutually depend on each other, so there is no reason to believe either of them, and consequently neither is contained in the answer set.
It is easy to check that program~$\programr\ref{prg:background}$ has a second answer set $\set{q,r}$.\qed
\end{example}

\paragraph{Well-founded model semantics.}
We introduce a definition of the well-founded model semantics for normal logic programs in terms of the least fixpoint of a $\Gamma_\P$ operator~\cite{van1989alternating}
which is, though equivalent, slightly different from the original definition by~\citeN{van1988unfounded} and \citeN{van1991well}.
Given a normal logic program~$\P$,
let $\Gamma_\P$ be the function mapping each set of atoms $M$ to the \changed{$\subseteq$-minimal closed set} of the program~$\P^M$
and let $\Gamma_\P^2$ be the operator mapping each set $M$ to $\Gamma_\P(\Gamma_\P(M))$.
Then, $\Gamma_\P$ and $\Gamma_\P^2$ are antimonotonic and monotonic, respectively, and, consequently, the latter has a least and greatest fixpoint, which we respectively denote by $\lfp(\Gamma_\P^2)$ and $\gfp(\Gamma_\P^2)$.
We also respectively denote by
\mbox{$\WF{\P}^{+} \eqdef \lfp(\Gamma_\P^2)$}
and
\mbox{$\WF{\P}^{-} \eqdef (\at\backslash\gfp(\Gamma_\P^2))$}
the set of true and false atoms in the well-founded model of $\P$.
The well-founded model of $\P$ can then be defined as the set of literals: $\WF{\P} \eqdef \WF{\P}^{+} \cup \Not \WF{\P}^{-}$.
The well-founded model is said to be \emph{complete} \ifonlyif\ 
\mbox{$\WF{\P}^{+} \cup \WF{\P}^{-} = \at$}.
We say that an atom $a$ is \emph{true} \wrt\ the well-founded model if $a \in \WF{\P}$, \emph{false} if $\Not a \in \WF{\P}$, and \emph{undefined} otherwise.

It is easy to see that, by definition, the answer sets of any normal program $\P$ coincide with the fixpoints of $\Gamma_P$ and, thus, every stable model is also a fixpoint of~$\Gamma_P^2$.
Hence, every stable model $M$ satisfies: 
$\WF{\P}^{+} \subseteq M$
and
$\WF{\P}^{-} \cap M = \emptyset$.
\changed{In other words, the well-founded model semantics is more sceptical than the answer set semantics in the sense that all atoms that are true (resp. false) in the well-founded model are also true (resp. false) in all answer sets.}

\begin{examplecont}{ex:background}
\CHANGED
Continuing with our running example, it is easy to see that
$\programr\ref{prg:background}^{\emptyset}$ is:
\begin{gather*}
\begin{IEEEeqnarraybox}{l C l}
p &\lparrow& q
\\
r &\lparrow&
\end{IEEEeqnarraybox}
\hspace{2cm}
\begin{IEEEeqnarraybox}{l C l}
s &\lparrow& t
\\
t &\lparrow& s
\end{IEEEeqnarraybox}
\hspace{2cm}
\begin{IEEEeqnarraybox}{l C l}
q
\\
\end{IEEEeqnarraybox}
\end{gather*}
and that its $\subseteq$-minimal model is $\set{p,q,r}$.
Hence, we have that 
$\Gamma_{\programr\ref{prg:background}}(\emptyset)= \set{p,q,r}$.
In a similar way, it can be checked that
$\Gamma^2_{\programr\ref{prg:background}}(\emptyset) = \Gamma^4_{\programr\ref{prg:background}}(\emptyset) = \set{q}$
is the least fixpoint of the $\Gamma^2_{\programr\ref{prg:background}}$ operator.
Hence, we have that 
$\WF{\programr\ref{prg:background}} = \set{q, \Not s, \Not t }$.
As expected, $q$ is true in all answer sets of $\programr\ref{prg:background}$ while $s$ and $t$ are false in all of them. Furthermore, $p$ and $r$ are true in one answer set but not in the other and are left undefined in the well-founded model.
Note that it is possible that an atom is true in all answer sets, but undefined in the well-founded model. For instance, $\answersetr\ref{as:1:prg:background} = \set{p,q}$ is the unique answer set of
$\programr\ref{prg:background} \cup \set{ u \lparrow r \wedge \Not u }$, but $p$ is still undefined in its well-founded model.
\qed
\end{examplecont}

\paragraph{Explicit negation.}
In addition to negation-as-failure, we use the operator $\negClas$ to denote \emph{explicit negation}.
For an atom $a$, $\negClas a$ denotes the contrary of $a$. 
By $\negSet S \eqdef \setm{\negClas a}{a \in S}$ we denote the explicitly negated atoms of a set $S \subseteq \at$
and, by $\ExpLit \eqdef \at \cup \negSet \at$ we denote the set of \emph{extended atoms} consisting of atoms and explicitly negated atoms.
By $\litExt \eqdef \ExpLit \cup \setm{ \Not \rA }{ \rA \in \ExpLit }$, we denote the set of \emph{extended literals} over $\at$. 
As for logic programs without explicit negation, extended literals $\negClas a$ and $\Not \negClas a$ are respectively called
\emph{positive} and {negative}.

\CHANGED
An \emph{extended rule} is an expression of the form~\eqref{eq:rule}
where each $\rH_i$, $\rB_i$ and $\rC_i$ is an extended atom.
An \emph{extended} (logic) program is a set of extended rules.
\END
The notions of head, body, etc. directly carry over from rules without explicit negation.
Note that we say that a program is positive when it does not contain negation-as-failure, even if it contains explicit negation.

The definition of answer sets and \changed{well-founded model\footnote{\CHANGED Even though this simply transfer is sufficient for the purpose of this paper, for the well-founded model semantics the property ensuring that the explicit negation of a formula implies its default negation \newChanged{is} lost. For a detailed study and solution of this problem we refer to the work of~\citeN{PereiraA92}.} are} easily transferred to extended logic programs by replacing $M \subseteq \at$ with $M \subseteq \ExtAt$.
If an answer set \changed{(resp. the well-founded model)} contains both an atom $a$ and its contrary $\negClas a$, the answer set is called \emph{contradictory} \cite{GelfondL1991,Gelfond2008}.
In some works \cite{GelfondL1991}, a contradictory answer set is only an answer set if the program has no other answer set and is, by definition, $\ExtAt$.

\begin{example}\label{ex:exp-neg}
\CHANGED
Let  $\newprogram\label{prg:exp-neg}$  be the logic program consisting of the following rules:
\begin{gather*}
\begin{IEEEeqnarraybox}{l C l}
p &\lparrow& q \wedge \Not r
\\
r &\lparrow& \Not p
\end{IEEEeqnarraybox}
\hspace{2cm}
\begin{IEEEeqnarraybox}{rl}
\neg& p
\\
&q
\end{IEEEeqnarraybox}
\end{gather*}
and let $\newanswerset\label{as:1:prg:exp-neg}$ be the set of extended atoms $\set{\neg p,q,r}$.
Then, the reduct of $\programr\ref{prg:exp-neg}$ \wrt\ $\answersetr\ref{as:1:prg:exp-neg}$
is the program $\programr\ref{prg:exp-neg}^{\answersetr\ref{as:1:prg:exp-neg}}$:
\begin{gather*}
\begin{IEEEeqnarraybox}{l C l}
\\
r &\lparrow&
\end{IEEEeqnarraybox}
\hspace{2cm}
\begin{IEEEeqnarraybox}{rl}
\neg& p
\\
&q
\end{IEEEeqnarraybox}
\end{gather*}
whose $\subseteq$-minimal closed set is precisely $\set{\neg p,q,r}$.
Hence, $\answersetr\ref{as:1:prg:exp-neg}$ is an answer set of~$\programr\ref{prg:exp-neg}$.
Note that there is a second answer set $\set{p,\neg p, q}$ which is contradictory.
According to the definition of \citeN{GelfondL1991}, $\answersetr\ref{as:1:prg:exp-neg}$ is thus the only answer set.\qed
\end{example}

\section{Justifications of Consistent Logic Programs}
\label{sec:justifications}
 
\CHANGED
In this section, we review the most prominent approaches for explaining \emph{consistent} logic programs under the answer set semantics.
All approaches reviewed here, except for the formal theory of justifications (Section~\ref{sec:formal}), aim to provide concise structures called justifications that provide a somewhat minimal explanation as to why a literal in question belongs to an answer set.
 
We start by introducing off-line (Section~\ref{sec:offline};~\citeNP{PontelliSE2009,PontelliS2006}),
LABAS (Section~\ref{sec:labas};~\citeNP{SchulzT2016,SchulzT2013})
and causal justifications (Section~\ref{sec:causal}; \citeNP{CabalarFF2014,CabalarF16}).
In these three approaches, justifications are represented as different kinds of dependency graphs between literals and/or rules.
Next, we review why-not provenance justifications (Section~\ref{sec:why-not};~\citeNP{DamasioAA2013}), which represent justifications as propositional formulas instead of graph structures.
It is interesting to note that why-not provenance and causal justifications share a multivalued semantic definition based on
a lattice.
Finally, we sketch the main idea of
rule-based justifications \cite{BeatrixLGS2016}
and the
formal theory of justifications \cite{DeneckerS1993,DeneckerBS2015}
in Section~\ref{sec:other_justifications}.
\END


\subsection{Off-line Justifications}
\label{sec:offline}

Off-line justifications~\cite{PontelliSE2009,PontelliS2006} are graph structures that describe the reason for the truth value of an atom with respect to a given answer set.
In particular, each off-line justification describes the derivation of the truth value (that is, true or false) of an atom using the rules in the program.
\CHANGED
Each vertex of such a graph represents an atom and each edge the fact that the two vertices that it joins are related by some rule in the program, with the edge pointing from the head of the rule to some atom in its body.
Atoms that are true with respect to a given answer set are labelled `$+$', whereas atoms that are false with respect to it are labelled `$-$' (see condition~\ref{item:atomlabels:def:off-line.explanation} in Definition~\ref{def:off-line.explanation} below).
Similarly, edges labelled `$+$' represent positive dependencies
while those labelled `$-$' represent negative ones.
This is reflected in conditions~\ref{item:bodyp:def:off-line.explanation} (a true atom is supported by a true atom through a positive dependency and by a false atom through a negative dependency)
and condition~\ref{item:bodyn:def:off-line.explanation} of Definition~\ref{def:off-line.explanation} below (a false atom is supported by a false atom through a positive dependency and by a true atom through a negative dependency). 

Before we technically describe off-line justifications, we need the following notation:
\END
for any set of atoms $S \subseteq \at$,
the sets of \emph{annotated atoms} are defined as
$S^{p} \eqdef \setm{ a^{+} }{ a \in S }$
and
$S^{n} \eqdef \setm{ a^{-} }{ a \in S }$.
Furthermore, given an annotated atom $a^{\pm}$ (that is, $a^{\pm} = a^{+}$ or $a^{\pm} = a^{-}$), by $\atom{a^{\pm}} = a$ we denote the atom associated with $a^{\pm}$.
Given a set of annotated atoms $S$, by
$\atoms{S} \eqdef \setm{ \atom{a^{\pm}} }{a^{\pm} \in S }$, we denote the set of atoms associated with the annotated atoms in $S$.

\begin{definition}[Off-line Explanation Graph]\label{def:off-line.explanation}
Let $\P$ be a normal logic program, let
$M, U \subseteq \at$ be two sets of atoms, and let
\mbox{$a^{\pm} \in (\atp \cup \atn)$} be an annotated atom%
\footnote{Off-line justifications were defined without using explicit negation, so we here stick to logic programs without explicit negation. However, it is easy to see that they can be applied to extended logic program by replacing $\at$ by $\ExtAt$.}.
An \emph{off-line explanation graph} 
of $a^{\pm}$ w.r.t.~$\P$, $M$ and~$U$
is
a labelled, directed graph \mbox{$G = \tuple{V,E}$} with a set of vertices
\mbox{$V \subseteq (\atp \cup \atn \cup \set{\assume,\top,\bot})$}
and a set of edges
\mbox{$E \subseteq (V \times V \times \set{+,-})$}, which
satisfies the following conditions:
\begin{enumerate}
\item $a^{\pm} \in V$ and every $b \in V$ is reachable from $a^{\pm}$,
\item the only sinks in the graph are: $\assume$, $\top$ and $\bot$,
\item $\atoms{V \cap \atp} \subseteq M$ and $\atoms{V \cap \atn} \subseteq (\overline{M} \cup U)$, \label{item:atomlabels:def:off-line.explanation}

\item The set of edges $E$ satisfies the following two conditions:
 \begin{enumerate}
 \item $\setm{ c }{ (b^{+}, c^{-},+) \in E} \cup \setm{ c }{ (b^{+}, c^{+},-) \in E} = \emptyset$
and
\item $\setm{ c }{ (b^{-}, c^{+},+) \in E} \cup  \setm{ c }{ (b^{-}, c^{+},-) \in E} = \emptyset$,
 \end{enumerate}
 \label{item:edges:def:off-line.explanation}

\item every
$b^{+} \in V$ satisfies that there is a rule~$r \in \P$ with $\head{r} = b$ s.t.
 \begin{enumerate}
 \item $\body{r} = \setm{ c }{ (b^{+}, c^{+},+) \in E} \cup  \setm{ \Not c }{ (b^{+}, c^{-},-) \in E}$, or
 \label{item:bodyp:def:off-line.explanation}

 \item both $\body{r} = \emptyset$ and $(b^{+}, \top,+)$ is the unique edge in $E$ with source $b^{+}$,
 \end{enumerate}
\item every $b^{-} \in V$ with $b \in U$ satisfies that $(b^{-},\assume,-)$ is the only edge with source $b^{-}$,
\label{item:assum:def:off-line.explanation}

\item every $b^{-} \in V$ with $b \notin U$ and no rule $r \in \P$ with $\head{r} = b$
 satisfies that $(b^{-},\bot,+)$ is the only edge with source $b^{-}$,

\item every $b^{-} \in V$ with $b \notin U$ and some rule $r \in \P$ with $\head{r} = b$ satisfies that
$S = \setm{ c }{ (b^{-}, c^{-},+) \in E} \cup \setm{ \Not c }{ (b^{-}, c^{+},-) \in E}$ is a minimal set of literals
such that every rule $r' \in \P$  with $\head{r'} = b$ satisfies
\mbox{$\body{r'} \cap S \neq \emptyset$}.\qed
\label{item:bodyn:def:off-line.explanation}
\end{enumerate}
\end{definition}	

\noindent
Intuitively, $M$ represents some answer set and $U$ represents a set of assumptions with respect to $M$.
These assumptions derive from the inherent `guessing' process involved in the definition and algorithmic construction of answer sets.
In this sense, the truth value of assumed atoms has no further justification while non-assumed atoms must be justified by the rules of the program.
This is reflected in condition~\ref{item:assum:def:off-line.explanation} of Definition~\ref{def:off-line.explanation}.
Note also that this condition ensures that true elements are not treated as assumptions, which follows from the intuition that any true atom in an answer set must be justified.
Condition~\ref{item:edges:def:off-line.explanation} ensures that a labelled atom is not supported by the wrong type of relation.

The following example illustrates how assumptions are used to justify
atoms that are false \wrt\ an answer set in question.

\begin{example}\label{ex:off-line.cycle}
Let $\newprogram\label{prg:cycle}$ be the program containing the following two rules:
\begin{gather*}
p \lparrow \Not q
\hspace{2cm}
q \lparrow \Not p
\end{gather*}
Program~$\programr\ref{prg:cycle}$ has two answer sets,
namely
\mbox{$\newanswerset\label{as:1:prg:cycle} = \set{p}$}
and
\mbox{$\newanswerset\label{as:2:prg:cycle} = \set{q}$}.
Figure~\ref{fig:off-line.cycle} depicts the off-line explanation graphs
justifying the truth of $p$ (annotated atom~$p^{+}$) and the falsity of $q$ (annotated atom~$q^{-}$) with respect to the program $\programr\ref{prg:cycle}$, the answer set
 $\answersetr\ref{as:1:prg:cycle}$ and the set of assumptions $\set{q}$.
Note that the falsity of $q$ is assumed in both justifications.\qed
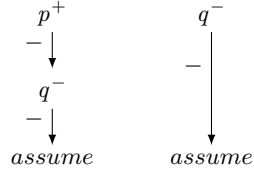
\begin{figure}\centering
\begin{tikzpicture}[tikzpict]
    \matrix[row sep=0.5cm,column sep=0.8cm,ampersand replacement=\&] {
      \node (p) {$p^{+}$};
      \&
      \node (q2) {$q^{-}$};
      \\
      \node (q) {$q^{-}$};
      \\
      \node (assume) {$\assume$};
      \&
      \node (assume2) {$\assume$};
      \\
     };
    \draw [->] (p) to node[pos=0.3,left]{$-$} (q);
    \draw [->] (q) to  node[pos=0.3,left]{$-$} (assume);
    \draw [->] (q2) to  node[pos=0.3,left]{$-$} (assume2);
\end{tikzpicture}%
\caption{Off-line justifications of $p^{+}$ and $q^{-}$ w.r.t. $\answersetr\ref{as:1:prg:cycle}= \set{p}$ 
in Example~\ref{ex:off-line.cycle}.
The assumption is $\set{q}$.
}
\label{fig:off-line.cycle}
\end{figure}
\end{example}

To ensure that the set of assumptions is meaningful with respect to the answer set being explained, it needs to be restricted.
In particular, it will be restricted to a subset of atoms that are false \wrt\ the answer set and undefined \wrt\ the well-founded model.
\CHANGED
As mentioned above, assumptions are restricted to be false atoms to follow the intuition that any true atom in an answer set must be justified.
Restricting the set of assumptions further to only those that are undefined \wrt\ the well-founded model 
ensures that false atoms that are also false \wrt\ to the well-founded model are justified by the constructive process of the well-founded model rather than being assumed.
\END
The following notation is needed to achieve this restriction:

\begin{definition}\label{def:negative.atoms}
Given a normal program $\P$, by
\mbox{$\NANT{\P} \eqdef \setm{ b \in \at }{ \exists r \in \P \text{ s.t. } b \in \bodyn{r} }$},
we denote the set of atoms that occur negated in $\P$.\qed
\end{definition}

\begin{definition}[Negative Reduct]\label{def:negative.reduct}
Given a normal program $\P$, by
\mbox{$\NR{\P}{U} \eqdef \setm{ r \in \P}{ \head{r} \notin U}$},
we denote the \emph{negative reduct} of $\P$ w.r.t. some set of atoms $U \subseteq \at$.\qed
\end{definition}

\begin{definition}[Assumptions]
Let $\P$ be a normal program and
$M$ an answer set of $\P$.
Let us denote by
\begin{gather*}
\TA{\P}{M}
	\ \eqdef \ \setm{ a \in \NANT{\P} }{ a \in \overline{M} \text{ and } a \notin (\WF{\P}^{+} \cup \WF{\P}^{-}) }
\end{gather*}
the \emph{tentative assumptions} of $\P$ w.r.t.~$M$.
Then, an \emph{assumption} w.r.t~$M$ is a set of atoms $U \subseteq \TA{\P}{M}$ such that
\changed{$\WF{\NR{\P}{U}}^{+} = M$}.
The set of all possible assumptions of $\P$ w.r.t. $M$ is denoted by $\Assumptions{\P}{M}$.\qed
\end{definition}

An interesting observation to make is that
$\TA{\P}{M}$ is
always an element of the set
$\Assumptions{\P}{M}$
and, therefore, the latter is never empty.
\CHANGED
Intuitively, an assumption is a set of atoms that are false w.r.t. the considered answer set and that, when `forced to be false' in the program,
 produces a complete \mbox{well-founded} model that coincides with this answer set.
 The negative reduct (see Definition~\ref{def:negative.reduct}), removing all rules whose head belongs to the assumption, \newChanged{can be} interpreted as `forcing atoms to be false' since it results in all atoms in the assumption being false in the well-founded model.
 Then, since the computation of the well-founded model is deterministic, no guessing is necessary.
\END
\CHANGED
Justifications relative to the \mbox{well-founded} model can thus be used for the explanation \wrt\ an answer set by adding edges that point out which atoms in the assumption were used to obtain the answer set.
\END
This is formalised as follows:

\begin{definition}[Off-line Justification]\label{def:off-line.justification}
Let $\P$ be a normal program,
$M$ an answer set of $\P$,
$U \in \Assumptions{\P}{M}$ an assumption w.r.t $M$ and $\P$, and
$a^{\pm} \in (\atp \cup \atn)$ an annotated atom.
Then, an \emph{\mbox{off-line} justification} of $a^{\pm}$ w.r.t.~$\P$, $M$ and~$U$ is an off-line explanation graph  w.r.t.~$\P$, $M$ and~$U$ (Definition~\ref{def:off-line.explanation}),
which satisfies that for all $b \in \at$,
$(b^{+},b^{+})$ does not belong to the transitive closure of $\setm{(c,e)}{ (c,e,+) \in E}$.
\qed
\end{definition}

\CHANGED
The last condition of Definition~\ref{def:off-line.justification} ensures that true atoms are not justified through positive cycles, thus ensuring that justifications of true atoms are rooted in some
rule without positive body, that is, either facts or rules whose body is a conjunction of negative literals.
We may also interpret the latter type of rules as a kind of `facts by default'.
\END
\CHANGED

\begin{example}\label{ex:off-line.positive.cycle}
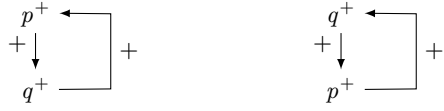
\begin{figure}[t]\centering
\begin{tikzpicture}[tikzpict]
    \matrix[row sep=0.5cm,column sep=0.8cm,ampersand replacement=\&] {
      \node (q) {$p^{+}$};
      \\
      \node (p) {$q^{+}$};
      \\
     };
    \draw [->] (q) to node[pos=0.3,left]{$+$} (p);
    \draw [->] (p) to (1,-0.5) to node[pos=0.5,right]{$+$} (1,0.5) to (q);
\end{tikzpicture}%
\hspace{2cm}
\begin{tikzpicture}[tikzpict]
    \matrix[row sep=0.5cm,column sep=0.8cm,ampersand replacement=\&] {
      \node (q) {$q^{+}$};
      \\
      \node (p) {$p^{+}$};
      \\
     };
    \draw [->] (q) to node[pos=0.3,left]{$+$} (p);
    \draw [->] (p) to (1,-0.5) to node[pos=0.5,right]{$+$} (1,0.5) to (q);
\end{tikzpicture}%
\caption{Off-line explanation graphs of $p^{+}$ and $q^{+}$ w.r.t. $\emptyset$, which are not off-line justifications.}\label{off-line.positive.cycle}
\end{figure}
Let $\newprogram\label{prg:positive.cycle}$ be the program containing the following two rules:
\begin{gather*}
p \lparrow q
\hspace{2cm}
q \lparrow p
\end{gather*}
It has a unique answer set that coincides with its complete well-founded model:
$\newanswerset\label{as:prg:positive.cycle} = \WF{\programr\ref{prg:positive.cycle}}^{+} = \emptyset$.
Figure~\ref{off-line.positive.cycle} depicts two cyclic off-line explanation graphs of~$p^{+}$ and~$q^{+}$,
which, as can be expected, are not off-line justifications since $p$ and $q$ are false \wrt\ $\answersetr\ref{as:prg:positive.cycle}$ and since positive cycles are allowed in explanation graphs, but not in off-line justifications.
\begin{figure}[ht]\centering
\begin{tikzpicture}[tikzpict]
    \matrix[row sep=0.5cm,column sep=0.8cm,ampersand replacement=\&] {
      \node (q) {$p^{-}$};
      \\
      \node (p) {$q^{-}$};
      \\
     };
    \draw [->] (q) to node[pos=0.3,left]{$+$} (p);
    \draw [->] (p) to (1,-0.5) to node[pos=0.5,right]{$+$} (1,0.5) to (q);
\end{tikzpicture}%
\hspace{2cm}
\begin{tikzpicture}[tikzpict]
    \matrix[row sep=0.5cm,column sep=0.8cm,ampersand replacement=\&] {
      \node (q) {$q^{-}$};
      \\
      \node (p) {$p^{-}$};
      \\
     };
    \draw [->] (q) to node[pos=0.3,left]{$+$} (p);
    \draw [->] (p) to (1,-0.5) to node[pos=0.5,right]{$+$} (1,0.5) to (q);
\end{tikzpicture}%
\caption{Off-line justifications of $p^{-}$ and $q^{-}$ w.r.t. $\answersetr\ref{as:prg:positive.cycle} = \emptyset$ and assumption $\emptyset$.}\label{off-line.positive.cycle.2}
\end{figure}
Figure~\ref{off-line.positive.cycle.2} depicts two cyclic off-line justifications explaining that $p$ and $q$ are false \wrt\ $\answersetr\ref{as:prg:positive.cycle}$ because they positively depend on each other.
Note that cycles between negatively annotated atoms are allowed in off-line justifications.\qed
\end{example}
\END

The following example illustrates how off-line justifications are built for a more complex program that has a complete well-founded model, in which case the unique assumption is the empty set.
Example~\ref{ex:off-line.cycle} is continued later, in Example~\ref{ex:off-line.cycle2}, where it is shown that the off-line explanation graphs in Figure~\ref{fig:off-line.cycle} are in fact off-line justifications.
Note that the program discussed in Example~\ref{ex:off-line.cycle} has a non-complete well-founded model and, thus, some atoms will need to be assumed to build the off-line justifications.

\begin{example}\label{ex:positive}
Let $\newprogram\label{prg:positive}$ be the program consisting of the following rules:
\begin{gather*}
p \lparrow q
\hspace{2cm}
q \lparrow r \wedge s
\hspace{2cm}
r \lparrow \Not t
\hspace{2cm}
s
\end{gather*}
This program has a unique answer set $\newanswerset\label{as:positive} = \set{p,q,r,s}$,
which coincides with its complete well-founded model.
As a result, we have an empty set of tentative assumptions
$\TA{\programr\ref{prg:positive}}{\answersetr\ref{as:positive}} = \emptyset$
and the empty set as the only valid assumption,
that is,
\mbox{$\Assumptions{\programr\ref{prg:positive}}{\answersetr\ref{as:positive}} = \set{ \emptyset }$}.
\begin{figure}\centering
\subfloat[]{%
\begin{tikzpicture}[tikzpict]
    \matrix[row sep=0.5cm,column sep=0.3cm,ampersand replacement=\&] {
      \&
      \node (p) {$p^{+}$};
      \\
      \&
      \node (q) {$q^{+}$};
      \\
      \node (r) {$r^{+}$};
      \&\&
      \node (s) {$s^{+}$};
      \\
      \node (t) {$t^{-}$};
      \\
      \node (bot) {$\bot$};
      \&\&
      \node (top) {$\top$};
      \\
     };
    \draw [->] (p) to node[pos=0.3,left]{$+$}  (q);
    \draw [->] (q) to node[pos=0.3,left]{$+$}  (r);
    \draw [->] (q) to node[pos=0.3,right]{$+$}  (s);
    \draw [->] (r) to node[pos=0.4,left]{$-$} (t);
    \draw [->] (t) to node[pos=0.4,right]{$+$} (bot);
    \draw [->] (s) to node[pos=0.4,right]{$+$} (top);
\end{tikzpicture}%
\label{fig:positive.off-line}
}
\hspace{2cm}
\subfloat[]{%
\begin{tikzpicture}[tikzpict]
    \matrix[row sep=0.5cm,column sep=0.3cm,ampersand replacement=\&] {
      \&
      \node (p) {$p^{+}$};
      \\
      \&
      \node (q) {$q^{+}$};
      \\
      \node (r) {$r^{+}$};
      \&\&
      \node (s) {$s^{+}$};
      \\
      \&
      \node (t) {$t^{-}$};
      \\
      \&
      \node (bot) {$\bot$};
      \\
     };
    \draw [->] (p) to node[pos=0.3,left]{$+$}  (q);
    \draw [->] (q) to node[pos=0.3,left]{$+$}  (r);
    \draw [->] (q) to node[pos=0.3,right]{$+$}  (s);
    \draw [->] (r) to node[pos=0.4,left]{$-$} (t);
    \draw [->] (t) to node[pos=0.4,right]{$+$} (bot);
    \draw [->] (s) to node[pos=0.4,right]{$-$} (t);
\end{tikzpicture}%
\label{fig:positive2.off-line}
}
\caption{Off-line justifications of $p^{+}$ w.r.t. $\programr\ref{prg:positive2}$, $\answersetr\ref{as:positive}$, and assumption $\emptyset$.
Figure~\ref{fig:positive.off-line} is also an off-line justification w.r.t. $\programr\ref{prg:positive}$, $\answersetr\ref{as:positive}$, and $\emptyset$ (see Examples~\ref{ex:positive} and~\ref{ex:positive2}).}
\end{figure}
Figure~\ref{fig:positive.off-line} depicts the unique off-line justification of $p^{+}$ w.r.t. program~$\programr\ref{prg:positive}$ and answer set~$\answersetr\ref{as:positive}$.
\color{black}
Intuitively, the edge $(t^{-},\bot,+)$ points out that $t$ is false because there is no rule in~$\programr\ref{prg:positive}$ with $t$ in the head.
Then, as a consequence of the closed world assumption, $t$ is considered to be false.
Similarly, edge $(s^{+},\top,+)$ indicates that $s$ is true because it is a fact.
Edge $(p^{+},q^{+},+)$ (resp. $(r^{+},t^{-},-)$) indicates that $p$ (resp. $r$) is true because it positively (resp. negatively) depends on $q$ (resp. $t$) which is true (resp. false).
Finally, edges $(q^{+},r^{+},+)$ and $(q^{+},s^{+},+)$ together point out that $q$ is true because it positively depends on both $r$ and $s$, which are true.
It is also worth noting that the subgraphs of this off-line justification rooted in $q^{+}$, $r^{+}$ and $s^{+}$ constitute the off-line justifications of $q$, $r$ and $s$ being true w.r.t.~$\programr\ref{prg:positive}$ and~$\answersetr\ref{as:positive}$.
Similarly, the subgraph rooted in $t^{-}$ represents the off-line justification for the atom~$t$ being false.
\qed
\end{example}

In the above example, there is a unique off-line justification for each true or false atom.
The following examples show that several justifications may exist for a given atom w.r.t. a given answer set.

\begin{examplecont}{ex:positive}\label{ex:positive2}
Let $\newprogram\label{prg:positive2}$ be the result of adding rule $s \lparrow \Not t$ to program~$\programr\ref{prg:positive}$.
It is easy to check that
$\answersetr\ref{as:positive}$ is also the unique answer set of~$\programr\ref{prg:positive2}$ (and $\emptyset$ the unique assumption), but now there is a second way to justify the truth of $s$, namely in terms of the falsity of $t$.
As a result, there are two off-line justification of $p^{+}$, respectively depicted in Figures~\ref{fig:positive.off-line} and~\ref{fig:positive2.off-line}.\qed
\end{examplecont}

\begin{examplecont}{ex:off-line.cycle}\label{ex:off-line.cycle2}
In contrast to~$\programr\ref{prg:positive}$ and~$\programr\ref{prg:positive2}$, program~$\programr\ref{prg:cycle}$ does not have a complete \mbox{well-founded} model.
In fact, its well-founded model leaves all atoms undefined.
Thus,
\mbox{$q \in \NANT{\programr\ref{prg:cycle}}$} implies that
$\TA{\programr\ref{prg:cycle}}{\answersetr\ref{as:1:prg:cycle}} = \set{q}$ which, in turn, implies
$\Assumptions{\programr\ref{prg:cycle}}{\answersetr\ref{as:1:prg:cycle}} = \setb{ \set{q} }$.
Note that $\emptyset$ is not a valid assumption because the \mbox{well-founded} model of $\NR{\P}{\emptyset}$ is not complete.
Then, since there is no cycle in Figure~\ref{fig:off-line.cycle},
it follows that these two off-line explanation graphs are also off-line justifications.
Note that edge $(q^{-},\assume,-)$ captures that atom~$q$ is false because of the inherent guessing involved in the definition of answer sets.
\qed
\end{examplecont}

\CHANGED
In Example~\ref{ex:off-line.positive.cycle}, we already illustrated the difference between off-line explanation graphs and off-line justifications. The following example shows this difference in a program without cycles.
\END

\begin{example}\label{ex:off-line.explantions.no-justifications}
\begin{figure}\centering
\subfloat[]{%
\begin{tikzpicture}[tikzpict]
    \matrix[row sep=0.5cm,column sep=0.8cm,ampersand replacement=\&] {
      \node (p) {$p^{+}$};
      \&
      \node (q2) {$q^{-}$};
      \\
      \node (q) {$q^{-}$};
      \\
      \node (assume) {$\assume$};
      \&
      \node (assume2) {$\assume$};
      \\
     };
    \draw [->] (p) to node[pos=0.3,left]{$-$} (q);
    \draw [->] (q) to  node[pos=0.3,left]{$-$} (assume);
    \draw [->] (q2) to  node[pos=0.3,left]{$-$} (assume2);
\end{tikzpicture}%
\label{fig:off-line.cycle5}
}
\hspace{2cm}
\subfloat[]{%
\begin{tikzpicture}[tikzpict]
    \matrix[row sep=0.5cm,column sep=0.8cm,ampersand replacement=\&] {
      \node (p) {$p^{+}$};
      \&
      \node (q2) {$q^{-}$};
      \\
      \node (q) {$q^{-}$};
      \\
      \node (assume) {$\bot$};
      \&
      \node (assume2) {$\bot$};
      \\
     };
    \draw [->] (p) to  node[pos=0.3,left]{$-$} (q);
    \draw [->] (q) to  node[pos=0.3,left]{$-$} (assume);
    \draw [->] (q2) to  node[pos=0.3,left]{$-$} (assume2);
\end{tikzpicture}%
\label{fig:off-line.negation5}
}
\caption{Off-line justifications of $p^{+}$ and $q^{-}$ w.r.t. $\answersetr\ref{as:1:prg:cycle} = \answersetr\ref{as:prg:negation} = \set{p}$ 
in Examples~\ref{ex:off-line.cycle} and~\ref{ex:off-line.explantions.no-justifications}, respectively.
Note that the assumption is respectively $\set{q}$ and $\emptyset$ in subfigures~\ref{fig:off-line.cycle5} and in \ref{fig:off-line.negation5}.
}
\label{fig:off-line.cycle+negation5}
\end{figure}
Let 
$\newprogram\label{prg:negation}$ be the program containing the single rule $p \lparrow \Not q$.
Program $\programr\ref{prg:negation}$ has a complete well-founded model, which consequently coincides with the unique answer set:
$\newanswerset\label{as:prg:negation} = \WF{\programr\ref{prg:negation}}^{+} = \set{p}$.
As in Example~\ref{ex:off-line.cycle}, it easy to see that graphs depicted in Figure~\ref{fig:off-line.cycle} (also depicted in Figure~\ref{fig:off-line.cycle5} to ease the comparison) are off-line explanation graphs of~$p^{+}$ and~$q^{-}$ with respect to the program $\programr\ref{prg:negation}$, the answer set $\answersetr\ref{as:prg:negation}$ and the assumption $\set{q}$.
Moreover,
since the well-founded model is complete, there are no tentative assumptions, that is,
$\TA{\programr\ref{prg:negation}}{\answersetr\ref{as:prg:negation}} = \emptyset$
and
\mbox{$\Assumptions{\programr\ref{prg:negation}}{\answersetr\ref{as:prg:negation}} = \setb{ \emptyset }$}.
Therefore, the \mbox{off-line} explanation graphs in Figure~\ref{fig:off-line.cycle5} are not valid off-line justifications.
Figure~\ref{fig:off-line.negation5} depicts the off-line justifications of $p^{+}$ and $q^{-}$ with respect to program~$\programr\ref{prg:negation}$, the answer set~$\answersetr\ref{as:prg:negation}$ and the assumption $\emptyset$.
Note that, since there is no rule with $q$ in the head, the falsity of $q$ can be justified without assumptions.
\qed
\end{example}

By adding the rule $q \lparrow \Not p$ to program~$\programr\ref{prg:negation}$ (Example~\ref{ex:off-line.explantions.no-justifications})
we create an even-length negative dependency cycle, that is, not only $p$ is dependent on $q$ being false, but also $q$ is dependent on $p$ being false 
(note that this is exactly program~$\programr\ref{prg:cycle}$ from Example~\ref{ex:off-line.cycle}).
This has the effect of replacing the edge $(q^{-},\bot,-)$ by $(q^{-},\assume,-)$ in the off-line justifications of $p^{+}$ and $q^{-}$ (see Figure~\ref{fig:off-line.cycle+negation5}). 
In other words, rather than $q$ being false by default, it is now \emph{assumed} to be false \wrt\ the answer set~$\set{p}$.
As shown by the following example this change from default to assuming is not always the case when creating an even-length negative dependency cycle:
for some programs, this may have the effect of introducing additional justifications.
\begin{example}\label{ex:off-line.empty-well-founded-model.2justifications}
Let $\newprogram\label{prg:wfm2j}$ be the program
\begin{gather*}
p \lparrow \Not q
\hspace{2cm}
r \lparrow \Not p
\hspace{2cm}
s \lparrow \Not r
\end{gather*}
As in Example~\ref{ex:off-line.explantions.no-justifications},
this program has a complete well-founded model and, thus, a unique answer set that coincides with the well-founded model:
$\newanswerset\label{as:prg:wfm2j} = \WF{\programr\ref{prg:wfm2j}}^{+} = \set{p,s}$.
Then, we have that
$\TA{\programr\ref{prg:wfm2j}}{\answersetr\ref{as:prg:wfm2j} } = \emptyset$
and
$\Assumptions{\programr\ref{prg:wfm2j}}{\answersetr\ref{as:prg:wfm2j}} = \setb{ \emptyset }$.
Figure~\ref{fig:off-line.empty-well-founded-model.2justifications.P3} depicts the unique \mbox{off-line} justification of $s^{+}$ with respect to program~$\programr\ref{prg:wfm2j}$, the answer set~$\answersetr\ref{as:prg:wfm2j}$ and assumption~$\emptyset$.
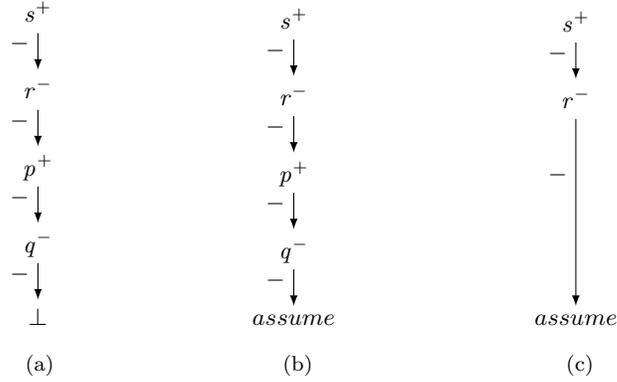
\begin{figure}\centering
\subfloat[]{%
\begin{tikzpicture}[tikzpict]
    \matrix[row sep=0.5cm,column sep=0.8cm,ampersand replacement=\&] {
      \node (s) {$s^{+}$};
      \\
      \node (r) {$r^{-}$};
      \\
      \node (p) {$p^{+}$};
      \\
      \node (q) {$q^{-}$};
      \\
      \node (assume) {$\bot$};
      \\
     };
    \draw [->] (s) to node[pos=0.3,left]{$-$} (r);
    \draw [->] (r) to node[pos=0.3,left]{$-$} (p);
    \draw [->] (p) to node[pos=0.3,left]{$-$} (q);
    \draw [->] (q) to node[pos=0.3,left]{$-$} (assume);
\end{tikzpicture}%
\label{fig:off-line.empty-well-founded-model.2justifications.P3}
}
\hspace{2cm}
\subfloat[]{%
\begin{tikzpicture}[tikzpict]
    \matrix[row sep=0.5cm,column sep=0.8cm,ampersand replacement=\&] {
      \node (s) {$s^{+}$};
      \\
      \node (r) {$r^{-}$};
      \\
      \node (p) {$p^{+}$};
      \\
      \node (q) {$q^{-}$};
      \\
      \node (assume) {$\assume$};
      \\
     };
    \draw [->] (s) to node[pos=0.3,left]{$-$} (r);
    \draw [->] (r) to node[pos=0.3,left]{$-$} (p);
    \draw [->] (p) to node[pos=0.3,left]{$-$} (q);
    \draw [->] (q) to node[pos=0.3,left]{$-$} (assume);
\end{tikzpicture}%
\label{fig:off-line.empty-well-founded-model.2justifications.P4.a}
}
\hspace{2cm}
\subfloat[]{%
\begin{tikzpicture}[tikzpict]
    \matrix[row sep=0.5cm,column sep=0.8cm,ampersand replacement=\&] {
      \node (s) {$s^{+}$};
      \\
      \node (r) {$r^{-}$};
      \\
      \\
      \\
      \\
      \\
      \node (assume) {$\assume$};
      \\
     };
    \draw [->] (s) to node[pos=0.3,left]{$-$} (r);
    \draw [->] (r) to node[pos=0.3,left]{$-$} (assume);
\end{tikzpicture}%
\label{fig:off-line.empty-well-founded-model.2justifications.P4.b}
}
\caption{Off-line justifications of $s^{+}$ w.r.t.
$\answersetr\ref{as:prg:wfm2j} = \answersetr\ref{as:1:prg:wfm2j.b} = \set{p,s}$ and 
the assumption $\emptyset$ (Subfigure~\ref{fig:off-line.empty-well-founded-model.2justifications.P3}), $\set{q}$ (Subfigure~\ref{fig:off-line.empty-well-founded-model.2justifications.P4.a}), and 
$\set{q,r}$ (Subfigure~\ref{fig:off-line.empty-well-founded-model.2justifications.P4.b})
in Example~\ref{ex:off-line.empty-well-founded-model.2justifications}.}
\label{fig:off-line.empty-well-founded-model.2justifications}
\end{figure}
Let now
$\newprogram\label{prg:wfm2j.b} = \programr\ref{prg:wfm2j} \cup \set{ q \lparrow \Not p }$.
As in Example~\ref{ex:off-line.cycle},
this program also has two answer sets,
namely
$\newanswerset\label{as:1:prg:wfm2j.b} = \set{p,s}$ and
\mbox{$\newanswerset\label{as:2:prg:wfm2j.b} = \set{q,r}$}, and an empty \mbox{well-founded} model $\WF{\programr\ref{prg:wfm2j.b}}^{+} = \WF{\programr\ref{prg:wfm2j.b}}^{-} = \emptyset$.
Then, it follows that
$\TA{\programr\ref{prg:wfm2j.b}}{\answersetr\ref{as:1:prg:wfm2j.b} } = \set{q,r}$
and
$\Assumptions{\programr\ref{prg:wfm2j.b}}{\answersetr\ref{as:1:prg:wfm2j.b}} = \setb{ \set{q},\, \set{q,r} }$.
Figures~\ref{fig:off-line.empty-well-founded-model.2justifications.P4.a} and~\ref{fig:off-line.empty-well-founded-model.2justifications.P4.b} 
depict the two off-line justifications of $s^{+}$
with respect to program~$\programr\ref{prg:wfm2j.b}$, $\answersetr\ref{as:1:prg:wfm2j.b}$
and assumptions $\set{q}$ and $\set{q,r}$, respectively.
As opposed to what happens in Example~\ref{ex:off-line.explantions.no-justifications},
adding the rule $q \lparrow \Not p$,
and thus creating an even-length negative dependency cycle, not only has the effect of replacing the edge $(q^{-},\bot,-)$ by $(q^{-},\assume,-)$, but it also produces a second off-line justification in which $r^{-}$ is assumed (Figure~\ref{fig:off-line.empty-well-founded-model.2justifications.P4.b}).
This difference disappears if we only take into account off-line justifications with respect to \mbox{$\subseteq$-minimal} assumptions,
in which case only Figures~\ref{fig:off-line.empty-well-founded-model.2justifications.P4.a} would be a justification.\qed
\end{example}

\CHANGED
As mentioned above, the last condition of Definition~\ref{def:off-line.justification} ensures that true atoms are not justified through positive cycles  (those in which all edges are labelled ~`$+$').
Still, there exist off-line justifications in which true atoms are justified by (non-positive) cycles, as illustrated by the following example.
\END

\begin{example}\label{ex:cyclic.justifications}
Let $\newprogram\label{prg:cyclic.justifications}$ be the program containing the following two rules:
\begin{gather*}
p \lparrow q \wedge \Not r
\hspace{2cm}
r \lparrow \Not p
\end{gather*}
This program has a complete well-founded model, which coincides with its unique answer set
$\WF{\programr\ref{prg:cyclic.justifications}}^{+} = \newanswerset\label{as:prg:cyclic.justifications}  = \set{r}$. 
Then,
$\Assumptions{\programr\ref{prg:cyclic.justifications}}{\answersetr\ref{as:prg:cyclic.justifications}} = \setb{ \emptyset }$.
\begin{figure}\centering
\subfloat[]{%
\begin{tikzpicture}[tikzpict]
    \matrix[row sep=0.5cm,column sep=0.8cm,ampersand replacement=\&] {
      \node (r) {$r^{+}$};
      \\
      \node (p) {$p^{-}$};
      \\
      \node (q) {$q^{-}$};
      \\
      \node (assume) {$\bot$};
      \\
     };
    \draw [->] (r) to node[pos=0.3,right]{$-$}  (p);
    \draw [->] (p) to node[pos=0.3,right]{$+$} (q);
    \draw [->] (q) to node[pos=0.3,right]{$+$} (assume);
\end{tikzpicture}%
\label{fig:cyclic.justifications.a}
}
\hspace{2cm}
\subfloat[]{%
\begin{tikzpicture}[tikzpict]
    \matrix[row sep=0.5cm,column sep=0.8cm,ampersand replacement=\&] {
      \node (r) {$r^{+}$};
      \\
      \node (p) {$p^{-}$};
      \\
      \\
      \\
      \\
      \\
     };
    \draw [->] (r) to node[pos=0.3,left]{$-$} (p);
    \draw [->] (p) to (1,0.5) to node[pos=0.5,right]{$-$} (1,1.5) to (r);
\end{tikzpicture}%
\label{fig:cyclic.justifications.b}
}
\caption{Off-line justifications of $r^{+}$ w.r.t. $\answersetr\ref{as:prg:cyclic.justifications} = \set{r}$ and assumption $\emptyset$ in Example~\ref{ex:cyclic.justifications}.}\label{fig:cyclic.justifications}
\end{figure}
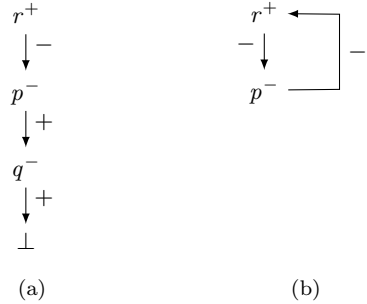
Figure~\ref{fig:cyclic.justifications} 
depicts the two off-line justifications of $r^{+}$
with respect to program~$\programr\ref{prg:cyclic.justifications}$, the answer set~$\answersetr\ref{as:prg:cyclic.justifications}$ and the assumption~$\emptyset$.\qed
\end{example}

\CHANGED
Though at first sight, cyclic justifications (like the one in Figure~\ref{fig:cyclic.justifications}) may seem to contradict the intuition that the justifications of true atoms must be rooted in \newChanged{a rule without positive body (facts or rules whose body is a conjunction of negative literals),}
we note that the existence of an acyclic off-line justification (Figure~\ref{fig:cyclic.justifications.a}) in Example~\ref{ex:cyclic.justifications} is not accidental. In fact, for every true atom, there always exists at least one acyclic justification~\cite[Proposition~2]{PontelliS2006}.
\END

\subsection{LABAS Justifications}\label{sec:labas}

LABAS justifications \cite{SchulzT2016,SchulzT2013} explain the truth value of an extended literal with respect to a given answer set of an extended normal logic program.\footnote{For simplicity, we use the term `literal' instead of `extended literal' throughout this section.}
They have been implemented in an online platform called \texttt{LABAS Justifier}.\footnote{\url{http://labas-justification.herokuapp.com/}}
In contrast to off-line justifications, where every rule application step used to derive a literal is included in a justification, LABAS justifications abstract away from intermediate rule applications in the derivation, only pointing out the literal in question and the facts and negative literals occurring in  rules used in the derivation.
In addition, the truth of negative literals $\Not l$ is not taken for granted or assumed, but is further explained in terms of the truth value of the respective positive literal $l$.

LABAS justifications have an \emph{argumentative} flavour as they are constructed from trees of conflicting \emph{arguments}.\footnote{\citeN{SchulzT2016} define arguments and attack trees with respect to the translation of a logic program into an \emph{\changed{Assumption-Based Argumentation (ABA)} framework} \cite{DungKT2009}. For simplicity, we here reformulate these definitions with respect to a logic program. Due to the semantic correspondence between logic programs and their translation into ABA frameworks \cite{SchulzT2016,SchulzT2015}, these definitions are equivalent to the original ones.}

\begin{definition}[Argument]
\label{def:argument}
Given an extended logic program $\P$, an \emph{argument} for $\rL \in \litExt$ is a finite tree, where every node holds a literal in $\litExt$, such that
 \begin{itemize}
  \item the root node holds $\rL$;
  \item for every node $N$
  \begin{itemize}
   \item if $N$ is a leaf then $N$ holds either a negative literal or a fact;
   \item if $N$ is not a leaf and $N$ holds the positive literal $\rH$, then there is a rule $\rH \lparrow \rB_1 \wedge \dotsc \wedge \rB_\npbody \wedge \Not \rC_1 \wedge \dotsc \Not \rC_\nnbody$ in $\P$ and
   $N$ has $\npbody + \nnbody$ children, holding $\rB_1, \dotsc, \rB_\npbody, \Not \rC_1, \dotsc \Not \rC_\nnbody$ respectively;
  \end{itemize}
  \item $AP$ is the set of all negative literals held by leaves;
  \item $FP$ is the set of all facts held by leaves.
 \end{itemize}
 An argument is denoted $A: \argument{AP}{FP}{\rL}$, where $A$ is a unique name, $AP$ is the set of \emph{assumption premises}, $FP$ the set of \emph{fact premises}, and $\rL$ the \emph{conclusion}.\qed
\end{definition}
Intuitively, an argument is a derivation where each rule is used and where only negative literals and facts are recorded.
Note however, that arguments are not necessarily minimal derivations and that they allow the repeated application of a rule.

\begin{example}
\label{ex:labas:arguments}
Let $\newprogram\label{prg:arguments}$ be the following logic program:
\begin{gather*}
p \lparrow q \wedge \Not r
\hspace{2cm}
q \lparrow q
\hspace{2cm}
q
\end{gather*}
There are \emph{infinitely} many arguments for $p$ (and $q$) since the second rule can be used infinitely many times before using the fact $q$. 
Figure~\ref{fig:labas:argument1} illustrates the argument~$A_1$ where the second rule is not used at all, Figure~\ref{fig:labas:argument2}
illustrates the argument $A_2$ where the second rule is used once, and Figure~\ref{fig:labas:argument3} illustrates arguments where the second rule is applied various times (indicated by the dots).
Note that all arguments with conclusion $p$ differ in their name and their tree representation, but they are all denoted $\argument{\{\Not r\}}{q}{p}$ in the shorthand notation. \qed
\end{example}

\begin{figure}
\centering
\subfloat[Argument $A_1$]{%
\begin{tikzpicture}[tikzpict]
    \matrix[row sep=0.5cm,column sep=0.3cm,ampersand replacement=\&] {
      \&
      \node (p) {$p$};
      \\
      \node (q) {$q$};
      \&\&
      \node (notr) {$\Not r$};
      \\
     };
    \draw (p) -- (q);
    \draw (p) -- (notr);
\end{tikzpicture}%
\label{fig:labas:argument1}
}
\hspace{0.5cm}
\subfloat[Argument $A_2$]{%
\begin{tikzpicture}[tikzpict]
    \matrix[row sep=0.5cm,column sep=0.3cm,ampersand replacement=\&] {
      \&
      \node (p) {$p$};
      \\
      \node (q) {$q$};
      \&\&
      \node (notr) {$\Not r$};
      \\
      \node (q2) {$q$};
      \\
     };
    \draw (p) -- (q);
    \draw (p) -- (notr);
    \draw (q) -- (q2);
\end{tikzpicture}%
\label{fig:labas:argument2}
}
\hspace{0.5cm}
\subfloat[Argument $A_n$]{%
\begin{tikzpicture}[tikzpict]
    \matrix[row sep=0.5cm,column sep=0.3cm,ampersand replacement=\&] {
      \&
      \node (p) {$p$};
      \\
      \node (q) {$q$};
      \&\&
      \node (notr) {$\Not r$};
      \\
      \node (q2) {$q$};
      \\
      \node (dots) {$\vdots$};
      \\
      \node (qn) {$q$};
      \\
     };
    \draw (p) -- (q);
    \draw (p) -- (notr);
    \draw (q) -- (q2);
    \draw (q2) -- (dots);
    \draw (dots) -- (qn);
\end{tikzpicture}%
\label{fig:labas:argument3}
}
\caption{Different arguments with conclusion $p$.}
\end{figure}
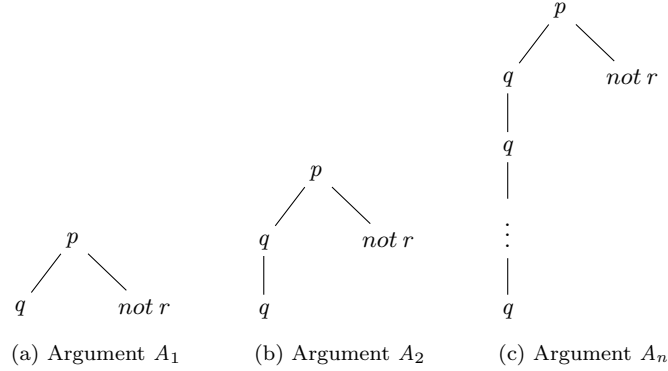

An argument for a literal only exists if all literals in the rules used in the derivation have an argument themselves.
That is, for a logic program with only one rule $p \lparrow q$, there is no argument for either $p$ or $q$ ($q$ is neither a negative literal nor a fact, so it cannot be the leaf of an argument tree).

If the conclusion of an argument is a positive literal $\rL$ then it \emph{attacks} every argument
that has $\Not \rL$ in its assumption premises. In other words, a derivation for $\rL$ provides a reason against any derivation using $\Not \rL$.
\begin{definition}[Attack]
\label{def:attack}
An argument $\argument{AP_{1}}{FP_{1}}{\rL_{1}}$
 \emph{attacks} an argument $\argument{AP_2}{FP_2}{l_2}$ \ifonlyif\
$\rL_1$ is a positive literal and $\Not \rL_1 \in AP_2$.\qed
\end{definition}
Note that attacks do not arise due to the existence of an atom $a$ and its contrary $\negClas a$ in two arguments.

\CHANGED
\begin{examplecontpage}{ex:off-line.cycle}\label{ex:labas_attack}
Four arguments can be constructed from $\programr\ref{prg:cycle}$:
\begin{gather*}
A_1: \argument{\{\Not p\}}{\{\}}{\Not p} \hspace{2cm}
A_3: \argument{\{\Not p\}}{\{\}}{q} \\
A_2: \argument{\{\Not q\}}{\{\}}{\Not q} \hspace{2cm}
A_4: \argument{\{\Not q\}}{\{\}}{p}
\end{gather*}
$A_3$ attacks $A_2$ and $A_4$ since its conclusion $q$ is the complement of the assumption premise $\Not q$ in the two attacked arguments.
Similarly, $A_4$ attacks $A_1$ and $A_3$.\qed
\end{examplecontpage}
\END

\subsubsection{Attack Trees}
LABAS justifications are constructed from trees of attacking arguments.
\begin{definition}[Attack Tree]
\label{def:attackTree}
 Given an extended program $\P$, an \emph{attack tree} of an argument $A: \argument{AP}{FP}{\rL}$ \wrt\ an answer set $M$ of $\P$,
 denoted $attTree_{M}(A)$, is a (possibly infinite) tree such that:
  \begin{enumerate}
  \item Every node in $attTree_{M}(A)$ holds an argument,
  labelled \posArg\ or \negArg.
  \item The root node is $A^{+}$ if $\forall \Not \rL' \in AP: \rL' \notin M$, or $A^{-}$ otherwise.
  \item\label{cond:attTree:posArg} For every node $B^{+}$ and for every argument $C$ attacking argument $B$,
  there exists a child node $C^{-}$ of $B^{+}$.
  \item\label{cond:attTree:negArg} Every node $B^{-}$ has exactly one child node $C^{+}$ for some argument\\
  \mbox{$C: \argument{AP_C}{FP_C}{l_C}$} attacking argument $B$ and satisfying that
  $\forall \Not \rL' \in AP_C$, $\rL' \notin M$.
  \item There are no other nodes in $attTree_{M}(A)$ except those given in 1-4.\qed
 \end{enumerate}
 \end{definition}
 
The intuition for labelling arguments in an attack tree is as follows: 
If an argument $A$ is based on  some negative literal $\Not \rL$ (i.e. it has $\Not \rL$ as an assumption premise) such that $\rL \in M$, then some rule used to construct $A$ is not applicable \wrt\ $M$ (namely the rule in which $\Not \rL$ occurs), so argument $A$ does not warrant that its conclusion is in $M$. Therefore, argument $A$ is labelled \negArg. Otherwise, all rules used to construct $A$ are applicable, so the conclusion of argument $A$ is in~$M$. Thus, argument $A$ is labelled \posArg.
 
\begin{examplecont}{ex:labas_attack}\label{ex:labas_tree:cycle}

The unique attack trees of $A_3$ and $A_4$ \wrt\ $\answersetr\ref{as:1:prg:cycle} = \{p\}$
are displayed in Figure~\ref{fig:labas_tree:cycle1} and~\ref{fig:labas_tree:cycle2}, respectively.
When inverting all \posArg\ and \negArg\ labels in the trees, the attack trees \wrt\ $\answersetr\ref{as:2:prg:cycle} = \{q\}$ are obtained.\qed
\end{examplecont} 

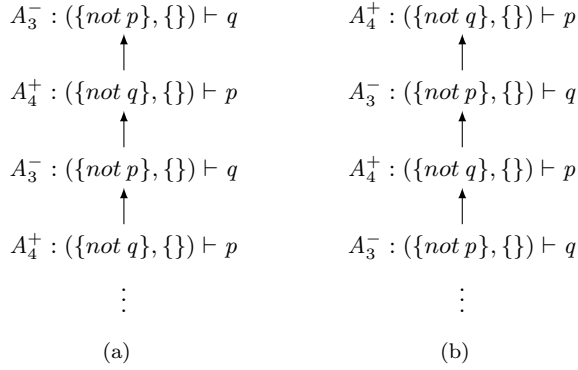
\begin{figure}
\centering
\subfloat[]{%
\begin{tikzpicture}[tikzpict]
    \matrix[row sep=0.5cm,column sep=0.3cm,ampersand replacement=\&] {
      \&
      \node (q) {$A_3^{-}: \argument{\{\Not p\}}{\{\}}{q}$};
      \\
      \&
      \node (p) {$A_4^{+}: \argument{\{\Not q\}}{\{\}}{p}$};
      \\
      \&
      \node (q2) {$A_3^{-}: \argument{\{\Not p\}}{\{\}}{q}$};
      \\
      \&
      \node (p2) {$A_4^{+}: \argument{\{\Not q\}}{\{\}}{p}$};
      \\[-0.5cm]
      \&
      \node (dots) {$\vdots$};
      \\
     };
    \draw[->] (p) -- (q);
    \draw[->] (q2) -- (p);
    \draw[->] (p2) -- (q2);
\end{tikzpicture}%
\label{fig:labas_tree:cycle1}
}
\hspace{0.5cm}
\centering
\subfloat[]{%
\begin{tikzpicture}[tikzpict]
    \matrix[row sep=0.5cm,column sep=0.3cm,ampersand replacement=\&] {
      \&
      \node (p) {$A_4^{+}: \argument{\{\Not q\}}{\{\}}{p}$};
      \\
      \&
      \node (q) {$A_3^{-}: \argument{\{\Not p\}}{\{\}}{q}$};
      \\
      \&
      \node (p2) {$A_4^{+}: \argument{\{\Not q\}}{\{\}}{p}$};
      \\
      \&
      \node (q2) {$A_3^{-}: \argument{\{\Not p\}}{\{\}}{q}$};
      \\[-0.5cm]
      \&
      \node (dots) {$\vdots$};
      \\
     };
    \draw[->] (q) -- (p);
    \draw[->] (p2) -- (q);
    \draw[->] (q2) -- (p2);
\end{tikzpicture}%
\label{fig:labas_tree:cycle2}
}
\caption{Attack trees of arguments $A_3$ and $A_4$ \wrt\ $\answersetr\ref{as:1:prg:cycle}$ of $\programr\ref{prg:cycle}$.}
\end{figure}
 
 An attack tree is thus made of layers of arguments for literals that are alternately true and false \wrt\ the answer set $M$.
Note the difference in Definition~\ref{def:attackTree} between arguments labelled \posArg, which have all attackers as child nodes, and arguments labelled \negArg, which have only one attacker as a child node.
This is in line with the definition of answer sets. 
To prove that a literal $\rL$ is in $M$, \emph{all} negative literals $\Not \rL'$ used in its derivation (i.e. in the argument $B$ in condition~\ref{cond:attTree:posArg}) need to be true, so for all $\rL'$ there must not be a derivation that concludes that $\rL'$ is true. Thus, all such derivations for $\rL'$ (i.e. all arguments $C$ attacking $B$ in condition~\ref{cond:attTree:posArg}) are explained in an attack tree.
In contrast, to prove that a derivation of a literal $\rL$ (argument $B$ in condition~\ref{cond:attTree:negArg}) does not lead to~$\rL$ being true \wrt\ $M$, it is sufficient that one negative literal $\Not \rL'$ used in this derivation is false, i.e. $\rL'$ is in $M$, so there exists some derivation for $\rL'$ (argument $C$ in condition~\ref{cond:attTree:negArg}) that warrants that $\rL'$ is true \wrt\ $M$.

\begin{example}\label{ex:labas_trees:multipleAttack}
Let $\newprogram\label{prg:multipleAttack}$ be the following logic program:
\begin{align*}
p &\lparrow \Not q \wedge \Not r
\hspace{2cm}
&&q \lparrow \Not s
\hspace{2cm}
s \\
r &\lparrow s \wedge \Not p
\hspace{2cm}
&&r \lparrow \Not s
\end{align*}
Program~$\programr\ref{prg:multipleAttack}$ has two answer sets,
namely
\mbox{$\newanswerset\label{as:1:prg:multipleAttack} = \set{s,p}$}
and
\mbox{$\newanswerset\label{as:2:prg:multipleAttack} = \set{s,r}$}.
The argument~$A_1: \argument{\{\Not q, \Not r\}}{\{\}}{p}$ has one attack tree \wrt\ $\answersetr\ref{as:1:prg:multipleAttack}$ and one \wrt~$\answersetr\ref{as:2:prg:multipleAttack}$, depicted in Figures~\ref{fig:labas_tree:multipleAttack1} and~\ref{fig:labas_tree:multipleAttack2}, respectively.
\changed{Note that in the attack tree of $A_1$ \wrt\ $\answersetr\ref{as:2:prg:multipleAttack}$, $A_2$ and $A_4$ cannot be chosen as the child nodes of $A_1$, even though they attack $A_1$, since 
they both have $\Not s$ as an assumption premise, where $s$ is contained in the answer set $\answersetr\ref{as:2:prg:multipleAttack}$
(they thus violate condition~4 in Definition~\ref{def:attackTree}). These arguments thus do not provide explanations as to why $r$ is true \wrt\ $\answersetr\ref{as:2:prg:multipleAttack}$ and consequently cannot be used to explain why $p$ is false.}\qed
\end{example}

\begin{figure}
\centering
\subfloat[]{%
\begin{tikzpicture}[tikzpict]
    \matrix[row sep=0.5cm,column sep=0.3cm,ampersand replacement=\&] {
      \&
      \node (p) {$A_1^{+}: \argument{\{\Not q, \Not r\}}{\{\}}{p}$};
      \\
      \node (q) {$A_2^{-}: \argument{\{\Not s\}}{\{\}}{q}$};
      \&
      \node (r1) {$A_3^{-}: \argument{\{\Not p\}}{\{s\}}{r}$};
      \&\&
      \node (r2) {$A_4^{-}: \argument{\{\Not s\}}{\{\}}{r}$};
      \\
      \node (s1) {$A_5^{+}: \argument{\{\}}{\{s\}}{s}$};
      \& 
      \node (p2) {$A_1^{+}: \argument{\{\Not q, \Not r\}}{\{\}}{p}$};
      \&\& 
      \node (s2) {$A_5^{+}: \argument{\{\}}{\{s\}}{s}$};
      \\[-0.5cm]
      \&
      \node (dots) {$\vdots$};
      \\
     };
    \draw[->] (q) -- (p);
    \draw[->] (r1) -- (p);
    \draw[->] (r2) -- (p);
    \draw[->] (s1) -- (q);
    \draw[->] (p2) -- (r1);
    \draw[->] (s2) -- (r2);
\end{tikzpicture}%
\label{fig:labas_tree:multipleAttack1}
}
\\
\vspace{0.5cm}
\centering
\subfloat[]{%
\begin{tikzpicture}[tikzpict]
    \matrix[row sep=0.5cm,column sep=0.3cm,ampersand replacement=\&] {
      \&
      \node (p) {$A_1^{-}: \argument{\{\Not q, \Not r\}}{\{\}}{p}$};
      \\
      \&
      \node (r) {$A_3^{+}: \argument{\{\Not p\}}{\{s\}}{r}$};
      \\
      \&
      \node (p2) {$A_1^{-}: \argument{\{\Not q, \Not r\}}{\{\}}{p}$};
      \\[-0.5cm]
      \&
      \node (dots) {$\vdots$};
      \\
     };
    \draw[->] (r) -- (p);
    \draw[->] (p2) -- (r);
\end{tikzpicture}%
\label{fig:labas_tree:multipleAttack2}
}
\caption{Attack trees of argument $A_1$ \wrt\ $\answersetr\ref{as:1:prg:multipleAttack}$ and $\answersetr\ref{as:2:prg:multipleAttack}$.}
\end{figure}
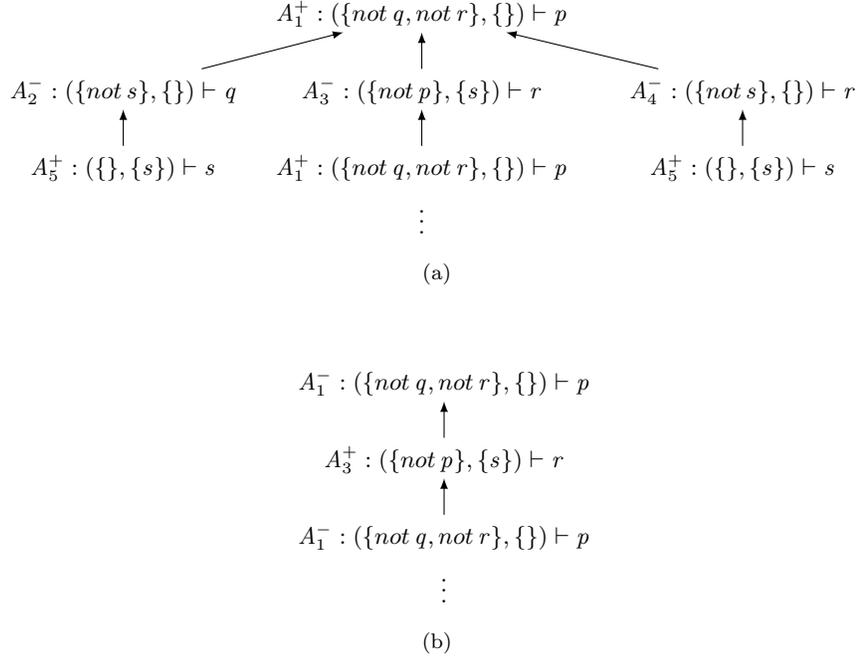

Attack trees are not only used to construct LABAS justifications, as explained in the following, but in fact constitute justifications of literals in their own right.

\begin{definition}[Attack Tree Justification]
Let $M$ be an answer set of an extended program $\P$, $\rL \in \litExt$, and $A$ an argument with conclusion $\rL$.
\begin{itemize}
 \item If $\rL$ is true \wrt\ $M$, then an $attTree_{M}(A)$ is a justification of $\rL$ if the root node is $A^{+}$.
 \item If $\rL$ is false \wrt\ $M$, then an $attTree_{M}(A)$ is a justification of $\rL$ if the root node is $A^{-}$.\qed
\end{itemize}
\end{definition}
In fact, in the second case \emph{any} attack tree for an argument with conclusion $\rL$ will have its root node labelled \negArg\
\cite[from Theorem~3 and Lemma~5]{SchulzT2016}.

\changed{Attack trees justify literals in terms of dependencies between arguments.
Next, we explain how dependencies between literals are extracted from attack trees to construct a justification in terms of literals.}

\subsubsection{Constructing LABAS Justifications}
Labelled ABA-Based Answer Set Justifications (``ABA'' stands for ``Assumption-Based Argumentation''), short LABAS justifications, are constructed from attack trees by extracting the relations between literals in arguments.
That is, literals occurring as assumption or fact premises in an argument of the attack tree are \emph{supporting} the conclusion literal, whereas the conclusion $\rL$ of an attacking argument \emph{attacks} the negative literal $\Not \rL$ occurring as an assumption premise of the attacked argument.

As a first step of the LABAS justification construction, an attack tree is transformed into a \emph{labelled justification}.
\changed{A labelled justification is a set of labelled relations between literals, which can thus be represented as a graph.
Each literal in a relation is labelled as \textquotesingle$+$\textquotesingle, meaning that it is true \wrt\ the answer set in question, or \textquotesingle$-$\textquotesingle, meaning that it is false \wrt\ the answer set in question. Support and attack relations are labelled the same as the respective source literals of the relation. The label \textquotesingle$+$\textquotesingle\ represents that the source label is able to effectively attack or support the target literal, whereas \textquotesingle$-$\textquotesingle\ represents an ineffective relation.
In addition, a literal is labelled with $fact$ or $asm$ if it is a fact or assumption premise, or else with its argument's name.}

\begin{definition}[Labelled Justification]
 \label{def:labas}
 Let $M$ be an answer set of an extended program~$\P$, $A$ an argument and $\Upsilon = attTree_{M}(A)$ an attack tree of $A$ \wrt\ $M$. For any node $B^{+/-}$ in $\Upsilon$, $children(B^{+/-})$ denotes the set of child nodes of $B^{+/-}$ and $conc(B^{+/-})$ the conclusion of argument~$B$.
 The \emph{labelled justification} of~$\Upsilon$,
 denoted $just(\Upsilon)$, is obtained as follows:\\
$\hspace*{10pt}just(\Upsilon) \eqdef  \\
\hspace*{20pt}\bigcup_{B^{+}: \argument{AP}{FP}{\rL} \textit{ in } \Upsilon}$
\begin{align*}
 &\{ supp\mathunderscore rel^{+}({\Not p}^{+}_{asm}, \rL_{B}^{+}) \; & &| \; \Not p \in AP \backslash \{\rL\}\} \; \cup\\
 &\{ supp\mathunderscore rel^{+}(f^{+}_{fact}, \rL_{B}^{+}) \; & &| \; f \in FP \backslash \{\rL\}\} \; \cup\\
 &\{ att\mathunderscore rel^{-}({k}_{C}^{-}, {\Not k}^{+}_{asm}) \; & &| \; C^{-} \in children(B^{+}),conc(C^{-}) = k \} \; \cup
  \end{align*}
$\hspace*{20pt}\bigcup_{B^{-}: \argument{AP}{FP}{\rL}  \textit{ in } \Upsilon}$
\begin{IEEEeqnarray*}{rl rl +x*}
 &\{ supp\mathunderscore rel^{-}({\Not p}^{-}_{asm}, \rL_{B}^{-}) \; & &| \; \Not p \in AP \backslash \{\rL\},children(B^{-}) = \{C^{+}\},\\
 & & &\;\;  conc(C^{+}) = p \} \; \cup\\
 &\{ att\mathunderscore rel^{+}(f^{+}_{fact}, {\Not f}^{-}_{asm}) \; & &| \; children(B^{-}) = 
 \{C^{+}: \argument{\{\}}{\{f\}}{f}\}\}\; \cup\\
 &\{ att\mathunderscore rel^{+}(k^{+}_{B}, {\Not k}^{-}_{asm}) \; & &| \; children(B^{-}) = \{C^{+}:\argument{AP_C}{FP_C}{k}\},
 \\ & & &\;\; AP_C \neq \{\} \textit{ or } FP_C \neq \{k\}\}
 & \qed
 \end{IEEEeqnarray*}
 \end{definition}
 
Note that a labelled justification does not extract \emph{all} relations from an attack tree but only those deemed relevant for justifying the conclusion of argument~$A$. For example, for an argument $B^{-}$ in the attack tree, only one negative literal is extracted as supporting the conclusion, namely the one that is attacked by the child node $C^{+}$ of $B^{-}$, since this negative literal provides the reason that the conclusion of $B$ is not in the answer set.
 
 Infinite attack trees, as for example shown in Figures~\ref{fig:labas_tree:cycle1} and~\ref{fig:labas_tree:cycle2}, may be represented by \emph{finite} LABAS justifications as
re-occurring arguments in an attack tree are only processed once (note that justifications are sets).
 
\begin{examplecont}{ex:labas_tree:cycle}\label{ex:labas_labelled:cycle}
Since the two attack trees $attTree_{\answersetr\ref{as:1:prg:cycle}}(A_3)$ and $attTree_{\answersetr\ref{as:1:prg:cycle}}(A_4)$ (Figures~\ref{fig:labas_tree:cycle1} and~\ref{fig:labas_tree:cycle2}) comprise the same nodes, their labelled justifications are the same, namely:
\begin{IEEEeqnarray*}{c +x*}
\{
supp\mathunderscore rel^{-}({\Not p}^{-}_{asm}, q^{-}_{A_3}), 
att\mathunderscore rel^{+}(p^{+}_{A_4},{\Not p}^{-}_{asm}),\\
\ \ supp\mathunderscore rel^{+}({\Not q}^{+}_{asm}, p^{+}_{A_4}), 
att\mathunderscore rel^{-}(q^{-}_{A_3},{\Not q}^{+}_{asm})
\}
&\qed
\end{IEEEeqnarray*}
\end{examplecont}

As illustrated by Example~\ref{ex:labas_labelled:cycle}, it is not obvious from a labelled justification, which literal is being justified.
A LABAS justification thus adds the literal being justified to labelled justifications.
It furthermore defines a justification in terms of \emph{one} labelled justification if a literal contained in the answer set is justified and in terms of \emph{all} labelled justifications if a literal not contained in the answer set is justified.
This is based on the idea that if a literal can be successfully derived in one way, it is in the answer set, but that it is not in the answer set only if all ways of deriving the literal are unsuccessful.

\begin{definition}[LABAS Justification]
 \label{def:labas_pos_neg}
 Let $M$ be an answer set of an extended program $\P$ and $\rL \in \litExt$.
 \begin{enumerate}
  \item Let $\rL$ be true \wrt\ $M$, let $A: \argument{AP}{FP}{l}$ be an argument,
   and $attTree_{M}(A)$ an attack tree with root node $A^{+}$.
  Let $lab(\rL) \eqdef \rL^{+}_{asm}$ if $\rL$ is a negative literal, $lab(\rL) \eqdef \rL^{+}_{fact}$ if $FP=\{\rL\}$ and $AP = \{\}$, and $lab(\rL) = \rL^{+}_A$ else.
 A \emph{(positive) LABAS justification} of $\rL$ with respect to $M$ is:\\
 $justLABAS_{M}^{+}(\rL) \eqdef \{lab(\rL)\} \; \cup \; just(attTree_{M}(A))$.
 \item Let $\rL$ be false \wrt\ $M$, let $A_1,\ldots, A_n$ be all arguments with conclusion $\rL$,
 and $\Upsilon_{11},\ldots, \Upsilon_{1m_1}, \ldots, \Upsilon_{n1}, \ldots, \Upsilon_{nm_n}$ all attack trees of $A_1,\ldots, A_n$ with root node labelled \negArg.
  \begin{enumerate}
    \item If $n = 0$, then the \emph{(negative) LABAS justification} of $\rL$ with respect to $M$ is:\\
    $justLABAS_{M}^{-}(\rL) \eqdef \emptyset$
    \item If $n > 0$, then let $lab(\rL_1) \eqdef \rL^{-}_{asm}$,  \dots , $lab(\rL_n) \eqdef \rL^{-}_{asm}$ if $\rL$ is a negative literal and
    $lab(\rL_1) \eqdef \rL^{-}_{A_1}$, \dots, $lab(\rL_n) \eqdef \rL^{-}_{A_n}$ else.
Then the \emph{(negative) LABAS justification} of $\rL$ with respect to $M$ is:\\
$justLABAS_{M}^{-}(\rL) \eqdef \{\{lab(\rL_1)\} \; \cup \; just(\Upsilon_{11}),\ldots, \{lab(\rL_n)\} \; \cup \; just(\Upsilon_{nm_n})\}$.\qed
   \end{enumerate}
 \end{enumerate}
 \end{definition}
Note that there may be various LABAS justifications of a literal that is true \wrt\ the answer set $M$, but only one LABAS justification of a literal that is false \wrt~$M$. 

\vspace{2cm}
 
\begin{examplecont}{ex:labas_labelled:cycle}\label{ex:labas_set:cycle}
Since there exists only one argument with conclusion $q \notin \answersetr\ref{as:1:prg:cycle}$, namely $A_3$,
and since this argument has a unique attack tree $attTree_{\answersetr\ref{as:1:prg:cycle}}(A_3)$, only
the labelled justification from Example~\ref{ex:labas_labelled:cycle} has to be taken into account for the LABAS justification
of~$q$ \wrt~$\answersetr\ref{as:1:prg:cycle}$. That is,
\begin{gather*}
justLABAS_{\answersetr\ref{as:1:prg:cycle}}^{-}(q) = 
\{\{q^{-}_{A_3}, 
supp\mathunderscore rel^{-}({\Not p}^{-}_{asm}, q^{-}_{A_3}), 
att\mathunderscore rel^{+}(p^{+}_{A_4},{\Not p}^{-}_{asm}),\\
supp\mathunderscore rel^{+}({\Not q}^{+}_{asm}, p^{+}_{A_4}), 
att\mathunderscore rel^{-}(q^{-}_{A_3},{\Not q}^{+}_{asm})
\}\}
\end{gather*}
Similarly, the only LABAS justification of $p$ \wrt\ $\answersetr\ref{as:1:prg:cycle}$ is
\begin{IEEEeqnarray*}{c +x*}
justLABAS_{\answersetr\ref{as:1:prg:cycle}}^{+}(p) = 
\{p^{-}_{A_4}, 
supp\mathunderscore rel^{-}({\Not p}^{-}_{asm}, q^{-}_{A_3}), 
att\mathunderscore rel^{+}(p^{+}_{A_4},{\Not p}^{-}_{asm}),\\
supp\mathunderscore rel^{+}({\Not q}^{+}_{asm}, p^{+}_{A_4}), 
att\mathunderscore rel^{-}(q^{-}_{A_3},{\Not q}^{+}_{asm})
\}
\end{IEEEeqnarray*}
Note that the first is a set of sets, whereas the second is a simple set.\qed
\end{examplecont}

LABAS justifications can be represented as directed graphs, where the justified literal is depicted as the top node of the graph, and all literals occurring in a relation as the other nodes. Support and attack relations form two different arcs:
here, dashed arcs represent support, whereas solid arcs represent attack.
Both types of arcs are labelled according to the label in the LABAS justification.

\begin{examplecont}{ex:labas_set:cycle}\label{ex:labas:cycle}
The graphical representations of the LABAS justifications in Example~\ref{ex:labas_set:cycle} are respectively illustrated in
Figures~\ref{fig:labas:cycle1} and~\ref{fig:labas:cycle2}.
Unsurprisingly, they have the same nodes and arcs. However, the respective orientation of the graph indicates the literal being justified.
Note the difference between the LABAS justification graphs and the off-line justifications in Figure~\ref{fig:off-line.cycle}. In particular, the LABAS justification graphs
\changed{explain the truth values of non-fact positive literals in terms of negative literals needed to derive the positive literal. Furthermore, the truth values of negative literals, which do not occur in off-line justifications at all, are explained in terms of their complement's truth value. Also note that $q$ being false \wrt\ $\answersetr\ref{as:1:prg:cycle}$ is explained as a truth value being assumed in the off-line justifications, whereas its truth value is further explained in terms of the ineffective support by $\Not p$ in the LABAS justifications.}
 \qed
\end{examplecont}

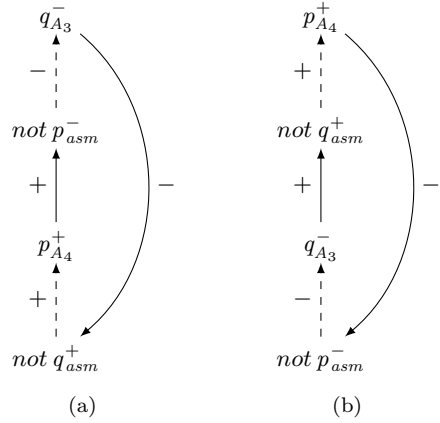
\begin{figure}
\centering
\subfloat[]{%
\begin{tikzpicture}[tikzpict]
    \matrix[row sep=0.5cm,column sep=0.3cm,ampersand replacement=\&] {
      \&
      \node (q) {$q^{-}_{A_3}$};
      \\
      \\
      \&
      \node (notp) {${\Not p}^{-}_{asm}$};
      \\
      \\
      \&
      \node (p) {$p^{+}_{A_4}$};
      \\
      \\
      \&
      \node (notq) {${\Not q}^{+}_{asm}$};
      \\
     };
    \draw[->] (p) to node[left]{$+$} (notp);
    \draw[->, dashed] (notp) to node[left] {$-$} (q);
    \draw[->, dashed] (notq) to node[left] {$+$}(p);
    \draw [->,bend left =50] (q) to node[right]{$-$} (notq);
\end{tikzpicture}%
\label{fig:labas:cycle1}
}
\hspace{0.5cm}
\centering
\subfloat[]{%
\begin{tikzpicture}[tikzpict]
    \matrix[row sep=0.5cm,column sep=0.3cm,ampersand replacement=\&] {
      \&
      \node (p) {$p^{+}_{A_4}$};
      \\
      \\
      \&
      \node (notq) {${\Not q}^{+}_{asm}$};
      \\
      \\
      \&
      \node (q) {$q^{-}_{A_3}$};
      \\
      \\
      \&
      \node (notp) {${\Not p}^{-}_{asm}$};
      \\
     };
    \draw[->] (q) to node[left]{$+$} (notq);
    \draw[->, dashed] (notp) to node[left] {$-$} (q);
    \draw[->, dashed] (notq) to node[left] {$+$}(p);
    \draw [->,bend left =50] (p) to node[right]{$-$} (notp);
\end{tikzpicture}%
\label{fig:labas:cycle2}
}
\caption{LABAS justifications of $q$ and $p$ \wrt\ $\answersetr\ref{as:1:prg:cycle}$:
dashed arcs represent support, whereas solid arcs represent attack.}
\end{figure}


\begin{examplecont}{ex:labas_trees:multipleAttack}\label{ex:labas:multipleAttack}
Figures~\ref{fig:labas:multipleAttack1} and~\ref{fig:labas:multipleAttack2} illustrate the LABAS justifications of $p$ \wrt\ $\answersetr\ref{as:1:prg:multipleAttack}$ and $\answersetr\ref{as:2:prg:multipleAttack}$ of $\programr\ref{prg:multipleAttack}$ (see Example~\ref{ex:labas_trees:multipleAttack}).
The first demonstrates the importance of labelling literals by their arguments for distinction.
If these labels did not exist, $r_{A_3}^{-}$ and~$r_{A_4}^{-}$ would collapse into one node $r^{-}$. The resulting graph would give the impression that there is only one derivation for $r$, which 
uses both $\Not p$ and $\Not s$. In contrast, the distinction achieved by labelling literals with their argument names (Figure~\ref{fig:labas:multipleAttack1}), expresses that there are two derivations for $r$, one using $\Not p$ and one using $\Not s$.
Note that off-line justifications use a non-labelling strategy, leading to the previously explained collapse of the two nodes holding atom $r$, as shown in Figures~\ref{fig:labas:multipleAttack1.off-line.1} and~\ref{fig:labas:multipleAttack1.off-line.2}.
\changed{Figure~\ref{fig:labas:multipleAttack1}, and in particular node $r_{A_3}^{-}$, furthermore shows that for nodes labelled \negArg\ in an attack tree, fact premises are not included in the LABAS justification ($A_3$ has a fact premise $s$).
In contrast, Figure~\ref{fig:labas:multipleAttack2}, and in particular node $r_{A_3}^{+}$, shows that for nodes labelled \posArg\ in an attack tree, all assumption and fact premises are included in a LABAS justification. Furthermore, for nodes labelled \negArg\ only the assumption premise that is attacked by the child node is included (only assumption premise $\Not r$ of $p$ is included and assumption premise $\Not q$ is neglected).}\qed
\end{examplecont}

\begin{figure}
\centering
\subfloat[]{%
\begin{tikzpicture}[tikzpict]
    \matrix[row sep=0.5cm,column sep=0.3cm,ampersand replacement=\&] {
      \&
      \node (p) {$p^{+}_{A_1}$};
      \\
      \\
      \node (notq) {${\Not q}^{+}_{asm}$};
      \&\&
      \node (notr) {${\Not r}^{+}_{asm}$};
      \\
      \\
      \node (q) {$q^{-}_{A_2}$};
      \&
      \node (r4) {$r^{-}_{A_4}$};
      \&
      \node (r3) {$r^{-}_{A_3}$};
      \\
      \\
      \node (nots) {${\Not s}^{-}_{asm}$};
      \&\&
      \node (notp) {${\Not p}^{-}_{asm}$};
      \\
      \\
      \node (s) {$s^{+}_{fact}$};
      \\
     };
    \draw[->, dashed] (notq) to node[left] {$+$} (p);
    \draw[->, dashed] (notr) to node[right] {$+$}(p);
    \draw [->] (q) to node[left]{$-$} (notq);
    \draw [->] (r3) to node[right]{$-$} (notr);
    \draw [->] (r4) to node[left]{$-$} (notr);
    \draw[->, dashed] (nots) to node[left] {$-$}(q);
    \draw[->, dashed] (nots) to node[right] {$-$}(r4);
    \draw [->] (s) to node[left]{$+$} (nots);
    \draw[->, dashed] (notp) to node[right] {$-$}(r3);
    \draw [->,bend left =110] (p) to node[right]{$+$} (notp);
\end{tikzpicture}%
\label{fig:labas:multipleAttack1}
}
\hspace{0.5cm}
\centering
\subfloat[]{%
\begin{tikzpicture}[tikzpict]
    \matrix[row sep=0.5cm,column sep=0.3cm,ampersand replacement=\&] {
      \&
      \node (p) {$p^{-}_{A_1}$};
      \\
      \\
      \&
      \node (notr) {${\Not r}^{-}_{asm}$};
      \\
      \\
      \&
      \node (r3) {$r^{+}_{A_3}$};
      \\
      \\
      \node (s) {$s^{+}_{fact}$};
      \&\&
      \node (notp) {${\Not p}^{+}_{asm}$};
      \\
     };
    \draw[->, dashed] (notr) to node[left] {$-$} (p);
    \draw[->] (r3) to node[left]{$+$} (notr);
    \draw[->, dashed] (notp) to node[right] {$+$}(r3);
    \draw[->, dashed] (s) to node[left] {$+$}(r3);
    \draw [->,bend left =30] (p) to node[right]{$-$} (notp);
\end{tikzpicture}%
\label{fig:labas:multipleAttack2}
}
\caption{LABAS justifications of $p$ \wrt\ $\answersetr\ref{as:1:prg:multipleAttack}$ and $\answersetr\ref{as:2:prg:multipleAttack}$.}
\end{figure}

\begin{figure}
\centering
\subfloat[]{%
\begin{tikzpicture}[tikzpict]
    \matrix[row sep=0.5cm,column sep=0.3cm,ampersand replacement=\&] {
      \&
      \node (p) {$p^{+}$};
      \\
      \\
      \node (q) {$q^{-}$};
      \&
      \&
      \node (r3) {$r^{-}$};
      \\
      \\
      \node (s) {$s^{+}$};
      \&\&
      \node (assume) {$\assume$};
      \\
     };
    \draw[<-] (q) to node[left] {$-$} (p);
    \draw[<-] (r3) to node[right] {$-$}(p);
    \draw[<-] (s) to node[left] {$-$}(q);
    \draw [<-] (assume) to node[right]{$-$} (r3);
\end{tikzpicture}%
\label{fig:labas:multipleAttack1.off-line.1}
}
\hspace{1.5cm}
\centering
\subfloat[]{%
\begin{tikzpicture}[tikzpict]
    \matrix[row sep=0.5cm,column sep=0.3cm,ampersand replacement=\&] {
      \&
      \node (p) {$p^{+}$};
      \\
      \\
      \node (q) {$q^{-}$};
      \&
      \&
      \node (r3) {$r^{-}$};
      \\
      \\
      \node (s) {$s^{+}$};
      \\
     };
    \draw[<-] (q) to node[left] {$-$} (p);
    \draw[<-] (r3) to node[right] {$-$}(p);
    \draw[<-] (s) to node[left] {$-$}(q);
    \draw[<-] (s) to node[left] {$-$}(r3);
    \draw [<-,bend left =110] (p) to node[right]{$-$} (r3);
\end{tikzpicture}%
\label{fig:labas:multipleAttack1.off-line.2}
}
\caption{Off-line justifications of $p$ \wrt\ $\answersetr\ref{as:1:prg:multipleAttack}$.}
\end{figure}

Comparing the LABAS justification in Figure~\ref{fig:labas:multipleAttack1} and the off-line justification in Figure~\ref{fig:labas:multipleAttack1.off-line.2}, we observe various similarities:
Deleting the nodes holding negative literals in the LABAS justification and collapsing the two nodes of atom $r$ results in the same nodes as in the off-line justification.
Note that this is because all derivations of atoms are ``one-step'' derivations, i.e. there is no chaining of rules involved. If the derivation of some atom involved the chaining of various rules, the \mbox{off-line} justification would include more nodes than the LABAS justification, even if nodes holding negative literals were deleted (see for example Figures~\ref{fig:dead.off-line} and~\ref{fig:dead.labas}).
Furthermore, `rerouting' the attack edges in the LABAS justification from the attacked negative literal to the atom supported by this negative literal (e.g. `rerouting' the attacking edge from $p^{+}$ to $\Not p$ instead to atom $r$, which is supported by $\Not p$) and then reverting them results, in this example, in the same edges as in the off-line justification. Note however that the labelling of edges is different in LABAS and off-line justifications.

The following examples point out some further differences between LABAS and off-line justifications.
In particular, LABAS justifications do not explicitly contain information about all rules applied in a derivation and 
there is no LABAS justification for literals that have no argument, i.e. literals that cannot be successfully derived.

\begin{examplecontpage}{ex:positive2}\label{ex:labas:positive2}
Figures~\ref{fig:positive2.labas} and \ref{fig:positive2.labas2} show the LABAS justifications of $p$ \wrt\ $\answersetr\ref{as:positive}$ of $\programr\ref{prg:positive2}$. 
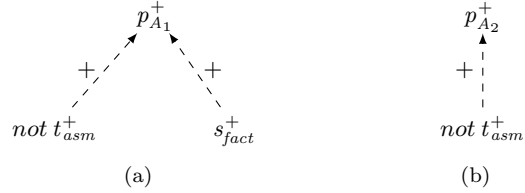
\begin{figure}[h]
\centering
\subfloat[]{%
\begin{tikzpicture}[tikzpict]
    \matrix[row sep=0.5cm,column sep=0.3cm,ampersand replacement=\&] {
      \&
      \node (p) {$p^{+}_{A_1}$};
      \\
      \\
      \node (nott) {${\Not t}^{+}_{asm}$};
      \&\&
      \node (s) {${s}^{+}_{fact}$};
      \\
     };
    \draw[->, dashed] (nott) to node[left] {$+$} (p);
    \draw[->, dashed] (s) to node[right] {$+$}(p);
\end{tikzpicture}%
\label{fig:positive2.labas}
}
\hspace{1.5cm}
\centering
\subfloat[]{%
\begin{tikzpicture}[tikzpict]
    \matrix[row sep=0.5cm,column sep=0.3cm,ampersand replacement=\&] {
      \&
      \node (p) {$p^{+}_{A_2}$};
      \\
      \\
      \&
      \node (nott) {${\Not t}^{+}_{asm}$};
      \\
     };
    \draw[->, dashed] (nott) to node[left] {$+$} (p);
\end{tikzpicture}%
\label{fig:positive2.labas2}
}
\caption{The two LABAS justifications of $p$ \wrt\ $\answersetr\ref{as:positive}$ of $\programr\ref{prg:positive2}$.}
\end{figure}
The difference between the two derivations of $p$ is not as explicit as in the off-line justifications
illustrated in Figures~\ref{fig:positive.off-line} and~\ref{fig:positive2.off-line} (page~\pageref{fig:positive.off-line}). It is merely indicated by the different argument labels of $p$.\qed
\end{examplecontpage}

\begin{examplecontpage}{ex:cyclic.justifications}\label{ex:labas:cyclic.justifications}
There are two off-line justifications of $r$ \wrt\ $\programr\ref{prg:cyclic.justifications}$ and
$\answersetr\ref{as:prg:cyclic.justifications}$ (see Figures~\ref{fig:cyclic.justifications.a} and~\ref{fig:cyclic.justifications.b} on page~\pageref{fig:cyclic.justifications.a}). In contrast, there is only \emph{one} LABAS justification of $r$,
shown in Figure~\ref{fig:cyclic.justifications:labas}. The reason is that there is no argument with conclusion $p$,
since no rule with head $p$ exists. Thus, $\Not p$ is not further explained as there is no way to prove~$p$.\qed
\end{examplecontpage}

\begin{figure}
\centering
\begin{tikzpicture}[tikzpict]
    \matrix[row sep=0.5cm,column sep=0.3cm,ampersand replacement=\&] {
      \&
      \node (r) {$r^{+}_{A_1}$};
      \\
      \\
      \&
      \node (notp) {${\Not p}^{+}_{asm}$};
      \\
     };
    \draw[->, dashed] (notp) to node[left] {$+$} (r);
\end{tikzpicture}%
\caption{The unique LABAS justification of $r$ \wrt\ $\answersetr\ref{as:prg:cyclic.justifications}$ of $\programr\ref{prg:cyclic.justifications}$.}
\label{fig:cyclic.justifications:labas}
\end{figure}

As previously pointed out, infinite attack trees may be represented by finite LABAS justifications.
However, this is only the case if the infinity is due to the repetition of the same arguments.
Instead, if the infinity is due to the existence of infinitely many arguments with the same conclusion, 
a LABAS justification may be infinite too.


\begin{example}\label{ex:labas:infinite}
Let $\newprogram\label{prg:infinite}$ be the following program with answer sets
$\newanswerset\label{as:1:infinite} = \{p,r\}$ and
\mbox{$\newanswerset\label{as:2:infinite} = \{q,r\}$}:
\begin{gather*}
p \lparrow \Not q \wedge r \hspace{2cm}
q \lparrow \Not p \wedge r \hspace{2cm}
r \lparrow r \hspace{2cm}
r
\end{gather*}
Note first that there are infinitely many arguments with conclusion $r$ of the form $A_{r_i}:\argument{\{\}}{\{r\}}{r}$, each applying the third rule a different number of times.
For the same reason, there are infinitely many arguments with conclusion $p$, of the form $A_{p_j}:\argument{\{\Not q\}}{\{r\}}{p}$, and with conclusion $q$, of the form $A_{q_k}:\argument{\{\Not p\}}{\{r\}}{q}$.
Since there are infinitely many arguments with conclusion $p$ (resp. $q$), there are also infinitely many attack trees explaining $p$ (resp. $q$) with respect to either of the two answer sets.
Similarly to the attack trees illustrated in Figures~\ref{fig:labas_tree:cycle1} and~\ref{fig:labas_tree:cycle2},
all attack trees for $p$ and $q$ are infinite in depth.
In addition, they are infinite in breadth since any of the $A_{p_j}$ attacks every $A_{q_k}$ and vice versa.
This means that whenever an argument for $p$ (resp. $q$) is labelled \posArg\ in an attack tree, all infinitely many arguments with conclusion $q$ (resp. $p$) are child nodes labelled \negArg.
\begin{figure}
\centering
\subfloat[]{%
\begin{tikzpicture}[tikzpict]
    \matrix[row sep=0.5cm,column sep=0.3cm,ampersand replacement=\&] {
      \&
      \node (p) {$A_{p_1}^{+}: \argument{\{\Not q\}}{\{r\}}{p}$};
      \\
      \node (q1) {$A_{q_1}^{-}: \argument{\{\Not p\}}{\{r\}}{q}$};
      \&
      \node (q2) {$A_{q_2}^{-}: \argument{\{\Not p\}}{\{r\}}{q}$};
      \&\&
      \node (q3) {$\mathbf{\ldots}$};
      \\
      \node (p2) {$A_{p_1}^{+}: \argument{\{\Not q\}}{\{r\}}{p}$};
      \&
      \node (p3) {$A_{p_1}^{+}: \argument{\{\Not q\}}{\{r\}}{p}$};
      \\[-0.5cm]
      \node (dots1) {$\vdots$};
      \&
      \node (dots2) {$\vdots$};
      \\
     };
    \draw[->] (q1) -- (p);
    \draw[->] (q2) -- (p);
    \draw[->] (p2) -- (q1);
    \draw[->] (p3) -- (q2);
\end{tikzpicture}%
\label{fig:labas:infinite.attacktree}
}
\hspace{0.5cm}
\centering
\subfloat[]{%
\begin{tikzpicture}[tikzpict]
    \matrix[row sep=0.5cm,column sep=0.3cm,ampersand replacement=\&] {
      \&
      \node (p) {$p^{+}_{A_{p_1}}$};
      \\
      \\
      \&
      \node (notq) {${\Not q}^{+}_{asm}$};
      \&\&
      \node (r) {$r^{+}_{fact}$};
      \\
      \\
      \node (q3) {$\mathbf{\ldots}$};
      \&
      \node (q2) {$q^{-}_{A_{q_2}}$};
      \&\&
      \node (q1) {$q^{-}_{A_{q_1}}$};
      \\
      \\
      \&
      \node (notp) {${\Not p}^{-}_{asm}$};
      \\
     };
    \draw[->,dashed] (notq) to node [left] {$+$}(p);
    \draw[->,dashed] (r) to node [right] {$+$}(p);
    \draw[->] (q1) to node [left] {$-$}(notq);
    \draw[->] (q2) to node [left] {$-$}(notq);
    \draw[->] (q3) to node [left] {$-$}(notq);
    \draw[->,dashed] (notp) to node [left] {$-$}(q1);
    \draw[->,dashed] (notp) to node [left] {$-$}(q2);
    \draw[->,dashed] (notp) to node [left] {$-$}(q3);
    \draw[->,bend left =100,looseness=2] (p) to node [left] {$+$}(notp);
\end{tikzpicture}%
\label{fig:labas:infinite}
}
\caption{One of the infinite attack trees and LABAS justifications of $p$ \wrt\ $\answersetr\ref{as:1:infinite}$ of $\programr\ref{prg:infinite}$.}
\end{figure}
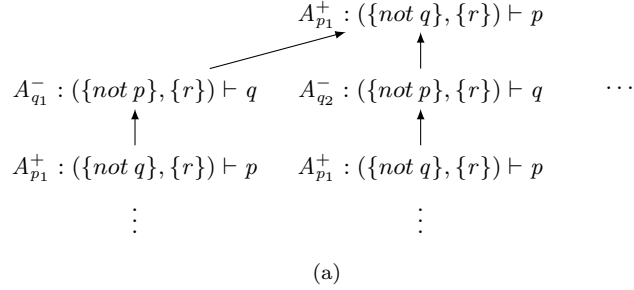
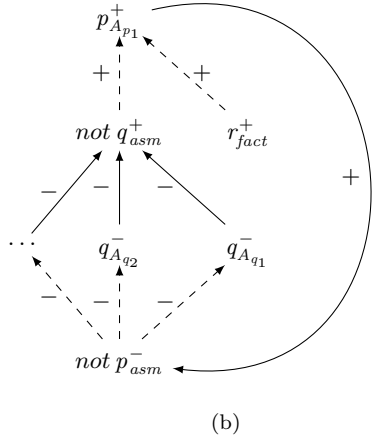
Figure~\ref{fig:labas:infinite.attacktree} illustrates an attack tree of one of the arguments with conclusion $p$ \wrt\ $\answersetr\ref{as:1:infinite}$. Note that in this particular attack tree, the argument $A_{p_1}^{+}:\argument{\{\Not q\}}{\{r\}}{p}$ is re-used to attack all the arguments with conclusion $q$ attacking the root node.
By exchanging any occurrence of $A_{p_1}^{+}:\argument{\{\Not q\}}{\{r\}}{p}$ by another argument with conclusion $p$, e.g. $A_{p_2}^{+}:\argument{\{\Not q\}}{\{r\}}{p}$, a different (infinite) attack tree explaining $p$ is obtained.
We observe that any of these attack trees yields an infinite LABAS justification.
For example, the attack tree from Figure~\ref{fig:labas:infinite.attacktree} results in a LABAS justification with infinitely many relations of the form $att\mathunderscore rel^{-}(q_{A_{q_k}}^{-}, p_{A_{p_j}}^{+})$ relations.
Assuming that the only argument with conclusion $p$ used in the attack tree in Figure~\ref{fig:labas:infinite.attacktree} is $A_{p_1}^{+}:\argument{\{\Not q\}}{\{r\}}{p}$, we obtain the infinite LABAS justification in Figure~\ref{fig:labas:infinite}.
\qed
\end{example}

This behaviour of infinity is dealt with in the \texttt{LABAS Justifier} by disallowing the repeated application of a rule when constructing an argument \cite{Schulz2017Thesis}.
In Example~\ref{ex:labas:infinite}, the \texttt{LABAS Justifier} thus only constructs two different arguments for $p$ and $q$.

\subsection{Causal Graph Justifications}\label{sec:causal}
\label{sec:cg}



\CHANGED
In contrast to the two previously discussed approaches (off-line and LABAS justifications), whose main purpose is to explain why a literal is (not) contained in an answer set, the approach outlined in this section -- called \emph{causal graph justifications}~\cite{CabalarFF2014,CabalarF16} -- is a reasoning formalism in its own right, which can additionally be used to explain why a literal is contained in an answer set:
the main goal of the causal justification approach is to formalise and reason with causal knowledge, so that sentences like ``whoever causes the death of somebody else will be imprisoned'' can be represented in an \emph{elaboration tolerant}\footnote{We recall that a representation is elaboration tolerant
if modifications of it can easily be taken into account.} manner~\cite{McC98}.
\END
\CHANGED
An online tool providing causal justifications and allowing this reasoning with causal knowledge~\cite{DFandinno16non-monotonic} is available at
\url{http://kr.irlab.org/cgraphs-solver/nmsolver}.
\END

The semantics used for causal justifications is a multi-valued extension of the answer set semantics, where each (true) literal in a model is associated with a set of \changed{causal values expressing causal reasons for its inclusion in the model. Each of these causal values represents a set of \emph{causal justifications}, each of which, in turn, can be depicted as a \emph{causal graph}}.
Regarding the causal literature, a \emph{causal graph} can be seen as an extension of Lewis's notion of \emph{causal chain}: ``let $c$, $d$, $e$, $\dotsc$ be a finite sequence of actual particular events such that $d$ causally depends on $c$, $e$ on~$d$, and so on throughout. Then, this sequence is a causal chain.''~(\citeNP{lewis1973causation}; see also \citeNP{hall2004} and \citeNP{hall2007structural}).
\changed{The following example illustrates the connection between causal chains and justifications in ASP.}

\begin{example}\label{ex:dead}
Consider a scenario in which Suzy pulls the trigger of her gun,
causing the gunpowder to explode.
This causes the bullet to leave the gun at a high speed, impacting on Billy's chest, provoking a massive haemorrhage and, consequently, Billy's death.
We can model this scenario as the following positive logic program~\newprogram\label{prg:dead}:
\begin{IEEEeqnarray}{l C ' l C l}
&& \dead &\lparrow&  \haemorrhage
\label{prg:dead.r1}
\\
&& \haemorrhage &\lparrow&  \impact
\label{prg:dead.r2}
\\
&& \impact &\lparrow&  \bullett
\label{prg:dead.r3}
\\
&& \bullett &\lparrow&  \gunpowder
\label{prg:dead.r4}
\\
&& \gunpowder &\lparrow&  \trigger(\suzy)
\label{prg:dead.r5}
\\
&& \trigger(\suzy)
\label{prg:dead.r6}
\end{IEEEeqnarray}
Then,
$\trigger(\suzy) \cdot \gunpowder \cdot \bullett \cdot \impact \cdot \haemorrhage \cdot \dead$ is a causal chain connecting $\trigger(\suzy)$ with $\dead$.\qed
\end{example}

\CHANGED
This example suggests
an intuitive correspondence between causal chains and the idea of justification.
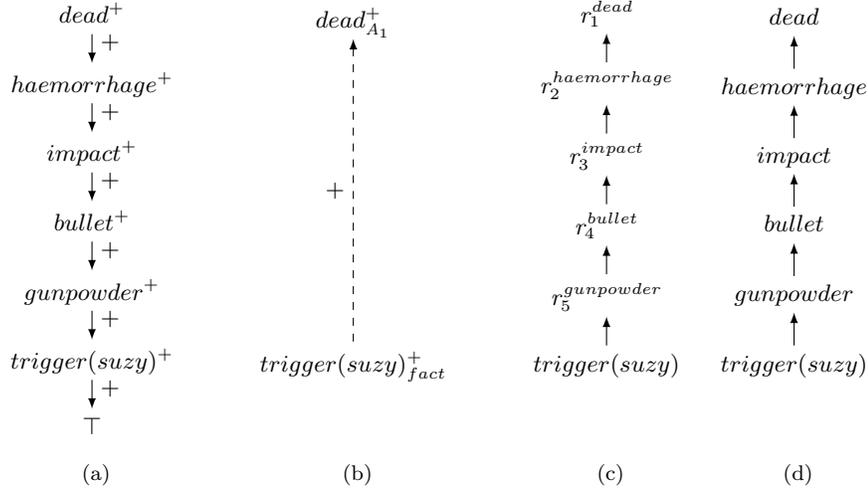
\begin{figure}[t]\centering
\subfloat[]{%
\begin{tikzpicture}[tikzpict]
    \matrix[row sep=0.4cm,column sep=0.8cm,ampersand replacement=\&] {
      \node (dead) {$\dead^{+}$};
      \\
      \node (haemorrhage) {$\haemorrhage^{+}$};
      \\
      \node (impact) {$\impact^{+}$};
      \\
      \node (bullett) {$\bullett^{+}$};
      \\
      \node (gunpowder) {$\gunpowder^{+}$};
      \\
      \node (trigger) {$\trigger(\suzy)^{+}$};
      \\
      \node (top) {$\top$};
      \\
     };
    \draw [->] (dead) to node[pos=0.3,right]{$+$}  (haemorrhage);
    \draw [->] (haemorrhage) to node[pos=0.3,right]{$+$} (impact);
    \draw [->] (impact) to node[pos=0.3,right]{$+$} (bullett);
    \draw [->] (bullett) to node[pos=0.3,right]{$+$} (gunpowder);
    \draw [->] (gunpowder) to node[pos=0.3,right]{$+$} (trigger);
    \draw [->] (trigger) to node[pos=0.3,right]{$+$} (top);
\end{tikzpicture}%
\label{fig:dead.off-line}
}
\hspace{0.5cm}
\subfloat[]{%
\begin{tikzpicture}[tikzpict]
    \matrix[row sep=0.4cm,column sep=0.8cm,ampersand replacement=\&] {
      \node (dead) {$\dead^{+}_{A_1}$};
      \\[30pt]
      \node (haemorrhage) {$ $};
      \\
      \node (impact) {$ $};
      \\
      \node (bullett) {$ $};
      \\
      \node (gunpowder) {$ $};
      \\
      \node (trigger) {$\trigger(\suzy)_{\mathit{fact}}^{+}$};
      \\[5pt]
      \node (top) {$ $};
      \\
     };
    \draw [<-, dashed] (dead) to node [left]{$+$} (trigger);
\end{tikzpicture}%
\label{fig:dead.labas}
}
\hspace{0.5cm}
\subfloat[]{%
\begin{tikzpicture}[tikzpict]
    \matrix[row sep=0.4cm,column sep=0.8cm,ampersand replacement=\&] {
      \node (dead) {$r_1^{\dead}$};
      \\
      \node (haemorrhage) {$r_2^{\haemorrhage}$};
      \\
      \node (impact) {$r_3^{\impact}$};
      \\
      \node (bullett) {$r_4^{\bullett}$};
      \\
      \node (gunpowder) {$r_5^{\gunpowder}$};
      \\
      \node (trigger) {$\trigger(\suzy)$};
      \\[5pt]
      \node (top) {$ $};
      \\
     };
    \draw [<-] (dead) to  (haemorrhage);
    \draw [<-] (haemorrhage) to (impact);
    \draw [<-] (impact) to (bullett);
    \draw [<-] (bullett) to (gunpowder);
    \draw [<-] (gunpowder) to (trigger);
\end{tikzpicture}%
\label{fig:dead.causal.just}
}
\subfloat[]{%
\begin{tikzpicture}[tikzpict]
    \matrix[row sep=0.4cm,column sep=0.8cm,ampersand replacement=\&] {
      \node (dead) {$\dead$};
      \\[1.5pt]
      \node (haemorrhage) {$\haemorrhage$};
      \\[1.5pt]
      \node (impact) {$\impact$};
      \\[1.5pt]
      \node (bullett) {$\bullett$};
      \\[1.5pt]
      \node (gunpowder) {$\gunpowder$};
      \\[1.5pt]
      \node (trigger) {$\trigger(\suzy)$};
      \\[5pt]
      \node (top) {$ $};
      \\
     };
    \draw [<-] (dead) to  (haemorrhage);
    \draw [<-] (haemorrhage) to (impact);
    \draw [<-] (impact) to (bullett);
    \draw [<-] (bullett) to (gunpowder);
    \draw [<-] (gunpowder) to (trigger);
\end{tikzpicture}%
\label{fig:dead.causal.chain}
}
\hspace{0.5cm}
\caption{Off-line justification, LABAS justification, causal justification and causal chain of $\dead$ in Example~\ref{ex:dead}.}\label{fig:dead.all.just}
\end{figure}
In particular, the causal chain that connects the fact $\trigger(\suzy)$ with $\dead$ can be written as the graph in Figure~\ref{fig:dead.causal.chain}.
It is easy to see the correspondence between this graph and the off-line justification of $\dead$, depicted in Figure~\ref{fig:dead.off-line}.
\END
For comparison, Figures~\ref{fig:dead.labas} and~\ref{fig:dead.causal.just} depict the LABAS justification and the causal 
graph
 (which will be defined later) of $\dead$.
Recall that LABAS justifications focus on facts and negative literals, precisely abstracting from the causal chain, which will be the focus of causal justifications.
\changed{In contrast, the causal graph expresses the same information as the causal chain.
This is due to the fact that no atom depends on more than one other atom.
More generally, causal chains coincide with the paths in causal graphs.}

\changed{In addition to the idealogical differences between causal justifications, which treat logic programs as causal knowledge, and off-line and LABAS justifications, which treat logic programs as declarative problem descriptions, causal justifications allow for causal reasoning, as they are based on a causal extension of the answer set semantics. More precisely,}
causal justifications are
\changed{defined in terms of}
 the \emph{causal value} that each \emph{causal answer set} associates to atoms (causal answer sets assign causal values instead of truth values to each atom).
These causal values form a completely distributive (complete) lattice that serves as the basis for a multi-valued extension of the answer set semantics.

Let us introduce causal terms as a suitable syntax to write causal values.

\begin{definition}[Causal Term]
\label{def:term}
Given a set of atoms $\at$ and a set of labels $\lb$, a \emph{\changed{(causal)} term} $t$ is recursively defined as one of the following expressions
\begin{gather*}
t \ \ ::= \ \ l \ \ \Big| \ \ \prod S \ \ \Big| \ \ \sum S \ \ \Big| \ \ t_1 \cdot t_2
\end{gather*}
where $l \in (\ExpLit \cup \lb)$ is an extended atom or a label, $t_1, t_2$ are in turn terms, and $S$ is a (possibly empty and possibly infinite) set of terms.\qed
\end{definition}

When $S = \set{t_1, \dotsc, t_n}$ is a finite set, we write $t_1 * \dotsc * t_n$ and $t_1 + \dotsc + t_n$ instead of~$\prod S$ and $\sum S$, respectively. 
The empty sum and empty product are respectively represented as $0$ and $1$.
We assume that \emph{application}~`$\cdot$' has higher priority than product~`$*$' and, in turn, product~`$*$' has higher priority than addition~`$+$'.
Intuitively, product `$*$' represents conjunction or joint causation,
sum~`$+$' represents alternative causes,
and \emph{application} `$\cdot$' is a non-commutative product that builds causal chains by capturing the successive application of rules.
\begin{figure}[htbp]
\begin{center}
\footnotesize
\newcommand{\titleSep}{-5pt}
\newcommand{\contentSep}{-10pt}
\newcommand{\rowSep}{5pt}
$
\begin{array}{c}
\hbox{\em Associativity}\vspace{\titleSep}\\
\hline\vspace{\contentSep}\\
\begin{array}{r@{\ }c@{\ }r@{}c@{}l c r@{}c@{}l@{\ }c@{\ }l@{\ }}
t & \cdot & (u & \cdot & w) & = & (t & \cdot & u) & \cdot & w\\
\\
\end{array}
\end{array}
$
\hspace{3pt}
$
\begin{array}{c}
\hbox{\em Absorption}\vspace{\titleSep}\\
\hline\vspace{\contentSep}\\
\begin{array}{r@{\ }c@{\ }c@{\ }c@{\ }l c r@{\ }c@{\ }r@{\ }c@{\ }c@{\ }c@{\ }c@{\ }l@{\ }}
&& t &&& = & t & + & u & \cdot & t & \cdot & w \\
u & \cdot & t & \cdot & w & = & t & * & u & \cdot & t & \cdot & w
\end{array}
\end{array}
$
\hspace{3pt}
$
\begin{array}{c}
\hbox{\em Identity}\vspace{\titleSep}\\
\hline\vspace{\contentSep}\\
\begin{array}{rc r@{\ }c@{\ }l@{\ }}
t & = & 1 & \cdot & t\\
t & = & t & \cdot & 1
\end{array}
\end{array}
$
\\
\vspace{\rowSep}
$
\begin{array}{c}
\hbox{\em Annihilator}\vspace{\titleSep}\\
\hline\vspace{\contentSep}\\
\begin{array}{rc r@{\ }c@{\ }l@{\ }}
0 & = & t & \cdot & 0\\
0 & = & 0 & \cdot & t\\
\end{array}
\end{array}
$
\hspace{3pt}
$
\begin{array}{c}
\hbox{\em Indempotence}\vspace{\titleSep}\\
\hline\vspace{\contentSep}\\
\begin{array}{r@{\ }c@{\ }l@{\ }c@{\ }l }
l & \cdot & l  & = & l\\
\\
\end{array}
\end{array}
$
\hspace{.05cm}
$
\begin{array}{c}
\hbox{\em Addition\ distributivity}\vspace{\titleSep}\\
\hline\vspace{\contentSep}\\
\begin{array}{r@{\ }c@{\ }r@{}c@{}l c r@{}c@{}l@{\ }c@{\ }r@{}c@{}l@{}}
t & \cdot & (u & + & w) & = & (t & \cdot & u) & + & (t & \cdot & w)\\
( t & + & u ) & \cdot & w & = & (t & \cdot & w) & + & (u & \cdot & w)\\
\end{array}
\end{array}
$
\\
\vspace{\rowSep}
$
\begin{array}{c}
\hbox{\em Product\ distributivity}\vspace{\titleSep}\\
\hline\vspace{\contentSep}\\
\begin{array}{rcl}
c \cdot d \cdot e & = & (c \cdot d) * (d \cdot e) \ \hbox{with} \ d \neq 1 \\
c \cdot (d*e)     & = & (c \cdot d) * (c \cdot e) \\
(c*d) \cdot e     & = & (c \cdot e) * (d \cdot e)
\end{array}
\end{array}
$
\end{center}
\vspace{-5pt}
\caption{Properties of the operators:
$t,u,w$ are terms, $l$ is a label or an extended atom and $c,d,e$ are terms without addition~`$+$'.
Addition and product distributivity are also satisfied over infinite sums and products.
A kind of absorption over infinite sums and products can also be derived from the finite absorption above and infinite distributivity.}
\label{fig:appl}
\end{figure}

\begin{definition}[\changed{Causal} Value]
\label{def:values}
\changed{\emph{(Causal) values}} are the equivalence classes of terms under the axioms for a completely distributive (complete) lattice with 
\changed{meet~`$\prod$' and join~`$\sum$'} plus the axioms in Figure~\ref{fig:appl}.
The set of values is denoted by~$\values$.
Furthermore, by $\causes$ we denote the subset of causal values with some representative term without addition~`$\sum$'.\qed
\end{definition}

As an example, the causal value $[a] = \set{ a ,\, a * a,\, a + a,\, a \cdotl a ,\, a * ( a + b ),\, \dotsc }$ is the (possibly infinite) set of causal terms that are equivalent to $a$ under the axioms for a completely distributive lattice with meet~`$\prod$' and join~`$\sum$' plus the axioms in Figure~\ref{fig:appl}.
\changed{Note} that there are no causal terms equivalent to $0$ or~$1$ besides themselves, that is, $[0] = \set{0}$ and $[1] = \set{1}$.
By abuse of notation, we will use any causal term belonging to a causal value to represent the value, that is, we write $a$ instead of $[a]$, $0$ instead of $[0]$, and so on.

Note that all three operations `$*$', `$+$' and `$\cdot$' are associative. Product~`$*$' and addition `$+$' are also commutative, and they satisfy the usual absorption and distributive laws with respect to infinite sums and products of a completely distributive lattice. 
As usual, the lattice order relation is defined as:
\begin{IEEEeqnarray*}{c"C"c"C"c}
t \leq u & \text{ iff } & t * u = t & \text{ iff } & t + u = u
\end{IEEEeqnarray*}
An immediate consequence of this definition is that the \mbox{$\leq$-relation} has the product as greatest lower bound, the addition as \changed{least upper bound,} $1$ as top element and~$0$ as bottom element.
The term~$1$ represents a value that holds by default, without an explicit cause, and will be assigned to the empty body.
The term~$0$ represents the absence of cause or the empty set of causes, and will be assigned to falsity.

Furthermore, applying distributivity (and absorption) of products and applications over addition, every term can be represented in a \emph{(minimal) disjunctive normal form} in which addition is not in the scope of any other operation and every pair of addends are pairwise $\leq$-incomparable.
As we will see in Example~\ref{ex:causal.sum},
this normal form emphasises the intuition that addition~`$+$' separates alternative causes.
Moreover, applying product distributivity, this normal form can be further rewritten into a \emph{graph normal form}
in which
the application operator~`$\cdot$'
 is only applied to \emph{pairs} of labels or extended atoms, thus representing the edges of a graph:
$v \cdotl v'$ with $v,v' \in (\ExpLit \cup \lb)$.
For instance, 
\CHANGED
applying priority rules,
\newChanged{the} causal terms $a * (((b \cdotl c) \cdotl e) + d)$ and $((a * ((b \cdotl c) \cdotl e)) + (a* d)$ can be rewritten as
$a * (b \cdotl c \cdotl e + d)$
and
$a * b \cdotl c \cdotl e + a* d$,
respectively.
Furthermore, it is easy to see that these two terms represent the same causal value since the former can be rewritten as the latter by applying distributivity of products over sums.
The latter is in disjunctive normal form
and can be further rewritten in graph normal form as
$a * b \cdotl c * c \cdotl e + a* d$ by applying distributivity of application over products. 
\END

Given any causal term without sums $c \in \causes$ in graph normal form, we can associate a graph $G_c = \tuple{V,E}$ where $V$ is the set of labels and extended atoms occurring in $c$ and~$E$ contains an edge $(v,v')$ for every subterm of the form $v \cdotl v'$.
By $\cgraph{c}$ we denote the transitive and reflexive reduction%
\footnote{Recall that the transitive and reflexive reduction of a graph $G$ is a graph~$G'$ whose transitive and reflexive closure is~$G$.
A causal graph \changed{(see Definition~\ref{def:causal.justification.graph})}, in which every cycle is a reflexive edge, has a unique transitive and reflexive reduction.}
of $G_c$.
Given this relation between application~`$\cdot$' and edges 
in such graphs
 it follows that application~`$\cdot$' must be non-commutative.
For any causal term in normal form~$t$, by $\cgraphs{t}$ we denote the set containing a graph $\cgraph{c}$ for each addend $c$ in~$t$.

\begin{examplecont}{ex:dead}
The causal chain of Example~\ref{ex:dead} is in disjunctive normal form (since it does not contain products nor sums), but not in graph normal form.
Using product distributivity, this causal chain can be rewritten in graph normal form as
$(\trigger(\suzy) \cdot \gunpowder) * (\gunpowder \cdot \bullett) * (\bullett \cdot \impact) * (\impact \cdot \haemorrhage) * (\haemorrhage \cdot \dead)$.
In this form, every subterm of the form $(v\cdotl v')$ corresponds to an edge in~Figure~\ref{fig:dead.causal.chain}.
\qed
\end{examplecont}

\CHANGED
So far, we have introduced causal values, which will be the semantic building blocks of causal justifications and the associated causal graphs. In the following, we define how these causal values are assigned to each atom to form causal answer sets and how causal justifications and graphs are obtained.
\END


\subsubsection{Causal Semantics for Programs without Negation-as-Failure}
 \changed{Semantics for logic programs usually assign truth values to atoms.
 In contrast, for the causal semantics of logic programs, causal interpretations assign causal values to atoms. Based on this, causal models and causal answer sets are defined. Causal justifications are then extracted using the causal value of atoms in a causal answer set corresponding to a standard answer set.}
 
A \emph{\changed{(causal)} interpretation} is a mapping \mbox{$\cI:\ExpLit \longrightarrow\values$} assigning a value to each extended atom and  satisfying $\cI(a) = 0$ or $\cI(\neg a) = 0$ for every atom \mbox{$a \in \at$}.
By
\mbox{$\Atoms(I)\eqdef \setm{ \rA \in \ExpLit }{ \cI(\rA) \neq 0}$}
we denote the set of extended atoms in an interpretation $I$.
For any pair of interpretations $\cI$ and $\cJ$,
we write $\cI\leq \cJ$ to represent the straightforward causal ordering, that is,
\mbox{$\cI(\rA) \leq \cJ(\rA)$}
for every atom $\rA \in \ExtAt$
and
we write $\cI \wleq \cJ$ when either $\cI \leq \cJ$ or $\Atoms(I) \subset \Atoms(J)$.
That is, $\cI \wleq \cJ$ is a weaker \changed{partial order,} since apart from the cases in which $\cI \leq \cJ$ holds, it also holds when true atoms in $I$ are a strict subset of true atoms in~$J$.
As usual, we write \mbox{$I<J$} (resp. $I \wless J$) iff $I\leq J$ (resp $I\wleq J$) and $I\neq J$.
Note that
\mbox{$\Atoms(I) \subset \Atoms(J)$} implies $I\neq J$ and so $I \wless J$.
We say that an interpretation $I$ is $\leq$-minimal (resp.~\mbox{$\wleq$-minimal}) satisfying some property when there is no $J < I$ (resp. $J \wless I)$ satisfying that property.
Note that there is a \mbox{$\leq$-bottom} and \mbox{$\wleq$-bottom} interpretation~\botI\ (resp. a $\leq$-top and $\wleq$-top interpretation~\topI) that stands for the interpretation mapping every extended atom~$\rA$ to the causal value~$0$ (resp. $1$).
\CHANGED
It is easy to see that $\wleq$-minimal models are also $\leq$-minimal models, though the converse is not necessarily true, as will be illustrated by Example~\ref{ex:disj.semantics.dif} (see page~\pageref{ex:disj.semantics.dif}).
\END
For every rule $r$ in the program,
we assign a label denoted by $\Label{\R}$.
We assume that $\Label{\rH} = \rH$ for every \changed{definite} fact~$\rH$
and that $\Label{\R} \neq \Label{\R'}$ for every pair of distinct rules $\R$ and $\R'$.
We also assume that $\lb$ contains all rule labels.


\begin{definition}[Causal Model]\label{def:causal.model.positive}
An interpretation $I$ \emph{satisfies} a positive rule $\R$ of the form~\eqref{eq:rule} (with $m=0$) iff
\begin{gather}
\big( \, I(\rB_1) * \dotsc * I(\rB_\npbody) \, \big) \cdot r_i \cdot \rH_j \ \ \leq \ \ I(\rH_j)
\end{gather}
for some atom $\rH_j \in \head{\R}$
and where $\R_i = \Label{\R}$ is the label associated with rule~$\R$.
We say that an interpretation~$I$ is a \emph{\changed{(causal)} model} of a positive extended program~$\P$, in symbols $I \models \P$, iff $I$ satisfies all rules in~$\P$.
\qed
\end{definition}

\begin{examplecont}{ex:dead}\label{ex:dead.3}
Let us assume that rules of $\programr\ref{prg:dead}$ are respectively labelled as $r_1$, $r_2$, $r_3$, $r_4$, $r_5$ and~$\trigger(\suzy)$.
Then, it is easy to check that the model $I$ of $\programr\ref{prg:dead}$
must satisfy
\begin{IEEEeqnarray*}{l ?C? l +x*}
I(\trigger(\suzy)) &\geq& \trigger(\suzy) \cdot \trigger(\suzy) \ \ = \ \ \trigger(\suzy)
\\
I(\gunpowder) &\geq& \trigger(\suzy) \cdot r_5 \cdot \gunpowder
&\qed
\end{IEEEeqnarray*}
\end{examplecont}

\begin{observation}{\label{obs:causa.fact}}
If $\R$ is a \changed{definite} fact $\rH$, that is, it has the form $(\rH \lparrow)$, then
$\Label{\R} = \rH$ and, thus, $I \models r$ iff $I(A) \geq \rH \cdotl \rH = \rH$ (by idempotence of application on labels).\qed
\end{observation}

\changed{Based on the definitions of causal values and models, the causal extension of the answer set semantics is defined as follows.}

\begin{definition}[Causal Answer Set without Negation-as-Failure]
\label{def:causal.smodel.positive}
Let $\P$ be a positive extended program.
A model~$I$ of~$\P$ is a \emph{causal answer set} iff it is \mbox{$\wleq$-minimal} among the models of~$\P$.
\qed
\end{definition}

\begin{examplecont}{ex:dead.3}\label{ex:dead.4}
Continuing with our running example, note that there is only one rule with atoms $\trigger(\suzy)$ and $\gunpowder$ in the head.
Then, any $\sqsubseteq$-minimal model $\newcausalanswerset\label{cas:prg:dead}$ of $\programr\ref{prg:dead}$
must satisfy equality instead of $\geq$, that is,
\begin{IEEEeqnarray*}{l ?C? l +x*}
\causalanswersetr\ref{cas:prg:dead}(\trigger(\suzy)) &=& \trigger(\suzy) \cdot \trigger(\suzy) \ \ = \ \ \trigger(\suzy)
\\
\causalanswersetr\ref{cas:prg:dead}(\gunpowder) &=& \trigger(\suzy) \cdot r_5 \cdot \gunpowder
\end{IEEEeqnarray*}
\CHANGED
Note that any $\sqsubseteq$-minimal model must also be a $\leq$-minimal model and, thus, $\causalanswersetr\ref{cas:prg:dead}(A)$ must be equal to the least upper bound of the terms corresponding to all rules with the atom $A$ in the head. Since here we only have one rule for each atom, this least upper bound coincides with the value corresponding to that rule.\qed
\end{examplecont}

\CHANGED
\begin{definition}[Causal Justification and Causal Graph]\label{def:causal.justification.graph}
Given a logic program $P$ and an answer set $M$ of $P$, a term without sums $c$ is a \emph{causal justification} of some atom $a$ \wrt~$P$ and~$M$ if there is some causal answer set $I$ of $P$ such that $\Atoms(I) = M$ and $c$ is an addend in the minimal disjunctive normal form of $I(a)$.
For any causal justification of $a$ \wrt~$P$ and~$M$,
$\cgraph{c}$ is a \emph{causal graph (justification)}.\qed
\end{definition}
\END

\begin{notation}
\CHANGED
In causal justifications, we will write $r_i^\rA$ instead of $r_i \cdotl \rA$ when $r_i \in \lb$ is a rule label and $\rA \in \ExpLit$ is an extended atom occurring in the head of the rule labelled~$r_i$.
Similarly, in causal graphs we write a single vertex $r_i^a$ instead of two vertices $r_i$ and $a$ and an edge connecting them.\qed
\end{notation}

\begin{examplecont}{ex:dead.4}\label{ex:dead.5}
Assuming the above notation, we may rewrite the causal value associated with $\gunpowder$, \changed{which is also 
its unique causal justifications,}
as
$\causalanswersetr\ref{cas:prg:dead}(\gunpowder) = \trigger(\suzy) \cdot r_5^{\gunpowder}$.
Similarly, 
it is also easy to check~that
\begin{gather*}
\causalanswersetr\ref{cas:prg:dead}(\dead)
  \ \ = \ \
    \trigger(\suzy) \cdot
    r_5^{\gunpowder} \cdot
    r_4^{\bullett} \cdot
    r_3^{\impact} \cdot
    r_2^{\haemorrhage} \cdot
    r_1^{\dead}
\end{gather*}
Figure~\ref{fig:dead.causal.just} depicts the causal graph associated with the causal justification $\causalanswersetr\ref{cas:prg:dead}(\dead)$. \qed
\end{examplecont}

Next, we give an example of causal justifications for non-normal programs taken from~\cite{CabalarF16}:

\begin{example}\label{ex:causal.harvey}
Assume that Harvey throws a coin and only shoots when he gets tails.
This scenario can be modelled as the following logic program~\newprogram\label{prg:coin}:
\begin{IEEEeqnarray}{l C ? c C l}
r_1 &:& \dead &\lparrow& \shoot
	\label{eq:harvey.r1}
\\
r_2 &:& \shoot &\lparrow& \tails
	\label{eq:harvey.r2}
	\\
r_3 &:& \heads \vee \tails &\lparrow& \harvey
	\label{eq:harvey.r3}
\\	
&&\harvey
\end{IEEEeqnarray}
where $r_1$, $r_2$ and $r_3$ represent the labels associated with the corresponding rules.
Then, this logic program has two (standard) answer sets:
$\newanswerset\label{as:harvey.1} = \set{\harvey, \heads}$
and
$\newanswerset\label{as:harvey.2} = \set{\harvey, \tails, \shoot, \dead}$.
Similarly, this program also has two causal answer sets satisfying
\begin{gather*}
\begin{IEEEeqnarraybox}{l C l}
\causalanswersetr\ref{as:harvey.1}(\harvey) &=& \harvey
\\
\causalanswersetr\ref{as:harvey.1}(\heads)  &=& \harvey \cdotl r_3^{\heads}
\\
\causalanswersetr\ref{as:harvey.1}(\tails)  &=& 0
\\
\causalanswersetr\ref{as:harvey.1}(\shoot)  &=& 0
\\
\causalanswersetr\ref{as:harvey.1}(\dead)   &=& 0
\end{IEEEeqnarraybox}
\hspace{1.5cm}
\begin{IEEEeqnarraybox}{l C l}
\causalanswersetr\ref{as:harvey.2}(\harvey) &=& \harvey
\\
\causalanswersetr\ref{as:harvey.2}(\heads)  &=& 0
\\
\causalanswersetr\ref{as:harvey.2}(\tails)  &=& \harvey \cdotl r_3^{\tails}
\\
\causalanswersetr\ref{as:harvey.2}(\shoot)  &=& \harvey \cdotl r_3^{\tails} \cdotl r_2^{\shoot}
\\
\causalanswersetr\ref{as:harvey.2}(\dead)   &=& \harvey \cdotl r_3^{\tails} \cdotl r_2^{\shoot} \cdotl r_1^{\dead}
\end{IEEEeqnarraybox}
\end{gather*}
\CHANGED
Here, the $\causalanswersetr\ref{as:harvey.2}(\dead)$ represents the causal justification of $dead$ \wrt~$\answersetr\ref{as:harvey.2}$
while
$\causalanswersetr\ref{as:harvey.1}(\dead) = 0$
states that there is no causal justifications for $\dead$ \wrt~$\answersetr\ref{as:harvey.1}$. \qed
\END
\end{example}

\CHANGED
Example~\ref{ex:causal.harvey} illustrates that a causal answer set assigns the value $0$ (that is, the absence of a justification) to an atom  iff the atom is false in its corresponding standard answer set.
\END

It is also worth to note that, for normal logic programs, there is a one-to-one correspondence between the standard answer sets of a program and their causal answer sets.
For programs with disjunctive rules, there also exists a
\mbox{one-to-one} correspondence, but in this case it relates each standard answer set with a class of causal answer sets that represent the same truth assignments, but different justifications
(see Example~\ref{ex:causal.non.one-to-one} below).
Furthermore, in the case of disjunctive rules, the superindex
of a disjunctive rule's label in the causal answer set
 indicates the disjunct that has been effectively applied.
For instance, in Example~\ref{ex:causal.harvey},
term $r_3^{\tails}$ points out that the disjunct $\tails$ in $r_3$ has been effectively applied.
In the case of normal rules, the superindex is somehow superfluous, as it is fully determined by the rule, and could easily be omitted as in~\cite{CabalarF16}.
Nevertheless, we decide to keep them to ease the comparison with the other justification approaches,
whose vertices are literals.

\begin{example}\label{ex:causal.non.one-to-one}
Consider a program~\newprogram\label{prg:causal.non.one-to-one} consisting of the following rules
\begin{gather*}
r_1 : \ \ a \vee b \lparrow
\hspace{2cm}
r_2 : \ \ a  \lparrow b
\hspace{2cm}
r_3 : \ \ b \lparrow a
\end{gather*}
which has a unique (standard) answer set~$\newanswerset\label{as:prg:causal.non.one-to-one} = \set{a, b}$,
but two causal ones that satisfy:
\vspace{-0.25cm}
\begin{gather*}
\begin{IEEEeqnarraybox}[][t]{l ; C ; l}
\causalanswersetr\ref{as:prg:causal.non.one-to-one}(a) &=& r_1^a
\\
\causalanswersetr\ref{as:prg:causal.non.one-to-one}(b) &=& r_1^a \cdotl r_3^b
\end{IEEEeqnarraybox}
\hspace{3cm}
\begin{IEEEeqnarraybox}[][t]{l ; C ; l}
\causalanswersetr\ref{as:prg:causal.non.one-to-one}'(a) &=& r_1^b \cdotl r_2^a
\\
\causalanswersetr\ref{as:prg:causal.non.one-to-one}'(b) &=& r_1^b
\end{IEEEeqnarraybox}
\end{gather*}
As we can see, the true atoms in both models,
$\Atoms(\causalanswersetr\ref{as:prg:causal.non.one-to-one})
    = \Atoms(\causalanswersetr\ref{as:prg:causal.non.one-to-one}')
    = \set{a,b}$,
coincide with the unique (standard) answer set $\answersetr\ref{as:prg:causal.non.one-to-one}$, but their \emph{justifications differ}.
In $\causalanswersetr\ref{as:prg:causal.non.one-to-one}$, atom $a$ is a (non-deterministic) effect of the disjunction $r_1$, while $b$ is derived from $a$ through $r_3$.
Analogously, $\causalanswersetr\ref{as:prg:causal.non.one-to-one}'$ makes $b$ true because of $r_1$ and then obtains $a$ from $b$ through $r_2$.
\CHANGED
It is interesting to point out that $\causalanswersetr\ref{as:prg:causal.non.one-to-one}''$
with
\begin{gather*}
\begin{IEEEeqnarraybox}[][t]{l ; C ; l}
\causalanswersetr\ref{as:prg:causal.non.one-to-one}''(a) &=& r_1^a + r_1^b \cdotl r_2^a
\\
\causalanswersetr\ref{as:prg:causal.non.one-to-one}''(b) &=& r_1^b + r_1^a \cdotl r_3^b
\end{IEEEeqnarraybox}
\end{gather*}
is also a model of the program, but not a $\sqsubseteq$-minimal one because we have $\causalanswersetr\ref{as:prg:causal.non.one-to-one} \sqsubset \causalanswersetr\ref{as:prg:causal.non.one-to-one}''$.
Intuitively, $\causalanswersetr\ref{as:prg:causal.non.one-to-one}''$ would represent a scenario in which both $a$ and $b$ are justified by rule $r_1$, which does not fit the intuitive understanding that rule $r_1$ can only justify one of its head atoms.
\qed
\end{example}

\CHANGED
Let us also recall that, for normal programs,~\cite{CabalarFF2014} defining causal answer sets as $\leq$-minimal models instead of $\sqsubseteq$-minimal ones. These two definitions agree for normal logic programs~\cite{CabalarF16} with the former being preferred for its simplicity.\footnote{This definition is also used in Section~\ref{sec:enablers-inhibitors} where the syntax is restricted to normal programs.} 
On the other hand, for disjunctive programs, there are \mbox{$\leq$-minimal} models that do not correspond to any standard stable model, thus the need for the latter. This is illustrated by the following example.

\begin{example}\label{ex:disj.semantics.dif}
Let \newprogram\label{prg:naive2} be the following logic program:
\begin{gather*}
r_1 : \ \ \heads \vee \tails
\hspace{2cm} \heads
\end{gather*}
which has two $\leq$-minimal models, one in which
$\causalanswersetr\ref{prg:naive2}(\heads) = \heads + r_1^{\heads}$
and
$\causalanswersetr\ref{prg:naive2}(\tails) = 0$,
plus another in which
$\causalanswersetr\ref{prg:naive2}'(\heads) = \heads$
and
$\causalanswersetr\ref{prg:naive2}'(\tails) = r_1^{\tails}$.
However, only the former is a $\sqsubseteq$-minimal one.
Note that this corresponds to the set of atoms $\Atoms(\causalanswersetr\ref{prg:naive2}) = \set{\heads }$
which is the unique standard answer set of the program.\qed
\end{example}
\END

The following example illustrates the fact that `$*$' is used to 
\CHANGED
represent joint causation, or in other words, that two or more atoms are needed to justify the conclusion of some rule.
\END

\begin{example}\label{ex:causal.prod}
\label{ex:causal.sum}
Consider the logic program~$\newprogram\label{prg:causal.prod}$ consisting of the following rules:
\begin{gather*}
r_1 : \ p \lparrow q
\hspace{1.25cm}
r_2 : \ q \lparrow r \wedge s
\hspace{1.25cm}
r
\hspace{1.25cm}
s
\end{gather*}
This program has a unique causal answer set~$\newcausalanswerset\label{cas:prg:causal.prod}$ that satisfies:
\begin{gather*}
\begin{IEEEeqnarraybox}{ l C l }
\causalanswersetr\ref{cas:prg:causal.prod}(p) &=& (r * s) \cdotl r_2^q \cdotl r_1^p
\\
\causalanswersetr\ref{cas:prg:causal.prod}(q) &=& (r * s) \cdotl r_2^q
\end{IEEEeqnarraybox}
\hspace{2cm}
\begin{IEEEeqnarraybox}{ l C l }
\causalanswersetr\ref{cas:prg:causal.prod}(r) &=& r
\\
\causalanswersetr\ref{cas:prg:causal.prod}(s) &=& s
\end{IEEEeqnarraybox}
\end{gather*}
As shown in Observation~\ref{obs:causa.fact},
we have
$\causalanswersetr\ref{cas:prg:causal.prod}(r) \geq r \cdotl r = r$.
Then, the value of
$\causalanswersetr\ref{cas:prg:causal.prod}(r)$
follows from the fact that causal answer sets are $\leq$-minimal models.
Similar \changed{reasoning applies} for the atom $s$.
Furthermore, from Definition~\ref{def:causal.model.positive}, it follows that
$\causalanswersetr\ref{cas:prg:causal.prod}(q) \geq (r * s)\cdotl r_2^q$
and, by minimality, that
$\causalanswersetr\ref{cas:prg:causal.prod}(q) = (r * s)\cdotl r_2^q$.
In a similar way, we obtain for $p$ that
$\causalanswersetr\ref{cas:prg:causal.prod}(p)
  = \causalanswersetr\ref{cas:prg:causal.prod}(q) \cdotl r_1^p
  = (r * s) \cdotl r_2^q \cdotl r_1^p$.
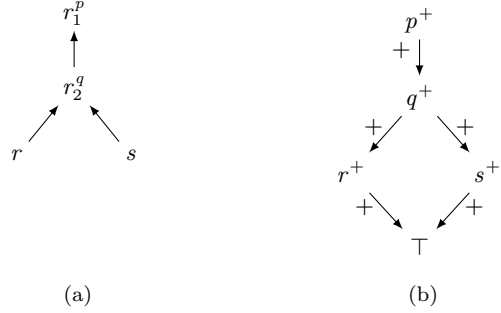
\begin{figure}\centering
\subfloat[]{%
\begin{tikzpicture}[tikzpict]
    \matrix[row sep=0.5cm,column sep=0.3cm,ampersand replacement=\&] {
      \&
      \node (p) {$r_1^p$};
      \\
      \&
      \node (q) {$r_2^q$};
      \\
      \node (r) {$r$};
      \&\&
      \node (s) {$s$};
      \\
      \node (a) {$ $};
      \\
      \\
     };
    \draw [<-] (p) to (q);
    \draw [<-] (q) to (r);
    \draw [<-] (q) to (s);
\end{tikzpicture}%
\label{fig:causal.prod}
}
\hspace{2cm}
\subfloat[]{%
\begin{tikzpicture}[tikzpict]
    \matrix[row sep=0.5cm,column sep=0.3cm,ampersand replacement=\&] {
      \&
      \node (p) {$p^{+}$};
      \\
      \&
      \node (q) {$q^{+}$};
      \\
      \node (r) {$r^{+}$};
      \&\&
      \node (s) {$s^{+}$};
      \\
      \&
      \node (top) {$\top$};
      \\
     };
    \draw [->] (p) to node[pos=0.3,left]{$+$}  (q);
    \draw [->] (q) to node[pos=0.3,left]{$+$}  (r);
    \draw [->] (q) to node[pos=0.3,right]{$+$}  (s);
    \draw [->] (r) to node[pos=0.4,left]{$+$} (top);
    \draw [->] (s) to node[pos=0.4,right]{$+$} (top);
\end{tikzpicture}%
\label{fig:causal.prod.off-line}
}
\caption{Causal graph and off-line justification of $p$ w.r.t. the unique answer set of $\programr\ref{prg:causal.prod}$ (see Examples~\ref{ex:causal.prod} and~\ref{ex:causal.prod}).}
\end{figure}
Figure~\ref{fig:causal.prod} depicts the causal graph associated with
$\causalanswersetr\ref{cas:prg:causal.prod}(p)$.
Note that product~`$*$' is translated in this causal graph (Figure~\ref{fig:causal.prod}) as two incoming edges to the vertex $r_2^q$.
\CHANGED
The causal graph associated with some causal value can be easily constructed by rewriting the causal value in graph normal form and representing each term of the form $v_1 \cdotl v_2$ with an edge from $v_1$ to $v_2$.
In particular, we can obtain the causal graph in Figure~\ref{fig:causal.prod} by rewriting $(r * s) \cdotl r_2^q \cdotl r_1^p$ in graph normal form as follows:
\begin{IEEEeqnarray*}{l ?C? l *x+}
(r * s) \cdotl r_2^q \cdotl r_1^p 
  &=&  r \cdotl r_2^q \cdotl r_1^p * s \cdotl r_2^q \cdotl r_1^p
\\
  &=&  r \cdotl r_2^q * r_2^q \cdotl r_1^p * s \cdotl r_2^q * r_2^q \cdotl r_1^p
\\
  &=&  r \cdotl r_2^q * r_2^q \cdotl r_1^p * s \cdotl r_2^q
\end{IEEEeqnarray*}
Then, the three edges of the causal graph in Figure~\ref{fig:causal.prod} correspond to the three subterms of the form $v_1 \cdotl v_2$ (that is, $r \cdotl r_2^q$,  $r_2^q \cdotl r_1^p$ and $s \cdotl r_2^q$) in the above causal term.
\END
\changed{For comparison,} Figure~\ref{fig:causal.prod.off-line} depicts the off-line justification of $p^{+}$.
It is easy to see that
this particular off-line justification can be obtained from the causal graph by replacing each vertex $r_i^\rA$ by $\rA$,
reversing edges, adding the label `$+$' to each vertex and resulting edge and adding edges of the form $(\rA,\top,+)$ for each resulting sink~$\rA$.
\qed
\end{example}

Next, we illustrate that `$+$' is used to separate alternative causal justifications
and the importance of addition distributivity to obtain such behaviour.


\begin{example}
Consider the logic program~$\newprogram\label{prg:causal.sum}$ consisting of the following rules:
\begin{gather*}
r_1 : \ p \lparrow q
\hspace{1.25cm}
r_2 : \ q \lparrow r
\hspace{1.25cm}
r_3 : \ q \lparrow s
\hspace{1.25cm}
r
\hspace{1.25cm}
s
\end{gather*}
This program has a unique causal answer set~$\newcausalanswerset\label{cas:prg:positive.sum}$ that satisfies:
\begin{gather*}
\begin{IEEEeqnarraybox}{ l C l }
\causalanswersetr\ref{cas:prg:positive.sum}(p) &=& r \cdotl r_2^q \cdotl r_1^p + s \cdotl r_3^q \cdotl r_1^p
\\
\causalanswersetr\ref{cas:prg:positive.sum}(q) &=& r \cdotl r_2^q + s \cdotl r_3^q
\end{IEEEeqnarraybox}
\hspace{2cm}
\begin{IEEEeqnarraybox}{ l C l }
\causalanswersetr\ref{cas:prg:positive.sum}(r) &=& r
\\
\causalanswersetr\ref{cas:prg:positive.sum}(s) &=& s
\end{IEEEeqnarraybox}
\end{gather*}
As in Example~\ref{ex:causal.prod},
we have that
$\causalanswersetr\ref{cas:prg:positive.sum}(r) = r$
and
$\causalanswersetr\ref{cas:prg:positive.sum}(s) = s$.
Furthermore, in this case, Definition~\ref{def:causal.model.positive} implies
$\causalanswersetr\ref{cas:prg:positive.sum}(q) \geq r \cdotl r_2^q$
and
$\causalanswersetr\ref{cas:prg:positive.sum}(q) \geq s \cdotl r_3^q$.
Then, the value of
$\causalanswersetr\ref{cas:prg:positive.sum}(q)$
follows from the fact that causal answer sets are $\leq$-minimal models and the fact that `$+$' is the least upper bound of the $\leq$ relation.
Finally,
$\causalanswersetr\ref{cas:prg:positive.sum}(p)
	= \causalanswersetr\ref{cas:prg:positive.sum}(q) \cdotl r_1^p
	= (r \cdotl r_2^q + s \cdotl r_3^q) \cdotl r_1^p$ follows in similar way.
The value of $\causalanswersetr\ref{cas:prg:positive.sum}(p)$ shown above is the disjunctive normal form of this term, and it is obtained by applying addition distributivity.
\changed{
Here, both addends in $\causalanswersetr\ref{cas:prg:positive.sum}(p)$, that is
$r \cdotl r_2^q \cdotl r_1^p$ and $s \cdotl r_3^q \cdotl r_1^p$, are causal justifications of $p$ \wrt~the unique answer set of the program.} \qed
\end{example}

\subsubsection{Causal Semantics for Programs with Negation-as-Failure}

\CHANGED
We now extend the causal answer set semantics to logic programs with negation-as-failure.
For this, the \emph{closed world assumption} is directly translated into the language of justifications, assuming that everything that has no justification is false by default.
Accordingly, negative literals are assumed to hold by default, without requiring further justification.
This contrasts with the previously presented off-line and LABAS justifications, which further explain why negative literals hold.
The next section shows how causal justifications can be extended in order to provide such information.
Let us start with an example motivating why omitting the justification of negative literals, thus treating them as defaults, may provide intuitive explanation in some scenarios.
\END

\begin{examplecont}{ex:dead}\label{ex:dead.exceptions}
Consider a variation of the scenario of Example~\ref{ex:dead} in which
shooting the victim may fail in several ways: the victim may be wearing a $\bulletproof$ vest, the gunpowder may be $\wet$, etc.
This is an instance of the well-known \emph{qualification problem}~\cite{McC77}:
any comprehensive knowledge base for general commonsense reasoning may contain hundreds or thousands of exceptions to any rule,
which may also be impossible to list in advance.
As usual in answer set programming, this problem can be solved by adding abnormality predicates to the body of all rules.
In particular, rules~(\ref{prg:dead.r1}-\ref{prg:dead.r6}) are rewritten as follows:
\begin{IEEEeqnarray}{l C ' l C l}
r_1 &:& \dead &\lparrow&  \haemorrhage \wedge \Not \ab_1
\label{prg:dead.r1b}
\\
r_2 &:& \haemorrhage &\lparrow&  \impact \wedge \Not \ab_2
\label{prg:dead.r2b}
\\
r_3 &:& \impact &\lparrow&  \bullett \wedge \Not \ab_2
\label{prg:dead.r3b}
\\
r_4 &:& \bullett &\lparrow&  \gunpowder \wedge \Not \ab_3
\label{prg:dead.r4b}
\\
r_5 &:& \gunpowder &\lparrow&  \trigger(\suzy) \wedge \Not \ab_4
\label{prg:dead.r5b}
\\
&& \trigger(\suzy)
\label{prg:dead.r6b}
\end{IEEEeqnarray}
Then, exceptions can be added in an \emph{elaboration tolerant} manner by adding new rules as follows:
\begin{IEEEeqnarray}{l C ? r C l}
r_6 &:& \ab_2 &\lparrow& \bulletproof
  \label{eq:ab.bulletproof}
\\
r_7 &:& \ab_4 &\lparrow& \wet
  \label{eq:ab.wet}
\end{IEEEeqnarray}
Let \newprogram\label{prg:dead.exceptions} be the program containing rules (\ref{prg:dead.r1b}-\ref{eq:ab.wet}).\qed
\end{examplecont}

For justifications, Example~\ref{ex:dead.exceptions} sets out a new challenge:
a justification for the lack of all exceptions may be much bigger than the justification for the conclusion without exceptions.
Furthermore, from a causal perspective, saying that the lack of an exception is part of a cause (e.g., for $\dead$) may seem rather counterintuitive.
It is not the case that the victim is $dead$ because the gunpowder was not $\wet$,
or because the victim was not wearing a $\bulletproof$ vest,
or whatever other possible exception might be added in the future.

This is a well-known problem in the causal literature~\cite{maudlin2004,hall2007structural,Halpern08a,hitchcock2009cause}:
in particular, \citeN{hitchcock2009cause} provides an extended discussion with several examples showing how people ordinarily understand causes as deviations from a normal or default behaviour.
In this sense, by understanding falsity of exceptions as the \emph{default situation}, we obtain that, when no exception is true with respect to the causal answer set,
the causal justifications for $\dead$ in programs~\programr\ref{prg:dead} and~\programr\ref{prg:dead.exceptions} are the same.
This interpretation of negation-as-failure can be captured by the following definitions:

\begin{definition}[Causal Program Reduct]\label{def:reduct}
The \emph{\changed{(causal)} reduct} of an extended program $P$ with respect to a \changed{causal} interpretation $I$, in symbols~$P^I$, is the result of
\begin{enumerate}[ topsep=2pt, leftmargin=15pt ]
\item removing all rules such that $I(\rB) \neq 0$ for some $\rB \in \bodyn{\R}$,
\item removing all the negative literals from the remaining rules.\qed
\end{enumerate}
\end{definition}

\begin{definition}[Causal Answer Set]\label{def:causal.smodel}
We say that a \changed{causal} interpretation $I$ is a \emph{causal answer set} of an extended program~$P$ iff $I$~is \changed{a causal answer set} of the positive program~$P^I$.\qed
\end{definition}

\begin{examplecont}{ex:dead.exceptions}\label{ex:dead.exceptions2}
Let
$\newcausalanswerset\label{cas:prg:dead.exceptions}$ be an interpretation
such that
$\causalanswersetr\ref{cas:prg:dead.exceptions}(A) = \causalanswersetr\ref{cas:prg:dead}(A)$ for all literals $A$ occurring in the program~\programr\ref{prg:dead}, and $\causalanswersetr\ref{cas:prg:dead.exceptions}(A) = 0$ for all other literals occurring in program~\programr\ref{prg:dead.exceptions}.
Then, it is easy to see that
$\programr\ref{prg:dead.exceptions}^{\causalanswersetr\ref{cas:prg:dead.exceptions}} = \programr\ref{prg:dead} \cup \set{ \eqref{eq:ab.bulletproof}, \eqref{eq:ab.wet}}$
and, thus, that
$\causalanswersetr\ref{cas:prg:dead.exceptions}$ is the \mbox{$\sqsubseteq$-minimal} model of $\programr\ref{prg:dead.exceptions}^{\causalanswersetr\ref{cas:prg:dead.exceptions}}$.
Note that $0$ is the bottom value and there are no rules assigning greater values to $\bulletproof$ or $\wet$ and, thus, neither to any of the~$\ab_i$.
That is, the unique answer sets of programs~$\programr\ref{prg:dead}$ and~$\programr\ref{prg:dead.exceptions}$ agree on the causal values assigned to all literals they have in common.\qed
\end{examplecont}

\changed{We note} that the behaviour of causal justifications in Example~\ref{ex:dead.exceptions} is similar to LABAS justifications in the sense that, in the latter, the defaults are not further explained either.
This happens because there are no derivations for any abnormality atom~$\ab_i$.
On the other hand, 
if exceptions could be derived, 
then the behaviour would be different.
For instance, let $ \newprogram\label{prg:dead.exceptions2}$ be the program obtained from 
$\programr\ref{prg:dead.exceptions}$ by replacing rule~\eqref{eq:ab.bulletproof} by the following two rules
\begin{IEEEeqnarray}{l C ? r C l}
r_6 &:& \ab_2 &\lparrow& \bulletproof \wedge \Not \ab_5
  \label{eq:ab.bulletproof2}
\\
r_8 &:& \ab_5 &\lparrow& \damaged
  \label{eq:damaged}
\end{IEEEeqnarray}
plus the facts $\bulletproof$ and $\damaged$.
In this case, $\ab_2$ is still false, so the causal justification of $\dead$ remains the same.
However, now there is a derivation for $\ab_2$ which is `attacked' by $\damaged$, so a LABAS justification further justifies the falsity of exception $\ab_2$ in terms of $\damaged$.
The following example illustrates some similarities and differences between causal and off-line justifications.

\begin{examplecontpage}{ex:positive}\label{ex:causal.positive}
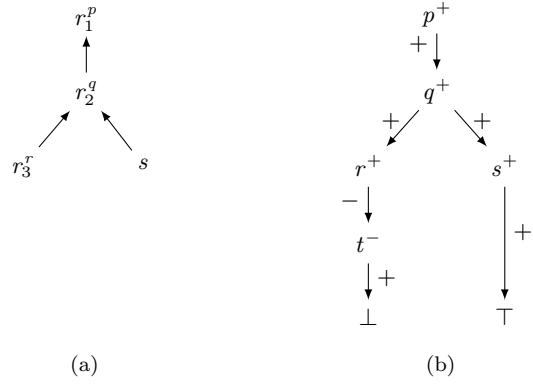
\begin{figure}\centering
\subfloat[]{%
\begin{tikzpicture}[tikzpict]
    \matrix[row sep=0.5cm,column sep=0.3cm,ampersand replacement=\&] {
      \&
      \node (p) {$r_1^p$};
      \\
      \&
      \node (q) {$r_2^q$};
      \\
      \node (r) {$r_3^r$};
      \&\&
      \node (s) {$s$};
      \\
      \node (a) {$ $};
      \\[15pt]
      \node (a) {$ $};
      \\
     };
    \draw [<-] (p) to (q);
    \draw [<-] (q) to (r);
    \draw [<-] (q) to (s);
\end{tikzpicture}%
\label{fig:causal.positive}
}
\hspace{2cm}
\subfloat[]{%
\begin{tikzpicture}[tikzpict]
    \matrix[row sep=0.5cm,column sep=0.3cm,ampersand replacement=\&] {
      \&
      \node (p) {$p^{+}$};
      \\
      \&
      \node (q) {$q^{+}$};
      \\
      \node (r) {$r^{+}$};
      \&\&
      \node (s) {$s^{+}$};
      \\
      \node (t) {$t^{-}$};
      \\
      \node (bot) {$\bot$};
      \&\&
      \node (top) {$\top$};
      \\
     };
    \draw [->] (p) to node[pos=0.3,left]{$+$}  (q);
    \draw [->] (q) to node[pos=0.3,left]{$+$}  (r);
    \draw [->] (q) to node[pos=0.3,right]{$+$}  (s);
    \draw [->] (r) to node[pos=0.4,left]{$-$} (t);
    \draw [->] (t) to node[pos=0.4,right]{$+$} (bot);
    \draw [->] (s) to node[pos=0.4,right]{$+$} (top);
\end{tikzpicture}%
\label{fig:causal.positive.off-line}
}
\caption{Causal graph and off-line justification of $p$ w.r.t. the unique answer set of $\programr\ref{prg:positive}$ (see Example~\ref{ex:positive} on page~\pageref{ex:positive} and Example~\ref{ex:causal.positive}).}
\end{figure}
Let us now consider the program~$\programr\ref{prg:positive}$ and the following labelling of its rules
\begin{gather*}
r_1 : \ p \lparrow q
\hspace{1.5cm}
r_2 : \ q \lparrow r \wedge s
\hspace{1.5cm}
r_3 : \ r \lparrow \Not t
\hspace{1.5cm}
s
\end{gather*}
Then, the unique causal answer set $\newcausalanswerset\label{cas:prg:positive}$ of program~$\programr\ref{prg:positive}$ satisfies:
\begin{gather*}
\begin{IEEEeqnarraybox}{ l C l }
\causalanswersetr\ref{cas:prg:positive}(p) &=& (r_3^r * s) \cdotl r_2^q \cdotl r_1^p
\\
\causalanswersetr\ref{cas:prg:positive}(q) &=& (r_3^r * s) \cdotl r_2^q
\\
\causalanswersetr\ref{cas:prg:positive}(r) &=& r_3^r
\end{IEEEeqnarraybox}
\hspace{2cm}
\begin{IEEEeqnarraybox}{ l C l }
\causalanswersetr\ref{cas:prg:positive}(s) &=& s
\\
\causalanswersetr\ref{cas:prg:positive}(t) &=& 0
\\
\end{IEEEeqnarraybox}
\end{gather*}
Figure~\ref{fig:causal.positive} depicts the \changed{causal} graph associated with
$\causalanswersetr\ref{cas:prg:positive}(p)$,
while Figure~\ref{fig:causal.positive.off-line} depicts the off-line justification of $p^{+}$ for the sake of comparison.
Note that the causal graph can be obtained from the off-line justification by removing the $\bot$, $\top$ and all negatively labelled vertices plus all the edges connected to these vertices (where the edges are inverted).
Note that the only change in the causal justification of~$p$ in this example with respect to that in program~$\programr\ref{prg:causal.prod}$ is the renaming of the node~$r$ as~$r_3^r$,
while off-line justifications of the two programs further differ in the subgraph rooted in~$r^{+}$.
\qed
\end{examplecontpage}

\begin{example}\label{ex:light}
\CHANGED
Let us consider a scenario where there is a light bulb that turns $\on$ whenever the switches $a$ and $b$
are pushed at the same time, and $\off$ whenever the switches $c$ and~$d$ are pushed at the same time.
Assume also that the light is currently $\off$ and the switches $a$ and $b$ are pushed (situation $0$). 
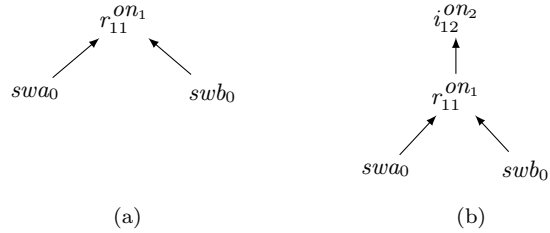
\begin{figure}[ht]\centering
\subfloat[]{%
\begin{tikzpicture}[tikzpict]
    \matrix[row sep=0.5cm,column sep=0.3cm,ampersand replacement=\&] {
      \&
      \node (on1) {$r_{11}^{\on_1}$};
      \\
      \node (swa0) {$\swa_0$};
      \&\&
      \node (swb0) {$\swb_0$};
      \\[15pt]
      \\
     };
    \draw [<-] (on1) to (swa0);
    \draw [<-] (on1) to (swb0);
\end{tikzpicture}%
\label{fig:on1:causal}
}
\hspace{1cm}
\subfloat[]{%
\begin{tikzpicture}[tikzpict]
    \matrix[row sep=0.5cm,column sep=0.05cm,ampersand replacement=\&] {
      \&\node (on2) {$i_{12}^{\on_2}$};
      \\
      \&
      \node (on1) {$r_{11}^{\on_1}$};
      \\
      \node (swa0) {$\swa_0$};
      \&\&
      \node (swb0) {$\swb_0$};
      \&
      \node (aa) {$ $};\&
      \\
     };
    \draw [<-] (on2) to  (on1);
    \draw [<-] (on1) to  (swa0);
    \draw [<-] (on1) to  (swb0);
\end{tikzpicture}%
\label{fig:on2:causal}
}
\caption{Causal justifications of the truth of $\on_1$ and $\on_2$.}
\end{figure}
This problem can be easily formalised as a logic program~$\newprogram\label{prg:light}$ consisting of rules\footnote{For the sake of simplicity, we avoid introducing a first order language here and indirectly use the propositional logic program that is produced through grounding.}:
\begin{gather}
r_{1t+1} : \ \on_{t+1}  \lparrow \swa_t \wedge \swb_t
\hspace{1.5cm}
r_{2t+1} : \ \off_{t+1} \lparrow \swc_t \wedge \swd_t
  \label{ex:light.off-line.effect.rules}
\end{gather}
for $t \geq 0$,
plus the facts $\off_{0}$, $\swa_0$ and $\swb_0$.
 As usual, inertia is represented by the following pair of rules:
\begin{IEEEeqnarray}{l C l}
i_{1t+1} : \ \on_{t+1}  &\lparrow& \on_{t} \wedge \Not \off_{t+1}\\
i_{2t+1} : \ \off_{t+1} &\lparrow& \off_{t} \wedge \Not \on_{t+1}
  \label{ex:light.off-line.inertia.rules}
\end{IEEEeqnarray}
for $t \geq 0$.
\CHANGED
We also have an integrity constraint 
\begin{gather}
\lparrow \on_t \wedge \off_t
\end{gather}
ensuring that $\on$ and $\off$ cannot hold at the same time.
\END
This program has a complete well-founded model and, thus, a unique answer set, in which $\on_t$ holds for every time $t > 0$.
Figures~\ref{fig:on1:causal} and~\ref{fig:on2:causal} respectively depict the causal justifications of $\on_1$
and $\on_2$ \wrt\ that answer set.\qed
\end{example}

As illustrated by the above example, understanding negation-as-failure as a default (which does not need to be further explained), allows that causal justifications are `preserved' by inertia in the following sense: at any situation $t+1$ if nothing happens, then the \changed{causal} justification of $\on_{t+1}$ can be obtained by adding to the \changed{causal} justification of $\on_t$, an edge from
$i_{1t}^{\on_{t}}$ to $i_{1t+1}^{\on_{t+1}}$.
True persistence of justifications, that is, exactly the same justification preserved by inertia, can be obtained by selecting some rule labels, in this case the labels associated with inertia
($i_{1t+1}$ and $i_{2t+1}$), as not forming part of the \changed{causal} justifications, and thus of the causal graphs.
In such case, the \changed{causal graph} for $\on_{t}$ at any situation $t$ would be the one depicted in Figure~\ref{fig:on1:causal}.
\changed{In contrast, the number} of off-line and LABAS justifications grows exponentially with the number of situations in which nothing happens.
This will be discussed in more detail in Section~\ref{sec:discussion}.

\subsubsection{Explaining Negative Literals in Causal Justifications}
\label{sec:enablers-inhibitors}

As we have seen, one major difference between causal justifications and the two previous approaches, off-line and LABAS justifications, is the way in which all negative literals 
that are true \wrt\ the answer set in question
are assumed to hold by default, so they do not need further justification.
This behaviour allows to get an important reduction in the number of justifications
in examples that involve exceptions or defaults like inertia (as \changed{was illustrated} in Example~\ref{ex:light}).
On the other hand,
there are scenarios in which justifications for negative literals are valuable.\footnote{A more detailed elaboration of this argument can be found in Section~\ref{sec:discussion}.}
Consider, for instance, the following example from~\cite{CabalarF2017}:

\begin{example}\label{ex:bond}
A drug~$d$ in James Bond's drink causes his paralysis $p$ provided that he was not given an antidote $a$ that day. We know that Bond's enemy, Dr. No, poured the drug
and that Bond is daily administered an antidote by the MI6, unless it is a holiday~$h$:
\begin{IEEEeqnarray}{l C ? l C l}
r_1 &:& p & \lparrow & d,\, \Not a 
\label{eq:bond.r1}
\\
r_2 &:& a &\lparrow& \Not h
\label{eq:bond.r3}
\\
&& d 
\label{eq:bond.d}
\end{IEEEeqnarray}
Then, $\set{a,d}$ is the unique answer set of the program consisting of rules
(\ref{eq:bond.r1}-\ref{eq:bond.d}).
Since $p$ is false with respect this answer set, the causal value associated to it is~$0$, that is, it has its value by default without further explanation.
On the other hand,
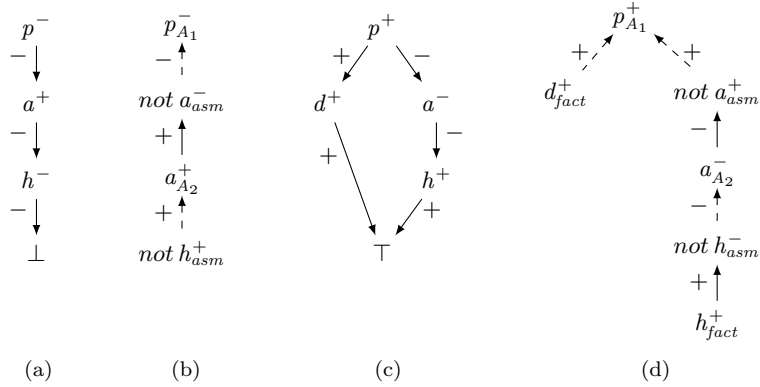
\begin{figure}\centering
\subfloat[]{%
\begin{tikzpicture}[tikzpict]
    \matrix[row sep=0.5cm,column sep=0.1cm,ampersand replacement=\&] {
      \node (p) {$p^{-}$};
      \\
      \node (a) {$a^{+}$};
      \\
      \node (h) {$h^{-}$};
      \\
      \node (bot) {$\bot$};
      \\[12pt]
      \\
     };
    \draw [->] (p) to node[pos=0.3,left]{$-$}  (a);
    \draw [->] (a) to node[pos=0.3,left]{$-$}  (h);
    \draw [->] (h) to node[pos=0.3,left]{$-$}  (bot);
\end{tikzpicture}%
\label{fig:bond1.off-line}
}
\hspace{0.5cm}
\subfloat[]{%
\begin{tikzpicture}[tikzpict]
    \matrix[row sep=0.5cm,column sep=0.1cm,ampersand replacement=\&] {
      \node (p) {$p^{-}_{A_1}$};
      \\
      \node (a) {$\Not a^{-}_{asm}$};
      \\
      \node (a2) {$a^{+}_{A_2}$};
      \\
      \node (h) {$\Not h^{+}_{asm}$};
      \\[12pt]
      \\
     };
    \draw [<-,dashed] (p) to node[left]{$-$}  (a);
    \draw [<-] (a) to node[left]{$+$}  (a2);
    \draw [<-,dashed] (a2) to node[left]{$+$}  (h);
\end{tikzpicture}%
\label{fig:bond1.labas}
}
\hspace{0.5cm}
\subfloat[]{%
\begin{tikzpicture}[tikzpict]
    \matrix[row sep=0.5cm,column sep=0.1cm,ampersand replacement=\&] {
      \&\node (p) {$p^{+}$};
      \\
      \node (d) {$d^{+}$};
      \&\&
      \node (a) {$a^{-}$};
      \\
      \&\&\node (h) {$h^{+}$};
      \\
      \&\node (top) {$\top$};
      \\[12pt]
      \\
     };
    \draw [->] (p) to node[pos=0.3,left]{$+$}  (d);
    \draw [->] (p) to node[pos=0.3,right]{$-$}  (a);
    \draw [->] (a) to node[pos=0.3,right]{$-$}  (h);
    \draw [->] (d) to node[pos=0.3,left]{$+$}  (top);
    \draw [->] (h) to node[pos=0.3,right]{$+$}  (top);
\end{tikzpicture}%
\label{fig:bond2.off-line}
}
\hspace{0.5cm}
\subfloat[]{%
\begin{tikzpicture}[tikzpict]
    \matrix[row sep=0.5cm,column sep=0.1cm,ampersand replacement=\&] {
      \&\node (p) {$p^{+}_{A_1}$};
      \\
      \node (d) {$d^{+}_{fact}$};
      \&\&
      \node (a) {$\Not a^{+}_{asm}$};
      \\
      \&\&\node (a2) {$a^{-}_{A_2}$};
      \\
      \&\&\node (h) {$\Not h^{-}_{asm}$};
      \\
      \&\&\node (h2) {$h^{+}_{fact}$};
      \\
     };
    \draw [<-,dashed] (p) to node[left]{$+$}  (d);
    \draw [<-,dashed] (p) to node[right]{$+$}  (a);
    \draw [<-] (a) to node[left]{$-$}  (a2);
    \draw [<-,dashed] (a2) to node[left]{$-$}  (h);
    \draw [<-] (h) to node[left]{$+$}  (h2);
\end{tikzpicture}%
\label{fig:bond2.labas}
}
\caption{Off-line and LABAS justifications of $p$ w.r.t. the unique answer set of~Example~\ref{ex:bond}.}\label{fig:bond}
\end{figure}
Figures~\ref{fig:bond1.off-line} and~\ref{fig:bond1.labas} respectively
depict the off-line and LABAS justifications explaining that $p$ does not hold because $a$ is somehow preventing it.
The extension of causal justifications, presented in this section, associates the causal value
$(\sneg r_2^a * d) \cdotl r_1^p$ to $p$ in this scenario, pointing out that rule~$r_2$ (and, thus $a$) is what prevents $p$ from becoming true.
A causal reading of this expression is that ``$a$ has prevented (through rule~$r_2$) $d$ to cause $p$ (through rule~$r_1$)''
or, equivalently,
``if it was not for rule~$r_2$ (implying $a$), $d$ would cause $p$ through rule~$r_1$''.
Suppose now that it is a holiday, so fact $h$ is added to the program
\eqref{eq:bond.r1}-\eqref{eq:bond.r3}. Then, $a$ is itself disabled and $d$
is free to cause~$p$.
The causal justification of $p$ in this case is
$d \cdotl r_1^p$ (which corresponds to the graph with a single edge from $d$ to $r_1^p$),
which reflects the fact that $d$ has caused~$p$, but without keeping any record about the fact that $h$ has also been necessary for this to happen.
On the other hand, we can see in Figures~\ref{fig:bond2.off-line} and~\ref{fig:bond2.labas} that both off-line and LABAS justifications keep track of this dependency.
The extended causal justifications also keep track of this dependency and associate the casual value
$(\sneg\sneg h  * d ) \cdotl r_1^p   +  (\sneg r_2^a  * d ) \cdotl r_1^p$
with $p$.
Here, the first addend can be informally read as ``$h$ has allowed $d$ to cause~$p$ (through rule~$r_1$).''
Double negation in front of $h$ is introduced to distinguish between the philosophically distinct concepts%
\footnote{A productive cause is an event connected to its effect by a causal chain as explained at the beginning of Section~\ref{sec:cg}.
For a thorough philosophical explanation about the differences between productive causes and contingently counterfactual dependencies we refer to~\cite{hall2004,hall2007structural}.}
of \emph{productive cause} (in this case~$d$) and other contingently counterfactual dependencies (in this case~$h$), though this distinction is not of particular relevance in the context of justifications.
As before, the second addend can be informally read as
``if it was not for rule~$r_2$ (implying $a$), $d$ would cause $p$ through rule~$r_1$'' (even without the presence of~$h$).
\qed
\end{example}

In order to introduce information about negative literals in causal justifications,
\citeNS{CabalarF2017} extended causal justifications with a negation inspired by why-not provenance justifications~(see Section~\ref{sec:why-not}; \citeNP{DamasioAA2013}).
We now review this extension, starting with the introduction of negation in causal terms as follows:

\begin{definition}[Extended Causal Terms]
\label{def:eterm}
Given a set of atoms $\at$ and a set of labels $\lb$, an \emph{extended \changed{(causal)} term} (\emph{e-term} for short),
$t$ is recursively defined as one of the following expressions
\begin{gather*}
t \ \ ::= \ \ l \ \ | \ \ \prod S \ \ | \ \ \sum S \ \ | \ \ t_1 \cdotl t_2 \ \ | \ \ \sneg t_1
\end{gather*}
where \mbox{$l \in \ExtLb \eqdef \setm{ r_i^a}{ r_i \in \lb \text{ and } a \in  \ExpLit}$, }
$t_1, t_2$ are in turn terms, and $S$ is a (possibly empty and possibly infinite) set of terms. 
An e-term is \emph{elementary} if it has the form $l$, $\sneg l$ or $\sneg\sneg l$ with
\mbox{$l\in \ExtLb$} being an extended label.\qed
\end{definition}

\begin{figure}[t]
\begin{center}
\footnotesize
\newcommand{\titleSep}{-5pt}
\newcommand{\contentSep}{-12pt}
\newcommand{\rowSep}{5pt}
$
\begin{array}{c}
\hbox{\em Pseudo-complement}\vspace{\titleSep}\\
\hline\vspace{\contentSep}\\
\begin{array}{c@{\ }c@{\ }r@{}  }
t \ * \ \sneg t     & = & 0
\\
\sneg\sneg\sneg t   & = & \sneg t
\end{array}
\end{array}
\hspace{0.3cm}
\begin{array}{c}
\hbox{\em De Morgan}\vspace{\titleSep}\\
\hline\vspace{\contentSep}\\
\begin{array}{r@{}r@{}c@{}l@{\ }c@{\ }l@{}c@{}l@{} }
\sneg & (t & + & u) & = & (\sneg t  & * & \sneg u)\\
\sneg & (t & * & u) & = & (\sneg t  & + & \sneg u)
\end{array}
\end{array}
\hspace{0.3cm}
\begin{array}{c}
\hbox{\em Weak excl. middle}\vspace{\titleSep}\\
\hline\vspace{\contentSep}\\
\begin{array}{c@{\ }c@{\ }r@{}  }
\sneg t \ + \ \sneg\sneg t     & = & 1
\\
\\
\end{array}
\end{array}
\hspace{0.3cm}
\begin{array}{c}
\hbox{\em appl. negation}\vspace{\titleSep}\\
\hline\vspace{\contentSep}\\
\begin{array}{c@{\ }c@{\ }r@{}  }
\sneg (t \ \cdotl \ u)   & = & \sneg (t * u) 
\\
\\
\end{array}
\end{array}
$
\end{center}
\vspace{-5pt}
\caption{Properties of the `$\sneg$' operator.}
\label{fig:neg}
\end{figure}

\begin{definition}[Extended Causal Values]
\label{def:evalues}
An \emph{extended \changed{(causal)} value} (\emph{e-value} for short) is each equivalence class of \mbox{e-terms} under axioms for a completely distributive (complete) lattice with meet `$*$' and join `$+$' plus the axioms of Figures~\ref{fig:appl} and~\ref{fig:neg}.
The set of e-values is denoted by $\evalues$.\qed
\end{definition}

As with causal values, we will use any of the members of the class as representative of the extended casual value.
Note that $[0] = \set{ 0, r_1^a * \sneg r_1^a, \dotsc, }$ and $[1] = \set{ 1 , \sneg r_1^a + \sneg\sneg r_1^a, \dotsc }$ are no longer singleton sets.
The definition of disjunctive and graph normal form is now strengthened by requiring that negation~`$\sneg$' or double negation `$\sneg\sneg$' only occurs in front of labels and extended atoms.
\CHANGED
Similarly, the graph normal form also requires now that negation~`$\sneg$' or double negation `$\sneg\sneg$' only occurs in front of labels and extended atoms.
\END

Interpretations are extended in a straightforward way:
an \emph{e-interpretation} is a mapping \mbox{$\eI:\ExpLit\longrightarrow\evalues$} assigning an e-value to each extended atom
such that $\eI(a) = 0$ or $\eI(\neg a) = 0$ for every atom $a \in \at$.
For interpretations $\eI$ and $\eJ$ we say that $\eI\leq \eJ$ when
\mbox{$\eI(\rA) \leq \eJ(\rA)$} for each atom $\rA \in \ExpLit$.
As above, there is a \mbox{$\leq$-bottom} \mbox{e-interpretation} \botI\ (resp. a $\leq$-top \mbox{e-interpretation}~\topI) that stands for the \mbox{e-interpretation} mapping each extended atom $\rA$ to $0$ (resp. $1$).
The value assigned to a negative literal $\Not \rA$ by an \mbox{e-interpretation}~$\eI$, denoted as $\eI(\Not \rA)$, is defined as $\eI(\Not \rA) \eqdef \sneg\eI(\rA)$, as expected.
Similarly, for any \mbox{e-term} $t$, its valuation $\eI(t)\eqdef [t]$ is the equivalence class of~$t$.

To define the semantics of logic programs for extended causal justifications a slight extension in the syntax is also needed:
we allow that $\rB_1,\dotsc,\rB_n$ in~\eqref{eq:rule}, are not only extended atoms, but also e-terms.
For instance, $p \lparrow q \wedge (a * \sneg b)$, with $p,q \in \ExtAt$ and $a,b \in \lb$, is a valid rule in this extended syntax.
Furthermore, only normal logic programs are considered.

\begin{definition}[E-Model]
\label{def:emodel}
A \mbox{e-interpretation} $\eI$ \emph{satisfies} a rule like \eqref{eq:rule} with $\nphead=1$ iff
\begin{IEEEeqnarray}{c+x*}
\big( \ \eI(\rB_1) * \dotsc * \eI(\rB_\npbody)
    * \eI(\Not \rC_1) * \dotsc * \eI(\Not \rC_\nnbody) \ \big) 
        \cdot r_i^{\rH_1} \ \leq \ \eI(\rH_1) &
     \label{eq:model}
\end{IEEEeqnarray}
and $\eI$ is an e-model of $\cP$, written $\eI\models \cP$, iff $\eI$ satisfies all rules in $\cP$.\qed
\end{definition}

\begin{definition}[E-Reduct]
\label{def:e.reduct}
Given a normal program $\cP$ and an interpretation $\eI$,
by $\ereduct{\cP}{\eI}$ we denote the positive program
containing a rule of the form\footnote{\CHANGED Note that $\eI(\Not \rC_i)$ is a possibly infinite causal term for each $\rC_i$.}:
\begin{gather}
\rH_1 \leftarrow \rB_1, \dotsc, \rB_\npbody, \
    \eI(\Not \rC_1), \dotsc, \ \eI(\Not \rC_\nnbody)
    \label{eq:e.rule}
\end{gather}
for each rule of the form $\eqref{eq:rule}$ in $\cP$.\qed
\end{definition}

\noindent
Program~$\cP^\eI$ is positive and it has a \emph{$\leq$-least \mbox{e-model}}\footnote{\CHANGED Here, we take $\leq$-minimal models instead of $\sqsubseteq$-minimal models as in earlier sections. These two concepts coincide for normal programs, so we use the former for simplicity.}.
By \changed{$\eWp(\cdot)$,} we denote the operator\footnote{\CHANGED The operator $\eWp(\cdot)$ is analogous to the operator $\Gamma_P(\cdot)$ defined in Section~\ref{sec:background}, but using e-interpretations instead of sets of atoms.} mapping each \mbox{e-interpretation} $\eI$ to the $\leq$-least \mbox{e-model} of 
program~$\ereduct{\cP}{\eI}$.
Furthermore, $\eWp^2(\cdot)$ denotes the operator over \mbox{e-interpretations} resulting of applying $\eWp$ to the result of is its application to any \mbox{e-interpretation}, that is, \mbox{$\eWp^2(\eI) \eqdef \eWp(\eWp(\eI))$}.
This operator $\eWp^2$ is monotonic and so, by Knaster-Tarski's theorem, it has a least fixpoint $\elfp$ and a greatest fixpoint~$\egfp\eqdef\eWp(\elfp)$.
These two fixpoints respectively correspond to the justifications for true and for non-false
\CHANGED
(that is, either true of undefined)
\END
extended atoms in the (standard) well-founded model.
To capture justifications with respect to answer sets, we use the negative reduct from Definition~\ref{def:negative.reduct}.

\begin{definition}[Extended Causal Answer Sets]
Given a normal extended program $\cP$ one of its standard answer sets $M$, and a set of assumptions
\mbox{$U \subseteq \overline{M}$} such that
\mbox{$\WF{\NR{\P}{U}} = M$},
the \emph{extended causal answer set} (\emph{e-answer set}) corresponding to $M$ and $U$ is a function mapping each literal to an e-value as follows:
\begin{IEEEeqnarray*}{c +x*}
\eM_U(\rA)  \ \eqdef \ \elfpP{Q}(\rA) 
\hspace{40pt}
\eM_U(\Not \rA) \ \eqdef \ \sneg\egfpP{Q}(\rA)
&
\end{IEEEeqnarray*}
with $Q = \NR{\P}{U}$.\qed
\end{definition}

\CHANGED
The notion of causal justification is extended as expected. 

\begin{definition}[Extended Causal Justification]
Given a logic program $P$, an answer set $M$ of $P$ and a set of assumptions $U \subseteq \overline{M}$, a term without sums $c$ is an \emph{extended causal justification} of some literal~$l \in \set{a , \Not a }$ \wrt~$P$,~$M$ and $U$ if $c$ is an addend in the minimal disjunctive normal form of $\eM_U(l)$.
For any causal justification of $l$ \wrt~$P$, ~$M$ and~$U$
$\cgraph{c}$ is an \emph{extended causal graph (justification)}. \qed
\end{definition}
\END

\begin{examplecont}{ex:bond}\label{ex:bond2}
Let $\newprogram\label{prg:bond}$ be the logic program containing rules~(\ref{eq:bond.r1}-\ref{eq:bond.r3}).
This program has a complete well-founded model which coincides with its unique answer set:
$\newanswerset\label{as:prg:bond} = \set{a,d}$.
Then, the possible assumptions with respect to this answer set are those $U$ such that $U \subseteq \set{h}$, that is, $\emptyset$ and $\set{h}$.
Usually $\subseteq$-minimal assumptions are used
and, thus, we have that $\programr\ref{prg:bond} = \NR{\programr\ref{prg:bond}}{\emptyset}$ and that
\begin{gather*}
\begin{IEEEeqnarraybox}{l C l}
\eWpP{ \programr\ref{prg:bond} }(\botI)(p) &=&  d \cdotl r_1^p
\\
\eWpP{ \programr\ref{prg:bond} }(\botI)(d) &=&  d
\\
\eWpP{ \programr\ref{prg:bond} }(\botI)(a) &=&  r_2^a
\\
\eWpP{ \programr\ref{prg:bond} }(\botI)(h) &=&  0
\end{IEEEeqnarraybox}
\hspace{0.65cm}
\begin{IEEEeqnarraybox}{l C l}
\eWpP{ \programr\ref{prg:bond} }^2(\botI)(p) &=&  (\sneg r_2^a * d) \cdotl r_1^p
\\
\eWpP{ \programr\ref{prg:bond} }^2(\botI)(d) &=&  d
\\
\eWpP{ \programr\ref{prg:bond} }^2(\botI)(a) &=&  r_2^a
\\
\eWpP{ \programr\ref{prg:bond} }^2(\botI)(h) &=&  0
\end{IEEEeqnarraybox}
\hspace{0.65cm}
\begin{IEEEeqnarraybox}{l C l}
\eWpP{ \programr\ref{prg:bond} }^3(\botI)(p) &=&  (\sneg r_2^a * d) \cdotl r_1^p
\\
\eWpP{ \programr\ref{prg:bond} }^3(\botI)(d) &=&  d
\\
\eWpP{ \programr\ref{prg:bond} }^3(\botI)(a) &=&  r_2^a
\\
\eWpP{ \programr\ref{prg:bond} }^3(\botI)(h) &=&  0
\end{IEEEeqnarraybox}
\end{gather*}
Note that 
$\eWpP{ \programr\ref{prg:bond} }^2(\botI) = \eWpP{ \programr\ref{prg:bond} }^3(\botI)$
also implies that
$\eWpP{ \programr\ref{prg:bond} }^2(\botI) = \eWpP{ \programr\ref{prg:bond} }^4(\botI)$
and, thus, $\eWpP{ \programr\ref{prg:bond} }^2(\botI)$ is the least fixpoint of the
$\eWpP{ \programr\ref{prg:bond} }^2(\botI)$ operator.
Note also that
$\eWpP{ \programr\ref{prg:bond} }^2(\botI)(p) = (\sneg r_2^a * d) \cdotl r_1^p$
is precisely the causal justification shown in Example~\ref{ex:bond} to be associated with $p$ in this scenario.
Let now
$\newprogram\label{prg:bond2} = \programr\ref{prg:bond} \cup \set{ h }$,
which also has a complete well-founded model and unique answer set:
$\newanswerset\label{as:prg:bond2} = \set{p,d,h}$.
In this case, we have
\begin{gather*}
\begin{IEEEeqnarraybox}{l C l}
\eWpP{ \programr\ref{prg:bond} }(\botI)(p) &=&  d \cdotl r_1^p
\\
\eWpP{ \programr\ref{prg:bond} }(\botI)(d) &=&  d
\\
\eWpP{ \programr\ref{prg:bond} }(\botI)(a) &=&  r_2^a
\\
\eWpP{ \programr\ref{prg:bond} }(\botI)(h) &=&  h
\end{IEEEeqnarraybox}
\hspace{0.75cm}
\begin{IEEEeqnarraybox}{l C l}
\eWpP{ \programr\ref{prg:bond} }^2(\botI)(p) &=&  (\sneg r_2^a * d) \cdotl r_1^p
\\
\eWpP{ \programr\ref{prg:bond} }^2(\botI)(d) &=&  d
\\
\eWpP{ \programr\ref{prg:bond} }^2(\botI)(a) &=&  \sneg h \cdotl r_2^a
\\
\eWpP{ \programr\ref{prg:bond} }^2(\botI)(h) &=&  h
\end{IEEEeqnarraybox}
\hspace{0.75cm}
\begin{IEEEeqnarraybox}{l C l}
\eWpP{ \programr\ref{prg:bond} }^3(\botI)(p) &=&  \dotsc
\\
\eWpP{ \programr\ref{prg:bond} }^3(\botI)(d) &=&  d
\\
\eWpP{ \programr\ref{prg:bond} }^3(\botI)(a) &=&  \sneg h \cdotl r_2^a
\\
\eWpP{ \programr\ref{prg:bond} }^3(\botI)(h) &=&  h
\end{IEEEeqnarraybox}
\end{gather*}
with
$\eWpP{ \programr\ref{prg:bond} }^4(\botI)(p) = \eWpP{ \programr\ref{prg:bond} }^3(\botI)(p) =  (\sneg\sneg h * d) \cdotl r_1^p +  (\sneg r_2^a * d) \cdotl r_1^p$ as also mentioned in Example~\ref{ex:bond}.\qed
\end{examplecont}

An extended causal justification is said to be \emph{inhibited} when it contains a negated label (non-double negated).
Inhibited justifications point out derivations that could have justified the truth value of the atom, but that have been prevented to do so.
The negated subterms are the inhibitors of the extended causal justification.
\emph{Actual} extended causal justifications are those that only contain non-negated and double negated subterms.
In Example~\ref{ex:bond2},
the casual term $(\sneg\sneg h * d) \cdotl r_1^p$ represents the actual extended causal justification of $p$, while
$(\sneg r_2^a * d) \cdotl r_1^p$ is an inhibited extended causal justification that points out that
``had it not been for rule $r_2$, then $d$ would cause $p$ to be true through rule $r_1$
(without the need of~$h$)''. 
Note that the presence of the negated subterm $\sneg r_2^a$
in the inhibited extended causal justification
$(\sneg r_2^a * d) \cdotl r_1^p$ is similar to the attack from the argument with conclusion $a$ to the argument with conclusion $p$ in the attack tree used to construct the LABAS justification.

\begin{examplecont}{ex:light}\label{ex:causal.extended.light}
Continuing with the problem introduced in Example~\ref{ex:light} (page~\pageref{ex:light}),
we can see that 
$\eWpP{ \programr\ref{prg:light} }^i(\botI)(\on_1) =  (\swa_0 * \swb_0) \cdotl r_{11}$
for all $i \geq 1$.
That is, the extended causal justification of $\on_1$ has precisely the same graph as the (non-extended) causal justification depicted in Figure~\ref{fig:on1:causal} (page~\pageref{fig:on1:causal}).
We also have that
$\eWpP{ \programr\ref{prg:light} }^i(\botI)(\off_1)
  =  (\sneg\swa_0* \off_0) \cdotl i_{22} + (\sneg\swb_0 * \off_0) \cdotl i_{22} + (\sneg r_{11} * \off_0) \cdotl i_{22}$ for all $i \geq 2$.
This points out that $\off_1$ would be true by inertia (rule~$i_{22}$) if any of the facts $\swa_0$ or $\swb_0$ or the rule $r_{11}$ had not been in the program.
It can be checked that $(\swa_0 * \swb_0) \cdotl r_{11} \cdotl i_{12}$ is the extended causal justification of $\on_2$.
Recall that this is the (non-extended) causal justification of $\on_2$, whose corresponding causal graph is depicted in Figure~\ref{fig:on2:causal} (page~\pageref{fig:on2:causal}).\qed
\end{examplecont}

\begin{examplecont}{ex:bond2}\label{ex:bond3}
Recall that, in the unique answer set
$\answersetr\ref{as:prg:bond} = \set{d,a}$
of program~$\programr\ref{prg:bond}$,
the atom $p$ is false.
Extended causal justifications also allow to justify negative literals
and we have that $\Not p$ is explained by the causal value
$\sneg\sneg r_2^a + \sneg d + \sneg r_1^p$.
Here, $\sneg\sneg r_2^a$ is the actual extended causal justification explaining why $p$ is false, while \changed{$\sneg d$} and $\sneg r_1^p$ are inhibited extended causal justifications that point out that $p$ would also be false if either \changed{$d$} or $r_1$ were removed from the program.\qed
\end{examplecont}

Note that in Example~\ref{ex:bond3}
the application operator~`$\cdot$' does not appear in the extended causal justification of $\Not p$.
In fact, this is the general case for negative literals and, thus, extended causal justifications for negative literals do not keep track of the derivation order among rules. An algebraic treatment that allows to keep track of this derivation order is still an open topic. It is also an open topic to explain negative literals for disjunctive programs.

\subsection{Why-not Provenance Justifications}
\label{sec:why-not}
%
%
%

Why-not provenance~\cite{DamasioAA2013} is a declarative logical approach, which extracts \mbox{non-graph} based justifications for the truth value of atoms with respect to the (complete) well-founded model of normal logic programs.
\changed{It can furthermore be used to explain the truth value of atoms with respect to the answer set semantics.}
The approach has been implemented in a meta-programming tool~\cite{damasioMA15} available at \url{http://cptkirk.sourceforge.net}.
As mentioned in Section~\ref{sec:enablers-inhibitors},
the way extended causal justifications have been defined is inspired by this approach, therefore, we here just introduce the differences between these two approaches, avoiding the overlapping material.

As already mentioned, the first major difference \changed{compared to extended causal justifications (and the other justifications approaches reviewed in Section~\ref{sec:justifications})} is the non-graph nature of \mbox{why-not} provenance.
Instead, why-not provenance justifications are sets of annotations,
each one expressing a
possible modification of the program
to achieve a particular truth value of the justified atom \wrt\ the well-founded model (of the modified program).
In other words, \mbox{why-not} provenance computes justifications expressing how the atom can be made true, false, or undefined \wrt\ the well-founded model or the answer set semantics.
The justifications for the \emph{actual} truth value of the atom are those that do not imply any modification on the program.
This can be achieved by adding the axiom
\begin{gather}
(t \cdotl u) \ \ = \ \ (t * v)
  \label{eq:why-not.axiom.dot.elimination}
\end{gather}
to those defining e-values (Definition~\ref{def:evalues}).
That is, the non-commutative operator~`$\cdot$' is replaced by the commutative one~`$*$', effectively removing the order of application of rules from the justifications.

The second
difference \changed{compared to extended causal justifications} is that why-not provenance \changed{does not distinguish} between productive causes and other counterfactual dependencies, which is achieved by adding the double negation elimination axiom:
\begin{gather}
\sneg\sneg t \ \ = \ \ t
  \label{eq:why-not.axiom.neg.elimination}
\end{gather}

\begin{definition}[Why-Not Provenance Values]
\label{def:wvalues}
A \emph{why-not provenance value} (\emph{w-value} for short) is each equivalence class of \mbox{e-terms} (Definition~\ref{def:eterm}, page~\ref{def:eterm}) under axioms for a completely distributive (complete) lattice with meet `$*$' and join `$+$' plus the axioms of Figures~\ref{fig:appl} and~\ref{fig:neg}
and the axioms $(t \cdotl u) = (t * v)$ and $\sneg\sneg t = t$.
The set of w-values is denoted $\wvalues$.\qed
\end{definition}

Due to the addition of axioms~\eqref{eq:why-not.axiom.dot.elimination} and~\eqref{eq:why-not.axiom.neg.elimination}, w-values form a free boolean algebra\footnote{In fact, the original definition relies on a free boolean algebra instead of causal terms and assumes the notation of logical formulas to represent its values (see Notation~\ref{not:why-not.syntax} below).} generated by $\lb$.
The definitions of w-interpretation, w-model and reduct are \changed{analogous to the ones} in Section~\ref{sec:enablers-inhibitors}, but replacing e-values by w-values. 
We will use $\wI$, $\wJ$ and their variations to denote w-interpretations.
By $\Wp(\wI)$ we denote the least \mbox{w-model} of 
program~$\ereduct{\cP}{\wI}$
and by~$\Wp^2(\eI) \eqdef \Wp(\Wp(\eI))$ we denote the result of applying $\Wp$ to the result of its application to $\wI$.
Let us denote by
\mbox{$\wlfpP{P}$}
and 
\mbox{$\wgfpP{P}$},
the least and greatest fixpoint of the operator $\WpP{P}^2$.

\begin{notation}\label{not:why-not.syntax}
In order to closely follow the notation used in~\cite{DamasioAA2013}, we will represent the meet as conjunction~`$\wedge$' instead of as product~`$*$' and the joint as disjunction `$\vee$' instead of~`$+$' when representing w-values.
We will also write negation as `$\neg$' instead of `$\sneg$' to strengthen the fact that it now acts as classical negation
and omit the superindex of labels.\qed
\end{notation}

Note that the intuition of the two former operators is as before:
conjunction~`$\wedge$' indicates joint interaction, disjunction `$\vee$' represents alternative justifications.
On the other hand, now negation~`$\neg$' denotes hypothetical changes to the program (either removal or addition) that may lead to the literal belonging to the \mbox{well-founded} model.

 \begin{examplecont}{ex:off-line.empty-well-founded-model.2justifications}
\label{ex:why-not.empty-well-founded-model.2justifications}
Let us label each rule in the program $\programr\ref{prg:wfm2j}$ as follows
\begin{gather*}
r_1 : \ p \lparrow \Not q
\hspace{1.75cm}
r_2 : \ r \lparrow \Not p
\hspace{1.75cm}
r_3 : \ s \lparrow \Not r
\end{gather*}
As mentioned in Example~\ref{ex:off-line.empty-well-founded-model.2justifications},
this program has a \changed{complete well-founded model:} 
$\answersetr\ref{as:prg:wfm2j} = \set{p,s}$.
We also have that the following extended causal justifications:
\begin{gather*}
\begin{IEEEeqnarraybox}{l C l}
\eWpP{ \programr\ref{prg:wfm2j} }(\botI)(p) &=&  r_1^p
\\
\eWpP{ \programr\ref{prg:wfm2j} }(\botI)(q) &=&  0
\\
\eWpP{ \programr\ref{prg:wfm2j} }(\botI)(r) &=&  r_2^r
\\
\eWpP{ \programr\ref{prg:wfm2j} }(\botI)(s) &=&  r_3^s
\end{IEEEeqnarraybox}
\hspace{1cm}
\begin{IEEEeqnarraybox}{l C l}
\eWpP{ \programr\ref{prg:wfm2j} }^2(\botI)(p) &=&  r_1^p
\\
\eWpP{ \programr\ref{prg:wfm2j} }^2(\botI)(q) &=&  0
\\
\eWpP{ \programr\ref{prg:wfm2j} }^2(\botI)(r) &=&  \sneg r_1^p \cdotl r_2^r
\\
\eWpP{ \programr\ref{prg:wfm2j} }^2(\botI)(s) &=&  \sneg r_2^r \cdotl r_3^s
\end{IEEEeqnarraybox}
\hspace{1cm}
\begin{IEEEeqnarraybox}{l C l}
\eWpP{ \programr\ref{prg:wfm2j} }^3(\botI)(p) &=&  r_1^p
\\
\eWpP{ \programr\ref{prg:wfm2j} }^3(\botI)(q) &=&  0
\\
\eWpP{ \programr\ref{prg:wfm2j} }^3(\botI)(r) &=&  \sneg r_1 \cdotl r_2^r
\\
\eWpP{ \programr\ref{prg:wfm2j} }^3(\botI)(s) &=&  \sneg\sneg r_1 \cdotl r_3^s + \sneg r_2^r \cdotl r_3^s
\end{IEEEeqnarraybox}
\end{gather*}
and, it can be checked that, $\eWpP{ \programr\ref{prg:wfm2j} }^4(\botI) = \eWpP{ \programr\ref{prg:wfm2j} }^2(\botI)$.
Then, applying the above two axioms~(\ref{eq:why-not.axiom.dot.elimination}-\ref{eq:why-not.axiom.neg.elimination}) and the rewriting of Notation~\ref{not:why-not.syntax},
we have that
\begin{gather*}
\begin{IEEEeqnarraybox}{l C l}
\WpP{ \programr\ref{prg:wfm2j} }^4(\botI)(p) &=&  r_1
\\
\WpP{ \programr\ref{prg:wfm2j} }^4(\botI)(q) &=&  0
\\
\WpP{ \programr\ref{prg:wfm2j} }^4(\botI)(r) &=&  \neg r_1 \wedge r_2
\\
\WpP{ \programr\ref{prg:wfm2j} }^4(\botI)(s) &=&  r_1 \wedge r_3 \vee \neg r_2 \wedge r_3
\end{IEEEeqnarraybox}
\end{gather*}
The intuition behind $r_1 \wedge r_3$ is similar to the one in extended causal justifications, but without derivation order, distinction between productive causes and other contingently counterfactual dependencies: $r_1 \wedge r_3$ means that ``$s$ is true because both $r_1$ and $r_3$ are in the program''.
\qed
\end{examplecont}


\noindent
In other words, the least fixpoint of $\Wp^2$ can be obtained from the least fixpoint of $\eWp^2$ by replacing applications~`$\cdot$' by products~`$*$', removing every double negation symbols `$\sneg\sneg$' and, then, applying the rewriting of Notation~\ref{not:why-not.syntax}.
More formally,
let \mbox{$\lambdap:\evalues\longrightarrow\wvalues$} be this transformation from \mbox{e-values} to \mbox{w-values}, that is,
$\lambdap$ is defined in the following recursive way:
\begin{align*}
\lambdap(t) \ &\eqdef \ \begin{cases}
    \lambdap(u) \wedge \lambdap(w)
        &\text{if } t=u \odot v \text{ with } \odot\in\set{*, \cdot}
    \\
    \lambdap(u)  \vee  \lambdap(w)
        &\text{if } t=u + v 
    \\
    \neg \lambdap(u)
        &\text{if } t= \sneg u
    \\
    l   &\text{if } t= l \text{ with } l \in (\lb \cup \ExpLit)
\end{cases}
\end{align*}
with $t$ in graph normal form.

\CHANGED
Note that, similar to LABAS justifications, there are no extended causal justifications for atoms for which there is no derivation.
For instance, there is no justification for the atom $p$ \wrt\ to a program consisting of a single rule $p \lparrow q$.
On the other hand, as in off-line justifications, there are \mbox{why-not} provenance justifications for those atoms.
In our running example, $p$ is associated with the why-not provenance information
$\neg\Not(p) \vee r_1  \wedge \neg\Not(q)$ where $r_1$ is the label associated to the rule $p \lparrow q$.
This difference is due to the use of an extended program to compute \mbox{why-not} provenance information.
\END

\begin{definition}[Provenance Program]\label{def:provenance.program}
Given a normal program $\cP$, the why-not provenance program is $\wP(P)\eqdef \cP \cup \cP'$, where $\cP'$ contains a labelled fact of the form 
$(\neg not(a) : a)$
for each extended atom \mbox{$a \in \ExpLit$} not occurring as a fact in $\cP$.\qed
\end{definition} 

We write $\wP$ instead of $\wP(\cP)$ when the program $\cP$ is clear from the context.
To compute the \mbox{why-not} provenance information of some normal program~$P$, we will be interested in the least and greatest fixpoints of the $\WpP{\wP}^2$ operator with respect to the provenance program $\wP$ (corresponding to $P$), instead of those of~$P$ itself.
That is, we will use the least and greatest fixpoints $\wlfp$ and $\wgfp$.
It is also worth noting that these fixpoints can be obtained from the fixpoints of extended causal operator with respect to the extended program,
 that is,
\mbox{$\wlfp = \lambdap(\elfpP{\wP})$}
and 
\mbox{$\wgfp = \lambdap(\egfpP{\wP})$}.

\begin{definition}[Provenance Information]\label{def:wellf.provenance}
Given a normal program $\cP$, \emph{why-not provenance information}
is defined as a mapping from literals\footnote{In this section, we use a more general notion of `literal', where an atom $a$ may not only be proceeded by $\Not$, but also by $\Undef$.} into w-values satisfying:
\begin{IEEEeqnarray*}{l C l " l }
Why_\cP(\rA) &\eqdef& \phantom{\neg}\wlfpP{\wP}(\rA)
\\
Why_\cP(\Not \rA) &\eqdef& \neg\wgfpP{\wP}(\rA)
\\
Why_\cP(\Undef \rA) &\eqdef& \neg Why_\cP(\rA) * \neg Why_\cP(\Not \rA)
\end{IEEEeqnarray*}
for each extended atom~\mbox{$\rA \in \ExpLit$}.\qed
\end{definition}

Intuitively, each disjunct in the minimal disjunctive normal form of provenance information corresponds to a justification about to why the atom does or does not have the respective truth value \wrt\ the well-founded model. That is, the disjunct in $Why_\cP(\rA)$, $Why_\cP(\Not \rA)$, and $Why_\cP(\Undef \rA)$ respectively explain why $\rA$ is (not) true, false, and undefined \wrt\ the well-founded model. The \emph{actual} truth value of $\rA$ can be spotted if a disjunct in the respective justification ($Why_\cP(\rA)$, $Why_\cP(\Not \rA)$, or $Why_\cP(\Undef \rA)$) does not contain any negation $\neg$.

\begin{examplecont}{ex:why-not.empty-well-founded-model.2justifications}\label{ex:why-not.empty-well-founded-model.2justifications.2}
Continuing with our running example, we have that $\wP_{\ref{prg:wfm2j}} = \wP(\programr\ref{prg:wfm2j})$
consists of the following rules:
\begin{gather*}
\begin{IEEEeqnarraybox}{ l C l C l}
r_1 &:& \ p &\lparrow& \Not q
\\
r_2 &:& \ r &\lparrow& \Not p
\\
r_3 &:& \ s &\lparrow& \Not r
\end{IEEEeqnarraybox}
\hspace{2cm}
\begin{IEEEeqnarraybox}{ l C l C l}
\neg not(p) &:& \ p
\\
\neg not(q) &:& \ q
\\
\neg not(p) &:& \ r
\end{IEEEeqnarraybox}
\hspace{2cm}
\begin{IEEEeqnarraybox}{ l C l C l}
\neg not(s) &:& \ s
\\
\\
\end{IEEEeqnarraybox}
\end{gather*}
Since there is no fact $q$ in $\programr\ref{prg:wfm2j}$,
we have that $(\neg not(q) : \ q)$ belongs to $\wP_{\ref{prg:wfm2j}}$.
Furthermore, this is the unique rule in $\wP_{\ref{prg:wfm2j}}$ with $q$ in the head
and, consequently,
we have that
$\eWpP{ \wP_{\ref{prg:wfm2j}} }^i(\botI)(q) = \neg not(q)$
for all $i \geq 1$.
This implies  that
$\wlfpP{\wP}(q) = \wgfpP{\wP}(q) = \neg not(q)$
and, thus, that
\begin{IEEEeqnarray}{l C l}
Why_{ \programr\ref{prg:wfm2j} }(q)        &=&  \neg not(q)
\label{eq:q:why-not.empty-well-founded-model.2justifications.2}
\\
Why_{ \programr\ref{prg:wfm2j} }(\Not q)   &=&  \phantom{\neg} not(q)
\label{eq:notq:why-not.empty-well-founded-model.2justifications.2}
\\
Why_{ \programr\ref{prg:wfm2j} }(\Undef q) &=&  \phantom{\neg} 0
\end{IEEEeqnarray}
Note that $Why_{ \programr\ref{prg:wfm2j} }(q) = \neg not(q)$ corresponds to the off-line justification of $q$ consisting of a unique edge $(q^{-},\bot,-)$.
On other hand, since there is no rule in $P$ with~$q$ in the head,
there is no LABAS nor (extended) causal justification of $q$.
Similarly, to the computation shown in Example~\ref{ex:why-not.empty-well-founded-model.2justifications},
we also have that
\CHANGED
\begin{gather*}
\begin{IEEEeqnarraybox}{l C l C l}
\WpP{ \programr\ref{prg:wfm2j} }^i(\botI)(p) &=& \neg not(p) &\vee& r_1 \wedge not(q)
\\
\WpP{ \programr\ref{prg:wfm2j} }^i(\botI)(r) &=& 
\neg not(r) &\vee&  r_2 \wedge not(p) \wedge \neg r_1
\ \vee \  r_2 \wedge not(p) \wedge \neg not(q)
\\
\WpP{ \programr\ref{prg:wfm2j} }^i(\botI)(s) &=&  
\neg not(s)
&\vee& r_3 \wedge not(r) \wedge \neg r_2
\ \vee \ r_3 \wedge not(r) \wedge \neg not(p)
\\
&& &\vee& r_3 \wedge not(r) \wedge r_1 \wedge not(q)
\end{IEEEeqnarraybox}
\end{gather*}
\END
for all $i \geq 2$.
This implies that
$\wlfpP{\wP}(p) = \wgfpP{\wP}(p) = \neg not(p) \vee r_1 \wedge not(q)$
and that
\begin{gather*}
\begin{IEEEeqnarraybox}{l C l c l c l }
Why_{ \programr\ref{prg:wfm2j} }(p)        &=&  \neg not(p) &\ \vee \ & r_1 \wedge not(q)
\\
Why_{ \programr\ref{prg:wfm2j} }(\Not p)   &=&  \phantom{\neg} not(p) \wedge \neg r_1 &\ \vee \ & not(p) \wedge \neg not(q)
\\
Why_{ \programr\ref{prg:wfm2j} }(\Undef p) &=&  \phantom{\neg} 0
\end{IEEEeqnarraybox}
\end{gather*}
Following a similar procedure, it can be checked that
\begin{gather*}
\begin{IEEEeqnarraybox}{l C l c l c l }
Why_{ \programr\ref{prg:wfm2j} }(r)        &=&  \neg not(r) &\ \vee \ & r_2 \wedge not(p) \wedge \neg r_1 &\ \vee \ & r_2 \wedge not(p) \wedge \neg not(q)
\\
Why_{ \programr\ref{prg:wfm2j} }(\Not r)   &=&  \phantom{\neg} not(r) \wedge \neg r_2 &\ \vee \ & not(r) \wedge \neg not(p) &\ \vee \ & not(r) \wedge r_1 \wedge not(q)
\\
Why_{ \programr\ref{prg:wfm2j} }(\Undef r) &=&  \phantom{\neg} 0
\end{IEEEeqnarraybox}
\end{gather*}
that $Why_{ \programr\ref{prg:wfm2j} }(s)$ is
\begin{IEEEeqnarray}{l C l}
&&\neg not(s)
\\
    &\vee& r_3 \!\wedge\! not(r) \!\wedge\! \neg r_2
    \label{eq:6:ex:why-not.empty-well-founded-model.2justifications.2}
\\
    &\vee& r_3 \!\wedge\! not(r) \!\wedge\! \neg not(p) 
\\
    &\vee& r_3 \!\wedge\! not(r) \!\wedge\! r_1 \!\wedge\! not(q)
     \label{eq:5:ex:why-not.empty-well-founded-model.2justifications.2}
\end{IEEEeqnarray}
and that $Why_{ \programr\ref{prg:wfm2j} }(\Not s)$ is
\begin{IEEEeqnarray}{l C l}
&&not(s) \!\wedge\! \neg r_3
    \label{eq:1:ex:why-not.empty-well-founded-model.2justifications.2}
\\
    &\vee& not(s) \!\wedge\! \neg not(r)
    \label{eq:2:ex:why-not.empty-well-founded-model.2justifications.2}
\\
    &\vee& not(s) \!\wedge\! r_2 \!\wedge\! not(p) \!\wedge\! \neg r_1
    \label{eq:3:ex:why-not.empty-well-founded-model.2justifications.2}
\\
    &\vee& not(s) \!\wedge\! r_2 \!\wedge\! not(p) \!\wedge\! \neg not(q)
    \label{eq:4:ex:why-not.empty-well-founded-model.2justifications.2}
\end{IEEEeqnarray}
Comparing the conjunction $r_1 \wedge r_3$ obtained in Example~\ref{ex:why-not.empty-well-founded-model.2justifications} with the conjunction~\eqref{eq:5:ex:why-not.empty-well-founded-model.2justifications.2},
we can observe that annotations $not(r)$ and $not(q)$ have been added.
This can be informally read as ``$s$ is true because both $r_1$ and $r_3$ are in the program and facts $r$ and $q$ are not.''
Note also, that $not(r) \wedge r_1 \wedge not(q)$ is one of the disjuncts of
$Why_{ \programr\ref{prg:wfm2j} }(\Not r)$.
This could be read as ``$r$ is false because of rule $r_1$ and the absence of facts $r$ and $q$ in the program.'' 
\qed
\end{examplecont}

The following definitions formalises the notion of why-not provenance \emph{justification}, \changed{\newChanged{i.e.} a disjunct in the why-not provenance information,} and the intuition behind the meaning of each annotation in a justification.
In particular, it expresses the idea that each justification describes a modification of the program after which the atom has the truth value of the respective justification. 

\CHANGED
\begin{definition}[Why-not Provenance Justification]
Let $\P$ be a normal program, let $a \in \ExtAt$ be an extended atom and let $\rL \in \set{a, \Not a, \Undef a}$ such that
$Why_{\P}(\rL) = c_1 \vee \dotsc \vee c_n$
is the why-not provenance information of $\rL$ in minimal disjunctive normal form.
Then, we say that each $c_i$ is a \emph{why-not provenance justification} of $l$ \wrt~$P$. \qed
\end{definition}

\begin{definition}
\label{def:why-not:changes}
Let $\P$ be a normal program,
$a \in \ExtAt$ be an extended atom and  $\rL \in \set{a, \Not a, \Undef a}$.
Let $c$ be some why-not provenance justification of $l$ \wrt~$P$
and $C$ a set of annotations such that $\bigwedge C = c$.
Then,
the following sets are defined, where $b \in \ExtAt$ and $r \in \P$:
\begin{IEEEeqnarray*}{l ? C ? l c l +x* }
\KeepFacts(c)   &\eqdef& \setm{ b &}{& \phantom{\neg} b \in C }
\\
\RemoveFacts(c)   &\eqdef& \setm{ b &}{& \neg b \in C }
\\
\MissingFacts(c)  &\eqdef& \setm{ b &}{& \neg not(b) \in C }
\\
\NoFacts(c)     &\eqdef& \setm{ b  &}{& \phantom{\neg} not(b) \in C }
\\
\KeepRules(c)     &\eqdef& \setm{ \R &}{& \phantom{\neg} r_i \in C  \text{ and }  \Label{\R} = r_i }
\\
\RemoveRules(c)   &\eqdef& \setm{ \R &}{& \neg r_i \in C  \text{ and }  \Label{\R} = r_i }
&\qed
\end{IEEEeqnarray*}
\end{definition}
\END

\noindent
Intuitively, any \mbox{disjunct} $c_j$ in the why-not provenance information of some literal~$\rL$ expresses a possible modification of the program such that $\rL$ belongs to the
 well-founded model of the resulting program.
These modifications are captured by the above sets.

\changed{For instance, $\MissingFacts(c_j)$ is a set of facts that would be necessary to add to the program in order to justify $\rL$, while 
$\NoFacts(c_j)$ is a set of facts that cannot be added in order to justify~$\rL$.}
As a consequence, $\rL$ will belong\footnote{This has been shown in~\cite[Theorem 3]{DamasioAA2013}.%
}
to the \mbox{well-founded} model of any program resulting from adding any superset $G$ of $\MissingFacts(c_j)$
that does not contain any fact from $\NoFacts(c_j)$
(assuming that $\RemoveRules(c_j) = \RemoveFacts(c_j) = \emptyset$).

\begin{examplecont}{ex:why-not.empty-well-founded-model.2justifications.2}\label{ex:why-not.empty-well-founded-model.2justifications.3}
Continuing with our running example, we have that $\Not s$ does not belong to the well-founded model of~$\programr\ref{prg:wfm2j}$ and that
$c = not(s) \!\wedge\! \neg not(r)$ is a \changed{why-not provenance justification of $\Not s$, i.e. it is a disjunct~\eqref{eq:2:ex:why-not.empty-well-founded-model.2justifications.2}
of the
\mbox{why-not} provenance information of $\Not s$}.
Then, we also have
\mbox{$\MissingFacts(c) = \set{ r }$}
and
\mbox{$\NoFacts(c) = \set{ s }$}.
This expresses that $\Not s$ would belong to the well-founded model of any program $P' = \programr\ref{prg:wfm2j} \cup G$ with $G$ any set of facts that includes $r$ but does not include~$s$.\qed
\end{examplecont}

Similarly, $\KeepFacts(c_j)$ and $\KeepRules(c_j)$ point out facts and rules that need to be kept in the program to justify the literal
while $\RemoveFacts(c_j)$ and $\RemoveRules(c_j)$ state facts and rules that need to be removed from the program. 
Note that, if a conjunction $c_j$ contains no negation, then it does not imply any change in the program and, thus, constitutes an \emph{actual} justification for the \emph{actual} value of the literal.

\begin{examplecont}{ex:why-not.empty-well-founded-model.2justifications.3}\label{ex:why-not.empty-well-founded-model.2justifications.4}
As a further example,
let
$c' \,=\, r_3 \wedge not(r) \wedge r_1 \wedge not(q)$
be
\changed{a why-not provenance justification of $s$}
 (the conjunction corresponding to the disjunct~\eqref{eq:5:ex:why-not.empty-well-founded-model.2justifications.2}
of the why-not provenance information of $s$).
Informally, this conjunction expresses that
``$s$ is true because both $r_1$ and $r_3$ are in the program and facts $r$ and $q$ are not.''
Note that
$\KeepRules(c') = \set{r_1, r_3}$
and
$\NoFacts(c') = \set{ r, q }$,
indicating that $s$ remains true 
as long as we keep these two rules and we add neither $r$ nor $q$, even if we remove other rules or remove or add other facts.
Note also that there is no negated annotation in $c'$ and, thus, 
$\RemoveFacts(c')
\!=\!
\MissingFacts(c')
\!=\!
\RemoveRules(c')
\!=\! \emptyset$.
In other words,
$c'$ points out a that no modification is required to make $s$ true and, thus,
it is an actual justification for the truth of $s$.\qed
\end{examplecont}

The following example illustrates how why-not provenance captures justifications of programs with even-length negative dependency cycles:

\begin{examplecont}{ex:off-line.cycle}\label{ex:why-not.cycle}
Let us define the following labelling for program~$\programr\ref{prg:cycle}$:
\begin{gather*}
r_1 : \ p \lparrow \Not q
\hspace{2cm}
r_2 : \ q \lparrow \Not p
\end{gather*}
As we have seen, program~$\programr\ref{prg:cycle}$ has two answer sets,
namely
\mbox{$\answersetr\ref{as:1:prg:cycle} = \set{p}$}
and
\mbox{$\answersetr\ref{as:2:prg:cycle} = \set{q}$},
and an empty well-founded model.
The computation of the why-not provenance information goes as follows:
\small
\begin{gather*}
\begin{IEEEeqnarraybox}{l C l}
\WpP{ \wP_{\ref{prg:cycle}} }^1(\botI)(p) &=&  \neg not(p) \vee r_1
\\
\WpP{ \wP_{\ref{prg:cycle}} }^2(\botI)(p) &=&  \neg not(p) \vee r_1 \!\wedge\! not(q) \!\wedge\! \neg r_2
\\
\WpP{ \wP_{\ref{prg:cycle}} }^3(\botI)(p) &=&  \neg not(p) \vee r_1 \!\wedge\! not(q)
\\
\WpP{ \wP_{\ref{prg:cycle}} }^4(\botI)(p) &=&  \neg not(p) \vee r_1 \!\wedge\! not(q) \!\wedge\! \neg r_2
\end{IEEEeqnarraybox}
\hspace{0.5cm}
\begin{IEEEeqnarraybox}{l C l}
\WpP{ \wP_{\ref{prg:cycle}} }^1(\botI)(\Not q) &=&  not(q) \wedge \neg r_2
\\
\WpP{ \wP_{\ref{prg:cycle}} }^2(\botI)(\Not q) &=&  not(q) \wedge (\neg r_2 \!\vee\! \neg not(p) \!\vee\! r_1)
\\
\WpP{ \wP_{\ref{prg:cycle}} }^3(\botI)(\Not q) &=&  not(q) \wedge (\neg r_2 \!\vee\! \neg not(p))
\\
\WpP{ \wP_{\ref{prg:cycle}} }^4(\botI)(\Not q) &=&  not(q) \wedge (\neg r_2 \!\vee\! \neg not(p) \!\vee\! r_1)
\end{IEEEeqnarraybox}
\end{gather*}
\normalsize
$\WpP{ \wP_{\ref{prg:cycle}} }^2(\botI)$ and
$\WpP{ \wP_{\ref{prg:cycle}} }^3(\botI)$
respectively are the least and greatest fixpoint of $\eWpP{ \wP_{\ref{prg:cycle}} }^2$.
The case for $q$ and $\Not p$ are symmetric.
Then, the why-not provenance information for $p$ is as follows:
\begin{gather*}
\begin{IEEEeqnarraybox}{l C l c l c l }
Why_{P_{\ref{prg:cycle}} }(p)        &=&  \neg not(p) &\ \vee \ & r_1 \!\wedge\! not(q) \!\wedge\! \neg r_2
\\
Why_{P_{\ref{prg:cycle}} }(\Not p)   &=&  \phantom{\neg} not(p) \!\wedge\! \neg r_1 &\ \vee \ & not(p) \!\wedge\! \neg not(q)
\\
Why_{P_{\ref{prg:cycle}} }(\Undef p) &=&  \IEEEeqnarraymulticol{3}{l}{\phantom{\neg}  not(p) \!\wedge\! not(q) \!\wedge\! r_1 \!\wedge\! r_2 }
\end{IEEEeqnarraybox}
\end{gather*}
Note that the only 
\changed{why-not provenance justification} without negation $\neg$ occurs in $Why_{ wP_{\ref{prg:cycle}} }(\Undef p)$, indicating that the actual truth value of $p$ \wrt\ the well-founded model is undefined.
The conjunction expresses that $p$ is undefined in the well-founded model of~$\programr\ref{prg:cycle}$ because of the rules $r_1$ and $r_2$ and the absence of the facts $p$ and $q$.\qed
\end{examplecont}

\CHANGED
\subsubsection{Answer Set Why-not Provenance}
The why-not justifications reviewed so far explain the truth value of literals with respect to the well-founded model. 
\END
\mbox{Why-not} provenance information of a literal \wrt\ the answer set semantics is defined in terms of the \mbox{why-not} provenance of that literal being true in the \mbox{well-founded model} and the non-existence of undefined atoms in it. In other words, a literal is justified \wrt\ the answer set semantics by referring to modifications that make the literal \emph{true} \wrt\ the complete well-founded model, which implies that it becomes the unique answer set.

\begin{definition}[Answer Set Provenance Information]\label{def:ans.provenance}
Given a normal program $\cP$, the \emph{answer set why-not provenance information} of a literal $\rL \in \litExt$ is defined as:
$AnsWhy_\cP(\rL)\eqdef Why_\cP(\rL) \ \wedge \ \bigwedge_{b \in \ExpLit} \neg Why_\cP(\Undef b)$.\qed
\end{definition}

\CHANGED
\begin{definition}[Answer Set why-not Provenance Justification]
Let $\P$ be a normal program, let $a \in \ExtAt$ be an extended atom and let $\rL \in \set{a, \Not a, \Undef a}$ such that
$AndWhy_{\P}(\rL) = c_1 \vee \dotsc \vee c_n$
is the answer set why-not provenance information of $\rL$ in minimal disjunctive normal form.
Then, we say that each $c_i$ is an \emph{answer set why-not provenance justification} of $l$ \wrt~$P$. \qed
\end{definition}
\END

Note that Definition~\ref{def:ans.provenance} characterises the major difference between this justification approach and the three previous ones: there is a unique provenance information of a literal with respect to the \emph{whole program}, not with respect to each answer set.

In the case of Example~\ref{ex:why-not.cycle} the answer set provenance (Definition~\ref{def:ans.provenance}) for $p$, $q$, $\Not p$ and $\Not q$ coincides with their respective provenance information (Definition~\ref{def:wellf.provenance}). 
Note that none of the disjuncts in the \mbox{why-not} provenance information of $p$ (resp.~$q$) is without negation, which seems to point out that $p$ is not true (can only be made true through modifications of the program). 
The reason is that, even though $p$ (resp.~$q$) is true in \emph{some} answer set, it is not true in the \mbox{well-founded} model (it could also be due to the \mbox{well-founded} model  not being complete).
The answer set provenance thus points out modifications that would yield a complete
\mbox{well-founded} model (and, thus, a \emph{unique} answer set) in which $p$ (resp.~$q$) is true.

The following example illustrates that even if an atom is true in the \emph{unique} answer set, the answer set provenance (as given by \changed{Definition~\ref{def:ans.provenance})} may still point out that modifications are needed to make the atom true. This is because a unique answer set may not be a complete well-founded model.

\begin{examplecont}{ex:why-not.cycle}\label{ex:why-not.cycle.constraint}
Let $\newprogram\label{prg:why-not.cycle.constraint}$ be the program
\begin{gather*}
r_1 : \ p \lparrow \Not q
\hspace{2cm}
r_2 : \ q \lparrow \Not p
\hspace{2cm}
r_3 : \ s \lparrow p \wedge \Not s
\end{gather*}
obtained by adding rule $r_3$ to program~$\programr\ref{prg:cycle}$.
This program has a unique answer set $\newanswerset\label{as:prg:why-not.cycle.constraint} = \set{q}$.
Furthermore, adding rule $r_3$ to program~$\programr\ref{prg:cycle}$ does not change the why-not provenance information of $p$ or $q$.
The computation of the why-not provenance information for $s$ goes as follows:
\begin{gather*}
\begin{IEEEeqnarraybox}{l C l}
\WpP{ \wP_{\ref{prg:why-not.cycle.constraint}} }^1(\botI)(s) &=&  \neg not(s) \vee r_3 \!\wedge\! \neg not(p) \vee r_3 \!\wedge\! r_1
\\
\WpP{ \wP_{\ref{prg:why-not.cycle.constraint}} }^2(\botI)(s)
  &=&  \neg not(s)
\\
\WpP{ \wP_{\ref{prg:why-not.cycle.constraint}} }^3(\botI)(s) 
  &=& \neg not(s) \vee r_3 \!\wedge\! \neg not(p) \vee r_3 \!\wedge\! r_1 \!\wedge\! not(q)
\\
\WpP{ \wP_{\ref{prg:why-not.cycle.constraint}} }^4(\botI)(s)
  &=& \neg not(s)
\end{IEEEeqnarraybox}
\end{gather*}
and we obtain
\begin{gather*}
\begin{IEEEeqnarraybox}{l C l l c l c l }
Why_{ \programr\ref{prg:why-not.cycle.constraint} }(s)       
  &=&  \neg& not(s)
\\
Why_{ \programr\ref{prg:why-not.cycle.constraint} }(\Not s) 
  &=&  & not(s) \!\wedge\! \neg r_3
  \vee
  not(s) \!\wedge\! not(p) \!\wedge\! \neg r_1
  \vee
  not(s) \!\wedge\! not(p) \!\wedge\! \neg not(q)
\\
Why_{ \programr\ref{prg:why-not.cycle.constraint} }(\Undef s)
  &=&  \IEEEeqnarraymulticol{3}{l}{
    \phantom{\neg} r_3 \!\wedge\! not(s) \!\wedge\! \neg not(p) \ \vee \ r_1 \!\wedge\! r_3 \!\wedge\! not(s) \!\wedge\! not(q)
    }
\end{IEEEeqnarraybox}
\end{gather*}
That is, $s$ is undefined in the well-founded model because of rules~$r_1$ and~$r_3$ and the absence of the facts~$s$ and~$q$.
It would also be undefined if we added the fact~$p$ while keeping the rule~$r_3$ and the absence of~$s$.
Furthermore,
$AnsWhy_{ \programr\ref{prg:why-not.cycle.constraint} }(\Undef p)
 = AnsWhy_{ \programr\ref{prg:why-not.cycle.constraint} }(\Undef q)$
and, thus,
$\neg AnsWhy_{ \programr\ref{prg:why-not.cycle.constraint} }(\Undef p) \wedge
\neg AnsWhy_{ \programr\ref{prg:why-not.cycle.constraint} }(\Undef q) \wedge
\neg AnsWhy_{ \programr\ref{prg:why-not.cycle.constraint} }(\Undef s) \ \ = \ \
\neg AnsWhy_{ \programr\ref{prg:why-not.cycle.constraint} }(\Undef p) \wedge
\neg AnsWhy_{ \programr\ref{prg:why-not.cycle.constraint} }(\Undef s)$
which corresponds to
\begin{gather*}
 \neg( not(p) \!\wedge\! not(q) \!\wedge\! r_1 \!\wedge\! r_2)
\ \wedge \
\neg(r_3 \!\wedge\! not(s) \!\wedge\! \neg not(p) \ \vee \ r_1 \!\wedge\! r_3 \!\wedge\! not(s) \!\wedge\! not(q))
\end{gather*}
We also have that
\begin{gather*}
\begin{IEEEeqnarraybox}{l C l c l c l }
Why_{ \programr\ref{prg:why-not.cycle.constraint} }(q)     
   &=&  \neg not(q) &\ \vee \ & r_2 \!\wedge\! not(p) \!\wedge\! \neg r_1
\end{IEEEeqnarraybox}
\end{gather*}
This implies that the answer set provenance information for $q$ is:
\begin{gather*}
\begin{IEEEeqnarraybox}{l ?C? C r l c l c l }
AnsWhy_{ \programr\ref{prg:why-not.cycle.constraint} }(q)
    &=& 
  &\neg& not(q) \wedge \neg r_3
\\
  &&\vee&\neg& not(q) \wedge \neg not(s)
\\
  &&\vee&\neg& not(q) \wedge not(p)
\\
  &&\vee&\neg& r_1 \wedge r_2 \wedge not(p)
\end{IEEEeqnarraybox}
\end{gather*}
The disjuncts represent \changed{different} modifications of the program leading to the existence of 
a complete well-founded model (and, thus, a unique answer set), in which $q$ is true.\qed
\end{examplecont}

\CHANGED
Example~\ref{ex:why-not.cycle.constraint} can be used to illustrate how the notion of assumption, as introduced in Section~\ref{sec:offline}, can be applied to why-not provenance justifications.
In particular, the disjunct
$\neg r_1 \wedge r_2 \wedge not(p)$ 
in 
$AnsWhy_{ \programr\ref{prg:why-not.cycle.constraint} }(q)$
suggests removing all rules with $p$ in the head (just~$r_1$) and not adding the fact $p$ to the program. This can be understood as ``$p$ needs to be assumed to be false'' in a similar way as done in off-line or extended causal justifications. In order to make this informal reading about this last disjunct, we need to know that $p$ is actually false in the answer set that we are considering, i.e. $\answersetr\ref{as:prg:why-not.cycle.constraint} = \set{q}$, because
$AnsWhy_{ \programr\ref{prg:why-not.cycle.constraint} }(p)$ 
contains a symmetric disjunct
$\neg r_2 \wedge r_1 \wedge not(q)$
whose informal reading does not correspond to an assumption but to an actual modification.
This is not a surprise because why-not provenance (as an unsimplified formula) can be computed in polynomial time, while deciding whether some atom is true in some answer set of some normal program is, in general, NP-complete.
Hence, unless the polynomial hierarchy collapses, it is obvious that why-not provenance cannot contain information about whether some atom is true or false in some answer set.
Note also that, though extended causal justifications (as an unsimplified causal term) can be computed in polynomial time, they are construed \wrt\ a program reduced \wrt\ the set of assumptions corresponding to this answer set.
Hence, they assume the information of true atoms in an answer set as a given.
The same approach used to define extended causal justifications \wrt\ an answer set could be applied to why-not provenance as well.
\END


\subsection{Other Justification Approaches}\label{sec:other_justifications}

In this section we informally review two other approaches that deal with justifications in answer set programming, namely \emph{justifications in rule-based answer set computation}~\cite{BeatrixLGS2016} and the \emph{formal theory of justifications}~\cite{DeneckerS1993,DeneckerBS2015}.
\CHANGED
Despite sharing a similar purpose with previous approaches, the formal definition of~\citeN{BeatrixLGS2016} heavily relies on the concept of \emph{\mbox{\tt ASPeRiX} computation}~\cite{LefevreBSG17} and is out of the scope of this survey.
On the other hand, the purpose of the works by~\citeN{DeneckerS1993} and~\citeN{DeneckerBS2015} is to study different semantics of logic programming from the point of view of justifications rather than to provide explanations that are ``intelligible and easily accessible'' by humans, as required by the new GDPR.
\END

\subsubsection{Justifications in Rule-Based Answer Set Computation}
\label{sec:rule-based}

 \citeN{BeatrixLGS2016} study the notion of justification from a rule-based point of view of answer set computation, that is, under the assumption that the inherent non-determinism of answer sets is due to the guessing of the application or non-application of rules rather than the guessing of the truth value of literals.
 Another interesting point to mention is that justifications in this approach, called \emph{reasons}, are sets of rules instead of graphs. The following example illustrates these two differences.

\begin{samepage}
\begin{example}\label{ex:rule.cylce}\nopagebreak
Consider the following program~$\newprogram\label{prg:rule.cycle}$:
\begin{gather*}
r_1 : \ p \lparrow t \wedge \Not q
\hspace{1.25cm}
r_2 : \ q \lparrow s
\hspace{1.25cm}
r_3 : \ s \lparrow \Not p
\hspace{1.25cm}
t : \ t
\end{gather*}
which has two answer sets: $\newanswerset\label{as:1:prg:rule.cycle} = \set{p,t}$ and $\newanswerset\label{as:2:prg:rule.cycle} = \set{q,s,t}$.
The rule-based reason for the truth of the atom $p$ with respect to the answer set~$\set{p,t}$ of the program~$\programr\ref{prg:rule.cycle}$
is the set~$\set{r_1, t}$.
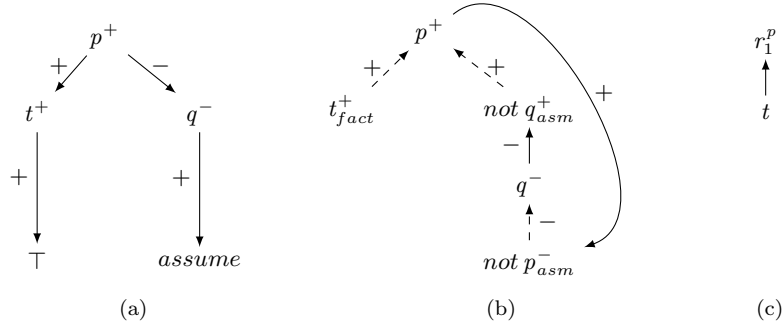
\begin{figure}\centering
\subfloat[]{%
\begin{tikzpicture}[tikzpict]
    \matrix[row sep=0.5cm,column sep=0.3cm,ampersand replacement=\&] {
      \&\node (p) {$p^{+}$};
      \\
      \node (t) {$t^{+}$};
      \&\&
      \node (q) {$q^{-}$};
      \\
      \\
      \\
      \node (top) {$\top$};
      \&\&
      \node (assume) {$\assume$};
      \\
     };
    \draw [->] (p) to node[pos=0.3,right]{$-$}  (q);
    \draw [->] (p) to node[pos=0.3,left]{$+$}  (t);
    \draw [->] (q) to node[pos=0.4,left]{$+$} (assume);
    \draw [->] (t) to node[pos=0.4,left]{$+$} (top);
\end{tikzpicture}%
\label{fig:rule.cylce.off-line}
}
\hspace{0.5cm}
\subfloat[]{%
\begin{tikzpicture}[tikzpict]
    \matrix[row sep=0.5cm,column sep=0.3cm,ampersand replacement=\&] {
      \&\node (p) {$p^{+}$};
      \\
      \node (t) {$t_{\mathit{fact}}^{+}$};
      \&\&
      \node (notq) {${\Not q}_{\mathit{asm}}^{+}$};
      \\
      \&\&\node (q) {$q^{-}$};
      \\
      \&\&
      \node (notp) {${\Not p}_{\mathit{asm}}^{-}$};
      \\
     };
     \draw [->,dashed] (t) to node[left]{$+$}  (p);
     \draw [->,dashed] (notq) to node[right]{$+$}  (p);
     \draw [<-] (notq) to node[left]{$-$}  (q);
     \draw [->,dashed] (notp) to node[right]{$-$}  (q);
     \draw [<-,bend right=100] (notp) to node[right]{$+$}  (p);
\end{tikzpicture}
\label{fig:rule.cylce.labas}
}
\hspace{0.5cm}
\subfloat[]{%
\begin{tikzpicture}[tikzpict]
    \matrix[row sep=0.5cm,column sep=0.3cm,ampersand replacement=\&] {
      \node (p) {$r_1^p$};
      \\
      \node (t) {$t$};
      \\
      \\
      \\
      \\
      \\
     };
    \draw [<-] (p) to (t);
\end{tikzpicture}%
\label{fig:rule.cylce.causal}
}
\caption{Off-line, LABAS, and causal justifications of $p$ w.r.t. $\set{p,t}$ and~$\programr\ref{prg:rule.cycle}$.}
\label{fig:rule.cylce.full}
\end{figure}
\qed
\end{example}
\end{samepage}

\CHANGED
We may use Example~\ref{ex:rule.cylce} to highlight some similarities and differences with previously discussed justification approaches.
It can be checked that the causal graph justification (Figure~\ref{fig:rule.cylce.causal}) for $p$ in this example has precisely vertices $t$ and $r_1^p$, corresponding to the rule-based reason.
\END
Correspondences with the \mbox{off-line} justification, shown in Figure~\ref{fig:rule.cylce.off-line}, are also easy to see: the application of rule $r_1$ is represented by the two outgoing edges from $p^{+}$ to $t^{+}$ and to $q^{-}$,
where assuming $q^{-}$ to be false ensures that $r_1$ is satisfied.
\CHANGED
Similarly, the answer set why-not provenance of $p$ includes the disjunct
$r_1 \wedge t \wedge not(q) \wedge \neg r_2$,
where $not(q) \wedge \neg r_2$ can be understood to mean that $q$ is assumed to be false.
The LABAS justification, 
shown in Figure~\ref{fig:rule.cylce.labas}, further explains that the falsity of $q$ depends on the truth of $p$, thus also using rules $r_2$ and $r_3$ for the explanation.
Interestingly, the answer set why-not provenance of $p$ has another
disjunct
$r_1 \wedge t \wedge not(q) \wedge \neg r_3$, which also uses rule $r_3$ to justify $p$.
\END

\changed{Note that the rule-based reason for the falsity of $q$ \wrt\ $\answersetr\ref{as:1:prg:rule.cycle}$ is a subset of the reason for $p$, namely $\set{r_1}$.}
This contrasts with off-line and causal justifications, in which $q$ is assumed to be false, and LABAS justifications, in which $q$ is explained in the same way as in the justification of $p$ (flipping the justification in Figure~\ref{fig:rule.cylce.labas} so that $q$ is at the top coincides with the LABAS justification of $q$), i.e. in terms of $r_3$ (and implicitly $r_2$) as well as $r_1$ and~$t$.
\CHANGED
The answer set why-not provenance of $\Not q$ includes the disjunct
$not(q) \wedge \neg r_2$ which, as mentioned before, can be understood as assuming that $q$ is false.
\END


\subsubsection{Formal Theory of Justifications}
\label{sec:formal}

\citeN{DeneckerS1993} and \citeN{DeneckerBS2015} present an abstract theory of justifications, suitable for describing the semantics of logics in knowledge representation and computational and mathematical logic.
In this theory, each program induces a semantic structure called justification frame, which embodies the potential \emph{reasons why} the program's conclusions are true.
Interestingly, the authors show that differences in various semantics can be traced back to a single difference, namely the way in which justifications with infinite branches are handled.
For instance, $p$ is justified w.r.t. program~$\programr\ref{prg:cycle} = \set{ p \lparrow \Not q,\ q \lparrow \Not p }$ by the following infinite branch:
\begin{gather*}
p \ \to \ \Not q \ \to \ p \ \to \ \Not q \ \to \ \dotsc
\end{gather*}
This is evaluated as undefined under the well-founded semantics (infinite branches altering positive and negative literals are always evaluated as undefined under the well-founded semantics).
In contrast, it takes the value of $\Not q$ under the answer set semantics
(under the answer set semantics infinite branches are evaluated to the truth value of  the first positive (resp. negative) literal whose predecessors are all negative (resp. positive) literals),
 which is true w.r.t. answer set $\set{p}$, but false w.r.t.~$\set{q}$.

Contrary to the other approaches surveyed here, this work focuses on exploiting justifications as mathematical objects to understand different semantics (and propose new ones) rather than as a means to answer in a compact way, \emph{why} a conclusion has been reached.
\changed{The \emph{complete justifications} defined in the formal theory of justifications are thus structures that contain information for all literals, even those that are not directly related to the derivation of a literal in question.
As an explanation in the sense of the new GDPR, complete justifications are thus not suitable as they are clearly not ``concise''
and likely not ``intelligible and easily accessible'', as they comprise information unnecessary for a user's understanding.}
Studying how
\changed{concise and intelligible}
 justifications can be obtained from this structures is an interesting open topic as it would be directly applicable to several logics and knowledge representation formalisms like argumentation.

\vspace{2cm}
\subsection{Summary and Discussion}
\label{sec:discussion}

 \begin{table}
 \caption{Comparison of explanation approaches for consistent logic programs under the answer set semantics.}
 \label{tab:comparison:explanation}
 \begin{tabular}{ l l l l l } 
  \multirow{2}{2.5cm}{\textbf{\changed{justification approach}}} & \multirow{2}{2cm}{\textbf{\changed{type of logic program}}} & \multirow{2}{2cm}{\textbf{explanation in terms of}} & \multirow{2}{1.5cm}{\textbf{derivation steps included}} & \multirow{2}{2cm}{\textbf{explains}} \\ \\ \\
  \hline
  \multirow{2}{2.5cm}{off-line justifications} & normal LP & \multirow{2}{2.5cm}{literal dependency} & all & \multirow{2}{2cm}{one literal (not) in answer set} \\ \\ \\
  \hline
  \multirow{2}{2.5cm}{LABAS justifications} & \multirow{2}{2cm}{normal extended LP} & \multirow{2}{2.5cm}{literal dependency} & some & \multirow{2}{2cm}{one literal (not) in answer set} \\ \\ \\
  \hline
  \multirow{2}{2.5cm}{causal justifications} &
  \multirow{2}{2.5cm}{extended LP with nested expressions in the body}
   & \multirow{2}{2.5cm}{rule-literal dependency} & all & \multirow{2}{2cm}{one literal in answer set} \\ \\ \\ \\
  \hline
  \multirow{2}{2.5cm}{extended causal justifications} & \multirow{2}{2cm}{normal extended LP} & \multirow{2}{2.5cm}{rule-literal dependency} & all &\multirow{2}{2cm}{one literal (not) in answer set}\\ \\ \\
  \hline

  \multirow{2}{2.5cm}{why-not provenance} & normal LP & \multirow{2}{2cm}{rule dependency} & all & \multirow{2}{2cm}{one literal (not) in the complete well-founded model\footnotemark} \\ \\ \\ \\ \\
  \hline
    \multirow{2}{2.5cm}{rule-based justifications} & normal LP & \multirow{2}{2cm}{rule dependency} & all & \multirow{2}{2cm}{one literal (not) in answer set}\\ \\ \\
  \hline
  \multirow{2}{2.5cm}{formal theory \changed{of justifications}} & normal LP & \multirow{2}{2cm}{literal dependency} & all & \multirow{2}{2cm}{whole answer set}\\ \\
  \hline
 \end{tabular}
 \end{table}

 \begin{table}
 \caption{Comparison of explanation approaches for consistent logic programs under the answer set semantics (continued).}
 \label{tab:comparison:explanation2}
 \begin{tabular}{ l l l l l } 
  \multirow{2}{2.5cm}{\textbf{\changed{justification approach}}} & \multirow{2}{2cm}{\textbf{computation uses other models}} & \multirow{2}{2cm}{\textbf{explanation of negative literals}} & \multirow{2}{2cm}{\textbf{infinite explanations}} & \multirow{2}{2cm}{\textbf{infinitely many explanations}} \\ \\ \\
  \hline
  \multirow{2}{2.5cm}{off-line justifications} & \multirow{2}{2cm}{well-founded model} & \multirow{2}{2cm}{assumed or further explained} & \multirow{2}{2cm}{no, if the program is finite} & \multirow{2}{2cm}{no, if the program is finite}\\ \\ \\
  \hline
  \multirow{2}{2.5cm}{LABAS justifications} & no & \multirow{2}{2cm}{further explained} & yes & yes\\ \\
  \hline
  \multirow{2}{2.5cm}{causal justifications} & no & assumed & no & \multirow{2}{2cm}{no, if the program is finite}\\ \\ \\
  \hline
  \multirow{2}{2.5cm}{extended causal justifications} & \multirow{2}{2cm}{well-founded model}  & \multirow{2}{2cm}{assumed or further explained} & \multirow{2}{2cm}{no, if the program is finite} & \multirow{2}{2cm}{no, if the program is finite}\\ \\ \\
  \hline
  \multirow{2}{2.5cm}{why-not provenance} &  \multirow{2}{2cm}{(do not need answer sets)} & \multirow{2}{2cm}{further explained} & \multirow{2}{2cm}{no, if the program is finite} & \multirow{2}{2cm}{no, if the program is finite}\\ \\ \\ \\
  \hline
    \multirow{2}{2.5cm}{rule-based justifications} & no  & \multirow{2}{2cm}{further explained} &  \multirow{2}{2cm}{no, if the program is finite} &  \multirow{2}{2cm}{no, if the program is finite} \\ \\ \\
  \hline
  formal theory & no  & \multirow{2}{2cm}{further explained} & yes &  \multirow{2}{2cm}{no, if the program is finite} \\ \\ \\
  \hline
 \end{tabular}
 \end{table}

In Sections~\ref{sec:offline} to~\ref{sec:other_justifications} we have surveyed the most prominent approaches for justifying the solutions to consistent logic programs under the answer set semantics.
Note that throughout these sections, by referencing an answer set to justify, we implicitly assumed that logic programs are consistent.
While explaining the justification approaches, we already pointed out differences and similarities between these approaches.
Some of these are reiterated in Tables~\ref{tab:comparison:explanation} and~\ref{tab:comparison:explanation2}, which provide a comparative overview of various features of the justification approaches.

\CHANGED
Table~\ref{tab:comparison:explanation} illustrates for which types of logic programs the different justification approaches are defined, in which terms they explain answer sets (i.e. dependencies between rules or literals), whether all parts of a literal's derivation are included in a justification, and what precisely is being explained, i.e. a literal in an answer set, a literal not contained in an answer set, or a whole answer set.
\footnotetext{\CHANGED The why-not provenance corresponding to each answer set can then be obtained by forcing the atoms not in the answer set as assumptions, similarly as done done for extended causal justifications.}
Table~\ref{tab:comparison:explanation2} complements this comparison, by showing whether the justification approaches make use of logic programming models other than the answer set in question when constructing a justification, whether negative literals occur in justifications and, if so, how their truth value is explained, whether justifications may be infinite, and whether there may be infinitely many justifications.
\END

In the following, we discuss some of the differences between the justification approaches in more detail and highlight some of their advantages and disadvantages.
\CHANGED
In particular, we focus on the philosophical ideas underpinning the different justifications approaches~ (Section~\ref{sec:explanatory.elements}), the problem of having exponentially many justifications (Section~\ref{sec:exponentiall}), how different approaches deal with negation-as-failure (Section~\ref{sec:negation}),
and the issues faced when dealing with large logic programs (Section~\ref{sec:large}).
\END

\subsubsection{Explanatory Elements}
\label{sec:explanatory.elements}
Due to the usage of different definitions of answer set, the different justifications embody distinct ideas. 
For instance, the intuition of \mbox{off-line} justifications (Section~\ref{sec:offline}) can be traced back to Prolog tabled justifications~\cite{roychoudhuryRR00},
LABAS justifications (Section~\ref{sec:labas}) have an argumentative flavour and are based on a correspondence between logic programs and their translation into argumentation frameworks \cite{SchulzT2015,SchulzT2016},
while causal justifications (Section~\ref{sec:causal}) rely on a causal interpretation of rules and the idea of causal chain~\cite{lewis1973causation}.
Despite their differences, these three approaches share the fact that they explain why a literal belongs to some answer set using a ``concise'' graph structure (in the sense that these graphs do not contain information not related to the literal in question).

The \mbox{why-not} provenance (Section~\ref{sec:why-not}), which is based on the concept of provenance inherited from the database literature~\cite{GreenKT07}, shares with these approaches the idea of building concise justifications for each literal.
However, why-not provenance justifications are set-based (instead of graph-based) and are built without referring 
to a specific answer set, so justifications are answer set independent.
\CHANGED
The justifications for a particular answer set can be obtained by ``forcing'' the appropriate assumptions as done in extended causal justifications.
\END

A similar point of view is also shared by rule-based justifications (Section~\ref{sec:rule-based}), which are based on the concept of an \emph{\mbox{\tt ASPeRiX} computation}~\cite{LefevreBSG17}.
Conceptually, the major difference between this and the previously mentioned approaches lies in what is considered as assumptions,
i.e. as elements that do not need to be further justified:
rules in the case of rule-based justifications and literals in the case of the other approaches.


Finally, the formal theory of justifications~(Section~\ref{sec:formal})
aims to explain the differences between different logic programming semantics by identifying how their conclusions are justified.
Contrary to the other approaches, it provides justifications for a whole answer set instead of concise justifications for each literal.
This is similar to debugging systems 
(which we will overview in Section~\ref{sec:debugging}), which explain why a whole set of literals is not an answer set, rather than explaining a specific literal.

\subsubsection{The Problem of Exponentially Many Justifications}
\label{sec:exponentiall}

As mentioned in the introduction, a key point for a human-understandable answer to the question of \emph{why} some conclusion is reached is its conciseness. 
Most justification approaches reviewed here have tackled this issue and provide justifications that only contain information related to the literal in question.
\CHANGED
However, a second issue related to conciseness is how many justifications there are.
In this section, we show that the number of justifications is in general exponential \wrt\ the size of the program.
Let us start by continuing here the discussion about the light bulb scenario introduced in Example~\ref{ex:light} (page~\pageref{ex:light}).
\END


\begin{examplecont}{ex:light}\label{ex:discussion.light}
\CHANGED
Recall that the program $\programr\ref{prg:light}$ representing this scenario consists of the following rules:
\begin{gather*}
\begin{IEEEeqnarraybox}{l C l}
r_{1t+1} : \ \on_{t+1}  &\lparrow& \swa_t \wedge \swb_t
\\
r_{2t+1} : \ \off_{t+1} &\lparrow& \swc_t \wedge \swd_t
\end{IEEEeqnarraybox}
\hspace{1.5cm}
\begin{IEEEeqnarraybox}{l C l}
i_{1t+1} : \ \on_{t+1}  &\lparrow& \on_{t} \wedge \Not \off_{t+1}\\
i_{2t+1} : \ \off_{t+1} &\lparrow& \off_{t} \wedge \Not \on_{t+1}
\end{IEEEeqnarraybox}
\end{gather*}
plus the integrity constraint 
$\lparrow \on_t \wedge \off_t$ 
for $t \geq 0$ and the facts $\off_0$, $\swa_0$ and $\swb_0$.
\begin{figure}[t]\centering
\subfloat[]{%
\begin{tikzpicture}[tikzpict]
    \matrix[row sep=0.5cm,column sep=0.3cm,ampersand replacement=\&] {
      \&
      \node (on1) {$\on_1^{+}$};
      \\
      \node (swa0) {$\swa_0^{+}$};
      \&\&
      \node (swb0) {$\swb_0^{+}$};
      \\
      \&
      \node (top) {$\top$};
      \\
     };
    \draw [->] (on1) to node[pos=0.3,left]{$+$}  (swa0);
    \draw [->] (on1) to node[pos=0.3,right]{$+$}  (swb0);
    \draw [->] (swa0) to node[pos=0.4,left]{$+$} (top);
    \draw [->] (swb0) to node[pos=0.4,right]{$+$} (top);
\end{tikzpicture}%
\label{fig:on1:off-line}
}
\hspace{0.4cm}
\subfloat[]{%
\begin{tikzpicture}[tikzpict]
    \matrix[row sep=0.5cm,column sep=0.3cm,ampersand replacement=\&] {
      \&
      \node (on1) {$\on_1^{+}$};
      \\
      \node (swa0) {${\swa_0^{+}}_{fact}$};
      \&\&
      \node (swb0) {${\swb_0^{+}}_{fact}$};
      \\[15pt]
      \\
     };
     \draw [<-,dashed] (on1) to node[pos=0.3,left]{$+$}  (swa0);
     \draw [<-,dashed] (on1) to node[pos=0.3,right]{$+$}  (swb0);
\end{tikzpicture}%
\label{fig:on1:labas}
}
\hspace{0.4cm}
\subfloat[]{%
\begin{tikzpicture}[tikzpict]
    \matrix[row sep=0.5cm,column sep=0.3cm,ampersand replacement=\&] {
      \&
      \node (on1) {$r_{11}^{\on_1}$};
      \\
      \node (swa0) {$\swa_0$};
      \&\&
      \node (swb0) {$\swb_0$};
      \\[15pt]
      \\
     };
    \draw [<-] (on1) to (swa0);
    \draw [<-] (on1) to (swb0);
\end{tikzpicture}%
\label{fig:on1:causal.discussion}
}
\caption{Off-line, LABAS, and causal justifications of the truth of $\on_1$.}
\end{figure}
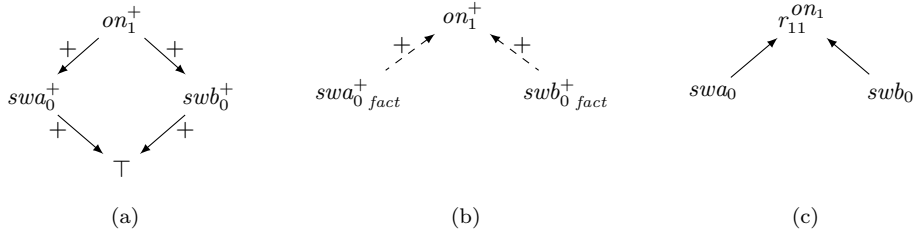
Recall also that this program has a complete well-founded model and, thus, a unique answer set, in which $\on_t$ holds for every time $t > 0$.
Figures~\ref{fig:on1:off-line},~\ref{fig:on1:labas} and~\ref{fig:on1:causal.discussion} respectively depict the off-line, the LABAS and the causal justification explaining why the light is $\on$ in situation~$1$.
We also have that the answer set why-not provenance of $\on_1$ corresponds to the following propositional formula:
\begin{IEEEeqnarray*}{ l C l}
AnsWhy_{ \programr\ref{prg:light} }(\on_1)
  &=&  \neg not(\on_1) \ \vee \ \neg not(\on_0) \wedge not(\off_1) \wedge i_{12} \ \vee \ \swa_0 \wedge \swb_0 \wedge r_{11}
\end{IEEEeqnarray*}
where
$\swa_0 \wedge \swb_0 \wedge r_{11}$
points out that $\on_1$ is true \wrt\ the unique answer set (which, here, coincides with the complete well-founded model) because of facts $\swa_0$ and~$\swb_0$ and rule~$r_{11}$.
It is easy to see the similarity with Figures~\ref{fig:on1:off-line},~\ref{fig:on1:labas} and~\ref{fig:on1:causal.discussion},
in particular that $\swa_0 \wedge \swb_0 \wedge r_{11}$ is precisely the conjunction of the three vertices in these justifications.
Informally, these justification can be read as ``because both switches $a$ and $b$ have been pushed in situation $0$''.

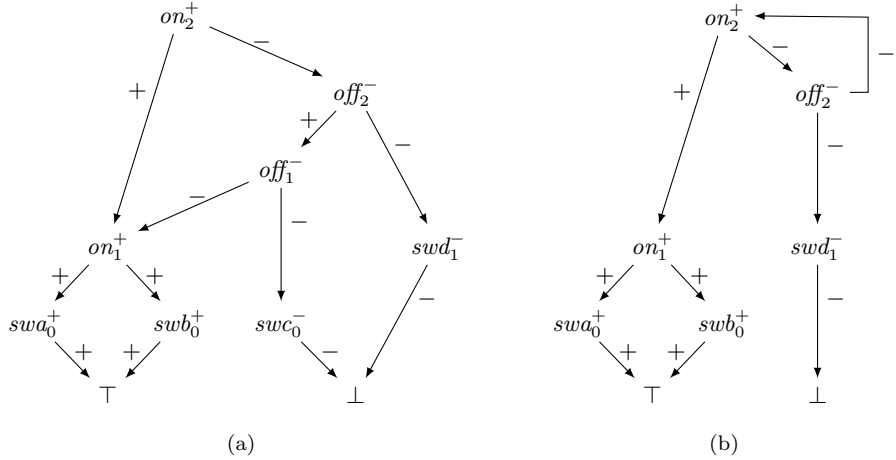
\begin{figure}[ht]\centering
\subfloat[]{%
\begin{tikzpicture}[tikzpict]
    \matrix[row sep=0.5cm,column sep=0.1cm,ampersand replacement=\&] {
      \&\& \node (on2) {$\on_2^{+}$};
      \\
      \&\&\&\&\&
      \node (noff2) {$\off_2^{\,-}$};
      \\
      \&\&\&\&
      \node (noff1) {$\off_1^{\,-}$};
      \&\&
      \\
      \&
      \node (on1) {$\on_1^{+}$};
      \&\&\&
      \&\&\&
      \node (swb1) {$\swd_1^{-}$};
      \\
      \node (swa0) {$\swa_0^{+}$};
      \&\&
      \node (swb0) {$\swb_0^{+}$};
      \&
      \node (aa) {$ $};\&
      \node (swc0) {$\swc_0^{-}$};
      \\
      \&
      \node (top) {$\top$};
      \&\&\&\&
      \node (bot) {$\bot$};
      \\
     };
    \draw [->] (on2) to node[pos=0.3,left]{$+$}  (on1);
    \draw [->] (on2) to node[pos=0.3,right]{$-$}  (noff2);
    \draw [->] (on1) to node[pos=0.3,left]{$+$}  (swa0);
    \draw [->] (on1) to node[pos=0.3,right]{$+$}  (swb0);
    \draw [->] (swa0) to node[pos=0.3,right]{$+$} (top);
    \draw [->] (swb0) to node[pos=0.3,left]{$+$} (top);
    \draw [->] (noff2) to node[pos=0.25,left]{$+$}  (noff1);
    \draw [->] (noff2) to node[pos=0.3,right]{$-$}  (swb1);
    \draw [->] (swb1) to node[pos=0.3,right]{$-$}  (bot);
    \draw [->] (noff1) to node[pos=0.3,left]{$-$}  (on1);
    \draw [->] (noff1) to node[pos=0.3,right]{$-$}  (swc0);
    \draw [->] (swc0) to node[pos=0.3,right]{$-$}  (bot);
\end{tikzpicture}%
\label{fig:inertia.a}
}
\hspace{0.5cm}
\subfloat[]{%
\begin{tikzpicture}[tikzpict]
    \matrix[row sep=0.5cm,column sep=0.1cm,ampersand replacement=\&] {
      \&\& \node (on2) {$\on_2^{+}$};
      \\
      \&\&\&\&\&
      \node (noff2) {$\off_2^{\,-}$};
      \\
      \\
      \\
      \&
      \node (on1) {$\on_1^{+}$};
      \&\&\&\&
      \node (swb1) {$\swd_1^{-}$};
      \\
      \node (swa0) {$\swa_0^{+}$};
      \&\&
      \node (swb0) {$\swb_0^{+}$};
      \\
      \&
      \node (top) {$\top$};
      \&\&\&\&
      \node (bot) {$\bot$};
      \\
     };
    \draw [->] (on2) to node[pos=0.3,left]{$+$}  (on1);
    \draw [->] (on2) to node[pos=0.3,right]{$-$}  (noff2);
    \draw [->] (on1) to node[pos=0.3,left]{$+$}  (swa0);
    \draw [->] (on1) to node[pos=0.3,right]{$+$}  (swb0);
    \draw [->] (swa0) to node[pos=0.3,right]{$+$} (top);
    \draw [->] (swb0) to node[pos=0.3,left]{$+$} (top);
    \draw [->] (noff2) to (2.25,1.5) to node[pos=0.5,right]{$-$} (2.25,2.5) to (on2);
    \draw [->] (noff2) to node[pos=0.3,right]{$-$}  (swb1);
    \draw [->] (swb1) to node[pos=0.3,right]{$-$}  (bot);
\end{tikzpicture}%
\label{fig:inertia.b}
}
\caption{Off-line justifications of $\on_2$ w.r.t. the unique answer set of~Example~\ref{ex:discussion.light}.}\label{fig:inertia}
\end{figure}

Let us now consider the justifications for the atom $on_2$, which is true \wrt\ the unique answer set.
Figure~\ref{fig:inertia} depicts two of the six possible off-line justifications for $on_2$.
Furthermore, 
by replacing $\swd_1^{-}$ with $\swc_1^{-}$
in Figures~\ref{fig:inertia.a} and~\ref{fig:inertia.b},
we obtain another two off-line justifications.
Similarly, by replacing $\swc_0^{-}$ with $\swd_0^{-}$ in Figure~\ref{fig:inertia.a},
we obtain another off-line justification
and, by replacing both $\swc_0^{-}$ and $\swd_1^{-}$ respectively with $\swd_0^{-}$ and $\swc_1^{-}$,
we obtain the sixth one.
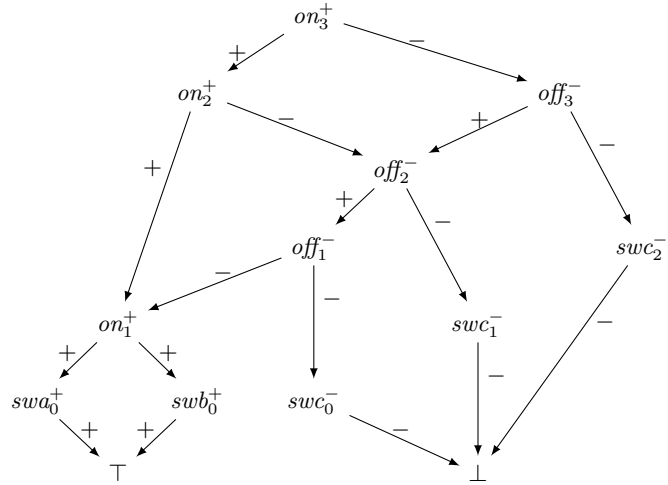
\begin{figure}[ht]\centering
\begin{tikzpicture}[tikzpict]
    \matrix[row sep=0.5cm,column sep=0.2cm,ampersand replacement=\&] {
      \&\&\&\&
      \node (on3) {$\on_3^{+}$};
      \\
      \&\& \node (on2) {$\on_2^{+}$};
      \&\&\&\&\&
      \node (noff3) {$\off_3^{\,-}$};
      \\
      \&\&\&\&\&
      \node (noff2) {$\off_2^{\,-}$};
      \\
      \&\&\&\&
      \node (noff1) {$\off_1^{\,-}$};
      \&\&
      \&\&
      \node (swb2) {$\swc_2^{-}$};
      \\
      \&
      \node (on1) {$\on_1^{+}$};
      \&\&\&
      \&\&
      \node (swb1) {$\swc_1^{-}$};
      \\
      \node (swa0) {$\swa_0^{+}$};
      \&\&
      \node (swb0) {$\swb_0^{+}$};
      \&
      \node (aa) {$ $};
      \&
      \node (swc0) {$\swc_0^{-}$};
      \\
      \&
      \node (top) {$\top$};
      \&\&\&\&\&
      \node (bot) {$\bot$};
      \\
     };
    \draw [->] (on3) to node[pos=0.5,left]{$+$}  (on2);
    \draw [->] (on2) to node[pos=0.3,left]{$+$}  (on1);
    \draw [->] (on2) to node[pos=0.3,right]{$-$}  (noff2);
    \draw [->] (on3) to node[pos=0.3,right]{$-$}  (noff3);
    \draw [->] (on1) to node[pos=0.3,left]{$+$}  (swa0);
    \draw [->] (on1) to node[pos=0.3,right]{$+$}  (swb0);
    \draw [->] (swa0) to node[pos=0.3,right]{$+$} (top);
    \draw [->] (swb0) to node[pos=0.3,left]{$+$} (top);
    \draw [->] (noff3) to node[pos=0.3,left]{$+$}  (noff2);
    \draw [->] (noff2) to node[pos=0.3,left]{$+$}  (noff1);
    \draw [->] (noff2) to node[pos=0.3,right]{$-$}  (swb1);
    \draw [->] (noff3) to node[pos=0.3,right]{$-$}  (swb2);
    \draw [->] (swb1) to node[pos=0.3,right]{$-$}  (bot);
    \draw [->] (noff1) to node[pos=0.3,left]{$-$}  (on1);
    \draw [->] (noff1) to node[pos=0.3,right]{$-$}  (swc0);
    \draw [->] (swc0) to node[pos=0.3,right]{$-$}  (bot);
    \draw [->] (swb2) to node[pos=0.3,right]{$-$}  (bot);
\end{tikzpicture}%
\caption{Off-line justification of $\on_3$ w.r.t. the unique answer set of~Example~\ref{ex:discussion.light}.}\label{fig:inertia.c}
\end{figure}
Figure~\ref{fig:inertia.c} depicts one of the off-line justifications of $\on_3^{+}$
and, by replacing any subset of $\set{\swc_1^{-}, \swc_2^{-}, \swc_3^{-}}$ by its corresponding subset of
$\set{\swd_1^{-}, \swd_2^{-}, \swd_3^{-}}$,
we obtain another $7$ alternative off-line justifications.
That is, the
number of off-line justifications grows exponentially with the number of situations in which nothing happens.
\begin{figure}\centering
\subfloat[]{%
\begin{tikzpicture}[tikzpict]
    \matrix[row sep=0.5cm,column sep=0.05cm,ampersand replacement=\&] {
      \&\node (on2) {${\on_2}_{A_1}^{+}$};
      \\
      \\
      \node (swa0) {${\swa_0}_{fact}^{+}$};
      \&
      \node (swb0) {${\swb_0}_{fact}^{+}$};
      \&
      \node (notOff2) {${\Not \off_2}_{asm}^{+}$};
      \\
      \\
      \&\&
      \node (off2) {${\off_2}_{A_2}^{-}$};
      \\
      \\
      \&\&
      \node (notOn1) {${\Not \on_1}_{asm}^{-}$};
      \\
      \\
      \&\&
      \node (on1) {${\on_1}_{A_3}^{+}$};
      \\
     };
    \draw [<-,dashed] (on2) to node [left]{$+$} (swa0);
    \draw [<-,dashed] (on2) to node [left]{$+$} (swb0);
    \draw [<-,dashed] (on2) to node [right]{$+$} (notOff2);
    \draw [->] (off2) to node [right]{$-$} (notOff2);
    \draw [->,dashed] (notOn1) to node [right] {$-$} (off2);
    \draw [->] (on1) to node [right] {$+$} (notOn1);
    \draw [<-,dashed] (on1) to node [left]{$+$} (swa0);
    \draw [<-,dashed] (on1) to node [left]{$+$} (swb0);
\end{tikzpicture}%
\label{fig:causal.inertia.labas}
}
\hspace{1cm}
\subfloat[]{%
\begin{tikzpicture}[tikzpict]
    \matrix[row sep=0.5cm,column sep=0.05cm,ampersand replacement=\&] {
      \&\node (on2) {${\on_2}_{A_1}^{+}$};
      \\
      \\
      \node (swa0) {${\swa_0}_{fact}^{+}$};
      \&
      \node (swb0) {${\swb_0}_{fact}^{+}$};
      \&
      \node (notOff2) {${\Not \off_2}_{asm}^{+}$};
      \\
      \\
      \&\&
      \node (off2) {${\off_2}_{A_2}^{-}$};
      \\
      \\
      \&\&
      \node (notOn2) {${\Not \on_2}_{asm}^{-}$};
      \\[28pt]
      \\
     };
    \draw [<-,dashed] (on2) to node [left]{$+$} (swa0);
    \draw [<-,dashed] (on2) to node [left]{$+$} (swb0);
    \draw [<-,dashed] (on2) to node [right]{$+$} (notOff2);
    \draw [->] (off2) to node [right]{$-$} (notOff2);
    \draw [->,dashed] (notOn2) to node [right] {$-$} (off2);
    \draw [->,bend left=80,looseness=1.5] (on2) to node [right] {$+$} (notOn2);
\end{tikzpicture}%
\label{fig:causal.inertia.labas2}
}
\caption{LABAS justifications of $\on_2$ w.r.t. the unique answer set of~Example~\ref{ex:discussion.light}.}\label{fig:causal.inertia}
\end{figure}
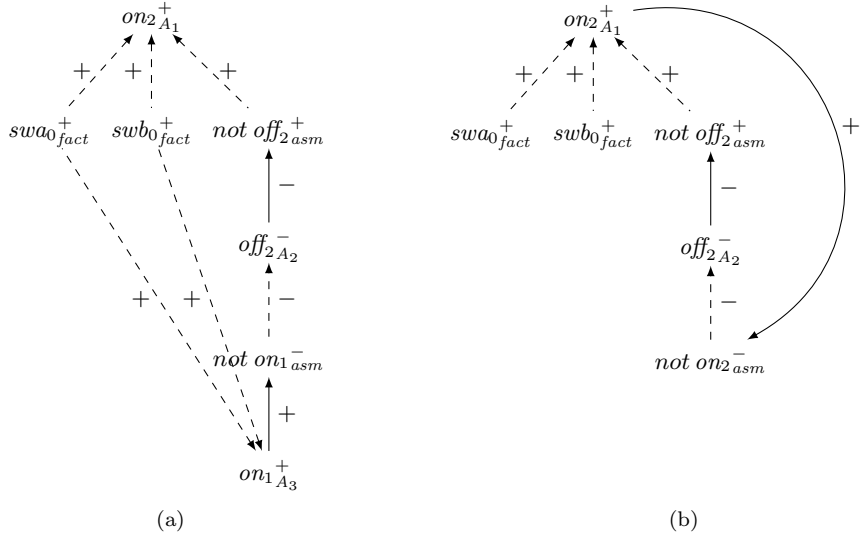
Similarly, the number of why-not justifications\footnote{\CHANGED Why-not information can be obtained in polynomial time and size \wrt\ the program. However, rewriting it as a disjunction of minimal conjuncts may require exponential space.} (i.e. disjuncts in the answer set provenance information) of~$\on_t$ grows exponentially, because the conjunction of all atoms in an off-line justification plus the rules used to derive those atoms form a why-not justification~\cite[Theorem~4]{DamasioAA2013}.
The number of LABAS justifications also grows exponentially. There are two LABAS justifications for $\on_2$, displayed in Figures~\ref{fig:causal.inertia.labas} and~\ref{fig:causal.inertia.labas2}. The reason for the exponential explosion is that $on_t$ can be justified through any $on_i$ with $i < t$. 
On the other hand, as explained in Section~\ref{sec:causal} (page~\pageref{ex:light}) (extended) causal justifications are somehow preserved by inertia in the sense that, at any situation $t+1$, if nothing happens, then the justification of $\on_{t+1}$ can be obtained by adding to the justification of $\on_t$ an edge from
$i_{1t}^{\on_{t}}$ to $i_{1t+1}^{\on_{t+1}}$.
For instance, Figure~\ref{fig:causal.inertia.causal.discussion} shows the unique (extended) causal justification of $\on_2$.
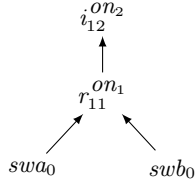
\begin{figure}\centering
\begin{tikzpicture}[tikzpict]
    \matrix[row sep=0.5cm,column sep=0.05cm,ampersand replacement=\&] {
      \&\node (on2) {$i_{12}^{\on_2}$};
      \\
      \&
      \node (on1) {$r_{11}^{\on_1}$};
      \\
      \node (swa0) {$\swa_0$};
      \&\&
      \node (swb0) {$\swb_0$};
      \&
      \node (aa) {$ $};\&
      \\
     };
    \draw [<-] (on2) to  (on1);
    \draw [<-] (on1) to  (swa0);
    \draw [<-] (on1) to  (swb0);
\end{tikzpicture}%
\caption{The unique causal justification of $\on_2$ w.r.t. the unique answer set of~Example~\ref{ex:discussion.light}.}\label{fig:causal.inertia.causal.discussion}
\end{figure}\qed
\end{examplecont}

Despite the fact that understanding negation-as-failure as a default allows to exponentially reduce the number of \changed{causal} justifications on some knowledge representation scenarios as illustrated by the above example, there still exist logic programs that \changed{produce} an exponential number of \changed{causal} justifications:



\begin{example}\label{ex:exponential}
Consider the following logic program adapted from~\cite{CabalarFF14}:
\begin{gather*}
\begin{IEEEeqnarraybox}[][t]{lCl"l}
p_1 & \leftarrow q_1
\\
p_1 & \leftarrow u_1
\end{IEEEeqnarraybox}
\hspace{1cm}
\begin{IEEEeqnarraybox}[][t]{lCl"l}
p_{i}		& \leftarrow &	p_{i-1} \wedge q_{i}
						&	\text{for }  i\in\set{2,\dotsc,n}
\\
p_{i}		& \leftarrow &	p_{i-1} \wedge u_{i}
						&	\text{for }  i\in\set{2,\dotsc,n}
\\
q_{i} && &  \text{for }  i\in\set{1,\dotsc,n}
\\
u_{i} && &  \text{for }  i\in\set{1,\dotsc,n}
\end{IEEEeqnarraybox}
\end{gather*}
whose unique answer set is $\set{p_1,q_1,u_1\dotsc,p_n,q_n,u_n}$.
Note that $p_1$ can be justified using the facts~$q_1$ or~$u_1$;
the atom $p_2$ can be justified using the sets of facts $\set{q_1,q_2}$,
$\set{q_1,u_2}$, $\set{u_1,q_2}$ or $\set{u_1,u_2}$; and so on.
It is easy to see that atom~$p_n$ can be justified using $2^n$ different sets of facts and, thus, that the number of justifications grows exponentially with respect to the size of the program.
\qed
\end{example}

Although this logic program has no deeper knowledge representation meaning, it points out a potential problem regarding the human-readability of the answers provided by current justification approaches.
\CHANGED
The issue of an exponential number of justifications illustrated by Example~\ref{ex:exponential} holds 
for any justification approach that records minimal sets of facts
used to derive the justified atom, in particular, all justification approaches reviewed here.
This does not mean that other kinds of polynomial justifications can be used.
For instance,
for causal justifications or \mbox{why-not} provenance, a non-simplified formula could be
returned and, if we consider such a formula as the justification, then it would be polynomial.
In our running example, we would have that $p_n$ is justified by the causal term
$(q_1 + u_1) * (q_2 + u_2) * \dotsc *  (q_n + u_n)$
or the why-not provenance formula
$(q_1 \vee u_1 \vee \neg not(p_1)) \wedge (q_2 \vee u_2 \vee \neg not(p_2)) \wedge \dotsc \wedge (q_n \vee u_n \vee \neg not(p_n))$.
On the other hand, these non-simplified expressions are not minimal and, thus, they do not adhere to the desired conciseness criterion for justifications.
Another alternative is to provide simplified justifications,  but selecting only some of them in case a some imposed preferences~\cite{CabalarFF14}.
\END
For instance, approaches in databases~\cite{Specht93} and Prolog~\cite{roychoudhuryRR00} implicitly impose such preferences by selecting only the first negative literal of a rule that fails as its unique justification.

\subsubsection{Interpreting Negation-as-Failure}
\label{sec:negation}
Related to the above exponentiality problem is the way in which different approaches interpret negative literals.
The definition of answer sets~\cite{GL88,GelfondL1991} is inherently non-deterministic:
a candidate set is (\mbox{non-deterministically}) selected and then checked against the program to see whether it is the minimal model of the reduct with respect to this candidate.
For normal logic programs, the checking part can be done deterministically in polynomial time, for instance, by iterating the well-known direct consequences operator introduced by~\mbox{\citeN{vEK76}};
but the non-determinism is still present in the selection of the candidate.
This non-determinism is handled by most justification approaches by considering some part of the justification as assumptions:
negative literals in the case of off-line, LABAS and causal justifications; and rules in the case of rule-based justifications
(formal theory of justifications takes a different approach, representing this by infinite branches).
Regarding the approaches that use negative literals as assumptions, a remarkable difference is how they do or do not justify those negative literals.
As the two extremes we have LABAS and causal justifications: the former justifies all negative literals (introducing cycles in the justifications when even-length negative dependency loops are present in the program),
while the latter treats all negative literals as assumptions, or rather defaults, that need no further explanation.
In the middle, we have off-line and extended causal justifications, which further explain some negative literals, while treating others as assumptions (when the set of assumptions is minimised, these approaches justify all negative literals that can be explained without introducing cycles in the justifications).

We have seen that treating negative literals as assumptions may help to (exponentially) reduce the number of justifications of some knowledge representation problems in which negation is used to express defaults.
Let us now illustrate the opposite case, with the following example from~\cite{SchulzT2016},
where justifications for negative literals are as important as those for positive literals:

\begin{example}
 \label{ex:doctor}
 The logic program $\newprogram\label{prg:doctor}$ in Figure~\ref{fig:prg.doctor} represents the decision support system used by an ophthalmologist.
 It encodes some general world knowledge as well as 
 an ophthalmologist's specialist knowledge about the possible treatments of shortsightedness.
  \begin{figure}[ht]
  \begin{align*}
  tightOnMoney &\lparrow student \wedge \Not  richParents\\
  caresAboutPracticality &\lparrow likesSports\\
  correctiveLens & \lparrow shortSighted \wedge \Not  laserSurgery\\
  laserSurgery &\lparrow shortSighted \wedge  \Not  tightOnMoney \wedge \Not  correctiveLens\\
  glasses &\lparrow correctiveLens \wedge \Not caresAboutPracticality \wedge\\ &\phantom{\lparrow \; \; } \Not contactLens\\
  contactLens &\lparrow correctiveLens \wedge \Not \textit{afraidToTouchEyes} \wedge\\ &\phantom{\lparrow \; \; } \Not longSighted \wedge \Not glasses\\
  intraocularLens &\lparrow correctiveLens \wedge \Not glasses \wedge \Not contactLens\\
  shortSighted &\lparrow\\
  \textit{afraidToTouchEyes} &\lparrow\\
  student &\lparrow\\
  likesSports &\lparrow
 \end{align*}
 \caption{Program~$\programr\ref{prg:doctor}$ from Example~\ref{ex:doctor}.}\label{fig:prg.doctor}
 \end{figure}
 $\programr\ref{prg:doctor}$ also captures the additional information that the ophthalmologist has about his shortsighted patient Peter.
 Program~$\programr\ref{prg:doctor}$ has a unique answer set 
\begin{align*}
 \newanswerset \ = \ \{ \ \textit{shortSighted},& \, \textit{afraidToTouchEyes},\,  \textit{student},\, \textit{likesSports},\, \textit{tightOnMoney},
 \\ &\hspace*{10pt}\textit{correctiveLens},\, \textit{caresAboutPracticality},\, \textit{intraocularLens} \ \}
\end{align*}
 Focusing on the positive dependencies on facts and not considering dependencies on negative literals, we can only say that Peter has been recommend to use an
 $\textit{intraocularLens}$ because he is $\textit{shortSighted}$.
 However, this reasoning could also lead to the recommendation of other treatments that have the same positive dependencies: $\textit{glasses}$,
 $\textit{contactLens}$ or $\textit{laserSurgey}$.
 Negative dependencies, on the other hand, tell us that 
 $\textit{intraocularLens}$ was recommended because all the other alternatives were discarded for different reasons: 
 $\textit{glasses}$ because Peter $\textit{likesSports}$,
 $\textit{contactLens}$ because he is $\textit{afraidToTouchEyes}$
 and
 $\textit{laserSurgey}$ because he is a $\textit{student}$ without $\textit{richParents}$.\qed
 \end{example}

The informal reading shown in the above example can be extracted from off-line, LABAS, extended causal, \CHANGED why-not provenace \END and rule-based justifications, but not from (non-extended) causal justifications.
A general approach to justifications should be able to effectively combine both interpretations of negation-as-failure, something which to the best of our knowledge has not been studied in the literature~yet.
 \color{black}
\subsubsection{Large Programs and Application-Oriented Considerations}
\label{sec:large}
Our comparison so far has concentrated on the theoretical, or even philosophical, nature of justification approaches.
Another important, and distinguishing, aspect of justification approaches is their applicability when solving real-world problems.
In such situations, various challenges arise.

Firstly, representing a real-world problem may result in a large logic program, where literals may have long derivations, i.e. their truth value depends on a large number of rules.
It is then not clear, which information a justification should comprise in order to be, on the one hand, succinct enough for humans to understand, but,
on the other hand, complete enough to provide all important information.
For example, justification approaches where all derivation steps are included in the justification, that is all approaches other than LABAS justifications, may struggle with the succinctness when explaining a large logic program, as explanations grow with longer derivations.
In contrast, LABAS justifications are independent of the derivation length. However, a large logic \changed{program} may also comprise more dependencies on negative literals, thus increasing the size of LABAS justifications.
More generally, it is an open problem how to effectively deal with the growing size (as well as the previously mentioned exponential number) of justifications.

In order to use justifications in real-world problems, they need to be automatically constructed.
Currently, only LABAS, causal and why-not provenance justifications have been implemented in working prototypes.\footnote{There also used to be an implementation of off-line justifications \cite{El-KhatibPS2005}, but this is not available anymore.}
\changed{A related issue is which type of logic programs can be explained. 
The only approach able to handle non-normal logic programs, i.e. logic programs with disjunctive heads, is the causal justification approach, which can also deal with nested expressions in the body.\footnote{In this survey, we have limited ourselves to normal extended logic programs. For a the definition of causal justifications for logic programs with nested expressions in the body, we refer to~\cite{Fandinno16}.}
Furthermore, in practice logic programs are rarely normal and often use additional language constructs, such as weight constraints, aggregates, and choice rules, which extend the syntax and/or semantics of logic programs under the answer set semantics.
Choice rules are handled by off-line justifications and in a limited way by causal justifications~\cite{CabalarF16}.
Note that explanations of additional language constructs have not been investigated so~far.}

As a last challenge, we mention variables. Even though the theory of most justification approaches can easily be applied to programs with variables by considering the complete grounding of the program, it is questionable if this method yields meaningful justifications in practice.
The difficulty of handling variables in explanations of inconsistent programs is a further indication that justifications involving variables are non-trivial, and therefore an interesting consideration for future work.

\section{Debugging of Inconsistent Logic Programs}\label{sec:debugging}
In this section, we review the most prominent approaches for explaining \emph{inconsistent} logic programs.
\changed{i.e. logic programs that have no answer set.}
Note that various approaches discussed in this section are not only applicable to inconsistent logic programs, but also to consistent ones.
More specifically, they can also be used to explain why a set of atoms of a \emph{consistent} logic program is not an answer set, or even why a set of atoms \emph{is} an answer set, and are 
thus closely related to the previously reviewed \changed{justification} approaches.

\changed{Finding errors that lead to a logic program being inconsistent} is often referred to as \emph{debugging}.
Errors can be roughly divided into \emph{syntactic} and \emph{semantic} ones.\footnote{Note that we here use these terms differently than e.g. \citeN{Syrjanen2006}.}
The first category, comprising for example misspelled literals and wrong rule layout, are handled by most IDEs (Integrated Development Environments) for ASP
such as \sealion\ \cite{BusoniuOPST2013}, ASPIDE~\cite{FebbraroRR2011}, and APE~\cite{SureshkumarVBF2007}.

Semantic errors are more difficult to identify due to the inherent declarative nature of the answer set semantics. 
In procedural programming languages, the cause of wrong program behaviour can be found by investigating the program procedure step-by-step.
This cannot be straightforwardly done for logic programs, as answer sets are computed in a `guess and check' fashion rather than procedurally.
\changed{Various approaches tackle this problem by searching for known error classes for inconsistent logic programs,} for example unfounded loops, unsupported atoms, and unsatisfied rules.
We review these approaches in Sections~\ref{sec:debugging:spock} to~\ref{sec:debugging:interactive}.
Another approach makes use of the unsatisfiable core feature of the ASP solver \wasp, which we review in Section~\ref{sec:debugging:dwasp}, and
Section~\ref{sec:debugging:stepping} outlines an approach for finding semantic errors that indeed applies a step-by-step procedure.
Finally, Section~\ref{sec:discussion.debugging} concludes the section with a discussion about similarities and differences between these debugging approaches.
\changed{Throughout this section, we will use the term `debugging' to refer to the task of finding and explaining \emph{semantic} errors in logic programs.}

\subsection{The \spock\ System -- Debugging with a Meta Program}\label{sec:debugging:spock}
The \spock\ system
explains why a \emph{potential} answer set, i.e. some set of atoms, is not an answer set of a the given 
program $\P$.
This is achieved by transforming $\P$ into a \emph{meta (logic) program}, expressing, for example,
conditions for the applicability of rules in $\P$.
Each answer set of this meta program contains the atoms of a potential answer set of $\P$
along with special atoms indicating reasons why this potential answer set is not an actual answer set of $\P$.
Thus, \spock\ uses \changed{answer sets of a meta logic program} for explaining the inconsistency of \changed{a given logic program}.

The \spock\ system is a command line tool\footnote{\url{http://www.kr.tuwien.ac.at/research/systems/debug/index.html}} usable with either the \texttt{DLV}~\cite{LeonePFEGPS2006} 
or \texttt{Smodels}~\cite{SyrjanenN2001} ASP solver.\footnote{\texttt{Smodels} is not maintained anymore and may thus not work on new systems. However, \spock\ should work fine on most systems using \texttt{DLV}.}
It implements two different approaches to transform $\P$ into a meta-program, where the second \cite{GebserPST08} was developed as a successor of the first \cite{BrainGPSTW2007b}. 
Both transformations distinguish three types of reasons for explaining why a set of atoms is not an answer set.
These reasons are different ways of violating the definition of answer sets as given by \citeNS{LinZ2004} and extended by \citeNS{Lee2005}.
\changed{Note that this definition of answer sets is equivalent to the one given in Section~\ref{sec:background}.}

\begin{definition}[Answer Set]\label{def:answerSet_LinZ_Lee}
 A set of atoms $M \subseteq \at$ is an answer set of a program $\P$ \ifonlyif
 \begin{enumerate}
  \item each rule $\R \in \P$ is \emph{satisfied} by $M$, i.e.
  \begin{itemize}
   \item $\head{\R} \cap M \neq \emptyset$ if $\R$ is applicable;
  \end{itemize}
  \item each atom $a \in M$ is \emph{supported} \wrt\ $M$, i.e. 
  \begin{itemize}
   \item there exists $\R \in \P$ such that $\R$ is applicable \wrt\ $M$ and $\head{\R} \cap M = \{a\}$;
  \end{itemize}
  \item each (positive dependency) loop $L \subseteq M$ is \emph{founded} \wrt\ $M$, where
  \begin{itemize}
   \item $L$ is a loop \ifonlyif\ for all $a \in L$ there is a chain of rules $\R_1, \ldots, \R_n \in \P$ ($n \geq 1$) such that 
  $a \in \head{\R_1} \cap \bodyp{\R_n}$, and if $n > 1$ then it holds for all $\R_i$ ($1 \leq i < n$) that $\exists b_i \in \bodyp{\R_i} \cap \head{\R_{i+1}}$ with $b_i \in L$, and 
   \item $L$ is founded \wrt\ $M$ \ifonlyif\ there exists $\R \in \P$ such that $\R$ is applicable and satisfied \wrt\ $M$,
   $\head{\R} \cap M \subseteq L$, and $\bodyp{\R} \cap L = \emptyset$.\qed
  \end{itemize}
 \end{enumerate}
\end{definition}

The third condition defines a loop as a set of atoms that positively depend on themselves, possibly via positive dependencies on other atoms in the loop.
Such a positive dependency loop is founded \wrt\ $M$ if there exists an applicable and satisfied rule that allows to derive some loop atoms without using other atoms in this loop.
An atom contained in an unfounded loop is said to be \emph{unfounded}.

Both transformation approaches of \spock\ generate reasons why a set of atoms $M$ is not an answer set in terms of violations of the three conditions in Definition~\ref{def:answerSet_LinZ_Lee}.
These reasons are:\footnote{\newChanged{\citeN{Lloyd1987} discusses a similar idea for diagnosing errors in Prolog programs in terms of incorrect rules (analogous to unsatisfied rules) and uncovered atoms (analogous to unsupported atoms).}}
\begin{enumerate}
 \item a rule $\R \in \P$ is not satisfied, 
 \item an atom $a \in M$ is not supported, 
 \item there exists an unfounded loop in $M$.
\end{enumerate}

In the following, we illustrate how the two transformation approaches generate these three reasons and point out some differences between the approaches.

\subsubsection{Transformation 1}\label{sec:debugging:spock1}
The first transformation approach \cite{BrainGPSTW2007a,BrainGPSTW2007b}, defined for \emph{normal} logic programs, can be used to explain
\begin{enumerate}
 \item \emph{why} a set of atoms \emph{is} an answer set, by referring to the
applicability and non-applicability of rules, and
 \item \emph{why} a set of atoms \emph{is not} an answer set, by referring to the
 violation (i.e. non-satisfaction) of rules, the unsupportedness of atoms, or the unfoundedness of atoms.
\end{enumerate}

To achieve the first, each rule $\R:  h \lparrow b_1 \wedge \dotsc \wedge b_n \wedge \Not c_1 \wedge \dotsc \Not c_m$ of a normal program $\P$ is transformed into two new rules\footnote{The transformed rules as originally
defined also have body literals $\ok{\R}$ and $\ko{\R}$ for fine-tuning the debugging process, which we omit as they do not play a role at this point.}
\begin{align}
\label{eq:spock1:appRule} \appRule{\R} &\lparrow b_1 \wedge \dotsc \wedge b_n \wedge \Not c_1 \wedge \dotsc \Not c_m \\
\label{eq:spock1:head} h &\lparrow \appRule{\R}
\end{align}
They respectively express that $\R$ is applicable if the body of $\R$ is true and that the head of $\R$ can be deduced if $\R$ is applicable.
Similarly, rules expressing conditions under which rule $\R$ is `blocked' are added, namely if one of its positive
body literals~$b$ or negative body literals $\Not c$ are false ($c^* \notin \at$ is a new atom).
\begin{align}
\label{eq:spock1:blockRuleP} \blockRule{\R} &\lparrow \Not b\\
\label{eq:spock1:blockRuleN} \blockRule{\R} &\lparrow \Not c^*\\
\label{eq:spock1:blockRuleN2} c^* &\lparrow \Not c
\end{align}
\changed{These transformed rules are added for each rule in the given program and each of its body literals.}

The transformation given by rules~\eqref{eq:spock1:appRule}-\eqref{eq:spock1:blockRuleN2} is called \emph{kernel transformation} of $\P$ and denoted $\transKern{\P}$.
For a consistent program $\P$, the answer sets of $\transKern{\P}$ coincide with the answer sets of $\P$,
but additionally contain the new \emph{tagging-atoms} $\appRule{\R}$ and $\blockRule{\R}$ \cite{BrainGPSTW2007b}.
This `explains' why a set of atoms is an answer set in the sense that it gives an insight into the rules that were used to derive the answer set.

\begin{examplecontpage}{ex:light}\label{ex:light.spock}
 The rules of the logic program from Example~\ref{ex:light} can be grounded for the first time step as follows,
 obtaining the logic program $\newprogram\label{prg:lights.spock}$:
 \begin{IEEEeqnarray*}{l C l C l}
 \R_1&: \ & \on_{1}  &\lparrow& \swa_0 \wedge \swb_0 \\
 \R_2&: \ & \off_{1} &\lparrow& \swc_0 \wedge \swd_0 \\
 \R_3&: \ & \on_{1}  &\lparrow& \on_{0} \wedge \Not \off_{1}\\
 \R_4&: \ & \off_{1} &\lparrow& \off_{0} \wedge \Not \on_{1}\\
 \R_5&: \ & \off_{0} &\lparrow& \\
 \R_6&: \ & \swa_0   &\lparrow& \\
 \R_7&: \ & \swb_0   &\lparrow& 
 \end{IEEEeqnarray*}
The only answer set of $\programr\ref{prg:lights.spock}$ is $\{\swa_0,\swb_0, \off_{0}, \on_{1}\}$.
 In comparison, the only answer set of $\transKern{\programr\ref{prg:lights.spock}}$ is
$\{\swa_0,\swb_0, \off_{0}, \on_{1}, \appRule{\R_1}, \appRule{\R_5}, \appRule{\R_6},\\ \appRule{\R_7}, \blockRule{\R_2}, \blockRule{\R_3}, \blockRule{\R_4}\}$,
pointing out that this answer set was obtained due to the applicability of rules $\R_1$, $\R_5$, $\R_6$, and $\R_7$, whereas the applicability of the other rules was blocked.\qed
\end{examplecontpage}

For explaining the inconsistency of a logic program,
three additional \emph{extrapolation transformations} are performed (rules~\eqref{eq:spock1:extraHead}-\eqref{eq:spock1:unfounded2}), denoted $\transEx{P}$.
They allow to generate potential answer sets, i.e. sets of atoms, that violate Definition~\ref{def:answerSet_LinZ_Lee} and thus provide an explanation of the inconsistency.
\changed{To generate potential answer sets \emph{choice}-rules are used, which allow to choose whether or not the head of this rule should be true if the rule is applicable. These rules have the form $\{\head{\R}\} \lparrow \body{\R}$ and are shorthand notation for}
\begin{align*}
\head{\R} &\lparrow \body{\R} \wedge \Not x \\
x &\lparrow \Not \head{\R}
\end{align*}
where $x \notin \at$ is a new atom.

Concerning the first inconsistency reason -- the violation of rules -- a new \emph{abnormality} tagging-atom $\abP{\R}$ is introduced and used to transform each rule $\R$, where $\head{\R} = h$.\footnote{We use the more intuitive
naming $\abP{\R}$ instead of the original $\mathsf{ab}_p(\R)$ \cite{BrainGPSTW2007b} (similarly for the tagging-atoms described in the rest of this section).}
\begin{align}
\label{eq:spock1:extraHead} \{h\} &\lparrow \appRule{\R} \\ 
 \abP{\R} &\lparrow \appRule{\R} \wedge \Not h \label{trans:abP}
\end{align}
When used for explaining inconsistent programs,
rule~\eqref{eq:spock1:extraHead} substitutes rule~\eqref{eq:spock1:head} from the kernel transformation.
This allows to exclude $h$ from an answer set, even if $\R$ is applicable.
\changed{This choice rule allows to generate potential answer sets and rule~\eqref{trans:abP} derives a respective reason why they may not be actual answer sets. In particular, this is the case if a rule is applicable \wrt\ a potential answer set but it head is not contained in this set.} 

The second extrapolation transformation is concerned with the supportedness of atoms.
It introduces a new abnormality tagging-atom $\abC{a}$ for each $a \in \at$,
used in a transformation as follows:
\begin{align}
\label{eq:spock1:atom} \{a\} &\lparrow \blockRule{\R_1} \wedge \dotsc \wedge \blockRule{\R_k} \\
\label{eq:spock1:unsupported} \abC{a} &\lparrow a, \blockRule{\R_1} \wedge \dotsc \wedge \blockRule{\R_k}
\end{align}
where $\R_1, \dotsc, \R_k$ are all the rules with head $a$.
Similarly to the first extrapolation transformation, rule~\eqref{eq:spock1:atom} allows to choose if $a$ is or is not included in a potential answer set being explained.
Rule~\eqref{eq:spock1:unsupported} derives $\abC{a}$ whenever $a$ is in the answer set without any rule to support it.

The third extrapolation transformation deals with unfounded atoms.
A new abnormality tagging-atom $\abL{a}$ is introduced for each atom $a \in \at$ and used as follows:
\begin{align}
\label{eq:spock1:unfounded} \{\abL{a}\} &\lparrow \Not \abC{a} \\
\label{eq:spock1:unfounded2} a &\lparrow \abL{a} 
\end{align}
This transformation gives a choice to include or exclude the abnormality atom $\abL{a}$, given that
\changed{there is no other reason for $a$to be causing the inconsistency, namely}
  being unsupported.
This is different from the previous transformations, where abnormality atoms are only derived if there is an actual violation of 
a condition in Definition~\ref{def:answerSet_LinZ_Lee}. 
Here, the abnormality atom may be derived even if the third condition in Definition~\ref{def:answerSet_LinZ_Lee} is not violated.
This means that unfounded loops cannot be identified with certainty. 

\begin{example}
 \label{ex:abnormality.spock}
 Consider the following inconsistent logic program $\newprogram\label{prg:abnormality.spock}$:
 \begin{align*}
  \R_1: \ a &\lparrow \Not b\\
  \R_2: \ b &\lparrow \Not b
 \end{align*}
 The answer sets of $\transKern{\programr\ref{prg:abnormality.spock}} \cup \transEx{\programr\ref{prg:abnormality.spock}}$
 (where rule~\eqref{eq:spock1:head} is not included since derivability of the head is 
 expressed through rule~\eqref{eq:spock1:extraHead} as previously explained)
 indicate potential answer sets and explain why these potential answers sets are not \emph{actual} answer sets by pointing out violations concerning the definition of answer sets.
 \begin{itemize}
  \item $\newanswerset\label{as:1:prg:abnormality.spock} = \{a, b, \abC{a}, \abC{b}, \blockRule{\R_1}, \blockRule{\R_2}\}$
  \item $\newanswerset\label{as:2:prg:abnormality.spock} = \{b, \abC{b}, \blockRule{\R_1}, \blockRule{\R_2}\}$
  \item $\newanswerset\label{as:3:prg:abnormality.spock} = \{a, \abL{a}, \abP{\R_2}, \appRule{\R_1}, \appRule{\R_2}\}$
  \item $\newanswerset\label{as:4:prg:abnormality.spock} = \{a, \abP{\R_2}, \appRule{\R_1}, \appRule{\R_2}\}$
  \item $\newanswerset\label{as:5:prg:abnormality.spock} = \{\abP{\R_1}, \abP{\R_2}, \appRule{\R_1}, \appRule{\R_2}\}$
 \end{itemize}
 $\answersetr\ref{as:1:prg:abnormality.spock}$ expresses that $\{a,b\}$ is not an answer set because neither of the two atoms are supported by an applicable rule.
 This is because both $\R_1$ and $\R_2$ are blocked \wrt\ $\{a,b\}$.
 In contrast $\answersetr\ref{as:5:prg:abnormality.spock}$ explains that \wrt\ $\{\}$ both $\R_1$ and $\R_2$ are applicable, but the head 
 of neither rule is included in $\{\}$.
 $\answersetr\ref{as:3:prg:abnormality.spock}$ illustrates the guessing of unfounded atoms. It states that $\{a\}$ is not an answer set
 because $a$ may be unfounded and because $\R_2$ is violated.
 Note that this guess is redundant, since answer set $\answersetr\ref{as:4:prg:abnormality.spock}$ explains $\{a\}$ by only referring to 
 the violation of $\R_2$.
\changed{In fact, $a$ is not unfounded here, as it is not part of an unfounded loop \wrt\ $\{a\}$ (it is not part of a loop at all).} 
 \qed
\end{example}

As shown by Example~\ref{ex:abnormality.spock}, there may be many explanations for the inconsistency of a logic program
and some of them may be redundant.
It is thus advisable to only consider explanations with a minimal number of abnormality tagging-atoms.
This also ensures that $\abL{a}$ only occurs if $a$ is indeed unfounded \cite{BrainGPSTW2007b}.
In Example~\ref{ex:abnormality.spock}, minimisation narrows the explanations down to sets $\answersetr\ref{as:2:prg:abnormality.spock}$ and~$\answersetr\ref{as:4:prg:abnormality.spock}$.

\CHANGED
\begin{samepage}
\begin{example}\label{ex:unfounded.spock}\nopagebreak
Let $\newprogram\label{prg:unfounded.spock}$ be the logic program \programr\ref{prg:abnormality.spock} with the two additional rules:
 \begin{align*}
  \R_3: \ a &\lparrow b\\
  \R_4: \ b &\lparrow a
 \end{align*}
 These rules induce an unfounded loop \wrt\ the set $\{a,b\}$.
 $\transKern{\programr\ref{prg:unfounded.spock}} \cup \transEx{\programr\ref{prg:unfounded.spock}}$ has three answer sets explaining why $\{a,b\}$ is not an answer set: one in terms of $a$ being an unfounded atom (comprising $\abL{a}$), one in terms of $b$ being an unfounded atom (comprising $\abL{b}$), and one in terms of both atoms being unfounded (comprising both $\abL{a}$ and $\abL{b}$). Similarly to Example~\ref{ex:abnormality.spock}, the last of these three answer sets provides a redundant explanation compared to the first two. 
However, here the explanations in terms of unfoundedness of atoms are correct, as there exists an unfounded loop. 
In addition, $\transKern{\programr\ref{prg:unfounded.spock}} \cup \transEx{\programr\ref{prg:unfounded.spock}}$ has four answer sets stating the same reasons as $\answersetr\ref{as:2:prg:abnormality.spock}-\answersetr\ref{as:5:prg:abnormality.spock}$. \qed
\end{example}
\end{samepage}
\END

Note that \spock\ does not suggest how to \emph{change} an inconsistent logic program to make it consistent.
However, based on the abnormality tagging-atoms in an answer set $M$ of $\transKern{\P} \cup \transEx{P}$ there is a straightforward way of turning the inconsistent
program $P$ into a consistent logic program:
\begin{itemize}
 \item if $\abP{\R} \in M$, then delete $\R$ from $P$;
 \item if $\abC{a} \in M$ or $\abL{a} \in M$, then add $a \lparrow$ to $P$.
\end{itemize}
If this is done for all abnormality-tagging atoms in $M$, the changed logic program has an answer set $M \cap \at$.
Note that even though this change results in a consistent program, there is no guarantee that this program captures the originally intended meaning.

\begin{examplecont}{ex:abnormality.spock}
Consider adding $b \lparrow$ to $\programr\ref{prg:abnormality.spock}$, based on $\answersetr\ref{as:2:prg:abnormality.spock}$.
This turns $\programr\ref{prg:abnormality.spock}$ into a consistent logic program with answer set $\{b\}$.
However, the intended meaning of the program may have been a choice between answer set $\{a\}$ and $\{b\}$, with the programmer's mistake being that $\Not b$ in $\R_2$ should have been $\Not a$.
In this case, the change does not capture the original meaning.\qed
\end{examplecont}

In addition to giving explanations of inconsistent programs with respect to automatically generated potential answer sets, the \spock\ system also allows for
more user-directed explanations. Among others, a user can specify a set of rules and atoms from which the explanations are drawn \cite{BrainGPSTW2007a}.
For example, in $\programr\ref{prg:abnormality.spock}$ we may be sure that rule $\R_2$ is correct and thus
restrict\footnote{In the \spock\ implementation this is achieved by using flags \texttt{exrules} and
\texttt{exatoms} for specifying rules and atoms to be debugged. This restricts the transformations to these rules and atoms.} abnormality tagging-atoms $\abP{\R}$ to rule $\R_1$.
This prevents the construction of answer set $\answersetr\ref{as:4:prg:abnormality.spock}$, thus resulting in $\answersetr\ref{as:2:prg:abnormality.spock}$ as the only explanation
(when using minimisation).
Furthermore, an atom $a$ that should be included in an answer set can be specified by adding the constraint $\lparrow \Not a$ to the kernel transformation of the given logic program.


\subsubsection{Transformation 2}\label{sec:debugging:spock2}
In the first transformation approach of \spock, an ASP solver is merely used to compute the answer sets of the kernel and extrapolation transformations, 
thus generating explanations.
That is, the kernel and extrapolation transformations are constructed externally (from the ASP solver).
In contrast, the second transformation approach of \spock\ \cite{GebserPST08} uses an ASP solver to both construct a transformation and compute explanations.
This is achieved by using a static non-ground \emph{meta-program} $\metaP$, which expresses violation conditions that can be instantiated with any given logic program.
The second transformation approach is furthermore defined for any logic program, i.e. the head of rules is a (possibly empty) disjunction of atoms.

In order to instantiate the meta-program with the rules and atoms of a given logic program $\P$, an \emph{input transformation} $\inputP{\P}$ is generated,
containing facts that express which rules $\R$ and atoms $a$ are contained in $\P$.
More specifically, for every atom $a \in \at$, every rule $\R \in \P$ (where $r$ is the label of the rule), and every $h \in \head{\R}$, $b \in \bodyp{\R}$, and
$c \in \bodyn{\R}$ the following facts are included in $\inputP{\P}$:
\begin{align}
\atomTrans{a} &\lparrow \\
\ruleTrans{\R} &\lparrow \\
\headTrans{\R}{h} &\lparrow \\
\bodypTrans{\R}{b} &\lparrow\\
\bodynTrans{\R}{c} &\lparrow
\end{align}
This input transformation $\inputP{\P}$ is combined with the static meta-program \metaP\
to compute explanations for inconsistent logic programs using an ASP solver.
The meta-program uses a more explicit way of constructing potential answers sets than the extrapolation transformations, namely, for every $\atomTrans{a}$ there is the choice to include or not include it in a potential answer set.\footnote{Throughout this section, we use uppercase letters to denote variables.}
\begin{align}
 \inAS{A} &\lparrow \atomTrans{A} \wedge \Not \notinAS{A}\\
 \notinAS{A} &\lparrow \atomTrans{A} \wedge \Not \inAS{A}
\end{align}
Thus, an answer set of $\inputP{\P} \cup \metaP$ comprises for each atom $a \in \at$ either $\inAS{a}$ or $\notinAS{a}$.
In contrast, an answer set of $\transKern{\P} \cup \transEx{\P}$ either does or does not contain $a \in \at$.

The other parts of the meta-program \metaP\ are similar to the kernel and extrapolation transformations.
The rule applicability conditions of the kernel transformation (rules~\eqref{eq:spock1:appRule}-\eqref{eq:spock1:blockRuleN2}) are expressed in \metaP\ as follows:
\begin{align}
\blockRule{R} &\lparrow \bodypTrans{R}{B} \wedge \notinAS{B} \\
\blockRule{R} &\lparrow \bodynTrans{R}{C} \wedge \inAS{C} \\
\appRule{R} &\lparrow \Not \blockRule{R}
\end{align}
In contrast to the first transformation approach, the applicability of a rule is here expressed in terms of the rule
not being blocked.

The following rules of the meta-program \metaP\ generalise the extrapolation transformations for rule satisfiability  
from normal rules to rules whose head may be empty or a disjunction of atoms.\footnote{The meta-program also contains rules explicitly handling unsatisfied constraints, tagging them with a different abnormality atom. For simplicity, and since rule~\eqref{eq:spock2:abP} also applies to constraints, we do not report these constraint rules.} In contrast to normal rules, here we check if at least \emph{one} of the head atoms of an applicable rule is satisfied.
\begin{align}
\trueHead{R} &\lparrow \headTrans{R}{A} \wedge \inAS{A} \\
\label{eq:spock2:abP} \abP{R} &\lparrow \appRule{R} \wedge \Not \trueHead{R}
\end{align}

For logic programs that are not normal, an atom may be unsupported even if there exists a rule
with $a$ in the head that is not blocked.
As stated in the second condition of Definition~\ref{def:answerSet_LinZ_Lee}, 
$a$ is supported if some rule is applicable and $a$ is the \emph{only} head atom that is in the potential answer set being explained.
Thus, for an atom to be unsupported, this condition has to be false.
\begin{align}
\otherHeadTrue{R}{A} &\lparrow \headTrans{R}{A2} \wedge A \neq A2 \wedge \inAS{A2} \\
\supported{A} &\lparrow \headTrans{R}{A} \wedge \appRule{R} \wedge \Not \otherHeadTrue{R}{A} \\
\abC{A} &\lparrow \inAS{A} \wedge \Not \supported{A}
\end{align}

The biggest difference between the first and second transformation approach concerns unfounded loops.
Just like the first approach, the second includes a choice as to whether or not an atom that is part of the potential answer set being explained is unfounded 
(see rules~\eqref{eq:spock2:unfounded} and~\eqref{eq:spock2:founded}).
The difference is that if an atom is guessed to be unfounded, there is a check (see rule~\eqref{eq:spock2:unfoundedCheck}) of the foundedness condition 
in Definition~\ref{def:answerSet_LinZ_Lee}.
That is, for an unfounded atom $a$ it is checked if there is an applicable rule $\R$ with $a$ in the head (if so, $\R$ is also satisfied since $\abL{a}$ only holds if $\inAS{a}$) that has no head atom that is founded (in the same loop) and no positive body atom that is unfounded (in the same loop). If such a rule exists, $a$ is by Definition~\ref{def:answerSet_LinZ_Lee} founded, which is why this check is implemented as a constraint in $\metaP$ (rule~\eqref{eq:spock2:unfoundedCheck}).
 This ensures that $\abL{a}$ is only part of an answer set of $\inputP{\P} \cup \metaP$, if $a$ is 
actually unfounded.
\begin{align}
 \label{eq:spock2:unfounded} \abL{A} &\lparrow \inAS{A} \wedge \supported{A} \wedge \Not \notAbL{A} \\
 \label{eq:spock2:founded} \notAbL{A} &\lparrow \inAS{A} \wedge \Not \abL{A} \\
 \label{eq:spock2:unfoundedCheck} &\lparrow \abL{A} \wedge \headTrans{R}{A} \wedge \appRule{R} \wedge \\ 
 &\phantom{\lparrow} \text{ $\Not$ } \headOutLoop{R} \wedge \text{ $\Not$ } \bodyInLoop{R} \notag \\
 \headOutLoop{R} &\lparrow \headTrans{R}{A} \wedge \notAbL{A} \\
 \bodyInLoop{R} &\lparrow \bodypTrans{R}{A} \wedge \abL{A}
\end{align}
Furthermore, there are rules ensuring that only one loop is considered at a time, i.e. $\abL{a}$ and $\abL{b}$ only hold
if $a$ and $b$ are part of the same loop.

Another main difference between the two \spock\ approaches is that the second transformation approach only explains sets of atoms that are \emph{not} answer sets of the given logic program, whereas the first also
explains \emph{actual} answer sets of the given logic program (if any exist).
This is due to the following rules in the meta-program \metaP, ensuring that at least one abnormality tagging-atom is
part of an answer~set:
\begin{align}
\noAS &\lparrow \abP{\R} \\
\noAS &\lparrow \abC{\R} \\
\noAS &\lparrow \abL{\R} \\
&\lparrow \Not \noAS
\end{align}

\begin{examplecont}{ex:abnormality.spock}\label{ex:abnormality:spock2}
Applying the second transformation approach to $\programr\ref{prg:abnormality.spock}$, \spock\ computes the answer sets of
$\inputP{\programr\ref{prg:abnormality.spock}} \cup \metaP$, yielding the following:
\begin{itemize}
 \item $\newanswerset\label{as:1:prg:abnormality.spock2} = \{ \inAS{a}, \inAS{b}, \notAbL{a}, \notAbL{b}, \abC{a}, \abC{b},\\
  \phantom{\answersetr\ref{as:1:prg:abnormality.spock2} = \{ } \blockRule{\R_1}, \blockRule{\R_2}, \trueHead{\R_1}, \trueHead{\R_2},\\
   \phantom{\answersetr\ref{as:1:prg:abnormality.spock2} = \{ } \headOutLoop{\R_1}, \headOutLoop{\R_2}\}$
 \item $\newanswerset\label{as:2:prg:abnormality.spock2} = \{\inAS{b}, \notinAS{a}, \notAbL{b}, \abC{b}, \blockRule{\R_1}, \blockRule{\R_2},\\
 \phantom{\answersetr\ref{as:2:prg:abnormality.spock2} = \{ } \trueHead{\R_2}, \headOutLoop{\R_2}\}$
 \item $\newanswerset\label{as:3:prg:abnormality.spock2} = \{ \inAS{a}, \notinAS{b}, \notAbL{a},\supported{a}, \supported{b}, \abP{\R_2}, \\
 \phantom{\answersetr\ref{as:3:prg:abnormality.spock2} = \{ } \appRule{\R_1}, \appRule{\R_2},  \trueHead{\R_1}, \headOutLoop{\R_1}\}$
 \item $\newanswerset\label{as:4:prg:abnormality.spock2} = \{ \notinAS{a}, \notinAS{b}, \supported{a}, \supported{b}, 
 \abP{\R_1}, \abP{\R_2},\\
 \phantom{\answersetr\ref{as:4:prg:abnormality.spock2} = \{ }   \appRule{\R_1}, \appRule{\R_2}\}$
\end{itemize}
Note that all answer sets also comprise the facts in $\inputP{\programr\ref{prg:abnormality.spock}}$,
such as $\atomTrans{a}$, $\headTrans{\R_1}{a}$, and $\ruleTrans{\R_1}$, as well as the atom $\noAS$,
\changed{which we omitted above for readability}.
Since the second transformation approach does not generate explanations containing unfoundedness as a reason when
an atom is in fact founded, there is no equivalent of answer set $\answersetr\ref{as:3:prg:abnormality.spock}$ from the first
transformation approach. All other answer sets of $\transKern{\programr\ref{prg:abnormality.spock}} \cup \transEx{\programr\ref{prg:abnormality.spock}}$ report the same reasons as the answer sets given above.\qed
\end{examplecont}

\changed{
\begin{examplecont}{ex:unfounded.spock}\label{ex:unfounded:spock2}
For the program $\programr\ref{prg:abnormality.spock}$, which comprises an unfounded loop \wrt\ $\{a,b\}$, even more redundant explanations are omitted when using the second transformation approach.
More precisely, as for $\programr\ref{prg:abnormality.spock}$ there is one explanation for each possible set of atoms, i.e. $\{\}$, $\{a\}$, $\{b\}$, and $\{a,b\}$.
The explanation as to why the last set is not an answer set is given by $\abL{a}$ and $\abL{b}$.
The explanations concerning the other three sets are analogue to the explanations of $\programr\ref{prg:abnormality.spock}$ in Example~\ref{ex:abnormality:spock2}. \qed
\end{examplecont}
}

Similarly to the first transformation approach, the user can specify constraints for debugging.
An atom $a$ can, for example, be forced to (not) be a part of an answer set by adding the constraint
$\lparrow \notinAS{a}$ (respectively $\lparrow \inAS{a}$) to the input transformation of the given logic program.
In the same way, constraints on the abnormality tagging-atoms can be specified, e.g. $\lparrow \abP{\R}$
enforces that rule~$\R$ is satisfied.

In conclusion, the second transformation approach requires less processing of the given logic program $\P$ performed outside the ASP solver
than the first transformation approach.
Furthermore, the two transformation approaches differ in the number of explanations given,
since the first approach may yield redundant explanations and explanations where unfoundedness is given as a reason even though the atom in question is founded.

\subsection{The \ouroboros\ System -- Debugging Non-ground Programs}\label{sec:debugging:ouroboros}
The two \spock\ approaches do not explicitly deal with variables occurring in the given logic program.
However, variables are important to consider for debugging approaches, since, in practice, logic programs
under the answer set semantics often contain first-order predicates and variables.
Handling variables when debugging thus requires an efficient grounding strategy.

Building upon the second \spock\ transformation, \citeNS{OetschPT10} develop a meta-program able to construct explanations of inconsistent \emph{extended} logic programs possibly comprising variables.
In contrast to the approach taken by \spock, which constructs various sets of atoms and explains why these are not answer sets,
\ouroboros\ requires an \emph{intended answer set}. 
It thus assumes that the user already has a solution in mind.
An explanation is then constructed for this anticipated solution.

Efficiently constructing explanations for logic programs with variables is non-trivial as it requires 
grounding (i.e. substituting variables with \newChanged{constants}).
First grounding a given logic program and then constructing explanations, for example using the \spock\ approach, requires exponential space and double-exponential time.
Instead, the \ouroboros\ approach requires only polynomial space and single-exponential time,
as it applies grounding to the input transformation and meta-program during the solving process rather than grounding the given logic program before transforming and solving it.

Similarly to the input transformation $\inputP{\P}$ of the second \spock\ approach,
\ouroboros\ constructs an \emph{input transformation} $\inputPOur{\P}$ of a given logic program~$\P$,
expressing which extended literals are part of the head and body of each rule.
Additionally, $\inputPOur{\P}$ includes facts expressing
which predicates occur in $\P$, 
what the position of variables and constants is in each predicate, and which variables occur in which rules.
Since \ouroboros\ requires a given set of atoms $M \subseteq \at$ to be explained,
this set is also transformed to make it applicable to the input transformation and the meta-program.
The \emph{interpretation transformation} $\interPOur{M}$ includes facts $\inAS{a}$ for each atom $a \in M$ as well as facts
stating which predicates occur in $M$ and what the position of constants is in predicates in $M$.

The \emph{meta-program} \metaPOur\ of \ouroboros\ follows the same ideas as \spock, expressing conditions
under which a rule is unsatisfied or a loop is unfounded.
Note that in contrast to \spock, \ouroboros\ does not explicitly point out unsupported atoms. Instead, unsupported atoms are handled as singleton loops that are unfounded.
The exact encoding of \metaPOur\ with its more than 160 rules can be found \newChanged{online\footnote{\url{www.kr.tuwien.ac.at/research/projects/mmdasp/encoding.tar.gz}}.}

When applying an ASP solver to $\inputPOur{\P} \cup \interPOur{M} \cup \metaPOur$ to compute explanations as to why $M$ is not an answer set, 
the automatic grounding of the solver allows for the efficient computation of ground answer sets if $P$ contains variables.

Just like \spock, \ouroboros\ only gives \emph{explanations} as to why a set of atoms is not an answer set.
The subsequent changing of the program to make it consistent is left to the user.
In addition to explicit negation, \ouroboros\ can also handle arithmetic operations with integers ($+$ and $*$)
and allows for comparison predicates ($=$, $\neq$, $\geq$, $\leq$, $>$, $<$).
\citeNS{PolleresFSF2013} further extend \ouroboros\ to deal with choice rules and cardinality and weight constraints
by translating these constructs into normal rules (possibly containing variables).
\citeNS{FruhstuckPF2013} integrate \ouroboros\ into the \sealion\ IDE.\footnote{Note that a special setup of ASP solvers is needed to make this integration work.}


\subsection{Interactive Debugging Based on \spock}\label{sec:debugging:interactive}
No matter which of the two transformations is used, the \spock\ approach may generate many different explanations,
since for every set of atoms that is not an answer set at least one explanation is constructed.
Even for the small logic program in Example~\ref{ex:abnormality.spock}, which has only two atoms, 
four explanations are generated using the second transformation (see Example~\ref{ex:abnormality:spock2}). 
\ouroboros\ tackles this problem by requiring the user to specify an intended answer set.
However, a user may not have a truth assignment for every atom in mind.
\citeNS{Shchekotykhin2015} therefore proposes an interactive method on top of the second \spock\ approach, where the user is \emph{queried} whether or not
an atom should be contained in an answer set. 
The user's answer narrows down the sets of atoms for which explanations are constructed to the ones \emph{relevant} to the user
and relieves the user of the burden to specify the whole intended answer set upfront.

As mentioned in previous sections, the user can force atoms to be contained or not contained in
explanation answer sets of \spock\ (using the second transformation) by adding facts $\inAS{a}$ or $\notinAS{a}$.
In the interactive debugging approach, such statements are explicitly used as \emph{test cases}.

\begin{definition}[Test Case and Background Theory]
Given a program $\P$, its input transformation $\inputP{\P}$, and the meta-program~$\metaP$
\begin{itemize}
\item $\posCase, \negCase \subseteq \setm{\inAS{a}, \notinAS{a}}{a \in \at}$ are sets of
positive and negative \emph{test cases},
\item $\backTheory \subseteq \P$ is a \emph{background theory}.\qed
\end{itemize}
\end{definition}
Positive test cases are atoms that have to be contained in ($\inAS{a}$) or excluded from ($\notinAS{a}$) \emph{all} answer sets.
In contrast, negative test cases are atoms that have to be contained in ($\inAS{a}$) or excluded from ($\notinAS{a}$) \emph{some} answer set.
A background theory consists of rules in the logic program that are assumed to be satisfied.

In contrast to the \spock\ approach, answer sets of $\inputP{\P} \cup \metaP$ that contain the same 
abnormality tagging-atoms are considered as the same explanation, even if the atoms in the respective explained answer sets are different.
The aim is to find sets of abnormality tagging-atoms that satisfy all given test cases and the given background theory.
In other words, we want to compute all answer sets of \mbox{$\inputP{\P} \cup \metaP$} containing \emph{only} abnormality tagging-atoms
satisfying the test cases and the background theory.
Sets of abnormality tagging-atoms satisfying this condition are called \emph{diagnoses}.

\begin{definition}[Diagnosis]
\label{def:diagnosisGround}
Let $\abAtoms{\P}$ be the set of all abnormality tagging-atoms over a program $\P$, that is,
$$\abAtoms{\P} \eqdef \setm{\abP{r}}{r \in \P} \cup \setm{\abC{a}, \abL{a}}{a \in \at}$$
A set $\diagnosis \subseteq \abAtoms{\P}$ is a \emph{diagnosis} for the \emph{problem instance}
$\langle \P, \backTheory, \posCase, \negCase \rangle$ if
\begin{enumerate}
 \item $\P^* = \inputP{\P} \cup \metaP \cup \setm{\lparrow d}{d \in \abAtoms{\P} \setminus \diagnosis} 
 \cup \setm{\lparrow \abP{\R}}{\R \in \backTheory} \cup \setm{p \lparrow}{p \in \posCase}$ has an answer set and
 \item for each $n \in \negCase$, $\P^* \cup \{n \lparrow\}$ has an answer set.\qed
\end{enumerate}
\end{definition}
Note that due to the constraints of the form $\lparrow d$, any answer set of $\P^*$ will 
only contain abnormality tagging-atoms from \diagnosis.

Diagnoses can be found by computing answer sets of the program $\inputP{\P} \cup \metaP \cup \setm{\lparrow \abP{\R}}{\R \in \backTheory}$ and then verifying whether the respective
sets of abnormality tagging-atoms contained in these answer sets satisfy the conditions for being a diagnosis.
Usually, only (subset) \emph{minimal} diagnoses will be considered.

\begin{example}\label{ex:light.inconsistent}
 Consider the logic program $\programr\ref{prg:lights.spock}$ (see Example~\ref{ex:light.spock}; page~\pageref{ex:light.spock}) with the additional constraint $\R_8: \lparrow \Not \off_{1}$.
 This program, called $\newprogram\label{prg:lights.inconsistent}$, is inconsistent.
 Using the second translation approach of \spock, 256 answer sets are computed for $\inputP{\programr\ref{prg:lights.inconsistent}} \cup \metaP$, each explaining a different set of atoms that is not an answer set.
 Let us now specify $\backTheory = \{\R_1,\R_2, \R_6, \R_7\}$, in other words, we are sure that the first two rules are correct and that switches $a$ and $b$ are $\on$ in situation 0.
 This narrows down the answer sets; 
 program~$\inputP{\programr\ref{prg:lights.inconsistent}} \cup \metaP \cup\, \{\lparrow \abP{\R_1},\lparrow \abP{\R_2},\\ \lparrow \abP{\R_6},\lparrow \abP{\R_7}\}$ has
 only 28 answer sets.
 Given positive test cases $\posCase = \{\notinAS{\swc_0}, \notinAS{\swd_0}\}$, only eight out of the 28 answer sets satisfy these, namely :
 \begin{itemize}
  \item $\newanswerset\label{as:1:prg:lights.inconsistent} = \{\notinAS{\on_0}, \notinAS{\off_0}, \inAS{\on_1}, \notinAS{\off_1}\}
  \cup \{\abP{\R_5}, \abP{\R_8}\}$
  \item $\newanswerset\label{as:2:prg:lights.inconsistent} = \{\inAS{\on_0}, \notinAS{\off_0}, \inAS{\on_1}, \notinAS{\off_1}\} \cup
  \{\abP{\R_5}, \abP{\R_8},\\ \phantom{\answersetr\ref{as:2:prg:lights.inconsistent} = \{} \abC{\on_0}\}$
  \item $\newanswerset\label{as:3:prg:lights.inconsistent} = \{\notinAS{\on_0}, \notinAS{\off_0}, \inAS{\on_1}, \inAS{\off_1}\} \cup
  \{\abP{\R_5}, \abC{\off_1}\}$
  \item $\newanswerset\label{as:4:prg:lights.inconsistent} = \{\inAS{\on_0}, \notinAS{\off_0}, \inAS{\on_1}, \inAS{\off_1}\}
  \cup \{\abP{\R_5}, \abC{\off_1},\\ \phantom{\answersetr\ref{as:4:prg:lights.inconsistent} = \{} \abC{\on_0}\}$
  \item $\newanswerset\label{as:5:prg:lights.inconsistent} = \{\notinAS{\on_0}, \inAS{\off_0}, \inAS{\on_1}, \notinAS{\off_1}\}
  \cup \{\abP{\R_8}\}$
  \item $\newanswerset\label{as:6:prg:lights.inconsistent} = \{\inAS{\on_0}, \inAS{\off_0}, \inAS{\on_1}, \notinAS{\off_1}\}
  \cup \{\abP{\R_8}, \abC{\on_0}\}$
  \item $\newanswerset\label{as:7:prg:lights.inconsistent} = \{\notinAS{\on_0}, \inAS{\off_0}, \inAS{\on_1}, \inAS{\off_1}\}
  \cup \{\abC{\off_1}\}$
  \item $\newanswerset\label{as:8:prg:lights.inconsistent} = \{\inAS{\on_0}, \inAS{\off_0}, \inAS{\on_1}, \inAS{\off_1}\}
  \cup \{\abC{\off_1}, \abC{\on_0}\}$
 \end{itemize}
Note that each answer set also contains $\inAS{\swa_0}$, $\inAS{\swb_0}$, $\notinAS{\swc_0}$, and $\notinAS{\swd_0}$, as well as the further tagging-atoms discussed in Section~\ref{sec:debugging:spock2}.
 Taking a closer look at these 8 answer sets, each of them defines a diagnosis when $\negCase = \{\}$, namely the second part of each answer set.
 Only $\answersetr\ref{as:5:prg:lights.inconsistent}$ and $\answersetr\ref{as:7:prg:lights.inconsistent}$ induce \emph{minimal} diagnoses.
 Now consider that $\negCase = \{\inAS{\on_0}, \inAS{\off_0}\}$. This rules out half of the diagnoses, leaving only the following four:
 \begin{itemize}
  \item $\diagnosis_1 = \{\abP{\R_5}, \abP{\R_8}, \abC{\on_0}\}$ (cf. $\answersetr\ref{as:2:prg:lights.inconsistent}$)
  \item $\diagnosis_2 = \{\abP{\R_5}, \abC{\off_1}, \abC{\on_0}\}$ (cf. $\answersetr\ref{as:4:prg:lights.inconsistent}$)
  \item $\diagnosis_3 = \{\abP{\R_8}, \abC{\on_0}\}$ (cf. $\answersetr\ref{as:6:prg:lights.inconsistent}$)
  \item $\diagnosis_4 = \{\abC{\off_1}, \abC{\on_0}\}$ (cf. $\answersetr\ref{as:8:prg:lights.inconsistent}$)
 \end{itemize}
 Even though $\inAS{\off_0} \notin \answersetr\ref{as:2:prg:lights.inconsistent}$, $\diagnosis_1$ is a diagnosis of the given problem instance since
 there are two answer sets of $\P^*$ \wrt\ $\diagnosis_1$, namely $\answersetr\ref{as:2:prg:lights.inconsistent}$ and $\answersetr\ref{as:6:prg:lights.inconsistent}$, and
\mbox{$\inAS{\off_0} \in \answersetr\ref{as:6:prg:lights.inconsistent}$}, thus satisfying the negative test case $\inAS{\off_0}$ \wrt\ $\diagnosis_1$.\qed
\end{example}

As illustrated in Example~\ref{ex:light.inconsistent}, positive and negative test cases can considerably reduce the number of diagnoses and, thus, of explanations as to why sets of atoms are not answer sets of $\P$.
If the user does not specify any test cases, it is therefore desirable to produce them automatically by querying the user.
That is, the user is asked whether an atom is expected to be contained in or excluded from all or some answer sets. Ideally, the debugging system chooses an atom as a query that helps to reduce the number of diagnoses as much as possible.

\begin{definition}[Query and Diagnosis Splitting]
Let $\diagnosisSet$ be the set of all diagnoses of the problem instance $\langle \P, \backTheory, \posCase, \negCase \rangle$
and let $\query \subseteq \at$ be a \emph{query}.
\query\ splits the diagnoses in \diagnosisSet\ into three sets, where for each $\diagnosis \in \diagnosisSet$:
\begin{itemize}
 \item $\diagnosis \in \diagnosisP$ if for all $a \in \query$, $\inAS{a}$ is in every answer set of $P^*$;
 \item $\diagnosis \in \diagnosisN$ if for all $a \in \query$, $\notinAS{a}$ is in every answer set of $P^*$;
 \item $\diagnosis \in \diagnosisO$ if $\diagnosis \notin (\diagnosisP \cup \diagnosisN)$.\qed
 \end{itemize}
\end{definition}
This means that $\diagnosisP$ and 
$\diagnosisN$
contain all diagnoses that are still diagnoses if the atoms in the query are added as positive test cases so as to force them to be, respectively, included in or excluded from all answer sets.
Thus, if the user's reply to a query is that the atoms should be included, then the diagnoses in $\diagnosisN$ can be 
disregarded. Likewise, if the user replies that the atoms should be excluded, the diagnoses in $\diagnosisP$ can be disregarded.

\begin{examplecont}{ex:light.inconsistent}\label{ex:light.inconsistent:queries}
 Consider the two atoms that are not part of positive or negative test cases yet, namely $\on_1$ and $\off_1$.
 For $\query_1 = \{\on_1\}$, all four diagnoses are in $\diagnosisP$, so $\diagnosisN = \diagnosisO = \{\}$. For example, the answer sets of $\P^*$ \wrt\ $\diagnosis_1$ are 
 $\answersetr\ref{as:2:prg:lights.inconsistent}$ and $\answersetr\ref{as:6:prg:lights.inconsistent}$, and both comprise $\inAS{\on_1}$.
 This means that if the user replies to the query, that $\on_1$ should be in the desired answer set, then no diagnoses can be disregarded.
 However, if the user replies that $\on_1$ should not be in the desired answer set, then all diagnoses would be disregarded and therefore no explanations given.
 This would imply, that the test cases specified could not be satisfied.
 In contrast, for $\query_2 = \{\off_1\}$ we get $\diagnosisP = \{\diagnosis_2, \diagnosis_4\}$, $\diagnosisN = \{\diagnosis_1, \diagnosis_3\}$, and $\diagnosisO = \{\}$.
 Note that if one of the negative test cases was used as a query, then
  $\diagnosisO \neq \{\}$.
 For instance, for $\query_3 = \{\off_0\}$ we get $\diagnosis_1 \in \diagnosisO$ since $\notinAS{\off_0} \in \answersetr\ref{as:2:prg:lights.inconsistent}$
 but $\inAS{\off_0} \in \answersetr\ref{as:6:prg:lights.inconsistent}$.\qed
\end{examplecont}

There may be a large number of queries, so
queries that yield a large information gain are desirable, i.e. queries that allow to disregard as many diagnoses as possible, independent of the user's answer, which clearly is not known when generating a query.
Thus, a useful query should at least yield a partition with $\diagnosisP, \diagnosisN \neq \emptyset$
so that independent of the user's answer, some diagnoses can be disregarded.


A straightforward selection method is the \emph{myopic} strategy, which prefers queries yielding sets $\diagnosisP$ 
and $\diagnosisN$ that have similar size and where $\diagnosisO$ is as small as possible. That is, a query that minimises
\begin{gather*}
\vert \mid \diagnosisP \mid - \mid \diagnosisN \mid \vert \hspace{3pt} + \mid \diagnosisO \mid
\end{gather*}

\begin{examplecont}{ex:light.inconsistent:queries}\label{ex:light.inconsistent:queriesOpt}
 According to the myopic strategy, $\query_2$ is preferable to $\query_1$ since independent of the answer of the user, the number of possible queries is reduced to two.\qed
\end{examplecont}

\changed{The idea of this interactive debugging approach is that queries are generated and presented to the user until only one diagnosis, or a specified maximal number of diagnoses, is~left.}

\subsection{The \dwasp\ System -- Interactive Debugging of Non-ground Programs}\label{sec:debugging:dwasp}

The interactive debugging approach discussed in the previous section only applies to logic programs without variables.
\citeNS{DodaroGMRS2015} and \citeNS{GasteigerDMRRS2016} extend the idea, of querying the user to find relevant explanations
of inconsistency, to non-ground programs. Instead of using an elaborate meta-program expressing possible reasons for inconsistencies as in \spock, they use the solving process of the ASP solver \wasp\ \cite{AlvianoDFLR2013,AlvianoDLR15} to find inconsistencies in a logic program. Their ASP debugger is thus called \dwasp.

Like \citeNS{Shchekotykhin2015}, \dwasp\ allows to define a background theory. If the background theory is not explicitly specified, the set of facts \changed{of the given logic program $\P$} is used.
Instead of applying abnormality tagging-atoms to indicate inconsistencies, the \dwasp\ system adds to each rule in $\P$ that is not part of the background theory a \emph{debug atom}, stating the name of the rule and the variables occurring in it.
\begin{definition}[Debugging Program]
Given a logic program $\P$ and a background theory $\backTheory \subseteq \P$, the \emph{debugging program}
is defined as:
\begin{gather}
\debP{\P} = \backTheory \cup \setm{h_1 \vee \dotsc \vee h_k \lparrow b_1 \wedge \dotsc \wedge b_n \wedge
\debugAtom{\R}{\vars{\R}}\\ \notag 
\wedge \Not c_1 \wedge \dotsc \Not c_m}{\R \in \P \setminus \backTheory, \head{\R} = \{h_1 \vee \dotsc \vee h_k\},\\ \notag 
\body{\R} = \{b_1, \dotsc, b_n, \Not c_1, \dotsc, \Not c_m\}}
\end{gather}
where $\vars{\R}$ is a tuple consisting of all variables in $\body{\R}$.\qed
\end{definition}

When applying the \wasp\ solver to the debugging program $\debP{\P}$, atoms can be \emph{assumed} to hold when computing answer sets.
That is, these assumed atoms do not need to be derived from rules or facts, they are true by default. 
Assumed atoms are thus similar to positive test cases in the approach of~\citeN{Shchekotykhin2015}.

If a debugging atom is \emph{not} assumed to hold, this amounts to ``blocking'' the respective rule specified in the atom, i.e. the rule is no longer applicable when computing answer sets, since a debugging atom cannot be derived using the rules in $\debP{\P}$. 
If all debugging atoms are assumed to hold, the answer sets of $\debP{\P}$ (minus the debugging atoms) coincide with the answer sets of $\P$.
If $\P$ is inconsistent, it therefore follows that $\debP{\P}$ is also inconsistent.

To find rules causing the inconsistency of a program, the \wasp\ solver allows to compute \emph{unsatisfiable cores}, i.e. sets of atoms such that if they are assumed to hold, no answer set exists.
In the \dwasp\ system, only debugging atoms are considered for unsatisfiable cores. Thus,
an unsatisfiable core points out a combination of rules causing the inconsistency.

\begin{definition}[Unsatisfiable Core]
Let $\debPGround{\P}$ be the grounding of $\debP{\P}$ and let
$\debAt{\P}$ be the set of all (ground) debugging atoms occurring in $\debPGround{\P}$.
$\core \subseteq \debAt{\P}$ is an \emph{unsatisfiable core} \ifonlyif\ $\debPGround{\P}$ is inconsistent when 
all debugging atoms in $\core$ are assumed to hold.\qed
\end{definition}
Note that this definition does not make any assumptions about other atoms assumed to hold.
Therefore, an unsatisfiable core is such that, no matter which other atoms are assumed to hold, $\debPGround{\P}$ is inconsistent.

If $\P$ is inconsistent, clearly $\debAt{\P}$ is an unsatisfiable core. However, there may be other unsatisfiable cores,
which are subsets of $\debAt{\P}$, and thus more useful for identifying the source of inconsistency.
Therefore, only (subset) \emph{minimal} unsatisfiable cores are of interest in \dwasp.

If there is only one unsatisfiable core, then deleting any of the atoms in the core from the atoms assumed to hold results in the existence of an answer set.
However, if there are various unsatisfiable cores, only a combination of atoms from the different cores will lead to the existence of an answer set.
\dwasp\ finds such sets of debugging atoms that, when no longer assumed to hold, ensure the existence of an answer set.
Such sets thus express which rules need to be ``blocked'' to obtain an answer set.

\begin{definition}[\dwasp\ Diagnosis]
\label{def:diagnosisNonGround}
Let $\debPGround{\P}$ be the grounding of $\debP{\P}$ and let
$\debAt{\P}$ be the set of all (ground) debugging atoms occurring in $\debPGround{\P}$.
$\diagnosisDwasp \subseteq \debAt{\P}$ is a \emph{diagnosis} \ifonlyif\ $\debPGround{\P}$ is consistent when 
none of the debugging atoms in $\diagnosisDwasp$ is assumed to hold.\qed
\end{definition}
The \dwasp\ system only considers \emph{minimal diagnoses}.
Even though the definition of diagnosis does not reference unsatisfiable cores, diagnoses are computed from unsatisfiable cores in \dwasp.

Note the difference between the notions of diagnosis used in \dwasp\ and in the approach of \citeNS{Shchekotykhin2015}.
In both cases, a diagnosis comprises atoms identifying the reason for inconsistency.
The difference is that in \dwasp\ a diagnosis is a set of atoms such that the debugging program is consistent if the atoms are \emph{not contained} in answer sets.
In contrast, a diagnosis according to Definition~\ref{def:diagnosisGround} is a set of abnormality tagging-atoms such that the transformed logic program is consistent if these are the only abnormality tagging-atoms \emph{contained} in answer sets.

As in the approach by \citeNS{Shchekotykhin2015}, there may be a large number of diagnoses and not all of them may be relevant to the user.
Thus, \dwasp\ uses the same strategy for querying the user as discussed in the previous section for the approach by \citeNS{Shchekotykhin2015}. 
That is, a \emph{query atom} $q \in \at$ is determined, i.e. a ground (non-debugging) atom, which partitions the set of all diagnoses
into $\diagnosisP$, $\diagnosisN$, and $\diagnosisO$, where:
\begin{itemize}
\item $\diagnosisDwasp \in \diagnosisP$ if $q$ is in every answer set of $\debPGround{\P}$ when none of the debugging atoms in $\diagnosisDwasp$ is assumed to hold;
\item $\diagnosisDwasp \in \diagnosisN$ if $q$ is in no answer set of $\debPGround{\P}$ when none of the debugging atoms in $\diagnosisDwasp$ is assumed to hold;
\item $\diagnosisDwasp \in \diagnosisO$ if $\diagnosisDwasp \notin (\diagnosisP \cup \diagnosisN)$.
\end{itemize}

The only difference in the usage of queries in \dwasp\ as compared to the approach of \citeNS{Shchekotykhin2015} is that, rather than adding test cases, 
the user's answer determines if $q$ (in case $q$ should hold) or $\Not q$ (in case $q$ should not hold) is added to the set of assumed atoms. 


\subsection{Stepping}\label{sec:debugging:stepping}
The debugging approach of \citeNS{OetschPT2017}, which extends previous work by \citeNS{OetschPT2011} and \citeNS{Puehrer2014}, 
tackles the problem of explaining why a set of atoms is or is not an answer set of a logic program
in a \emph{procedural} manner.
Inspired by debugging in \changed{procedural} programming languages, where the step-wise execution of a program can be traced,
the \stepping\ approach allows to apply rules and assign literals to be true or false with respect to a potential answer set step by step. In contrast to the execution of a \changed{procedural} program, the sequence of steps in the execution of a logic program is not \newChanged{predetermined}, due to the declarative nature of the answer set semantics.
Thus, the \emph{user} chooses the step sequence in the \stepping\ approach.
This debugging approach has been implemented in the \sealion\ IDE \cite{BusoniuOPST2013},
a logic programming plugin of the Eclipse platform.

Starting with the empty set as the potential answer set, in each computation step the user is presented with all rules that are applicable \wrt\
the current potential answer set. To satisfy the chosen rule, a head of the rule is then added to the current potential answer set and any atoms that \changed{thus cannot} be in the potential answer set (because they occur in the negative body of the rule) are recorded as being false \wrt\ the potential answer set.

\begin{examplecontpage}{ex:abnormality.spock}\label{ex:abnormality:stepping}
\begin{figure}[t]
\includegraphics[width=\textwidth]{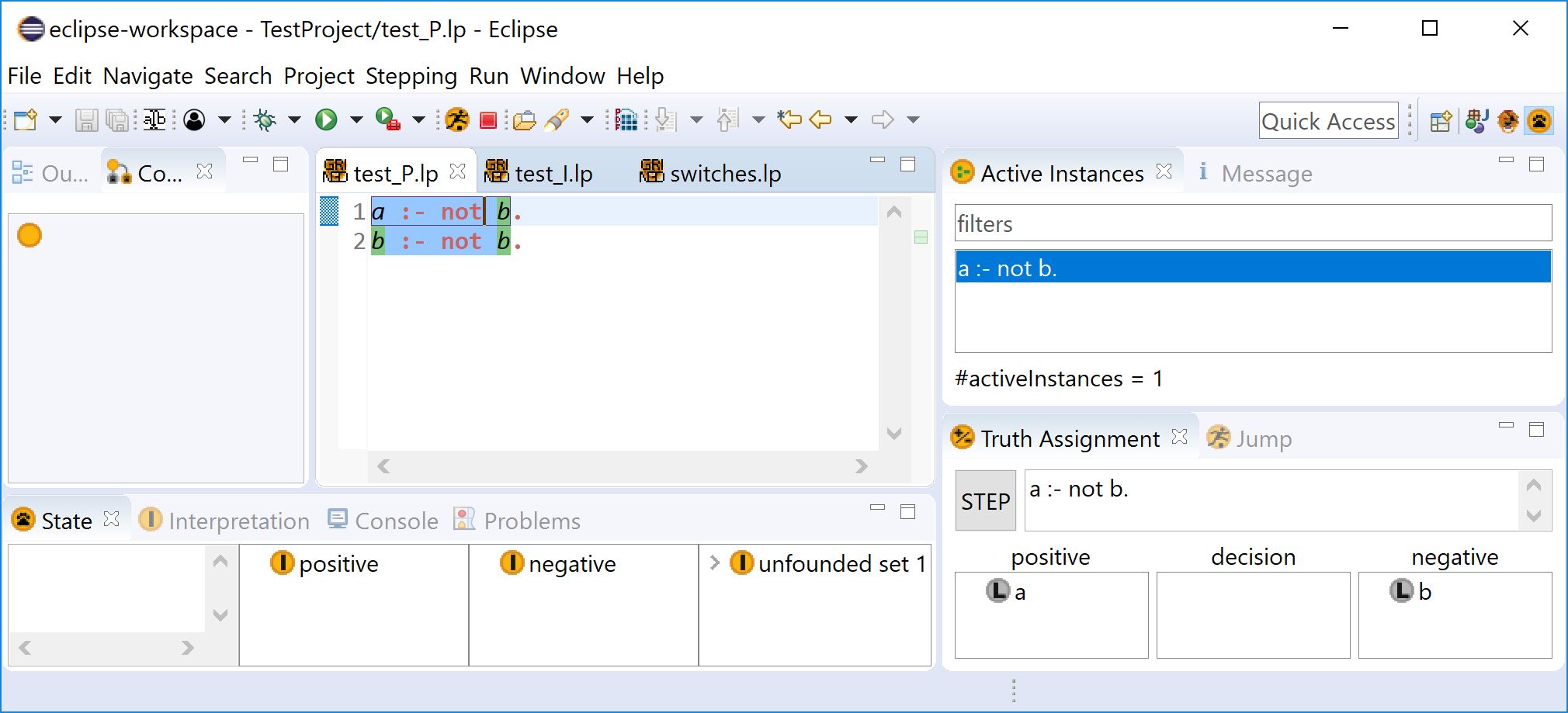}
\caption{The first rule of $\programr\ref{prg:lights.spock}$ is chosen for stepping. The `truth assignment' tab
shows the assignment of truth values to the atoms $a$ and $b$ if a step is performed on the chosen rule.}
\label{fig:abnormality:stepping:r1_1}
\end{figure}
Recall the logic program $\programr\ref{prg:lights.spock}$:
 \begin{align*}
  \R_1: \ a &\lparrow \Not b\\
  \R_2: \ b &\lparrow \Not b
 \end{align*}
 The stepping starts with 
no atoms recorded as being true or false \wrt\ the potential answer set.
Thus, both $\R_1$ and $\R_2$ are applicable since $b$ is not recorded as being in the potential answer set, so $\Not b$ may be true \wrt\ the current potential answer set.
The user can therefore choose which of the two rules to apply.
Figure~\ref{fig:abnormality:stepping:r1_1} illustrates this scenario in the \stepping\ component of \sealion,
where all applicable rules are marked in blue.
The user chooses $\R_1$ to proceed, so $\R_1$ is the only `active instance' of the chosen rule shown in the respective tab (if $\R_1$ contained variables, all applicable grounded versions would be shown in this tab).
The active instance $\R_1$ is then used for the `truth assignment', which is performed by clicking the `step' button.
This records $a$ as being true and $b$ as being false \wrt\ the potential answer set $M$, as illustrated in the `state' tab
at the bottom of Figure~\ref{fig:abnormality:stepping:r1_2}.
After this first step, rule~$\R_2$ is still applicable, so it is chosen for the next `truth assignment'.
However, as indicated by the red X in Figure~\ref{fig:abnormality:stepping:r1_2}, the truth assignment that would satisfy $\R_2$ cannot be performed.
Thus, the stepping computation fails before being completed, indicating to the user that the assignment of truth values performed so far does not lead to an answer set. Note that the reason why $\R_2$ cannot be used for the next step is not pointed out to the user
explicitly, i.e. that $b$ is recorded as false, but to satisfy $\R_2$ it would also have to be true. 
If $\R_2$ was chosen in the first step, the stepping would fail straight away, i.e. 
the scenario from Figure~\ref{fig:abnormality:stepping:r1_2} would apply, but without the truth assignments shown in the `state' tab at the bottom.\qed
\end{examplecontpage}

\begin{figure}[t]
\includegraphics[width=\textwidth]{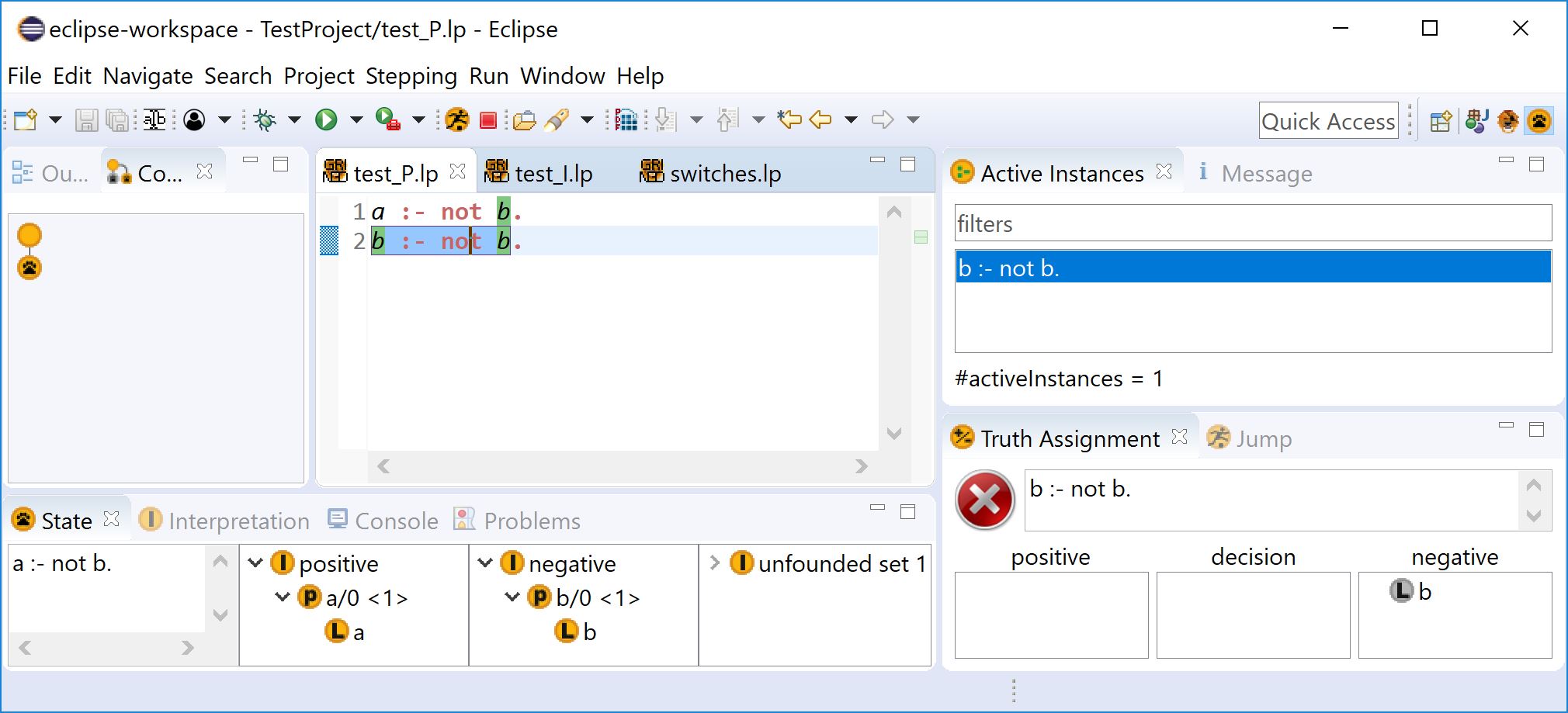}
\caption{After the first step, the second rule is active but a step cannot be performed.}
\label{fig:abnormality:stepping:r1_2}
\end{figure}

As illustrated in Example~\ref{ex:abnormality:stepping}, the \stepping\ approach gives the user an insight into 
the answer set computation in terms of truth assignments to atoms, rather than providing an explicit explanation of the cause of inconsistency like 
the previously discussed debugging approaches.
It also does not make any suggestions on how to change the logic program to make it consistent.
Whereas in \ouroboros\ the user needs to explicitly specify an intended answer set, the \stepping\ approach indirectly allows this but does not require it. In other words, if a user expects a certain answer set, but the logic program is inconsistent or has different answer sets, the stepping can be targeted towards the intended answer set, until it becomes clear why certain atoms in the intended answer set are false or why atoms not expected to be in the answer set are true.
However, the \stepping\ approach can also be applied if a logic program is inconsistent and the user does not know what the answer set should be. In this case, the user can simply step through applicable rules until the stepping computation fails, thus providing an insight into how the inconsistency of the logic program arises.
Note that the stepping approach can also be used to find out how consistent answer sets are derived, in line with 
the approaches discussed in Section~\ref{sec:justifications}.

Like \ouroboros\ and \dwasp, the \stepping\ system can handle logic programs with variables and supports language constructs such as constraints, choice rules, and aggregates. Furthermore, it can easily be used with different ASP solvers.

The theory behind the \stepping\ approach is based on an extension of the \mbox{FLP-semantics} \cite{FaberPL2011} by \citeNS{OetschPT2012},
which coincides with the answer set semantics.
This guarantees that the computation of answer sets using \stepping\ is sound and complete, that is, any answer set can be reached through
the step-wise application of rules and truth assignment of atoms, and any successfully terminated step-wise computation results in an assignment of truth values to atoms forming an answer set.
Thus, if the step-wise computation does not terminate successfully, the current assignment of truth values cannot be extended to an answer set.

To speed up the step-wise computation, especially in large logic programs with variables, where rules have various groundings that can be applied in different steps, the user can perform \emph{jumps}.
A jump is the automatic application of various specified rules in such a way that they are satisfied.
This is useful if the user is not interested in the exact workings of these rules and their influence on a potential answer set.
Note that it only makes sense to use a jump if the chosen rules can be satisfied given the current truth assignment, so the user should be sure that the chosen jumping rules do not pose a problem.

\begin{figure}[th]
\includegraphics[width=\textwidth]{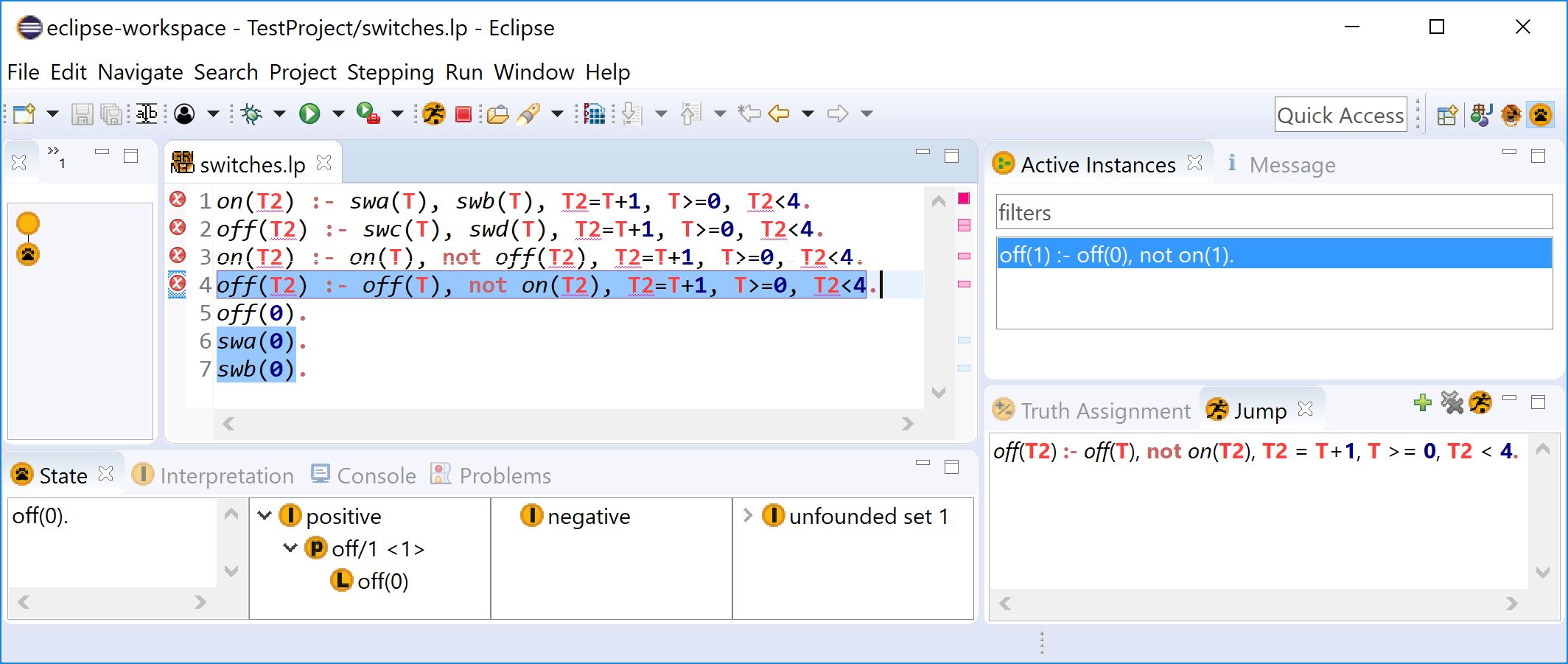}
\caption{The user chooses $\R_4$ as a rule for jumping.}
\label{fig:lights.stepping_addJump}
\end{figure}

\begin{examplecontpage}{ex:light}\label{ex:light.stepping}
Consider again the logic program about a light bulb and the four switches to turn the light on and off.
We encode this in $\newprogram\label{prg:lights.stepping}$ for the time steps $t=0 \ldots 3$.
Figure~\ref{fig:lights.stepping_addJump} illustrates $\programr\ref{prg:lights.stepping}$ and the scenario where the user
chose the fact $\off(0)$ in the first step and now decides to perform a jump on~$\R_4$ (see the `jump' tab).
Since the jump only considers the current assignment of truth values and the chosen rule(s), 
 it makes $\off(1)$, $\off(2)$, and $\off(3)$ true and $\on(1)$, $\on(2)$, and $\on(3)$ false by repeatedly applying $\R_4$.
\begin{figure}[th]
\includegraphics[width=\textwidth]{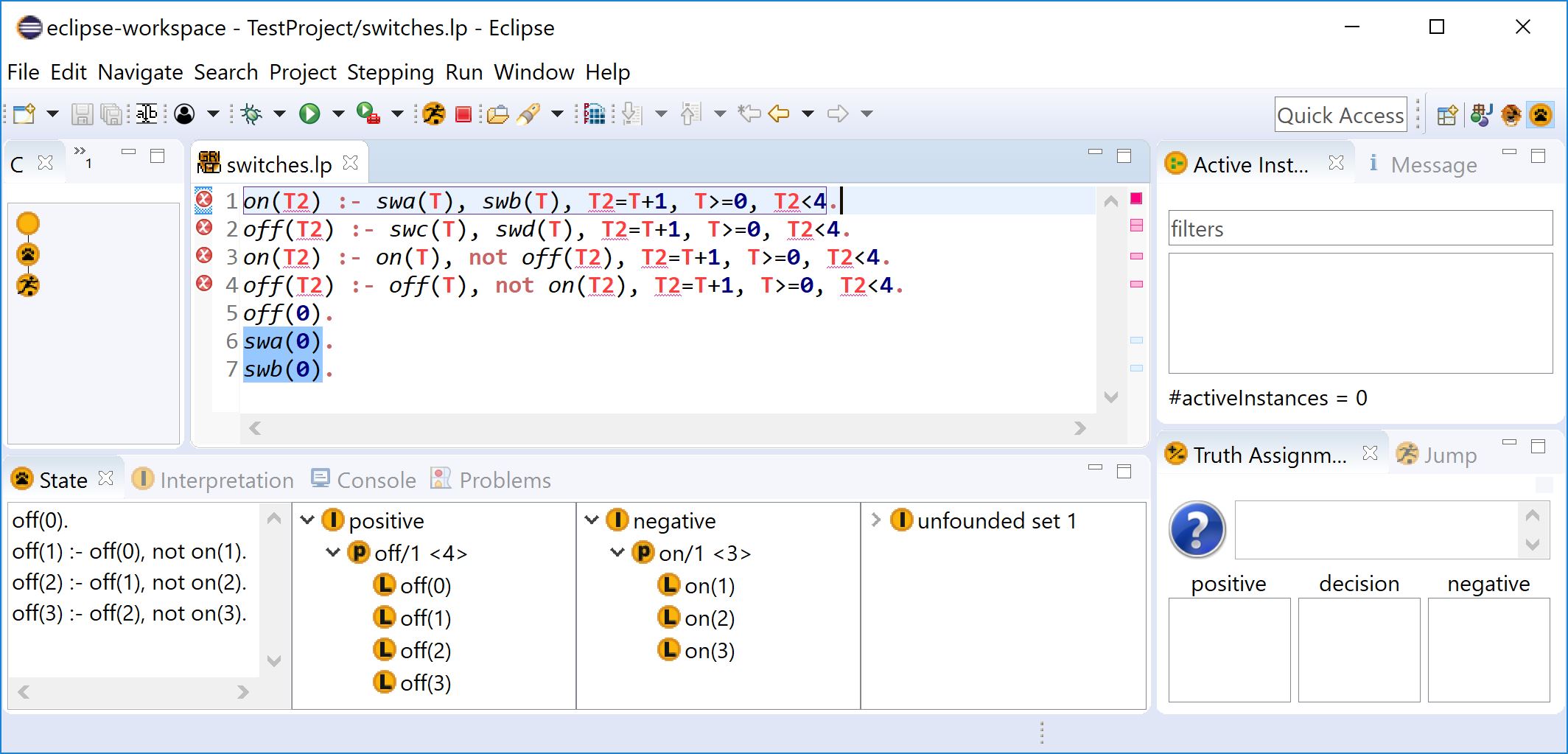}
\caption{Truth assignment and applicable facts (highlighted blue) after the jump.}
\label{fig:lights.stepping_jumped}
\end{figure}
 This automatic assignment is shown in the `state' tab in Figure~\ref{fig:lights.stepping_jumped}, along with the 
 grounded rules used in the automatic steps of the jump. As illustrated by the blue highlighting, at this point
 only facts $\swa(0)$ and $\swb(0)$ are applicable. Performing steps on these two facts results in $\R_1$ being applicable,
\begin{figure}[th]
\includegraphics[width=\textwidth]{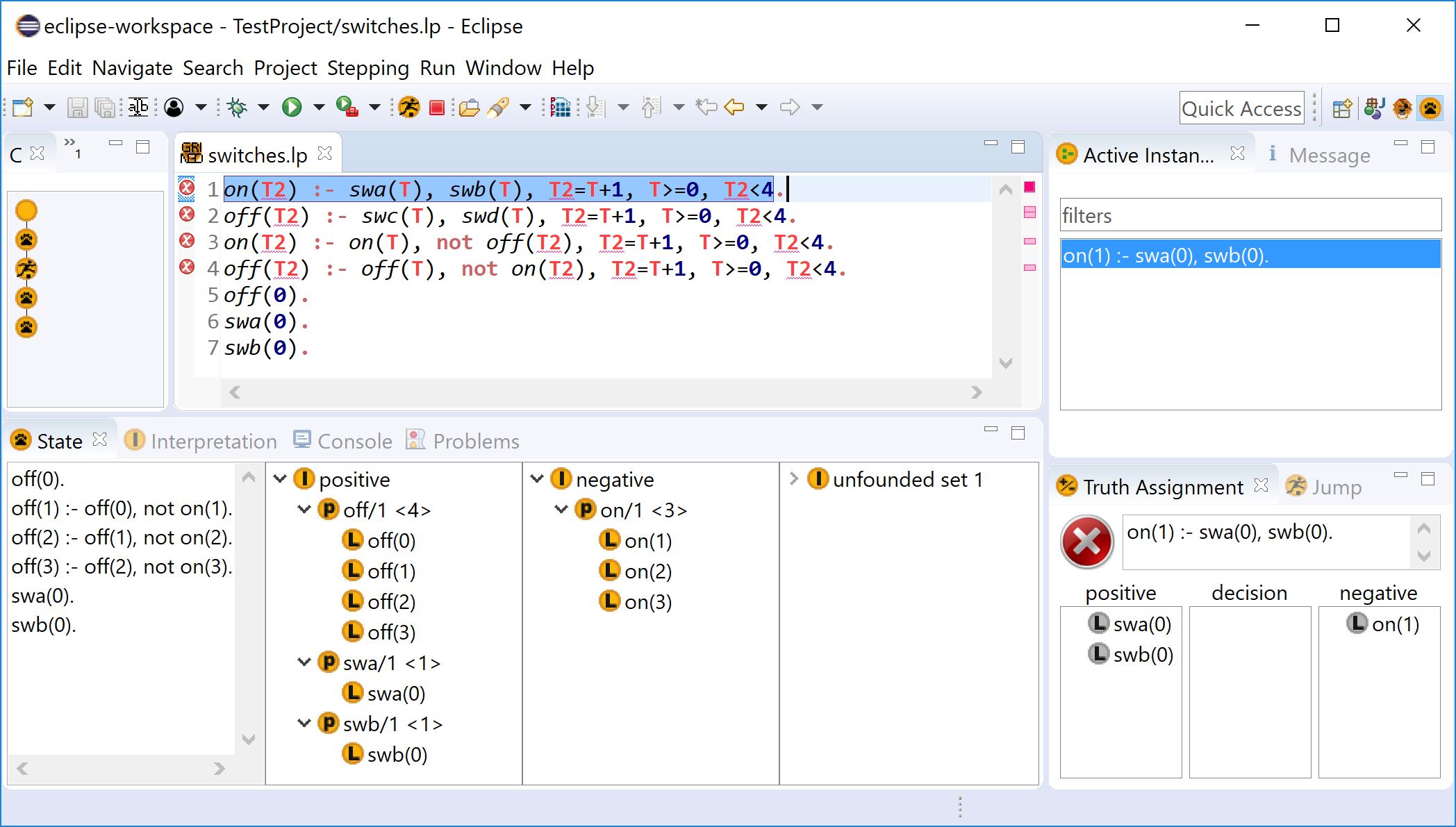}
\caption{Failure of the \stepping\ computation.}
\label{fig:lights.stepping_jumped_fail}
\end{figure}
  but the rule cannot be satisfied \wrt\ the current truth assignment, as shown in Figure~\ref{fig:lights.stepping_jumped_fail}.
  The failure provides insights as to why there is no answer set in which the bulb is turned $\off$ at $t \geq 0$. Namely, 
  the reason it may be turned $\off$ is inertia (application of rule~$\R_4$), however, since switches $\swa$ and $\swb$ are 
  pushed, it follows that the light bulb must be turned on at $t=1$. This conflicts with the previous inertia assumption
  that the light is not turned on ($\Not \on(1)$ in $\R_4$ when deriving $\off(1)$).\qed
\end{examplecontpage}

%
%
%

\vspace{2cm}
\CHANGED
\subsection{Summary and Discussion}
\label{sec:discussion.debugging}

In Sections~\ref{sec:debugging:spock} to~\ref{sec:debugging:stepping}, we outlined the most prominent approaches to ASP debugging, i.e. the explanation of non-existence of answer sets in terms of semantic errors.
In contrast to the justification approaches discussed Section~\ref{sec:justifications}, where the truth value of literals is explained in detail by referring to truth values of other literals used in their derivation, the explanations provided by debugging approaches can seem rather minimalistic. Indeed, debugging aims at providing a \emph{pointer} to the cause of inconsistency rather than a full-fledged explanation.
Furthermore, 
we
\END
have seen that these approaches follow different ideas as to what an explanation should encompass and that they use different methodologies to achieve this.
Tables~\ref{tab:comparison:debugging} and~\ref{tab:comparison:debugging2} provide a comparative overview of the differences and similarities of the surveyed debugging approaches.
\CHANGED
In particular, Table~\ref{tab:comparison:debugging} compares debugging approaches concerning the type of logic programs that can be debugged, whether or not logic programs with variables as well as with language constructs such as aggregates or arithmetic terms can be debugged, and whether the approach can also be used to explain consistent logic programs.
Table~\ref{tab:comparison:debugging2} complements this by illustrating whether the debugging approaches require an intended answer set, or rather, whether they detect mistakes with respect to potentially intended answer sets, which types of errors in a logic program the debugging approaches distinguish, and whether the user can or has to interact with the debugger.
\END

In the following, we discuss some of the distinguishing features in more detail, to facilitate users to choose the appropriate debugging approach for their application.
  
 \begin{table}
  \caption{Comparison of explanation approaches for inconsistent logic programs.}
 \label{tab:comparison:debugging}

 \begin{tabular}{ l l l l l} 
  \multirow{2}{2.5cm}{\textbf{\changed{debugging approach}}} & \multirow{2}{2cm}{\textbf{\changed{type of logic program}}} & \multirow{2}{2cm}{\textbf{variables supported}} & \multirow{2}{2.5cm}{\textbf{additional language constructs}} & \multirow{2}{2cm}{\textbf{explains consistent LPs}} \\\\ \\
  \hline
  \multirow{2}{2.5cm}{\spock\ transformation 1} & normal LP & no & no & yes \\ \\
  \hline 
  \multirow{2}{2.5cm}{\spock\ transformation 2} & LP & no & no & \multirow{2}{2cm}{only non answer sets}\\ \\
  \hline
  \ouroboros & extended LP & yes & \multirow{2}{2.5cm}{arithmetic, comparison} & \multirow{2}{2cm}{only non answer sets}\\ \\
  \hline
  interactive \spock & LP & no & no & \multirow{2}{2cm}{only non answer sets}\\ \\
  \hline
  \dwasp & LP & yes & no & no\\
  \hline
  \stepping & \changed{LP\footnote{\changed{The earlier version of the \stepping\ approach \cite{OetschPT2011} uses extended normal programs.}}} & yes & \multirow{2}{2.5cm}{aggregates, weight constraints, external atoms} & yes \\ \\ \\ \\
  \hline
 \end{tabular}
 \end{table}

 \begin{table}
  \caption{Comparison of explanation approaches for inconsistent logic programs (continued).}
 \label{tab:comparison:debugging2}
 \begin{tabular}{  l l l l l  } 
  \multirow{2}{3cm}{\textbf{debugging approach}} & \multirow{2}{3cm}{\textbf{intended answer set}} & \multirow{2}{3cm}{\textbf{error types}} & \multirow{2}{2.5cm}{\textbf{user interaction}}  \\\\ \\
  \hline
  \multirow{2}{3cm}{\spock\ transformation 1} & \multirow{2}{3cm}{possible but not required (automatically generated)} & \multirow{2}{3cm}{unsatisfied rule, unsupported atom, unfounded atom} &  possible \\ \\ \\ \\
  \hline 
  \multirow{2}{3cm}{\spock\ transformation 2} & \multirow{2}{3cm}{possible but not required (automatically generated)} & \multirow{2}{3cm}{unsatisfied rule/constraint, unsupported atom, unfounded atom} & possible \\ \\ \\ \\
  \hline
  \ouroboros & required & \multirow{2}{3cm}{unsatisfied rule/constraint, unfounded atom} &  \multirow{2}{2.5cm}{required for intended answer set} \\ \\ \\
  \hline
  interactive \spock & \multirow{2}{3cm}{possible but not required} & \multirow{2}{3cm}{unsatisfied rule/constraint, unsupported atom, unfounded atom} & required \\ \\ \\ \\
  \hline
  \dwasp & \multirow{2}{3cm}{possible but not required} & \multirow{2}{3cm}{minimal unsatisfiable core} & required \\ \\ 
  \hline
  \stepping & \multirow{2}{3cm}{not required but (indirectly) possible} & \multirow{2}{3cm}{unsatisfiability of rules, conflicting truth value of atoms} & required \\ \\ \\
  \hline
 \end{tabular}
 \end{table}

 \subsubsection{Knowledge Representation versus Programming}
 As discussed by \citeNS{Cabalar2011}, \changed{logic programs under the answer set semantics} are seen as a pure knowledge representation and reasoning formalism by some and as a programming language by others.
 It is therefore not surprising that explanation and debugging approaches reflect this difference.
 Seeing ASP as a knowledge representation formalism, a user represents knowledge in terms of a logic program and uses the answer set semantics to find out which conclusions can be drawn from this knowledge.
 The user may also represent a problematic situation and compute answer sets to find a solution to the problem.
 Especially in the latter of these two cases, the user most likely has no idea what the solution may be, in other words, there is no answer set intended by the user.
 On the other hand, if ASP is seen as a programming language, the user may well have an idea as to what the solution, i.e. the answer sets, should look like.
 
 Taking these considerations into account, the \spock\ approach (Section~\ref{sec:debugging:spock}) may be more suitable for knowledge representation applications, as it does not require that the user specifies an intended answer set.
 Sets of literals are generated automatically as potential answer sets, which are then justified as to \emph{why} they are not actual answer sets.
 Similarly, the \stepping\ approach (Section~\ref{sec:debugging:stepping}) does not require the user to have an answer set in mind as applicable rules are automatically determined and the user can then freely choose which one to use.
 However, both approaches allow the user to guide the explanation towards specific literals that may be expected in an answer set.
 
 The interactive debugging approaches (Sections~\ref{sec:debugging:interactive} and~\ref{sec:debugging:dwasp}) take a programming language rather than a knowledge representation view on ASP, as they assume that the user has at least some idea as to what an answer set should look like, querying the user about the expected truth values of some literals. The user can certainly choose these truth values at random, making the interactive approaches applicable even if the user has no answer set in mind. However, this \changed{is not} their intended usage. Note also that in order to know the truth value of a literal chosen by the debugging approach, the user essentially has to have an answer set in mind, as the user does not know upfront which literal will be chosen as a query.
 
 The \ouroboros\ approach (Section~\ref{sec:debugging:ouroboros}) is clearly on the programming language end of the spectrum as it requires the user to specify a complete intended answer set.
 The user could of course choose an `intended' answer set at random, but, again, this is not the usage envisaged by this approach.

 \subsubsection{Error Classification}
 As in the case of justifications for consistent logic programs, the debugging approaches also differ regarding the elements used for explaining the inconsistency.
 More precisely, they identify different types of `errors' causing a set of literals to not be an answer set.
 Broadly speaking, two different ideas towards errors can be distinguished: the classification of errors into different classes or the reduction of all errors two one `class'.
 
  \dwasp\ and the \stepping\ approach do not use any named error classes, thus following the latter idea.
  In \dwasp\ errors are sets of rules that, when blocked, make the program consistent.
  However, there is no further explanation as to \emph{why} this is the case.
  On the other hand, errors in the \stepping\ approach are only indirectly specified. They are indicated by (partial) truth assignments to literals, which lead to a contradiction.
  Again, there is no further explanation, other than the rule causing the contradiction.
\changed{In contrast, the other approaches reviewed here distinguish different classes of errors.}
 
 The \spock\ system and the two approaches based on it (interactive debugging and \ouroboros) use mostly the same classes of errors.
 As previously explained, these are violations of the definition of answer sets by 
 \citeNS{LinZ2004} and \citeNS{Lee2005} (see Definition~\ref{def:answerSet_LinZ_Lee} on page~\pageref{def:answerSet_LinZ_Lee}), namely
 unsatisfied rules, unsupported atoms, and unfounded atoms. 
 
\changed{Interestingly, one reason for inconsistency of logic programs often discussed in the literature \cite{YouY1994,Syrjanen2006,Costantini2006,SchulzST2015} is not explicitly pointed out by \spock, namely \emph{odd-length (negative dependency) cycles}.
In Examples~\ref{ex:abnormality.spock} and~\ref{ex:abnormality:spock2} (see pages~\pageref{ex:abnormality.spock} and~\pageref{ex:abnormality:spock2}), the odd-length cycle in $\R_2$ of $\programr\ref{prg:abnormality.spock}$ is only indirectly pointed out:
$\answersetr\ref{as:2:prg:abnormality.spock}$ expresses that $\{b\}$ is not an answer set of $\programr\ref{prg:abnormality.spock}$ since all rules with head $b$ are blocked by $\{b\}$.
Taking a closer look at $\programr\ref{prg:abnormality.spock}$, we realise that the only rule with head $b$ is $\R_2$ and that the reason for it being blocked is that $\Not b$ is in the body of $\R_2$.
However, if $\programr\ref{prg:abnormality.spock}$ was a large logic program, it would be infeasible to check all rules with head $b$ to find out that one of them may comprise an odd-length cycle, causing the rule to be blocked.
Similarly, $\answersetr\ref{as:4:prg:abnormality.spock}$ indirectly points out the odd-length cycle by stating that $\R_2$ is applicable but its head is not contained in the set $\{a\}$. 
We then realise that the reason for $\R_2$ not being satisfied is the odd-length cycle.

\begin{samepage}
\begin{example}\label{ex:oddcycle}\nopagebreak
Let $\newprogram\label{prg:oddcycle}$ be the inconsistent logic program with:
\begin{gather}
 \R_1: \ a \lparrow b
 \hspace{2cm}
 \R_2: \ b \lparrow \Not a
\end{gather}
The answer sets of $\transKern{\programr\ref{prg:oddcycle}} \cup \transEx{\programr\ref{prg:oddcycle}}$ (when using minimisation) are:
\begin{itemize}
 \item $\newanswerset\label{as:1:prg:oddcyle.spock} = \{a,b, \abC{b}, \appRule{\R_1}, \blockRule{\R_2}\}$
 \item $\newanswerset\label{as:2:prg:oddcyle.spock} = \{a, \abC{a}, \blockRule{\R_1}, \blockRule{\R_2}\}$
 \item $\newanswerset\label{as:3:prg:oddcyle.spock} = \{b, \abP{\R_1}, \appRule{\R_1}, \appRule{\R_2}\}$
 \item $\newanswerset\label{as:4:prg:oddcyle.spock} = \{\abP{\R_2}, \blockRule{\R_1}, \appRule{\R_2}\}$
\end{itemize}
None of the answer sets captures the fact that there is an odd-length cycle
\mbox{$a \leftarrow \Not a$}.
For a similarly structured logic program with more rules and derivation steps between $a$ and $\Not a$ it
would therefore be difficult to identify that the reason of the inconsistency is an odd-length cycle.\qed
\end{example}
\end{samepage}

A debugging approach related to \spock\ \cite{Syrjanen2006} explicitly points out inconsistencies due to odd-length cycles. The approach also uses the input transformation $\inputP{\P}$ of a logic program together with a meta-encoding of two types of errors: odd-length cycles and violated constraints.
However, \emph{all} odd-length cycles are considered as faulty, even though some odd-length cycles do not cause a logic program to be inconsistent. In contrast to the \spock\ system, faults are pointed out independent of intended or potential answer sets.} 
 

\CHANGED
Another class of `errors' not considered in any of the debugging approaches are those of contradictory answer sets.
In fact, none of the debugging approaches reviewed here deals with contradictory atoms in an answer set.
\citeNS{SchulzST2015} show that logic programs with contradictory answer sets include different types of semantic errors than inconsistent logic programs. This is also taken into account in the inconsistency measurements of \citeNS{UlbrichtTB2016}.
\END

 \subsubsection{Large and Real-World Logic Programs}
 We already hinted at the fact that the different debugging approaches require various levels of user interaction to obtain an explanation.
 In particular, some approaches require the user to specify an intended answer set before starting the debugging process, especially the \ouroboros\ system.
 This can be difficult if faced with a large logic program, potentially comprising hundreds of atoms.
 Furthermore, using the \stepping\ approach, the user has to step through every single applicable rule, unless being sure that some rules are not problematic, in which case the jumping feature can be used.
\CHANGED
Assuming that the user does not have any idea why the logic program is inconsistent, thus ruling out jumping, the stepping approach can take a long time and also be prone to errors for these large programs.%
\END

 In contrast, for approaches requiring only little user interaction, first and foremost the \spock\ system, the amount of
interaction does not increase when dealing with large logic programs.
 However, note that the more literals occur in a program, the more explanations are computed by \spock, namely one for each potential answer set. 
 The user interaction is thus implicitly required after explanations are computed, since the user then has to decide which explanations to take into account.
 It follows, that, just like the \ouroboros\ and \stepping\ approaches, using spock with large logic programs may
\CHANGED
take a long time.
\END
 
The two interactive approaches (the one based on \spock\ and the \dwasp\ system) 
\CHANGED
are the ones that require least user interaction
\END when handling large logic programs.
This is because queries are determined in such a way that the user's answer provides maximal information gain. Consequently, the total number of queries generated is as small as possible.
From a user's point of view, answering a query on the expected truth value of a single literal may furthermore be easier than specifying the truth value of all literals at once or choosing a meaningful explanation from all the ones generated.
 
 When using ASP in practice, logic programs often include additional language constructs, make use of variables, and are seldom limited to normal rules.
 These are important consideration when choosing a debugging approach. 
 Currently, \ouroboros\ and the \stepping\ approach are the only ones to handle both negation-as-failure and explicit negation, variables, and additional language constructs, where the \stepping\ approach supports more constructs than \ouroboros.
 \dwasp\ supports variables, but to the best of our knowledge no explicit negation or additional language constructs. Nevertheless, is to be assumed that these will be supported in the future since \dwasp\ is implemented in terms of the ASP solver \wasp, which is able to handle these.

\vspace{2cm}
\changed{\section{Related Work}}
\label{sec:related}

\CHANGED
In this survey, we focussed on justification and debugging approaches for logic programs under the \emph{answer set} semantics.
Historically, the concept of justifications can be traced back to the works of~\citeN{Shapiro1983} and~\citeN{sterling1986explanation}, where they have been used as a means for identifying bugs in programs.
Later, \citeN{Lloyd1987} introduced the notions of uncovered atoms and incorrect rules under the completion semantics~\cite{clark1978negation} while \citeN{sterling1989explaining} explained Prolog expert systems using a meta-interpreter.

An important notion for understanding errors in ASP is the concept of a supported set of atoms, which
was introduced by~\citeN{Pereira1991ContradictionRW} and further elaborated by~\citeN{PEREIRA1993}.
Another important concept is the  notion of assumptions, which was introduced for truth maintenance systems by~\citeN{DEKLEER1986} and developed for logic programming by~\citeN{pereira1993debugging}.
\citeN{Specht93} presented one of the first techniques to compute complete proof trees
for bottom-up evaluation of database systems by means of a program transformation.
Further techniques for computing justifications or explanations for Prolog by means of meta-interpreters or program transformations can be found in~\cite{sterling1994art} and~\cite{bratko2001prolog}.
Furthermore, explanation approaches have been developed for knowledge representation paradigms related to~ASP.
For instance, \citeNS{AroraRRSS1993} present explanations for deductive databases and \citeNS{FerrandLT2006} for constraint logic programs and constraint satisfaction problems.

Regarding justifications for logic programs under the answer set semantics, \citeNS{BrainV2005} were one of the first to tackle this issue, by presenting two algorithms for producing natural language explanations as to why a (set of) literal(s) is or is not part of an answer set. 
In the first case, applicable rules are provided as an explanation, whereas in the second case contradictions (concerning the truth values of atoms) are pointed out.

Off-line justifications \cite{PontelliS2006,PontelliSE2009}, as reviewed in Section~\ref{sec:offline},
use graphs as justifications, expressing why an atoms is (not) contained in a given answer set.
This approach can be traced back to tabled justifications for Prolog~\cite{roychoudhuryRR00,PemmasaniGDRR03}.
\citeNS{AlbrechtKK2013} further show how off-line explanation graphs can be constructed from a graphical representation of logic programs called extended dependency graph.
The root of causal justifications can be traced back to~\cite{Cabalar2011},
where an extension of the stable semantics with causal proofs was introduced,
and~\cite{CabalarF13}, where an algebraic characterisation of this semantics was developed.
Argumentation-based answer set justifications \cite{SchulzST2013} are a predecessor of LABAS justifications. 
They share the argumentative flavour of LABAS justifications but use a slightly different way of constructing arguments and justifications.

\citeNS{ErdemO2015} use ASP to construct explanations for biomedical queries.
These explanations have a tree structure expressing derivations of a literal in question and have a close relationship with off-line justifications.
\citeNS{Lifschitz17} introduces a methodology that facilitates
the design of encodings that are easy to understand and provably correct.
In addition to the implementations of justification and debugging approaches reviewed here, \citeNS{PerriRTCV2007} integrate an explanation and debugging component into the \texttt{DLV} solver. 

As we saw throughout this survey, many justification approaches construct a graphical explanation.
Graph representations of logic programs have also been extensively studied for other purposes \cite{CostantiniDP2002,CostantiniP2010}.
Graphs can for instance be useful for the computation of answer sets, as is the purpose of attack graphs \cite{DimopoulosT1996}, rule graphs \cite{Dimopoulos1996}, and block graphs \cite{Linke2001}
and their extensions \cite{LinkeS2004,KonczakLS2006}.
Furthermore, \citeNS{Costantini2001} and \citeNS{CostantiniP2011} study desirable properties of graphs representing logic programs and \citeNS{Costantini2006} uses cycle graphs to prove conditions for the existence of answer sets.

Various IDEs for ASP also make use of graphical representations of logic programs or visualise dependencies between literals to help the user understand
a problem represented as a logic program.
For example, for the \texttt{DLV} solver a visual computation tracing feature \cite{CalimeriLRV2009} as well as a dependency graph feature \cite{FebbraroRR2011} have been developed.
Furthermore, the \texttt{VIDEAS} system \cite{OetschPSTZ2011} uses entity relationship graphs of logic programs for model-driven engineering in ASP and, in the `Visual ASP' system \cite{FebbraroRR2010}, the user can draw a graph, which is then translated into a logic program.
\END

\section{Conclusion}\label{sec:conclusion}
\changed{\citeNS{Lifschitz10} lists thirteen different definitions of the concept of answer set (and points out that even more exist).
These definitions are equivalent (at least for normal programs), but provide alternative points of view on the intuitive meaning of logic programs or lead to different algorithms for generating answer sets.
In this sense, it is not surprising that there exist several ways of explaining the solutions to consistent programs and the errors in inconsistent ones.}
In this survey, we have reviewed and compared the most prominent explanation approaches for both consistent and inconsistent logic programs \changed{under the answer set semantics and pointed out their differences and similarities.}
\changed{These approaches try to answer important `why'-questions regarding answer sets, namely \emph{why} a set of literals is or is not an answer set, or \emph{why} a logic program is inconsistent. 
Approaches aiming at answering the first question for consistent logic programs are referred to as \emph{justification} approaches, while
explanation approaches trying to answer the second question for inconsistent logic programs are referred to as \emph{debugging} approaches.}
The latter take a more global view than justification approaches:
in debugging approaches the explanation is \wrt\ a whole \emph{set} that is not an answer set, whereas in most justification approaches the explanation is \wrt\ one \emph{literal} that is (not) in an answer set.

As we have seen in Sections~\ref{sec:discussion} and~\ref{sec:discussion.debugging}, the different justification and debugging approaches suffer from various issues.
Building upon these observations, in the following we suggest some considerations for future research that are mainly independent of
philosophical choices made by different approaches.
\changed{These are particularly important in the light of the European Union's new General Data Protection Regulation (GDPR), which states that explanations should consist of ``meaningful information about the logic involved'' and be ``concise, intelligible and easily accessible'' \cite{Goodman2016european}. Since the approaches discussed here construct explanations based on the logical connection between  rules and literals leading to the existence of a particular answer set or to inconsistency, at least the first part of the first GDPR condition, i.e. ``information about the logic involved'', can be deemed satisfied by these approaches. The proposed directions of research are as follows:}

\begin{itemize}
 \item Number of explanations \changed{(tackling the conciseness and intelligibility required by the GDPR):} As previously discussed, most justification and debugging approaches suffer from a large number of possible explanations when dealing with large programs with, potentially, many (and long) dependencies between literals. This is not feasible in practice, so a method for choosing the most suitable explanation(s) is needed. This could for example be tackled by querying the user as in \dwasp\ and the interactive \spock\ approach.

 \item Size of explanations \changed{(tackling meaningfulness of information, conciseness, intelligibility, and easy accessibility required by the GDPR):} A related problem is the growth in size, from which many of the justification approaches suffer. Large explanations are infeasible in many practical applications, since they make it difficult for the user to understand the explanation. The development of techniques for collapsing less important parts of an explanation provides a challenging topic for the future.
 \item Language constructs and variables: We have seen that, especially among the justification approaches, there is little support for logic programs that contain language constructs such as aggregates, weight constraints, etc. Many approaches are not even able to efficiently handle variables. In order to apply explanations in practice, these issues will have to be addressed.
 \item Cross-fertilisation of justification and debugging: Most current approaches either focus on justifying consistent programs or debugging inconsistent programs.
 A first step towards the cross-fertilisation of the two was made by \mbox{\citeNS{DamasioMA2015}}, who combine the second \spock\ transformation approach with why-not provenance justifications.
 \item Going beyond debugging: Current debugging approaches merely point out errors in a program, leaving the fixing of these errors to the user.
 The automatic revision of inconsistent logic programs is thus an interesting, and challenging, topic for future investigations. 
 A first step in this direction was made by \citeNS{LiVPSB2015}, who use inductive logic programming to achieve a semi-automatic revision of logic programs.
\end{itemize}

\CHANGED
\noindent
Meeting the requirements of the GDPR will be a challenging task, especially since conditions like meaningfulness and intelligibility of information may have to be realised differently for ASP experts and non-experts.
Applications of ASP explanation approaches will thus determine whether or not they meet the required conditions.
In this sense, an exciting prospect for the future is the
combination of the advantages and minimisation the disadvantages of all the different approaches for answering a `why'-question in answer set programming.
\END

\paragraph{Acknowledgements} We are thankful to the anonymous reviewers for their valuable feedback, which helped to improve the paper.

\bibliography{justificationSurvey}

\begin{thebibliography}{}

\bibitem[\protect\citeauthoryear{Albrecht, Kr{\"{u}}mpelmann, and
  Kern-Isberner}{Albrecht et~al\mbox{.}}{2013}]{AlbrechtKK2013}
{\sc Albrecht, E.}, {\sc Kr{\"{u}}mpelmann, P.}, {\sc and} {\sc Kern-Isberner,
  G.} 2013.
\newblock {Construction of Explanation Graphs from Extended Dependency Graphs
  for Answer Set Programs}.
\newblock In {\em Revised Selected Papers of the Kiel Declarative Programming
  Days (KDPD'13)}. 1--16.

\bibitem[\protect\citeauthoryear{Alviano, Dodaro, Faber, Leone, and
  Ricca}{Alviano et~al\mbox{.}}{2013}]{AlvianoDFLR2013}
{\sc Alviano, M.}, {\sc Dodaro, C.}, {\sc Faber, W.}, {\sc Leone, N.}, {\sc
  and} {\sc Ricca, F.} 2013.
\newblock {WASP: A Native ASP Solver Based on Constraint Learning}.
\newblock In {\em Proceedings of the 12th International Conference on Logic
  Programming and Nonmonotonic Reasoning (LPNMR'13)}. 54--66.

\bibitem[\protect\citeauthoryear{Alviano, Dodaro, Leone, and Ricca}{Alviano
  et~al\mbox{.}}{2015}]{AlvianoDLR15}
{\sc Alviano, M.}, {\sc Dodaro, C.}, {\sc Leone, N.}, {\sc and} {\sc Ricca, F.}
  2015.
\newblock {Advances in WASP}.
\newblock In {\em Proceedings of the 13th International Conference on Logic
  Programming and Nonmonotonic Reasoning (LPNMR'15)}. 40--54.

\bibitem[\protect\citeauthoryear{Arora, Ramakrishnan, Roth, Seshadri, and
  Srivastava}{Arora et~al\mbox{.}}{1993}]{AroraRRSS1993}
{\sc Arora, T.}, {\sc Ramakrishnan, R.}, {\sc Roth, W.~G.}, {\sc Seshadri, P.},
  {\sc and} {\sc Srivastava, D.} 1993.
\newblock {Explaining Program Execution in Deductive Systems}.
\newblock In {\em Proceedings of the 3rd International Conference on Deductive
  and Object-Oriented Databases (DOOD'93)}. 101--119.

\bibitem[\protect\citeauthoryear{Balduccini and Girotto}{Balduccini and
  Girotto}{2010}]{BalducciniG2010}
{\sc Balduccini, M.} {\sc and} {\sc Girotto, S.} 2010.
\newblock {Formalization of Psychological Knowledge in Answer Set Programming
  and its Application}.
\newblock {\em Theory and Practice of Logic Programming\/}~{\em 10,\/}~4-6,
  725--740.

\bibitem[\protect\citeauthoryear{B{\'{e}}atrix, Lef{\`{e}}vre, Garcia, and
  St{\'{e}}phan}{B{\'{e}}atrix et~al\mbox{.}}{2016}]{BeatrixLGS2016}
{\sc B{\'{e}}atrix, C.}, {\sc Lef{\`{e}}vre, C.}, {\sc Garcia, L.}, {\sc and}
  {\sc St{\'{e}}phan, I.} 2016.
\newblock {Justifications and Blocking Sets in a Rule-Based Answer Set
  Computation}.
\newblock In {\em Technical Communications of the 32nd International Conference
  on Logic Programming (ICLP'16)}. 6:1--6:15.

\bibitem[\protect\citeauthoryear{Boenn, Brain, {De Vos}, and Fitch}{Boenn
  et~al\mbox{.}}{2011}]{BoennBVF2011}
{\sc Boenn, G.}, {\sc Brain, M.}, {\sc {De Vos}, M.}, {\sc and} {\sc Fitch,
  J.~P.} 2011.
\newblock {Automatic Music Composition Using Answer Set Programming}.
\newblock {\em Theory and Practice of Logic Programming\/}~{\em 11,\/}~2-3,
  397--427.

\bibitem[\protect\citeauthoryear{Brain and {De Vos}}{Brain and {De
  Vos}}{2005}]{BrainV2005}
{\sc Brain, M.} {\sc and} {\sc {De Vos}, M.} 2005.
\newblock {Debugging Logic Programs under the Answer Set Semantics}.
\newblock In {\em Proceedings of the 3rd Workshop on Answer Set Programming,
  Advances in Theory and Implementation (ASP'05)}.

\bibitem[\protect\citeauthoryear{Brain and {De Vos}}{Brain and {De
  Vos}}{2008}]{BrainV2008}
{\sc Brain, M.} {\sc and} {\sc {De Vos}, M.} 2008.
\newblock {Answer Set Programming - a Domain in Need of Explanation: A Position
  Paper}.
\newblock In {\em Proceedomgs of the 3rd International Workshop on
  Explanation-aware Computing (ExaCt'08)}. 37--48.

\bibitem[\protect\citeauthoryear{Brain, Gebser, P{\"{u}}hrer, Schaub, Tompits,
  and Woltran}{Brain et~al\mbox{.}}{2007a}]{BrainGPSTW2007b}
{\sc Brain, M.}, {\sc Gebser, M.}, {\sc P{\"{u}}hrer, J.}, {\sc Schaub, T.},
  {\sc Tompits, H.}, {\sc and} {\sc Woltran, S.} 2007a.
\newblock {Debugging ASP Programs by Means of ASP}.
\newblock In {\em Proceedings of the 9th International Conference on Logic
  Programming and Nonmonotonic Reasoning (LPNMR'07)}. 31--43.

\bibitem[\protect\citeauthoryear{Brain, Gebser, P{\"{u}}hrer, Schaub, Tompits,
  and Woltran}{Brain et~al\mbox{.}}{2007b}]{BrainGPSTW2007a}
{\sc Brain, M.}, {\sc Gebser, M.}, {\sc P{\"{u}}hrer, J.}, {\sc Schaub, T.},
  {\sc Tompits, H.}, {\sc and} {\sc Woltran, S.} 2007b.
\newblock {"That is illogical captain!" - The debugging support tool spock for
  answer-set programs: System description}.
\newblock In {\em Proceedings of the 1st International Workshop on Software
  Engineering for Answer Set Programming (SEA'07)}. 71--85.

\bibitem[\protect\citeauthoryear{Bratko}{Bratko}{2001}]{bratko2001prolog}
{\sc Bratko, I.} 2001.
\newblock {\em Prolog programming for artificial intelligence}.
\newblock Pearson education.

\bibitem[\protect\citeauthoryear{Brewka, Eiter, and Truszczynski}{Brewka
  et~al\mbox{.}}{2011}]{BrewkaET2011}
{\sc Brewka, G.}, {\sc Eiter, T.}, {\sc and} {\sc Truszczynski, M.} 2011.
\newblock {Answer Set Programming at a Glance}.
\newblock {\em Communications of the ACM\/}~{\em 54,\/}~12, 92--103.

\bibitem[\protect\citeauthoryear{Busoniu, Oetsch, P{\"{u}}hrer, Skocovsky, and
  Tompits}{Busoniu et~al\mbox{.}}{2013}]{BusoniuOPST2013}
{\sc Busoniu, P.-A.}, {\sc Oetsch, J.}, {\sc P{\"{u}}hrer, J.}, {\sc Skocovsky,
  P.}, {\sc and} {\sc Tompits, H.} 2013.
\newblock {SeaLion: An Eclipse-Based IDE for Answer-Set Programming with
  Advanced Debugging Support}.
\newblock {\em Theory and Practice of Logic Progrmming\/}~{\em 13,\/}~4-5,
  657--673.

\bibitem[\protect\citeauthoryear{Cabalar}{Cabalar}{2011}]{Cabalar2011}
{\sc Cabalar, P.} 2011.
\newblock {Answer Set; Programming?}
\newblock In {\em Logic Programming, Knowledge Representation, and Nonmonotonic
  Reasoning - Essays Dedicated to Michael Gelfond on the Occasion of His 65th
  Birthday}. 334--343.

\bibitem[\protect\citeauthoryear{Cabalar and Fandinno}{Cabalar and
  Fandinno}{2013}]{CabalarF13}
{\sc Cabalar, P.} {\sc and} {\sc Fandinno, J.} 2013.
\newblock An algebra of causal chains.
\newblock {\em CoRR\/}~{\em abs/1312.6134}.

\bibitem[\protect\citeauthoryear{Cabalar and Fandinno}{Cabalar and
  Fandinno}{2016}]{CabalarF16}
{\sc Cabalar, P.} {\sc and} {\sc Fandinno, J.} 2016.
\newblock Justifications for programs with disjunctive and causal-choice rules.
\newblock {\em {Theory and Practice of Logic Programming}\/}~{\em 16,\/}~5-6,
  587--603.

\bibitem[\protect\citeauthoryear{Cabalar and Fandinno}{Cabalar and
  Fandinno}{2017}]{CabalarF2017}
{\sc Cabalar, P.} {\sc and} {\sc Fandinno, J.} 2017.
\newblock {Enablers and Inhibitors in Causal Justifications of Logic Programs}.
\newblock {\em Theory and Practice of Logic Programming\/}~{\em 17,\/}~1,
  49--74.

\bibitem[\protect\citeauthoryear{Cabalar, Fandinno, and Fink}{Cabalar
  et~al\mbox{.}}{2014}]{CabalarFF2014}
{\sc Cabalar, P.}, {\sc Fandinno, J.}, {\sc and} {\sc Fink, M.} 2014.
\newblock {Causal Graph Justifications of Logic Programs}.
\newblock {\em Theory and Practice of Logic Programming\/}~{\em 14,\/}~4-5,
  603--618.

\bibitem[\protect\citeauthoryear{Cabalar, Fandi{\~{n}}o, and Fink}{Cabalar
  et~al\mbox{.}}{2014}]{CabalarFF14}
{\sc Cabalar, P.}, {\sc Fandi{\~{n}}o, J.}, {\sc and} {\sc Fink, M.} 2014.
\newblock A complexity assessment for queries involving sufficient and
  necessary causes.
\newblock In {\em Proceedings of the 14th European Conference on Logics in
  Artificial Intelligence (JELIA'14)}. Lecture Notes in Computer Science, vol.
  8761. Springer, 297--310.

\bibitem[\protect\citeauthoryear{Calimeri, Leone, Ricca, and Veltri}{Calimeri
  et~al\mbox{.}}{2009}]{CalimeriLRV2009}
{\sc Calimeri, F.}, {\sc Leone, N.}, {\sc Ricca, F.}, {\sc and} {\sc Veltri,
  P.} 2009.
\newblock {A Visual Tracer for DLV}.
\newblock In {\em Proceedings of the 2nd International Workshop on Software
  Engineering for Answer Set Programming (SEA'09)}. 79--93.

\bibitem[\protect\citeauthoryear{Clark}{Clark}{1978}]{clark1978negation}
{\sc Clark, K.~L.} 1978.
\newblock Negation as failure.
\newblock In {\em Logic and data bases}. Springer, 293--322.

\bibitem[\protect\citeauthoryear{Costantini}{Costantini}{2001}]{Costantini2001}
{\sc Costantini, S.} 2001.
\newblock {Comparing Different Graph Representations of Logic Programs under
  the Answer Set Semantics}.
\newblock In {\em Proceedings of the 1st International Workshop on Answer Set
  Programming: Towards Efficient and Scalable Knowledge Representation and
  Reasoning (ASP'01)}.

\bibitem[\protect\citeauthoryear{Costantini}{Costantini}{2006}]{Costantini2006}
{\sc Costantini, S.} 2006.
\newblock {On the Existence of Stable Models of Non-Stratified Logic Programs}.
\newblock {\em Theory and Practice of Logic Programming\/}~{\em 6,\/}~1-2,
  169--212.

\bibitem[\protect\citeauthoryear{Costantini, D'Antona, and Provetti}{Costantini
  et~al\mbox{.}}{2002}]{CostantiniDP2002}
{\sc Costantini, S.}, {\sc D'Antona, O.}, {\sc and} {\sc Provetti, A.} 2002.
\newblock {On the Equivalence and Range of Applicability of Graph-based
  Representations of Logic Programs}.
\newblock {\em Information Processing Letters\/}~{\em 84,\/}~5, 241--249.

\bibitem[\protect\citeauthoryear{Costantini and Provetti}{Costantini and
  Provetti}{2010}]{CostantiniP2010}
{\sc Costantini, S.} {\sc and} {\sc Provetti, A.} 2010.
\newblock {Graph Representations of Logic Programs: Properties and Comparison}.
\newblock In {\em Proceedings of the 6th Latin American Workshop on
  Non-Monotonic Reasoning}. 1--14.

\bibitem[\protect\citeauthoryear{Costantini and Provetti}{Costantini and
  Provetti}{2011}]{CostantiniP2011}
{\sc Costantini, S.} {\sc and} {\sc Provetti, A.} 2011.
\newblock {Conflict, Consistency and Truth-Dependencies in Graph
  Representations of Answer Set Logic Programs}.
\newblock In {\em Revised Selected Papers of the 2nd International Workshop on
  Graph Structures for Knowledge Representation and Reasoning (GKR'11)}.
  68--90.

\bibitem[\protect\citeauthoryear{Dam{\'{a}}sio, Analyti, and
  Antoniou}{Dam{\'{a}}sio et~al\mbox{.}}{2013}]{DamasioAA2013}
{\sc Dam{\'{a}}sio, C.~V.}, {\sc Analyti, A.}, {\sc and} {\sc Antoniou, G.}
  2013.
\newblock {Justifications for Logic Programming}.
\newblock In {\em Proceedings of the 12th International Conference on Logic
  Programming and Nonmonotonic Reasoning (LPNMR'13)}. 530--542.

\bibitem[\protect\citeauthoryear{Dam{\'{a}}sio, Moura, and
  Analyti}{Dam{\'{a}}sio et~al\mbox{.}}{2015}]{DamasioMA2015}
{\sc Dam{\'{a}}sio, C.~V.}, {\sc Moura, J.}, {\sc and} {\sc Analyti, A.} 2015.
\newblock {Unifying Justifications and Debugging for Answer-Set Programs}.
\newblock In {\em Technical Communications of the 31st International Conference
  on Logic Programming (ICLP'15)}.

\bibitem[\protect\citeauthoryear{Dam{\'{a}}sio, Pires, and
  Analyti}{Dam{\'{a}}sio et~al\mbox{.}}{2015}]{damasioMA15}
{\sc Dam{\'{a}}sio, C.~V.}, {\sc Pires, J.~M.}, {\sc and} {\sc Analyti, A.}
  2015.
\newblock Unifying justifications and debugging for answer-set programs.
\newblock In {\em Proceedings of the Technical Communications of the 31st
  International Conference on Logic Programming {(ICLP}'15)}, {M.~D. Vos},
  {T.~Eiter}, {Y.~Lierler}, {and} {F.~Toni}, Eds. {CEUR} Workshop Proceedings,
  vol. 1433. CEUR-WS.org.

\bibitem[\protect\citeauthoryear{de~Kleer}{de~Kleer}{1986}]{DEKLEER1986}
{\sc de~Kleer, J.} 1986.
\newblock An assumption-based tms.
\newblock {\em Artificial Intelligence\/}~{\em 28,\/}~2, 127 -- 162.

\bibitem[\protect\citeauthoryear{Denecker, Brewka, and Strass}{Denecker
  et~al\mbox{.}}{2015}]{DeneckerBS2015}
{\sc Denecker, M.}, {\sc Brewka, G.}, {\sc and} {\sc Strass, H.} 2015.
\newblock {A Formal Theory of Justifications}.
\newblock In {\em Proceedings of the 13th International Conference on Logic
  Programming and Nonmonotonic Reasoning (LPNMR'15)}. 250--264.

\bibitem[\protect\citeauthoryear{Denecker and {De Schreye}}{Denecker and {De
  Schreye}}{1993}]{DeneckerS1993}
{\sc Denecker, M.} {\sc and} {\sc {De Schreye}, D.} 1993.
\newblock {Justification Semantics: A Unifiying Framework for the Semantics of
  Logic Programs}.
\newblock In {\em Proceedings of the 2nd International Workshop on Logic
  Programming and Non-monotonic Reasoning (LPNMR'93)}. 365--379.

\bibitem[\protect\citeauthoryear{Dimopoulos}{Dimopoulos}{1996}]{Dimopoulos1996}
{\sc Dimopoulos, Y.} 1996.
\newblock {On Computing Logic Programs}.
\newblock {\em Journal of Automated Reasoning\/}~{\em 17,\/}~3, 259--289.

\bibitem[\protect\citeauthoryear{Dimopoulos and Torres}{Dimopoulos and
  Torres}{1996}]{DimopoulosT1996}
{\sc Dimopoulos, Y.} {\sc and} {\sc Torres, A.} 1996.
\newblock {Graph Theoretical Structures in Logic Programs and Default
  Theories}.
\newblock {\em Theoretical Computer Science\/}~{\em 170,\/}~1-2, 209--244.

\bibitem[\protect\citeauthoryear{Dodaro, Gasteiger, Musitsch, Ricca, and
  Shchekotykhin}{Dodaro et~al\mbox{.}}{2015}]{DodaroGMRS2015}
{\sc Dodaro, C.}, {\sc Gasteiger, P.}, {\sc Musitsch, B.}, {\sc Ricca, F.},
  {\sc and} {\sc Shchekotykhin, K.~M.} 2015.
\newblock {Interactive Debugging of Non-ground ASP Programs}.
\newblock In {\em Proceedings of the 13th International Conference on Logic
  Programming and Nonmonotonic Reasoning (LPNMR'15)}. 279--293.

\bibitem[\protect\citeauthoryear{Dung, Kowalski, and Toni}{Dung
  et~al\mbox{.}}{2009}]{DungKT2009}
{\sc Dung, P.~M.}, {\sc Kowalski, R.~A.}, {\sc and} {\sc Toni, F.} 2009.
\newblock {Assumption-Based Argumentation}.
\newblock In {\em Argumentation in Artificial Intelligence}, {G.~R. Simari}
  {and} {I.~Rahwan}, Eds. Springer US, 199--218.

\bibitem[\protect\citeauthoryear{El-Khatib, Pontelli, and Son}{El-Khatib
  et~al\mbox{.}}{2005}]{El-KhatibPS2005}
{\sc El-Khatib, O.}, {\sc Pontelli, E.}, {\sc and} {\sc Son, T.~C.} 2005.
\newblock {Justification and Debugging of Answer Set Programs in ASP - Prolog}.
\newblock In {\em Proceedings of the 6th International Workshop on Automated
  Debugging (AADEBUG'05)}. 49--58.

\bibitem[\protect\citeauthoryear{Erdem and {\"{O}}ztok}{Erdem and
  {\"{O}}ztok}{2015}]{ErdemO2015}
{\sc Erdem, E.} {\sc and} {\sc {\"{O}}ztok, U.} 2015.
\newblock {Generating Explanations for Biomedical Queries}.
\newblock {\em Theory and Practice of Logic Programming\/}~{\em 15,\/}~1,
  35--78.

\bibitem[\protect\citeauthoryear{Faber, Pfeifer, and Leone}{Faber
  et~al\mbox{.}}{2011}]{FaberPL2011}
{\sc Faber, W.}, {\sc Pfeifer, G.}, {\sc and} {\sc Leone, N.} 2011.
\newblock {Semantics and Complexity of Recursive Aggregates in Answer Set
  Programming}.
\newblock {\em Artificial Intelligence\/}~{\em 175,\/}~1, 278--298.

\bibitem[\protect\citeauthoryear{Fandinno}{Fandinno}{2016a}]{DFandinno16non-monotonic}
{\sc Fandinno, J.} 2016a.
\newblock Deriving conclusions from non-monotonic cause-effect relations.
\newblock {\em {Theory and Practice of Logic Programming}\/}~{\em 16,\/}~5-6,
  670--687.

\bibitem[\protect\citeauthoryear{Fandinno}{Fandinno}{2016b}]{Fandinno16}
{\sc Fandinno, J.} 2016b.
\newblock Towards deriving conclusions from cause-effect relations.
\newblock {\em Fundamenta Informaticae\/}~{\em 147,\/}~1, 93--131.

\bibitem[\protect\citeauthoryear{Febbraro, Reale, and Ricca}{Febbraro
  et~al\mbox{.}}{2010}]{FebbraroRR2010}
{\sc Febbraro, O.}, {\sc Reale, K.}, {\sc and} {\sc Ricca, F.} 2010.
\newblock {A Visual Interface for Drawing ASP Programs}.
\newblock In {\em Proceedings of the 25th Italian Conference on Computational
  Logic (CILC'10)}.

\bibitem[\protect\citeauthoryear{Febbraro, Reale, and Ricca}{Febbraro
  et~al\mbox{.}}{2011}]{FebbraroRR2011}
{\sc Febbraro, O.}, {\sc Reale, K.}, {\sc and} {\sc Ricca, F.} 2011.
\newblock {ASPIDE: Integrated Development Environment for Answer Set
  Programming}.
\newblock In {\em Proceedings of the 11th International Conference on Logic
  Programming and Nonmonotonic Reasoning (LPNMR'11)}. 317--330.

\bibitem[\protect\citeauthoryear{Ferrand, Lesaint, and Tessier}{Ferrand
  et~al\mbox{.}}{2006}]{FerrandLT2006}
{\sc Ferrand, G.}, {\sc Lesaint, W.}, {\sc and} {\sc Tessier, A.} 2006.
\newblock {Explanations and Proof Trees}.
\newblock {\em Computers and Informatics\/}~{\em 25,\/}~2-3, 105--122.

\bibitem[\protect\citeauthoryear{Fr{\"{u}}hst{\"{u}}ck, P{\"{u}}hrer, and
  Friedrich}{Fr{\"{u}}hst{\"{u}}ck et~al\mbox{.}}{2013}]{FruhstuckPF2013}
{\sc Fr{\"{u}}hst{\"{u}}ck, M.}, {\sc P{\"{u}}hrer, J.}, {\sc and} {\sc
  Friedrich, G.} 2013.
\newblock {Debugging Answer-Set Programs with Ouroboros - Extending the SeaLion
  Plugin}.
\newblock In {\em Proceedings of the 12th International Conference on Logic
  Programming and Nonmonotonic Reasoning (LPNMR'13)}. 323--328.

\bibitem[\protect\citeauthoryear{Gasteiger, Dodaro, Musitsch, Reale, Ricca, and
  Schekotihin}{Gasteiger et~al\mbox{.}}{2016}]{GasteigerDMRRS2016}
{\sc Gasteiger, P.}, {\sc Dodaro, C.}, {\sc Musitsch, B.}, {\sc Reale, K.},
  {\sc Ricca, F.}, {\sc and} {\sc Schekotihin, K.} 2016.
\newblock {An integrated Graphical User Interface for Debugging Answer Set
  Programs}.
\newblock In {\em Proceedings of the Workshop on Trends and Applications of
  Answer Set Programming (TAASP'16)}.

\bibitem[\protect\citeauthoryear{Gebser, P{\"{u}}hrer, Schaub, and
  Tompits}{Gebser et~al\mbox{.}}{2008}]{GebserPST08}
{\sc Gebser, M.}, {\sc P{\"{u}}hrer, J.}, {\sc Schaub, T.}, {\sc and} {\sc
  Tompits, H.} 2008.
\newblock A meta-programming technique for debugging answer-set programs.
\newblock In {\em Proceedings of the 23rd {AAAI} Conference on Artificial
  Intelligence (AAAI'18)}, {D.~Fox} {and} {C.~P. Gomes}, Eds. {AAAI} Press,
  448--453.

\bibitem[\protect\citeauthoryear{Gebser, Schaub, Thiele, and Veber}{Gebser
  et~al\mbox{.}}{2011}]{GebserSTV2011}
{\sc Gebser, M.}, {\sc Schaub, T.}, {\sc Thiele, S.}, {\sc and} {\sc Veber, P.}
  2011.
\newblock {Detecting Inconsistencies in Large Biological Networks with Answer
  Set Programming}.
\newblock {\em Theory and Practice of Logic Programming\/}~{\em 11,\/}~2-3,
  323--360.

\bibitem[\protect\citeauthoryear{Gelfond}{Gelfond}{2008}]{Gelfond2008}
{\sc Gelfond, M.} 2008.
\newblock {Answer Sets}.
\newblock In {\em Handbook of Knowledge Representation}. 285--316.

\bibitem[\protect\citeauthoryear{Gelfond and Lifschitz}{Gelfond and
  Lifschitz}{1988}]{GL88}
{\sc Gelfond, M.} {\sc and} {\sc Lifschitz, V.} 1988.
\newblock The stable model semantics for logic programming.
\newblock In {\em Logic Programming: Proceedings of the 5th International
  Conference and Symposium (Volume 2)}.

\bibitem[\protect\citeauthoryear{Gelfond and Lifschitz}{Gelfond and
  Lifschitz}{1991}]{GelfondL1991}
{\sc Gelfond, M.} {\sc and} {\sc Lifschitz, V.} 1991.
\newblock {Classical Negation in Logic Programs and Disjunctive Databases}.
\newblock {\em New Generation Computing\/}~{\em 9,\/}~3/4, 365--386.

\bibitem[\protect\citeauthoryear{Goodman and Flaxman}{Goodman and
  Flaxman}{2016}]{Goodman2016european}
{\sc Goodman, B.} {\sc and} {\sc Flaxman, S.} 2016.
\newblock European union regulations on algorithmic decision-making and a
  "right to explanation".
\newblock {\em arXiv preprint arXiv:1606.08813\/}.

\bibitem[\protect\citeauthoryear{Green, Karvounarakis, and Tannen}{Green
  et~al\mbox{.}}{2007}]{GreenKT07}
{\sc Green, T.~J.}, {\sc Karvounarakis, G.}, {\sc and} {\sc Tannen, V.} 2007.
\newblock Provenance semirings.
\newblock In {\em Proceedings of the 26th {ACM} {SIGACT-SIGMOD-SIGART}
  Symposium on Principles of Database Systems}, {L.~Libkin}, Ed. {ACM}, 31--40.

\bibitem[\protect\citeauthoryear{Hall}{Hall}{2004}]{hall2004}
{\sc Hall, N.} 2004.
\newblock Two concepts of causation.
\newblock In {\em Causation and counterfactuals}, {J.~Collins}, {N.~Hall},
  {and} {L.~A. Paul}, Eds. Cambridge, MA: MIT Press, 225--276.

\bibitem[\protect\citeauthoryear{Hall}{Hall}{2007}]{hall2007structural}
{\sc Hall, N.} 2007.
\newblock Structural equations and causation.
\newblock {\em Philosophical Studies\/}~{\em 132,\/}~1, 109--136.

\bibitem[\protect\citeauthoryear{Halpern}{Halpern}{2008}]{Halpern08a}
{\sc Halpern, J.~Y.} 2008.
\newblock Defaults and normality in causal structures.
\newblock In {\em Proceedings of the 11th International Conference on
  Principles of Knowledge Representation and Reasoning (KR'08)}, {G.~Brewka}
  {and} {J.~Lang}, Eds. {AAAI} Press, 198--208.

\bibitem[\protect\citeauthoryear{Hitchcock and Knobe}{Hitchcock and
  Knobe}{2009}]{hitchcock2009cause}
{\sc Hitchcock, C.} {\sc and} {\sc Knobe, J.} 2009.
\newblock Cause and norm.
\newblock {\em Journal of Philosophy\/}~{\em 11}, 587--612.

\bibitem[\protect\citeauthoryear{Inclezan}{Inclezan}{2015}]{Inclezan2015a}
{\sc Inclezan, D.} 2015.
\newblock {An Application of Answer Set Programming to the Field of Second
  Language Acquisition}.
\newblock {\em Theory and Practice of Logic Programming\/}~{\em 15,\/}~01,
  1--17.

\bibitem[\protect\citeauthoryear{Konczak, Linke, and Schaub}{Konczak
  et~al\mbox{.}}{2006}]{KonczakLS2006}
{\sc Konczak, K.}, {\sc Linke, T.}, {\sc and} {\sc Schaub, T.} 2006.
\newblock {Graphs and Colorings for Answer Set Programming}.
\newblock {\em Theory and Practice of Logic Programming\/}~{\em 6,\/}~1-2,
  61--106.

\bibitem[\protect\citeauthoryear{Lee}{Lee}{2005}]{Lee2005}
{\sc Lee, J.} 2005.
\newblock {A Model-Theoretic Counterpart of Loop formulas}.
\newblock In {\em Proceedings of the 19th International Joint Conference on
  Artificial Intelligence (IJCAI'05)}. 503--508.

\bibitem[\protect\citeauthoryear{Lef{\`{e}}vre, B{\'{e}}atrix, St{\'{e}}phan,
  and Garcia}{Lef{\`{e}}vre et~al\mbox{.}}{2017}]{LefevreBSG17}
{\sc Lef{\`{e}}vre, C.}, {\sc B{\'{e}}atrix, C.}, {\sc St{\'{e}}phan, I.}, {\sc
  and} {\sc Garcia, L.} 2017.
\newblock Asperix, a first-order forward chaining approach for answer set
  computing.
\newblock {\em {Theory and Practice of Logic Programming}\/}~{\em 17,\/}~3,
  266--310.

\bibitem[\protect\citeauthoryear{Leone, Pfeifer, Faber, Eiter, Gottlob, Perri,
  and Scarcello}{Leone et~al\mbox{.}}{2006}]{LeonePFEGPS2006}
{\sc Leone, N.}, {\sc Pfeifer, G.}, {\sc Faber, W.}, {\sc Eiter, T.}, {\sc
  Gottlob, G.}, {\sc Perri, S.}, {\sc and} {\sc Scarcello, F.} 2006.
\newblock {The DLV System for Knowledge Representation and Reasoning}.
\newblock {\em ACM Transactions on Computational Logic\/}~{\em 7,\/}~3,
  499--562.

\bibitem[\protect\citeauthoryear{Lewis}{Lewis}{1973}]{lewis1973causation}
{\sc Lewis, D.~K.} 1973.
\newblock Causation.
\newblock {\em The journal of philosophy\/}~{\em 70,\/}~17, 556--567.

\bibitem[\protect\citeauthoryear{Li, {De Vos}, Padget, Satoh, and Balke}{Li
  et~al\mbox{.}}{2015}]{LiVPSB2015}
{\sc Li, T.}, {\sc {De Vos}, M.}, {\sc Padget, J.}, {\sc Satoh, K.}, {\sc and}
  {\sc Balke, T.} 2015.
\newblock {Debugging ASP using ILP}.
\newblock In {\em Proceedings of the Technical Communications of the 31st
  International Conference on Logic Programming (ICLP'15)}.

\bibitem[\protect\citeauthoryear{Lifschitz}{Lifschitz}{2008}]{Lifschitz08}
{\sc Lifschitz, V.} 2008.
\newblock {What Is Answer Set Programming?}
\newblock In {\em Proceedings of the 23rd AAAI Conference on Artificial
  Intelligence (AAAI'08)}. 1594--1597.

\bibitem[\protect\citeauthoryear{Lifschitz}{Lifschitz}{2010}]{Lifschitz10}
{\sc Lifschitz, V.} 2010.
\newblock Thirteen definitions of a stable model.
\newblock In {\em Fields of Logic and Computation, Essays Dedicated to Yuri
  Gurevich on the Occasion of His 70th Birthday}, {A.~Blass}, {N.~Dershowitz},
  {and} {W.~Reisig}, Eds. Lecture Notes in Computer Science, vol. 6300.
  Springer, 488--503.

\bibitem[\protect\citeauthoryear{Lifschitz}{Lifschitz}{2017}]{Lifschitz17}
{\sc Lifschitz, V.} 2017.
\newblock Achievements in answer set programming.
\newblock {\em {Theory and Practice of Logic Programming}\/}~{\em 17,\/}~5-6,
  961--973.

\bibitem[\protect\citeauthoryear{Lin and Zhao}{Lin and Zhao}{2004}]{LinZ2004}
{\sc Lin, F.} {\sc and} {\sc Zhao, Y.} 2004.
\newblock {ASSAT: Computing Answer Sets of a Logic Program by SAT Solvers}.
\newblock {\em Artificial Intelligence\/}~{\em 157,\/}~1-2, 115--137.

\bibitem[\protect\citeauthoryear{Linke}{Linke}{2001}]{Linke2001}
{\sc Linke, T.} 2001.
\newblock {Graph Theoretical Characterization and Computation of Answer Sets}.
\newblock In {\em Proceedings of the 7th International Joint Conference on
  Artificial Intelligence (IJCAI'01)}. 641--648.

\bibitem[\protect\citeauthoryear{Linke and Sarsakov}{Linke and
  Sarsakov}{2004}]{LinkeS2004}
{\sc Linke, T.} {\sc and} {\sc Sarsakov, V.} 2004.
\newblock {Suitable Graphs for Answer Set Programming}.
\newblock In {\em Proceedings of the 11th International Conference on Logic for
  Programming, Artificial Intelligence, and Reasoning (LPAR'04)}. 154--168.

\bibitem[\protect\citeauthoryear{Lloyd}{Lloyd}{1987}]{Lloyd1987}
{\sc Lloyd, J.~W.} 1987.
\newblock Declarative error diagnosis.
\newblock {\em New Generation Computing\/}~{\em 5,\/}~2 (Jun), 133--154.

\bibitem[\protect\citeauthoryear{Maudlin}{Maudlin}{2004}]{maudlin2004}
{\sc Maudlin, T.} 2004.
\newblock Causation, counterfactuals, and the third factor.
\newblock In {\em Causation and Counterfactuals}, {J.~Collins}, {E.~J. Hall},
  {and} {L.~A. Paul}, Eds. MIT Press.

\bibitem[\protect\citeauthoryear{McCarthy}{McCarthy}{1977}]{McC77}
{\sc McCarthy, J.} 1977.
\newblock Epistemological problems of {A}rtificial {I}ntelligence.
\newblock In {\em Proceedings of the International Joint Conference on
  Artificial Intelligence (IJCAI)}. {MIT} Press, Cambridge, MA, 1038--1044.

\bibitem[\protect\citeauthoryear{McCarthy}{McCarthy}{1998}]{McC98}
{\sc McCarthy, J.} 1998.
\newblock Elaboration tolerance.
\newblock In {\em Proceedings of the 4th Symposium on Logical Formalizations of
  Commonsense Reasoning (Commonsense'98)}. London, UK, 198--217.
\newblock Updated version at\\
  \verb+http://www-formal.stanford.edu/jmc/elaboration.ps+.

\bibitem[\protect\citeauthoryear{Oetsch, P{\"{u}}hrer, Seidl, Tompits, and
  Zwickl}{Oetsch et~al\mbox{.}}{2011}]{OetschPSTZ2011}
{\sc Oetsch, J.}, {\sc P{\"{u}}hrer, J.}, {\sc Seidl, M.}, {\sc Tompits, H.},
  {\sc and} {\sc Zwickl, P.} 2011.
\newblock {VIDEAS: A development tool for answer-set programs based on
  model-driven engineering technology}.
\newblock In {\em Proceedings of the 11th International Conference on Logic
  Programming and Nonmonotonic Reasoning (LPNMR'11)}. 382--387.

\bibitem[\protect\citeauthoryear{Oetsch, P{\"{u}}hrer, and Tompits}{Oetsch
  et~al\mbox{.}}{2010}]{OetschPT10}
{\sc Oetsch, J.}, {\sc P{\"{u}}hrer, J.}, {\sc and} {\sc Tompits, H.} 2010.
\newblock Catching the ouroboros: On debugging non-ground answer-set programs.
\newblock {\em {Theory and Practice of Logic Programming}\/}~{\em 10,\/}~4-6,
  513--529.

\bibitem[\protect\citeauthoryear{Oetsch, P{\"{u}}hrer, and Tompits}{Oetsch
  et~al\mbox{.}}{2011}]{OetschPT2011}
{\sc Oetsch, J.}, {\sc P{\"{u}}hrer, J.}, {\sc and} {\sc Tompits, H.} 2011.
\newblock {Stepping through an Answer-Set Program}.
\newblock In {\em Proceedings of the 11th International Conference on Logic
  Programming and Nonmonotonic Reasoning (LPNMR'11)}. 134--147.

\bibitem[\protect\citeauthoryear{Oetsch, P{\"{u}}hrer, and Tompits}{Oetsch
  et~al\mbox{.}}{2012}]{OetschPT2012}
{\sc Oetsch, J.}, {\sc P{\"{u}}hrer, J.}, {\sc and} {\sc Tompits, H.} 2012.
\newblock {An FLP-Style Answer-Set Semantics for Abstract-Constraint Programs
  with Disjunctions}.
\newblock In {\em Technical Communications of the 28th International Conference
  on Logic Programming (ICLP'12)}. 222--234.

\bibitem[\protect\citeauthoryear{Oetsch, P{\"{u}}hrer, and Tompits}{Oetsch
  et~al\mbox{.}}{2018}]{OetschPT2017}
{\sc Oetsch, J.}, {\sc P{\"{u}}hrer, J.}, {\sc and} {\sc Tompits, H.} 2018.
\newblock {Stepwise Debugging of Answer-Set Programs}.
\newblock {\em {Theory and Practice of Logic Programming}\/}~{\em 18,\/}~1,
  30--80.

\bibitem[\protect\citeauthoryear{{Parliament and Council of the European
  Union}}{{Parliament and Council of the European Union}}{2016}]{GDPregulation}
{\sc {Parliament and Council of the European Union}}. 2016.
\newblock {\em Regulation (EU) 2016/679: General Data Protection Regulation}.

\bibitem[\protect\citeauthoryear{Pemmasani, Guo, Dong, Ramakrishnan, and
  Ramakrishnan}{Pemmasani et~al\mbox{.}}{2003}]{PemmasaniGDRR03}
{\sc Pemmasani, G.}, {\sc Guo, H.}, {\sc Dong, Y.}, {\sc Ramakrishnan, C.~R.},
  {\sc and} {\sc Ramakrishnan, I.~V.} 2003.
\newblock Online justification for tabled logic programs.
\newblock In {\em Proceedings of the 19th International Conference on Logic
  Programming (ICLP'03)}, {C.~Palamidessi}, Ed. Lecture Notes in Computer
  Science, vol. 2916. Springer, 500--501.

\bibitem[\protect\citeauthoryear{Pereira and Alferes}{Pereira and
  Alferes}{1992}]{PereiraA92}
{\sc Pereira, L.~M.} {\sc and} {\sc Alferes, J.~J.} 1992.
\newblock Well founded semantics for logic programs with explicit negation.
\newblock In {\em Proceedings of the 10th European Conference on Artificial
  Intelligence (ECAI'92)}. 102--106.

\bibitem[\protect\citeauthoryear{Pereira, Alferes, and Apar{\'i}cio}{Pereira
  et~al\mbox{.}}{1991}]{Pereira1991ContradictionRW}
{\sc Pereira, L.~M.}, {\sc Alferes, J.~J.}, {\sc and} {\sc Apar{\'i}cio, J.~N.}
  1991.
\newblock Contradiction removal within well founded semantics.
\newblock In {\em Proceedings of the 1st International Workshop on Logic
  Programming and Non-monotonic Reasonin (LPNMR'91)}.

\bibitem[\protect\citeauthoryear{Pereira, Aparício, and Alferes}{Pereira
  et~al\mbox{.}}{1993}]{PEREIRA1993}
{\sc Pereira, L.~M.}, {\sc Aparício, J.~N.}, {\sc and} {\sc Alferes, J.} 1993.
\newblock Non-monotonic reasoning with logic programming.
\newblock {\em The Journal of Logic Programming\/}~{\em 17,\/}~2, 227 -- 263.
\newblock Special Issue: Non-Monotonic Reasoning and Logic Programming.

\bibitem[\protect\citeauthoryear{Pereira, Dam{\'a}sio, and Alferes}{Pereira
  et~al\mbox{.}}{1993}]{pereira1993debugging}
{\sc Pereira, L.~M.}, {\sc Dam{\'a}sio, C.~V.}, {\sc and} {\sc Alferes, J.~J.}
  1993.
\newblock Debugging by diagnosing assumptions.
\newblock In {\em International Workshop on Automated and Algorithmic
  Debugging}. Springer, 58--74.

\bibitem[\protect\citeauthoryear{Perri, Ricca, Terracina, Cianni, and
  Veltri}{Perri et~al\mbox{.}}{2007}]{PerriRTCV2007}
{\sc Perri, S.}, {\sc Ricca, F.}, {\sc Terracina, G.}, {\sc Cianni, D.}, {\sc
  and} {\sc Veltri, P.} 2007.
\newblock {An Integrated Graphic Tool for Developing and Testing DLV Programs}.
\newblock In {\em Proceedings of the 1st International Workshop on Software
  Engineering for Answer Set Programming (SEA'07)}. 86--100.

\bibitem[\protect\citeauthoryear{Polleres, Fr{\"{u}}hst{\"{u}}ck, Schenner, and
  Friedrich}{Polleres et~al\mbox{.}}{2013}]{PolleresFSF2013}
{\sc Polleres, A.}, {\sc Fr{\"{u}}hst{\"{u}}ck, M.}, {\sc Schenner, G.}, {\sc
  and} {\sc Friedrich, G.} 2013.
\newblock {Debugging Non-ground ASP Programs with Choice Rules, Cardinality and
  Weight Constraints}.
\newblock In {\em Proceedings of the 12th International Conference on Logic
  Programming and Nonmonotonic Reasoning (LPNMR'13)}. 452--464.

\bibitem[\protect\citeauthoryear{Pontelli and Son}{Pontelli and
  Son}{2006}]{PontelliS2006}
{\sc Pontelli, E.} {\sc and} {\sc Son, T.~C.} 2006.
\newblock {Justifications for Logic Programs Under Answer Set Semantics}.
\newblock In {\em Proceedings of the 22nd International Conference on Logic
  Programming (ICLP'06)}. 196--210.

\bibitem[\protect\citeauthoryear{Pontelli, Son, and El-Khatib}{Pontelli
  et~al\mbox{.}}{2009}]{PontelliSE2009}
{\sc Pontelli, E.}, {\sc Son, T.~C.}, {\sc and} {\sc El-Khatib, O.} 2009.
\newblock {Justifications for Logic Programs under Answer Set Semantics}.
\newblock {\em Theory and Practice of Logic Programming\/}~{\em 9,\/}~1, 1--56.

\bibitem[\protect\citeauthoryear{P{\"{u}}hrer}{P{\"{u}}hrer}{2014}]{Puehrer2014}
{\sc P{\"{u}}hrer, J.} 2014.
\newblock {Stepwise Debugging in Answer-Set Programming: Theoretical
  Foundations and Practical Realisation}.
\newblock Ph.D. thesis.

\bibitem[\protect\citeauthoryear{Ricca, Grasso, Alviano, Manna, Lio, Iiritano,
  and Leone}{Ricca et~al\mbox{.}}{2012}]{RiccaGAMLIL2012}
{\sc Ricca, F.}, {\sc Grasso, G.}, {\sc Alviano, M.}, {\sc Manna, M.}, {\sc
  Lio, V.}, {\sc Iiritano, S.}, {\sc and} {\sc Leone, N.} 2012.
\newblock {Team-Building with Answer Set Programming in the Gioia-Tauro
  Seaport}.
\newblock {\em Theory and Practice of Logic Programming\/}~{\em 12,\/}~3,
  361--381.

\bibitem[\protect\citeauthoryear{Roychoudhury, Ramakrishnan, and
  Ramakrishnan}{Roychoudhury et~al\mbox{.}}{2000}]{roychoudhuryRR00}
{\sc Roychoudhury, A.}, {\sc Ramakrishnan, C.~R.}, {\sc and} {\sc Ramakrishnan,
  I.~V.} 2000.
\newblock Justifying proofs using memo tables.
\newblock In {\em Proceedings of the 2nd ACM SIGPLAN International Conference
  on Principles and Practice of Declarative Programming (PPDP'00)}. 178--189.

\bibitem[\protect\citeauthoryear{Schulz}{Schulz}{2017}]{Schulz2017Thesis}
{\sc Schulz, C.} 2017.
\newblock Developments in abstract and assumption-based argumentation and their
  application in logic programming.
\newblock Ph.D. thesis, Imperial College London.

\bibitem[\protect\citeauthoryear{Schulz, Satoh, and Toni}{Schulz
  et~al\mbox{.}}{2015}]{SchulzST2015}
{\sc Schulz, C.}, {\sc Satoh, K.}, {\sc and} {\sc Toni, F.} 2015.
\newblock {Characterising and Explaining Inconsistency in Logic Programs}.
\newblock In {\em Proceedings of the 13th International Conference on Logic
  Programming and Nonmonotonic Reasoning (LPNMR'15)}. 467--479.

\bibitem[\protect\citeauthoryear{Schulz, Sergot, and Toni}{Schulz
  et~al\mbox{.}}{2013}]{SchulzST2013}
{\sc Schulz, C.}, {\sc Sergot, M.}, {\sc and} {\sc Toni, F.} 2013.
\newblock {Argumentation-Based Answer Set Justification}.
\newblock In {\em Proceedings of the 11th International Symposium on Logical
  Formalizations of Commonsense Reasoning (Commonsense'13)}.

\bibitem[\protect\citeauthoryear{Schulz and Toni}{Schulz and
  Toni}{2013}]{SchulzT2013}
{\sc Schulz, C.} {\sc and} {\sc Toni, F.} 2013.
\newblock {ABA-Based Answer Set Justification}.
\newblock {\em Theory and Practice of Logic Programming\/}~{\em
  13,\/}~4-5-Online-Supplement.

\bibitem[\protect\citeauthoryear{Schulz and Toni}{Schulz and
  Toni}{2015}]{SchulzT2015}
{\sc Schulz, C.} {\sc and} {\sc Toni, F.} 2015.
\newblock {Logic Programming in Assumption-Based Argumentation Revisited -
  Semantics and Graphical Representation}.
\newblock In {\em Proceedings of the 29th AAAI Conference on Artificial
  Intelligence (AAAI'15)}. 1569--1575.

\bibitem[\protect\citeauthoryear{Schulz and Toni}{Schulz and
  Toni}{2016}]{SchulzT2016}
{\sc Schulz, C.} {\sc and} {\sc Toni, F.} 2016.
\newblock {Justifying Answer Sets using Argumentation}.
\newblock {\em Theory and Practice of Logic Programming\/}~{\em 16,\/}~01,
  59--110.

\bibitem[\protect\citeauthoryear{Shapiro}{Shapiro}{1983}]{Shapiro1983}
{\sc Shapiro, E.~Y.} 1983.
\newblock {\em Algorithmic Program DeBugging}.
\newblock MIT Press, Cambridge, MA, USA.

\bibitem[\protect\citeauthoryear{Shchekotykhin}{Shchekotykhin}{2015}]{Shchekotykhin2015}
{\sc Shchekotykhin, K.~M.} 2015.
\newblock {Interactive Query-Based Debugging of ASP Programs}.
\newblock In {\em Proceedings of the 29th AAAI Conference on Artificial
  Intelligence (AAAI'15)}. 1597--1603.

\bibitem[\protect\citeauthoryear{Specht}{Specht}{1993}]{Specht93}
{\sc Specht, G.} 1993.
\newblock Generating explanation trees even for negations in deductive database
  systems.
\newblock In {\em Proceedings of the 5th Workshop on Logic Programming
  Environments (LPE'93)}, {M.~Ducass{\'{e}}}, {B.~L. Charlier}, {Y.~Lin}, {and}
  {L.~{\"{U}}. Yal{\c{c}}inalp}, Eds. IRISA, Campus de Beaulieu, France, 8--13.

\bibitem[\protect\citeauthoryear{Sterling and Lalee}{Sterling and
  Lalee}{1986}]{sterling1986explanation}
{\sc Sterling, L.} {\sc and} {\sc Lalee, M.} 1986.
\newblock An explanation shell for expert systems.
\newblock {\em Computational Intelligence\/}~{\em 2,\/}~1, 136--141.

\bibitem[\protect\citeauthoryear{Sterling and Shapiro}{Sterling and
  Shapiro}{1994}]{sterling1994art}
{\sc Sterling, L.} {\sc and} {\sc Shapiro, E.~Y.} 1994.
\newblock {\em The art of Prolog: advanced programming techniques}.
\newblock MIT press.

\bibitem[\protect\citeauthoryear{Sterling and Yal{\c{c}}inalp}{Sterling and
  Yal{\c{c}}inalp}{1989}]{sterling1989explaining}
{\sc Sterling, L.} {\sc and} {\sc Yal{\c{c}}inalp, L.~{\"U}.} 1989.
\newblock Explaining prolog based expert systems using a layered
  meta-interpreter.
\newblock In {\em Proceedings of the 11th International Joint Conference on
  Artificial Intelligence (IJCAI'89)}. 66--71.

\bibitem[\protect\citeauthoryear{Sureshkumar, {De Vos}, Brain, and
  Fitch}{Sureshkumar et~al\mbox{.}}{2007}]{SureshkumarVBF2007}
{\sc Sureshkumar, A.}, {\sc {De Vos}, M.}, {\sc Brain, M.}, {\sc and} {\sc
  Fitch, J.} 2007.
\newblock {APE: An AnsProlog* environment}.
\newblock In {\em Proceedings of the 1st International Workshop on Software
  Engineering for Answer Set Programming (SEA'07)}. 101--115.

\bibitem[\protect\citeauthoryear{Syrj{\"{a}}nen}{Syrj{\"{a}}nen}{2006}]{Syrjanen2006}
{\sc Syrj{\"{a}}nen, T.} 2006.
\newblock {Debugging Inconsistent Answer Set Programs}.
\newblock In {\em Proceedings of the 11th International Workshop on
  Non-Monotonic Reasoning (NMR'06)}. 77--84.

\bibitem[\protect\citeauthoryear{Syrj{\"{a}}nen and
  Niemel{\"{a}}}{Syrj{\"{a}}nen and Niemel{\"{a}}}{2001}]{SyrjanenN2001}
{\sc Syrj{\"{a}}nen, T.} {\sc and} {\sc Niemel{\"{a}}, I.} 2001.
\newblock {The Smodels System}.
\newblock In {\em Proceedings of the 6th International Conference on Logic
  Programming and Nonmonotonic Reasoning (LPNMR'01)}. 434--438.

\bibitem[\protect\citeauthoryear{Ulbricht, Thimm, and Brewka}{Ulbricht
  et~al\mbox{.}}{2016}]{UlbrichtTB2016}
{\sc Ulbricht, M.}, {\sc Thimm, M.}, {\sc and} {\sc Brewka, G.} 2016.
\newblock {Measuring Inconsistency in Answer Set Programs}.
\newblock In {\em Proceedings of the 15th European Conference on Logics in
  Artificial Intelligence (JELIA'16)}. 577--583.

\bibitem[\protect\citeauthoryear{van Emden and Kowalski}{van Emden and
  Kowalski}{1976}]{vEK76}
{\sc van Emden, M.~H.} {\sc and} {\sc Kowalski, R.~A.} 1976.
\newblock The semantics of predicate logic as a programming language.
\newblock {\em Journal of the ACM\/}~{\em 23,\/}~4, 733--742.

\bibitem[\protect\citeauthoryear{Van~Gelder}{Van~Gelder}{1989}]{van1989alternating}
{\sc Van~Gelder, A.} 1989.
\newblock The alternating fixpoint of logic programs with negation.
\newblock In {\em Proceedings of the 8th ACM SIGACT-SIGMOD-SIGART Symposium on
  Principles of Database Systems}. ACM, 1--10.

\bibitem[\protect\citeauthoryear{Van~Gelder, Ross, and Schlipf}{Van~Gelder
  et~al\mbox{.}}{1988}]{van1988unfounded}
{\sc Van~Gelder, A.}, {\sc Ross, K.}, {\sc and} {\sc Schlipf, J.~S.} 1988.
\newblock Unfounded sets and well-founded semantics for general logic programs.
\newblock In {\em Proceedings of the 7th ACM SIGACT-SIGMOD-SIGART Symposium on
  Principles of Database Systems}. ACM, 221--230.

\bibitem[\protect\citeauthoryear{Van~Gelder, Ross, and Schlipf}{Van~Gelder
  et~al\mbox{.}}{1991}]{van1991well}
{\sc Van~Gelder, A.}, {\sc Ross, K.~A.}, {\sc and} {\sc Schlipf, J.~S.} 1991.
\newblock The well-founded semantics for general logic programs.
\newblock {\em Journal of the ACM (JACM)\/}~{\em 38,\/}~3, 619--649.

\bibitem[\protect\citeauthoryear{You and Yuan}{You and Yuan}{1994}]{YouY1994}
{\sc You, J.-H.} {\sc and} {\sc Yuan, L.~Y.} 1994.
\newblock {A Three-Valued Semantics for Deductive Databases and Logic
  Programs}.
\newblock {\em Journal of Computer and System Sciences\/}~{\em 49,\/}~2,
  334--361.

\end{thebibliography}

\end{document}